\pdfoutput=1
\documentclass[sigconf,nonacm]{acmart}

\setcopyright{none}
\renewcommand\footnotetextcopyrightpermission[1]{}

\usepackage{multirow}
\usepackage{colortbl}
\usepackage{adjustbox}
\usepackage{tabularx}
\usepackage{xspace}
\usepackage{algorithm}
\usepackage{algpseudocode}
\usepackage{enumitem}

\makeatletter
\newcommand{\res}[1]{\ifcsname res@#1\endcsname\csname res@#1\endcsname\else\textbf{??\detokenize{#1}}\PackageWarning{results}{Undefined result key #1}\fi}
\makeatother
\expandafter\def\csname res@beta.cf.n-runs\endcsname{7{,}650}
\expandafter\def\csname res@beta.fc.gnn.real.nse.mean\endcsname{0.934}
\expandafter\def\csname res@beta.fc.hybrid.q0.nse.mean\endcsname{0.479}
\expandafter\def\csname res@beta.fc.hybrid.real.nse.mean\endcsname{0.944}
\expandafter\def\csname res@beta.fc.persistence.nse\endcsname{0.267}
\expandafter\def\csname res@beta.fc.physonly.q0.nse.mean\endcsname{0.305}
\expandafter\def\csname res@beta.fig2a.r.gnn-untrained.signed-err\endcsname{$-$48.4}
\expandafter\def\csname res@beta.fig2a.r.gnn.rho.boot-hi\endcsname{0.94}
\expandafter\def\csname res@beta.fig2a.r.gnn.rho.boot-lo\endcsname{0.93}
\expandafter\def\csname res@beta.fig2a.r.gnn.rho.mean\endcsname{0.94}
\expandafter\def\csname res@beta.fig2a.r.gnn.signed-err.mean\endcsname{$-$3.0}
\expandafter\def\csname res@beta.fig2a.r.hybrid-untrained.rho\endcsname{0.15}
\expandafter\def\csname res@beta.fig2a.r.hybrid.rho.boot-hi\endcsname{0.95}
\expandafter\def\csname res@beta.fig2a.r.hybrid.rho.boot-lo\endcsname{0.93}
\expandafter\def\csname res@beta.fig2a.r.hybrid.rho.mean\endcsname{0.94}
\expandafter\def\csname res@beta.fig2a.r.hybrid.signed-err.mean\endcsname{$-$8.0}
\expandafter\def\csname res@beta.fig2a.r.physonly.rho.boot-hi\endcsname{0.78}
\expandafter\def\csname res@beta.fig2a.r.physonly.rho.boot-lo\endcsname{0.76}
\expandafter\def\csname res@beta.fig2a.r.physonly.rho.mean\endcsname{0.77}
\expandafter\def\csname res@beta.fig2a.r.physonly.signed-err.mean\endcsname{$-$14.5}
\expandafter\def\csname res@beta.fig2a.r.physres-untrained.rho\endcsname{0.39}
\expandafter\def\csname res@beta.fig2a.r.physres-untrained.signed-err\endcsname{$-$48.2}
\expandafter\def\csname res@beta.fig2a.r.physres.rho.boot-hi\endcsname{0.84}
\expandafter\def\csname res@beta.fig2a.r.physres.rho.boot-lo\endcsname{0.81}
\expandafter\def\csname res@beta.fig2a.r.physres.rho.mean\endcsname{0.83}
\expandafter\def\csname res@beta.fig2a.r.physres.signed-err.mean\endcsname{$-$13.6}
\expandafter\def\csname res@beta.fig2a.s.gnn.abs-err.mean\endcsname{2.31}
\expandafter\def\csname res@beta.fig2a.s.gnn.rho.mean\endcsname{0.03}
\expandafter\def\csname res@beta.fig2a.s.hybrid.abs-err.mean\endcsname{2.28}
\expandafter\def\csname res@beta.fig2a.s.hybrid.rho.mean\endcsname{0.02}
\expandafter\def\csname res@beta.fig2a.s.physonly-untrained.rho\endcsname{0.13}
\expandafter\def\csname res@beta.fig2a.s.physonly.abs-err.mean\endcsname{2.33}
\expandafter\def\csname res@beta.fig2a.s.physonly.rho.boot-lo\endcsname{0.15}
\expandafter\def\csname res@beta.fig2a.s.physonly.rho.mean\endcsname{0.24}
\expandafter\def\csname res@beta.fig2a.s.physres-untrained.rho\endcsname{0.17}
\expandafter\def\csname res@beta.fig2a.s.physres.abs-err.mean\endcsname{2.33}
\expandafter\def\csname res@beta.fig2a.s.physres.rho.boot-lo\endcsname{0.16}
\expandafter\def\csname res@beta.fig2a.s.physres.rho.mean\endcsname{0.29}
\expandafter\def\csname res@beta.fig2a.share-disc.r\endcsname{96}
\expandafter\def\csname res@beta.fig2a.share-disc.s\endcsname{9}
\expandafter\def\csname res@beta.fig2a.true-abs-phi-disc.r\endcsname{48.5}
\expandafter\def\csname res@beta.fig2a.true-abs-phi-disc.s\endcsname{2.36}
\expandafter\def\csname res@beta.fig3.d9wm4.n_nodes_mean_gt1cm_s1\endcsname{1}
\expandafter\def\csname res@beta.fig3.gconv.n_nodes_mean_gt1cm_s1\endcsname{4}
\expandafter\def\csname res@beta.fig3.grid.gconv.cross_a25_s30\endcsname{26}
\expandafter\def\csname res@beta.fig3.grid.gconv.mean_s30\endcsname{1.2}
\expandafter\def\csname res@beta.fig3.grid.n_unwarned_near\endcsname{1{,}735}
\expandafter\def\csname res@beta.fig3.grid.sim_cross_a25_s30\endcsname{74}
\expandafter\def\csname res@beta.fig3.grid.sim_cross_rain_only_max\endcsname{48}
\expandafter\def\csname res@beta.fig3.grid.sim_cross_stage_only_max\endcsname{37}
\expandafter\def\csname res@beta.fig3.grid.sim_mean_a25\endcsname{8.4}
\expandafter\def\csname res@beta.fig3.grid.sim_mean_a25_s30\endcsname{18.6}
\expandafter\def\csname res@beta.fig3.grid.sim_mean_interaction_a25_s30\endcsname{2.2}
\expandafter\def\csname res@beta.fig3.grid.sim_mean_s30\endcsname{8.1}
\expandafter\def\csname res@beta.fig3.grid.sim_median_a0_s8\endcsname{0.21}
\expandafter\def\csname res@beta.fig3.grid.sim_median_a6_s0\endcsname{0.85}
\expandafter\def\csname res@beta.fig3.grid.sim_median_a6_s8\endcsname{5.74}
\expandafter\def\csname res@beta.fig3.grid.wm3fpd.cross_max\endcsname{32}
\expandafter\def\csname res@beta.fig3.grid.wm3fpd.mean_s30\endcsname{0.7}
\expandafter\def\csname res@beta.fig3.irs.coalition.gconv.recall\endcsname{3.5}
\expandafter\def\csname res@beta.fig3.irs.coalition.n_issue_flagged\endcsname{24}
\expandafter\def\csname res@beta.fig3.irs.coalition.share_det\endcsname{5.2}
\expandafter\def\csname res@beta.fig3.irs.coalition.share_det_quality\endcsname{4.8}
\expandafter\def\csname res@beta.fig3.irs.coalition.share_gt_2nf\endcsname{10.6}
\expandafter\def\csname res@beta.fig3.irs.d9wm4.trained.mae_all\endcsname{1.79}
\expandafter\def\csname res@beta.fig3.irs.gconv.trained.mae_all\endcsname{1.80}
\expandafter\def\csname res@beta.fig3.irs.nointeraction.mae_all\endcsname{1.79}
\expandafter\def\csname res@beta.fig3.irs.nointeraction.mae_det\endcsname{6.5}
\expandafter\def\csname res@beta.fig3.irs.sim.share_det\endcsname{23.7}
\expandafter\def\csname res@beta.fig3.irs.sim.share_det_2nf\endcsname{40.0}
\expandafter\def\csname res@beta.fig3.irs.sim.share_det_4nf_nofloor\endcsname{34.9}
\expandafter\def\csname res@beta.fig3.irs.sim.share_positive\endcsname{64}
\expandafter\def\csname res@beta.fig3.sim.maxpath_affected_m\endcsname{1{,}458}
\expandafter\def\csname res@beta.fig3.sim.n_nodes_affected_s1\endcsname{23}
\expandafter\def\csname res@beta.fig3.sim.n_nodes_mean_gt1cm_s1\endcsname{86}
\expandafter\def\csname res@beta.fig3.sim.near_mean_s1\endcsname{8.1}
\expandafter\def\csname res@beta.fig3.sim.near_median_s1\endcsname{2.0}
\expandafter\def\csname res@beta.fig3.wm3fpd.n_nodes_mean_gt1cm_s1\endcsname{3--4}
\expandafter\def\csname res@beta.gen.antecedent-max-mm\endcsname{12.7}
\expandafter\def\csname res@beta.gen.antecedent-prob\endcsname{0.5}
\expandafter\def\csname res@beta.gen.antecedent-window-h\endcsname{6}
\expandafter\def\csname res@beta.gen.main-depth-max-mm\endcsname{177.8}
\expandafter\def\csname res@beta.gen.main-depth-min-mm\endcsname{12.7}
\expandafter\def\csname res@beta.gen.main-start-max-h\endcsname{30}
\expandafter\def\csname res@beta.gen.main-start-min-h\endcsname{12}
\expandafter\def\csname res@beta.gen.surge-amp-max-cm\endcsname{61.0}
\expandafter\def\csname res@beta.gen.surge-centre-max-h\endcsname{42}
\expandafter\def\csname res@beta.gen.surge-centre-min-h\endcsname{6}
\expandafter\def\csname res@beta.gen.surge-prob\endcsname{0.6}
\expandafter\def\csname res@beta.gen.surge-sigma-max-h\endcsname{10}
\expandafter\def\csname res@beta.gen.surge-sigma-min-h\endcsname{3}
\expandafter\def\csname res@beta.gen.tide-offset-max-cm\endcsname{106.7}
\expandafter\def\csname res@beta.gen.tide-range-scale-max\endcsname{1.3}
\expandafter\def\csname res@beta.gen.tide-range-scale-min\endcsname{0.8}
\expandafter\def\csname res@beta.hist.agree.gnn.mean\endcsname{0.826}
\expandafter\def\csname res@beta.hist.agree.physonly.mean\endcsname{0.701}
\expandafter\def\csname res@beta.k.rain.gnn-half.k-median.mean\endcsname{0.33}
\expandafter\def\csname res@beta.k.rain.gnn.k-median.mean\endcsname{0.71}
\expandafter\def\csname res@beta.k.rain.gnn.k-median.sd\endcsname{0.13}
\expandafter\def\csname res@beta.k.rain.gnn.k-p10.mean\endcsname{$-$0.51}
\expandafter\def\csname res@beta.k.rain.gnn.share-negative.mean\endcsname{19.5}
\expandafter\def\csname res@beta.k.rain.hybrid-half.k-median.mean\endcsname{0.58}
\expandafter\def\csname res@beta.k.rain.hybrid-reversed.k-median.mean\endcsname{$-$0.18}
\expandafter\def\csname res@beta.k.rain.hybrid.k-median.mean\endcsname{0.92}
\expandafter\def\csname res@beta.k.rain.hybrid.share-negative.mean\endcsname{1.9}
\expandafter\def\csname res@beta.k.rain.n-cells\endcsname{17{,}543}
\expandafter\def\csname res@beta.k.rain.physonly-half.k-median.mean\endcsname{0.47}
\expandafter\def\csname res@beta.k.rain.physonly.k-median.mean\endcsname{0.86}
\expandafter\def\csname res@beta.k.rain.physonly.share-negative.mean\endcsname{0.3}
\expandafter\def\csname res@beta.k.rain.physres-half.k-median.mean\endcsname{0.58}
\expandafter\def\csname res@beta.k.rain.physres-untrained.k-median\endcsname{0.01}
\expandafter\def\csname res@beta.k.rain.physres.k-median.mean\endcsname{0.97}
\expandafter\def\csname res@beta.k.rain.physres.share-negative.mean\endcsname{0.1}
\expandafter\def\csname res@beta.k.rain.zero-response.mae\endcsname{10.82}
\expandafter\def\csname res@beta.k.stage.gnn.k-median.mean\endcsname{0.00}
\expandafter\def\csname res@beta.k.stage.hybrid.k-median.mean\endcsname{0.00}
\expandafter\def\csname res@beta.k.stage.mae.reduction-vs-zero.max\endcsname{4.7}
\expandafter\def\csname res@beta.k.stage.n-cells\endcsname{1{,}824}
\expandafter\def\csname res@beta.k.stage.physonly.k-median.mean\endcsname{0.00}
\expandafter\def\csname res@beta.k.stage.physres-untrained.k-median\endcsname{0.00}
\expandafter\def\csname res@beta.k.stage.physres.k-median.mean\endcsname{0.00}
\expandafter\def\csname res@beta.k.stage.zero-response.mae\endcsname{8.14}
\expandafter\def\csname res@beta.nodes.near\endcsname{24}
\expandafter\def\csname res@beta.nodes.rain\endcsname{163}
\expandafter\def\csname res@beta.nodes.scored\endcsname{207}
\expandafter\def\csname res@beta.noise.pert.all.max\endcsname{65.9}
\expandafter\def\csname res@beta.noise.pert.all.median\endcsname{0.023}
\expandafter\def\csname res@beta.noise.pert.all.p90\endcsname{1.30}
\expandafter\def\csname res@beta.noise.pert.all.share-le-1cm\endcsname{86.4}
\expandafter\def\csname res@beta.noise.pert.j74.median\endcsname{10.51}
\expandafter\def\csname res@beta.noise.pert.j74.p90\endcsname{55.51}
\expandafter\def\csname res@beta.noise.pert.j76.median\endcsname{5.25}
\expandafter\def\csname res@beta.noise.pert.j76.p90\endcsname{28.35}
\expandafter\def\csname res@beta.noise.pert.near.median\endcsname{0.16}
\expandafter\def\csname res@beta.noise.pert.states6.median\endcsname{0.064}
\expandafter\def\csname res@beta.noise.pertpost.all.median\endcsname{0.016}
\expandafter\def\csname res@beta.quality.cont-max\endcsname{9.2}
\expandafter\def\csname res@beta.quality.cont-median\endcsname{0.36}
\expandafter\def\csname res@beta.quality.cont-min\endcsname{$-$25.7}
\expandafter\def\csname res@beta.quality.n-abs-gt-2pct\endcsname{90}
\expandafter\def\csname res@beta.quality.n-abs-gt-2pct-norain-coalitions\endcsname{84}
\expandafter\def\csname res@beta.quality.n-abs-gt-5pct\endcsname{38}
\expandafter\def\csname res@beta.quality.preissue-identical-runs\endcsname{7{,}500}
\expandafter\def\csname res@beta.quality.readback-rain-mm-max\endcsname{$3\times10^{-8}$}
\expandafter\def\csname res@beta.quality.readback-stage-cm-max\endcsname{$8\times10^{-14}$}
\expandafter\def\csname res@beta.quality.shapley-usable-2pct\endcsname{125}
\expandafter\def\csname res@beta.quality.shapley-usable-primary\endcsname{139}
\expandafter\def\csname res@beta.rf.gnn.beyond_aff23\endcsname{19}
\expandafter\def\csname res@beta.rf.gnn.beyond_aff86\endcsname{82}
\expandafter\def\csname res@beta.rf.gnn.field_size\endcsname{4}
\expandafter\def\csname res@beta.rf.gnn.k_stage_in_reach.mean\endcsname{0.43}
\expandafter\def\csname res@beta.rf.gnn.k_stage_in_reach.n_cells\endcsname{394}
\expandafter\def\csname res@beta.rf.gnn.k_stage_in_reach.untrained\endcsname{0.00}
\expandafter\def\csname res@beta.rf.gnn.kstage_cells_in_reach_pct\endcsname{21.6}
\expandafter\def\csname res@beta.rf.gnn.layers\endcsname{3}
\expandafter\def\csname res@beta.rf.gnn.n_responding\endcsname{4}
\expandafter\def\csname res@beta.rf.hybrid.beyond_aff23\endcsname{17}
\expandafter\def\csname res@beta.rf.hybrid.beyond_aff86\endcsname{80}
\expandafter\def\csname res@beta.rf.hybrid.enc_layers\endcsname{4}
\expandafter\def\csname res@beta.rf.hybrid.field_size\endcsname{6}
\expandafter\def\csname res@beta.rf.hybrid.k_stage_in_reach.mean\endcsname{0.08}
\expandafter\def\csname res@beta.rf.hybrid.k_stage_in_reach.n_cells\endcsname{592}
\expandafter\def\csname res@beta.rf.hybrid.kstage_cells_in_reach_pct\endcsname{32.5}
\expandafter\def\csname res@beta.rf.hybrid.n_responding\endcsname{3--4}
\expandafter\def\csname res@beta.rf.hybrid.phys_reach_links\endcsname{1}
\expandafter\def\csname res@beta.rf.max_change_outside_reach\endcsname{$1.4\times10^{-4}$}
\expandafter\def\csname res@beta.rf.phys.rollout_steps\endcsname{24}
\expandafter\def\csname res@beta.rf.physres.beyond_aff23\endcsname{4}
\expandafter\def\csname res@beta.rf.physres.beyond_aff86\endcsname{11}
\expandafter\def\csname res@beta.rf.physres.field_size\endcsname{93}
\expandafter\def\csname res@beta.rf.physres.k_stage_in_reach.mean\endcsname{0.00}
\expandafter\def\csname res@beta.rf.physres.k_stage_in_reach.n_cells\endcsname{5{,}036}
\expandafter\def\csname res@beta.rf.stage.physres.affected_in_reach_share\endcsname{83}
\expandafter\def\csname res@beta.s10.stage.gnn.ba.mean\endcsname{0.65}
\expandafter\def\csname res@beta.s10.stage.gnn.n-both.mean\endcsname{57}
\expandafter\def\csname res@beta.s10.stage.gnn.tp-from-ifa.mean\endcsname{84}
\expandafter\def\csname res@beta.s10.stage.hybrid.tp-from-ifa.mean\endcsname{94}
\expandafter\def\csname res@beta.s10.stage.physonly.n-both.mean\endcsname{0}
\expandafter\def\csname res@beta.s10.stage.physonly.tp-from-ifa.mean\endcsname{100}
\expandafter\def\csname res@beta.s10.stage.physres.tp-from-ifa.mean\endcsname{99}
\expandafter\def\csname res@beta.s10.stage.pos-rate\endcsname{7.8}
\expandafter\def\csname res@beta.s10.stage.strict.gnn.ba.mean\endcsname{0.54}
\expandafter\def\csname res@beta.s10.stage.strict.hybrid.ba.mean\endcsname{0.50}
\expandafter\def\csname res@beta.s12.k.rain.affected.coverage\endcsname{82}
\expandafter\def\csname res@beta.s12.k.rain.affected.d9phys\endcsname{0.91}
\expandafter\def\csname res@beta.s12.k.rain.affected.d9wm4\endcsname{1.02}
\expandafter\def\csname res@beta.s12.k.rain.affected.gconv\endcsname{0.80}
\expandafter\def\csname res@beta.s12.k.rain.affected.wm3fpd\endcsname{1.00}
\expandafter\def\csname res@beta.s12.k.stage.affected.coverage\endcsname{60}
\expandafter\def\csname res@beta.s12.k.stage.max-over-amounts\endcsname{0.00}
\expandafter\def\csname res@beta.s12.k.stage.near.coverage\endcsname{51}
\expandafter\def\csname res@beta.s12.layer.affected_rain.n\endcsname{164}
\expandafter\def\csname res@beta.s12.layer.affected_stage.n\endcsname{23}
\expandafter\def\csname res@beta.s12.nonmono.rain.allscored.d9phys\endcsname{1.1}
\expandafter\def\csname res@beta.s12.nonmono.rain.allscored.d9wm4\endcsname{0.4}
\expandafter\def\csname res@beta.s12.nonmono.rain.allscored.gconv\endcsname{38.2}
\expandafter\def\csname res@beta.s12.nonmono.rain.allscored.simulator\endcsname{2.8}
\expandafter\def\csname res@beta.s12.nonmono.rain.allscored.wm3fpd\endcsname{4.9}
\expandafter\def\csname res@beta.s12.nonmono.rain.nearoutlet.simulator\endcsname{10.0}
\expandafter\def\csname res@beta.s12.nonmono.stage.nearoutlet.gconv\endcsname{5.0}
\expandafter\def\csname res@beta.s13.base_peak_boundary\endcsname{11}
\expandafter\def\csname res@beta.s13.grid.recall_max.d9phys\endcsname{0.0}
\expandafter\def\csname res@beta.s13.grid.recall_max.d9wm4\endcsname{0.1}
\expandafter\def\csname res@beta.s13.grid.recall_max.gconv\endcsname{1.0}
\expandafter\def\csname res@beta.s13.grid.recall_max.wm3fpd\endcsname{0.6}
\expandafter\def\csname res@beta.s13.grid.share_det.max\endcsname{32}
\expandafter\def\csname res@beta.s13.grid.share_det.min\endcsname{5}
\expandafter\def\csname res@beta.s13.joint_peak_time_new\endcsname{42}
\expandafter\def\csname res@beta.s13.peak_det_also_fixed\endcsname{70}
\expandafter\def\csname res@beta.s13.peakshift.joint.any\endcsname{35}
\expandafter\def\csname res@beta.s13.peakshift.joint.det\endcsname{66}
\expandafter\def\csname res@beta.s13.share_det.fixed\endcsname{18.3}
\expandafter\def\csname res@beta.s13.share_det.maxt\endcsname{44.6}
\expandafter\def\csname res@beta.s13.share_det.peak\endcsname{23.7}
\expandafter\def\csname res@beta.s13.sign_peak_vs_fixed\endcsname{95}
\expandafter\def\csname res@beta.s14.cont-1.max-abs-delta-ba\endcsname{0.016}
\expandafter\def\csname res@beta.s14.cont-1.max-abs-delta-k\endcsname{0.05}
\expandafter\def\csname res@beta.s14.m1.det-irs\endcsname{13.5}
\expandafter\def\csname res@beta.s14.m5.det-irs\endcsname{7.7}
\expandafter\def\csname res@beta.s14.no-bsens-2nf.max-abs-delta-ba\endcsname{0.004}
\expandafter\def\csname res@beta.s14.primary.det-irs\endcsname{10.5}
\expandafter\def\csname res@beta.s4.3a.d9phys.k_stage\endcsname{0.000}
\expandafter\def\csname res@beta.s4.3a.d9phys.share_stage\endcsname{0.042}
\expandafter\def\csname res@beta.s4.3a.d9wm4.share_stage\endcsname{0.032}
\expandafter\def\csname res@beta.s4.3a.gconv.adj4.mean_range\endcsname{4.9--9.2}
\expandafter\def\csname res@beta.s4.3a.gconv.adj4.share_range\endcsname{40--75}
\expandafter\def\csname res@beta.s4.3a.gconv.k_rain\endcsname{0.71}
\expandafter\def\csname res@beta.s4.3a.gconv.share_stage\endcsname{0.188}
\expandafter\def\csname res@beta.s4.3a.wm3fpd.adj4.mean_range\endcsname{1.7--7.9}
\expandafter\def\csname res@beta.s4.3a.wm3fpd.adj4.share_range\endcsname{13--64}
\expandafter\def\csname res@beta.s4.3a.wm3fpd.share_stage\endcsname{0.102}
\expandafter\def\csname res@beta.s4.dastro.median\endcsname{35.1}
\expandafter\def\csname res@beta.s4.dsurge.median\endcsname{0.03}
\expandafter\def\csname res@beta.s4.dsurge.p99\endcsname{18.8}
\expandafter\def\csname res@beta.s4.dsurge.share_gt10\endcsname{7.2}
\expandafter\def\csname res@beta.s4.h1.gconv.k_stage\endcsname{0.000}
\expandafter\def\csname res@beta.s4.h1.gconv.share_stage\endcsname{0.178}
\expandafter\def\csname res@beta.s4.h1.wm3fpd.share_stage\endcsname{0.049}
\expandafter\def\csname res@beta.s4.h2.gconv.k_rain\endcsname{0.58}
\expandafter\def\csname res@beta.s4.h2.gconv.share_stage\endcsname{0.226}
\expandafter\def\csname res@beta.s4.h2.wm3fpd.share_stage\endcsname{0.145}
\expandafter\def\csname res@beta.s4.h3.d9phys.share_stage\endcsname{0.001}
\expandafter\def\csname res@beta.s4.h3.d9wm4.share_stage\endcsname{0.006}
\expandafter\def\csname res@beta.s4.pr2.astro.max\endcsname{0.10}
\expandafter\def\csname res@beta.s4.pr2.astro.median\endcsname{0.04}
\expandafter\def\csname res@beta.s4.pr2.rain.max\endcsname{0.92}
\expandafter\def\csname res@beta.s4.pr2.rain.median\endcsname{0.64}
\expandafter\def\csname res@beta.s4.pr2.surge.max\endcsname{0.0007}
\expandafter\def\csname res@beta.s4.pr2.surge.median\endcsname{0.00006}
\expandafter\def\csname res@beta.s4.sim.adj4.mean_range\endcsname{12.2--12.4}
\expandafter\def\csname res@beta.s4.sim.n_near_median_gt1cm\endcsname{16}
\expandafter\def\csname res@beta.s4.sim.other20.mean_range\endcsname{1.1--11.8}
\expandafter\def\csname res@beta.s4.support.h1.surge_rise.p99\endcsname{48.5}
\expandafter\def\csname res@beta.s4.support.h2.surge_rise.p99\endcsname{37.5}
\expandafter\def\csname res@beta.s4.support.test.rain76.above_max\endcsname{25}
\expandafter\def\csname res@beta.s4.support.test.rain76.above_p99\endcsname{42}
\expandafter\def\csname res@beta.s4.support.test.stage30.above_max\endcsname{0}
\expandafter\def\csname res@beta.s4.support.test.stage30.above_p99\endcsname{91}
\expandafter\def\csname res@beta.s4.support.test.stage61.above_max\endcsname{99}
\expandafter\def\csname res@beta.s4.support.v5.rain_total.max\endcsname{149}
\expandafter\def\csname res@beta.s4.support.v5.rain_total.p99\endcsname{119}
\expandafter\def\csname res@beta.s4.support.v5.surge_rise.max\endcsname{52.5}
\expandafter\def\csname res@beta.s4.support.v5.surge_rise.p99\endcsname{28.3}
\expandafter\def\csname res@beta.s4.train_windows\endcsname{43{,}800}
\expandafter\def\csname res@beta.s9.rain.gnn-half.slope-fe.mean\endcsname{1.45}
\expandafter\def\csname res@beta.s9.rain.gnn-reversed.slope-pooled.mean\endcsname{0.16}
\expandafter\def\csname res@beta.s9.rain.gnn.slope-fe.mean\endcsname{0.98}
\expandafter\def\csname res@beta.s9.rain.gnn.slope-pooled.mean\endcsname{0.98}
\expandafter\def\csname res@beta.s9.rain.half.slope-pooled.max\endcsname{1.87}
\expandafter\def\csname res@beta.s9.rain.half.slope-pooled.min\endcsname{1.46}
\expandafter\def\csname res@beta.s9.rain.hybrid-reversed.slope-fe.mean\endcsname{1.50}
\expandafter\def\csname res@beta.s9.rain.hybrid-reversed.slope-pooled.mean\endcsname{1.51}
\expandafter\def\csname res@beta.s9.rain.physonly-reversed.slope-pooled.mean\endcsname{$-$2.44}
\expandafter\def\csname res@beta.s9.rain.physres-half.slope-fe.mean\endcsname{1.90}
\expandafter\def\csname res@beta.s9.rain.physres-reversed.slope-pooled.mean\endcsname{$-$1.36}
\expandafter\def\csname res@beta.s9.rain.physres.slope-fe.mean\endcsname{1.20}
\expandafter\def\csname res@beta.s9.surge.hybrid.slope-pooled.mean\endcsname{0.90}
\expandafter\def\csname res@beta.s9.surge.mean-abs-dpred.max\endcsname{0.08}
\expandafter\def\csname res@beta.s9.surge.mean-abs-dpred.min\endcsname{0.01}
\expandafter\def\csname res@beta.s9.surge.physres.slope-pooled.mean\endcsname{$-$1.67}
\expandafter\def\csname res@beta.tab3.always.ba\endcsname{0.50}
\expandafter\def\csname res@beta.tab3.boot.gnn-minus-linear.ba.hi\endcsname{0.08}
\expandafter\def\csname res@beta.tab3.boot.gnn-minus-linear.ba.lo\endcsname{0.03}
\expandafter\def\csname res@beta.tab3.boot.gnn-minus-linear.ba.point\endcsname{0.05}
\expandafter\def\csname res@beta.tab3.boot.hybrid-minus-gnn.ba.hi\endcsname{0.10}
\expandafter\def\csname res@beta.tab3.boot.hybrid-minus-gnn.ba.lo\endcsname{0.07}
\expandafter\def\csname res@beta.tab3.boot.hybrid-minus-gnn.ba.point\endcsname{0.09}
\expandafter\def\csname res@beta.tab3.boot.hybrid-minus-linear.ba.hi\endcsname{0.16}
\expandafter\def\csname res@beta.tab3.boot.hybrid-minus-linear.ba.lo\endcsname{0.12}
\expandafter\def\csname res@beta.tab3.boot.hybrid-minus-linear.ba.point\endcsname{0.14}
\expandafter\def\csname res@beta.tab3.boot.physonly-minus-linear.ba.hi\endcsname{0.11}
\expandafter\def\csname res@beta.tab3.boot.physonly-minus-linear.ba.lo\endcsname{0.07}
\expandafter\def\csname res@beta.tab3.boot.physonly-minus-linear.ba.point\endcsname{0.09}
\expandafter\def\csname res@beta.tab3.boot.physres-minus-linear.ba.hi\endcsname{0.14}
\expandafter\def\csname res@beta.tab3.boot.physres-minus-linear.ba.lo\endcsname{0.09}
\expandafter\def\csname res@beta.tab3.boot.physres-minus-linear.ba.point\endcsname{0.11}
\expandafter\def\csname res@beta.tab3.common-both.gnn.err-mm.mean\endcsname{8.8}
\expandafter\def\csname res@beta.tab3.common-both.hybrid.err-mm.mean\endcsname{11.0}
\expandafter\def\csname res@beta.tab3.common-both.n\endcsname{3{,}249}
\expandafter\def\csname res@beta.tab3.common-both.physonly.err-mm.mean\endcsname{12.6}
\expandafter\def\csname res@beta.tab3.common-both.physres.err-mm.mean\endcsname{12.3}
\expandafter\def\csname res@beta.tab3.gnn-untrained.ba\endcsname{0.44}
\expandafter\def\csname res@beta.tab3.gnn.ba.mean\endcsname{0.63}
\expandafter\def\csname res@beta.tab3.gnn.init-fa-share.mean\endcsname{8.0}
\expandafter\def\csname res@beta.tab3.hybrid-half.ba.mean\endcsname{0.52}
\expandafter\def\csname res@beta.tab3.hybrid.ba.mean\endcsname{0.72}
\expandafter\def\csname res@beta.tab3.hybrid.tp-from-ifa.mean\endcsname{9}
\expandafter\def\csname res@beta.tab3.linear.ba\endcsname{0.58}
\expandafter\def\csname res@beta.tab3.n\endcsname{12{,}778}
\expandafter\def\csname res@beta.tab3.physonly.ba.mean\endcsname{0.67}
\expandafter\def\csname res@beta.tab3.physonly.init-fa-share.mean\endcsname{5.9}
\expandafter\def\csname res@beta.tab3.physres.ba.mean\endcsname{0.69}
\expandafter\def\csname res@beta.tab3.physres.tp-from-ifa.mean\endcsname{7}
\expandafter\def\csname res@beta.tab3.pos-rate\endcsname{88.8}
\expandafter\def\csname res@beta.tab3.positives\endcsname{11{,}341}
\expandafter\def\csname res@beta.tab3.strict.gnn.ba.mean\endcsname{0.68}
\expandafter\def\csname res@beta.tab3.strict.hybrid.ba.mean\endcsname{0.78}
\expandafter\def\csname res@beta.tab3.strict.linear.ba\endcsname{0.62}
\expandafter\def\csname res@beta.tab3.strict.n\endcsname{10{,}913}
\expandafter\def\csname res@beta.tab3.strict.physonly.ba.mean\endcsname{0.74}
\expandafter\def\csname res@beta.tab3.strict.physres.ba.mean\endcsname{0.76}
\expandafter\def\csname res@crms.acc.test.gconv.dnse\endcsname{$-$0.019}
\expandafter\def\csname res@crms.conv3.gnn.converged-seeds\endcsname{3}
\expandafter\def\csname res@crms.conv3.test.dnse\endcsname{$-$0.005}
\expandafter\def\csname res@crms.data.n_basins\endcsname{9}
\expandafter\def\csname res@crms.data.n_candidates\endcsname{466}
\expandafter\def\csname res@crms.data.n_rain_inputs\endcsname{168}
\expandafter\def\csname res@crms.data.n_stations\endcsname{285}
\expandafter\def\csname res@crms.data.n_tide_gauges\endcsname{11}
\expandafter\def\csname res@crms.graph.check_corr_hourly_change.edges_median\endcsname{0.50}
\expandafter\def\csname res@crms.graph.check_corr_hourly_change.same_basin_nonedge_pairs_le_maxkm_median\endcsname{0.42}
\expandafter\def\csname res@crms.phys.acc.test.gnn.dnse\endcsname{0.004}
\expandafter\def\csname res@crms.phys.acc.test.hybrid.dnse\endcsname{$-$0.011}
\expandafter\def\csname res@crms.phys.cf.a_max_mm\endcsname{89.9}
\expandafter\def\csname res@crms.phys.graph.corr_edges_median\endcsname{0.17}
\expandafter\def\csname res@crms.phys.graph.corr_nonedge_median\endcsname{0.37}
\expandafter\def\csname res@crms.phys.graph.n_channel_edges\endcsname{36}
\expandafter\def\csname res@crms.phys.graph.n_no_edge\endcsname{161}
\expandafter\def\csname res@crms.phys.graph.n_snapped\endcsname{146}
\expandafter\def\csname res@crms.phys.graph.n_stations\endcsname{285}
\expandafter\def\csname res@crms.phys.graph.n_tide_edges\endcsname{92}
\expandafter\def\csname res@crms.phys.graph.tidal_r_tide_edges_median\endcsname{0.88}
\expandafter\def\csname res@crms.phys.vsinf.test.gnn.sc.q0.dnse\endcsname{0.011}
\expandafter\def\csname res@crms.rainfc.median_test_r2\endcsname{$-$0.01}
\expandafter\def\csname res@ctrl.check.base-vs-dataset.cm\endcsname{0.00}
\expandafter\def\csname res@ctrl.check.pre-issue.cm\endcsname{0.00}
\expandafter\def\csname res@ctrl.check.readback.max\endcsname{0}
\expandafter\def\csname res@ctrl.class.r2.gnn.trained\endcsname{shown wrong}
\expandafter\def\csname res@ctrl.dec.p0.gnn.lift.ba\endcsname{0.58}
\expandafter\def\csname res@ctrl.dec.p0.physres.avoid.ba\endcsname{0.86}
\expandafter\def\csname res@ctrl.dec.p0.physres.avoid.ba-level\endcsname{0.86}
\expandafter\def\csname res@ctrl.dec.r2.physres.lift.ba\endcsname{0.51}
\expandafter\def\csname res@ctrl.dec.w0.gnn.avoid.ba\endcsname{0.76}
\expandafter\def\csname res@ctrl.dec.w0.gnn.avoid.ba-level\endcsname{0.74}
\expandafter\def\csname res@ctrl.dec.w0.physonly.avoid.ba\endcsname{0.61}
\expandafter\def\csname res@ctrl.dec.w0.physonly.avoid.ba-level\endcsname{0.61}
\expandafter\def\csname res@ctrl.int-rain.r2.gnn.untrained-recall\endcsname{0.0}
\expandafter\def\csname res@ctrl.int-rain.r2.physonly.recall\endcsname{6.6}
\expandafter\def\csname res@ctrl.int-stage.w0.gnn.recall\endcsname{1.0}
\expandafter\def\csname res@ctrl.k.p0.gnn.trained.mean\endcsname{0.00}
\expandafter\def\csname res@ctrl.k.p0.gnn.trained.sd\endcsname{0.00}
\expandafter\def\csname res@ctrl.k.p0.gnn.untrained\endcsname{0.00}
\expandafter\def\csname res@ctrl.k.p0.hybrid.trained.mean\endcsname{0.00}
\expandafter\def\csname res@ctrl.k.p0.hybrid.trained.sd\endcsname{0.00}
\expandafter\def\csname res@ctrl.k.p0.hybrid.untrained\endcsname{0.00}
\expandafter\def\csname res@ctrl.k.p0.physonly.trained.mean\endcsname{0.00}
\expandafter\def\csname res@ctrl.k.p0.physonly.trained.sd\endcsname{0.00}
\expandafter\def\csname res@ctrl.k.p0.physonly.untrained\endcsname{0.00}
\expandafter\def\csname res@ctrl.k.p0.physres.trained.mean\endcsname{0.00}
\expandafter\def\csname res@ctrl.k.p0.physres.trained.sd\endcsname{0.00}
\expandafter\def\csname res@ctrl.k.p0.physres.untrained\endcsname{0.00}
\expandafter\def\csname res@ctrl.k.r2.gnn.trained.mean\endcsname{0.00}
\expandafter\def\csname res@ctrl.k.r2.gnn.trained.sd\endcsname{0.00}
\expandafter\def\csname res@ctrl.k.r2.gnn.untrained\endcsname{0.00}
\expandafter\def\csname res@ctrl.k.r2.hybrid.trained.mean\endcsname{0.00}
\expandafter\def\csname res@ctrl.k.r2.hybrid.trained.sd\endcsname{0.00}
\expandafter\def\csname res@ctrl.k.r2.hybrid.untrained\endcsname{0.00}
\expandafter\def\csname res@ctrl.k.r2.physonly.trained.mean\endcsname{0.00}
\expandafter\def\csname res@ctrl.k.r2.physonly.trained.sd\endcsname{0.00}
\expandafter\def\csname res@ctrl.k.r2.physonly.untrained\endcsname{0.00}
\expandafter\def\csname res@ctrl.k.r2.physres.trained.mean\endcsname{0.00}
\expandafter\def\csname res@ctrl.k.r2.physres.trained.sd\endcsname{0.00}
\expandafter\def\csname res@ctrl.k.r2.physres.untrained\endcsname{0.00}
\expandafter\def\csname res@ctrl.k.rain.gnn.trained.mean\endcsname{0.50}
\expandafter\def\csname res@ctrl.k.rain.physres.trained.mean\endcsname{0.96}
\expandafter\def\csname res@ctrl.k.stage.gnn.trained.mean\endcsname{0.00}
\expandafter\def\csname res@ctrl.k.w0.gnn.trained.mean\endcsname{0.00}
\expandafter\def\csname res@ctrl.k.w0.gnn.trained.sd\endcsname{0.00}
\expandafter\def\csname res@ctrl.k.w0.gnn.untrained\endcsname{0.00}
\expandafter\def\csname res@ctrl.k.w0.hybrid.trained.mean\endcsname{0.00}
\expandafter\def\csname res@ctrl.k.w0.hybrid.trained.sd\endcsname{0.00}
\expandafter\def\csname res@ctrl.k.w0.hybrid.untrained\endcsname{0.00}
\expandafter\def\csname res@ctrl.k.w0.physonly.trained.mean\endcsname{0.00}
\expandafter\def\csname res@ctrl.k.w0.physonly.trained.sd\endcsname{0.00}
\expandafter\def\csname res@ctrl.k.w0.physonly.untrained\endcsname{0.00}
\expandafter\def\csname res@ctrl.k.w0.physres.trained.mean\endcsname{0.00}
\expandafter\def\csname res@ctrl.k.w0.physres.trained.sd\endcsname{0.00}
\expandafter\def\csname res@ctrl.k.w0.physres.untrained\endcsname{0.00}
\expandafter\def\csname res@ctrl.n-issue\endcsname{150}
\expandafter\def\csname res@ctrl.n-runs\endcsname{3{,}450}
\expandafter\def\csname res@ctrl.n-runs-per-issue\endcsname{23}
\expandafter\def\csname res@ctrl.n-scored-nodes\endcsname{207}
\expandafter\def\csname res@ctrl.n-threshold-nodes\endcsname{206}
\expandafter\def\csname res@ctrl.noise.median.cm\endcsname{0.01}
\expandafter\def\csname res@ctrl.noise.p90.cm\endcsname{0.94}
\expandafter\def\csname res@ctrl.noise.share-le-1cm\endcsname{91}
\expandafter\def\csname res@ctrl.obs.ctrl.gnn.half.orig.n-seeds-shuffle-fail\endcsname{1}
\expandafter\def\csname res@ctrl.obs.ctrl.gnn.half.orig.slope\endcsname{0.90}
\expandafter\def\csname res@ctrl.obs.ctrl.gnn.trained.orig.n-seeds-shuffle-fail\endcsname{1}
\expandafter\def\csname res@ctrl.obs.ctrl.gnn.trained.orig.slope\endcsname{0.69}
\expandafter\def\csname res@ctrl.obs.ctrl.gnn.trained.v2.slope\endcsname{0.67}
\expandafter\def\csname res@ctrl.obs.ctrl.hybrid.half.orig.slope\endcsname{2.46}
\expandafter\def\csname res@ctrl.obs.ctrl.hybrid.reversed.orig.slope\endcsname{$-$0.49}
\expandafter\def\csname res@ctrl.obs.ctrl.hybrid.trained.orig.n-seeds-shuffle-fail\endcsname{1}
\expandafter\def\csname res@ctrl.obs.ctrl.hybrid.trained.orig.slope\endcsname{1.46}
\expandafter\def\csname res@ctrl.obs.ctrl.mean-abs-dpred.max\endcsname{0.13}
\expandafter\def\csname res@ctrl.obs.ctrl.mean-abs-dpred.min\endcsname{0.05}
\expandafter\def\csname res@ctrl.obs.ctrl.physonly.half.orig.n-seeds-shuffle-fail\endcsname{3}
\expandafter\def\csname res@ctrl.obs.ctrl.physonly.half.orig.slope\endcsname{1.71}
\expandafter\def\csname res@ctrl.obs.ctrl.physonly.reversed.orig.slope\endcsname{$-$0.04}
\expandafter\def\csname res@ctrl.obs.ctrl.physonly.trained.orig.slope\endcsname{1.15}
\expandafter\def\csname res@ctrl.obs.ctrl.physonly.untrained.orig.slope\endcsname{1.12}
\expandafter\def\csname res@ctrl.obs.ctrl.physres.half.orig.n-seeds-shuffle-fail\endcsname{3}
\expandafter\def\csname res@ctrl.obs.ctrl.physres.trained.orig.slope\endcsname{1.13}
\expandafter\def\csname res@ctrl.obs.ctrl.physres.untrained.orig.slope\endcsname{1.12}
\expandafter\def\csname res@ctrl.obs.perasset.n-pairs\endcsname{12}
\expandafter\def\csname res@ctrl.obs.perasset.trained.orig.n-pass\endcsname{7}
\expandafter\def\csname res@ctrl.obs.r2.hybrid.trained.orig.slope\endcsname{1.24}
\expandafter\def\csname res@ctrl.obs.r2.physres.trained.orig.slope\endcsname{1.20}
\expandafter\def\csname res@ctrl.obs.w0.gnn.trained.orig.slope\endcsname{0.77}
\expandafter\def\csname res@ctrl.obs.w0.hybrid.trained.orig.slope\endcsname{1.53}
\expandafter\def\csname res@ctrl.ph.k.p0.physonly.untrained.le1link\endcsname{0.64}
\expandafter\def\csname res@ctrl.ph.k.p0.physres.trained.le1link\endcsname{0.10}
\expandafter\def\csname res@ctrl.ph.k.p0.physres.untrained.le1link\endcsname{0.81}
\expandafter\def\csname res@ctrl.ph.k.r2.gnn.trained.le1link\endcsname{0.19}
\expandafter\def\csname res@ctrl.ph.k.r2.hybrid.trained.le1link\endcsname{0.13}
\expandafter\def\csname res@ctrl.ph.k.r2.physonly.trained.ge2links.maxabs\endcsname{0.05}
\expandafter\def\csname res@ctrl.ph.k.r2.physonly.trained.le1link\endcsname{0.78}
\expandafter\def\csname res@ctrl.ph.k.r2.physres.trained.le1link\endcsname{0.55}
\expandafter\def\csname res@ctrl.ph.k.r2.physres.untrained.le1link\endcsname{0.02}
\expandafter\def\csname res@ctrl.ph.k.w0.physres.trained.le1link\endcsname{0.09}
\expandafter\def\csname res@ctrl.ph.p0.n-le1link\endcsname{303}
\expandafter\def\csname res@ctrl.ph.p0.share-ge2links\endcsname{96}
\expandafter\def\csname res@ctrl.ph.p0.share-gt1cm.2to4links.max\endcsname{7}
\expandafter\def\csname res@ctrl.ph.p0.share-gt1cm.2to4links.min\endcsname{0}
\expandafter\def\csname res@ctrl.ph.r2.gnn.share-gt1cm.ge5links.max\endcsname{0}
\expandafter\def\csname res@ctrl.ph.r2.n-le1link\endcsname{1{,}094}
\expandafter\def\csname res@ctrl.ph.r2.share-ge2links\endcsname{91}
\expandafter\def\csname res@ctrl.ph.r2.share-gt1cm.2to4links.max\endcsname{40}
\expandafter\def\csname res@ctrl.ph.r2.share-gt1cm.2to4links.min\endcsname{15}
\expandafter\def\csname res@ctrl.ph.reach.p0.model.maxhops\endcsname{2}
\expandafter\def\csname res@ctrl.ph.reach.p0.model.n-max\endcsname{5}
\expandafter\def\csname res@ctrl.ph.reach.r2.model.maxhops\endcsname{3}
\expandafter\def\csname res@ctrl.ph.reach.r2.model.n-max\endcsname{13}
\expandafter\def\csname res@ctrl.ph.reach.w0.model.maxhops\endcsname{2}
\expandafter\def\csname res@ctrl.ph.reach.w0.model.n-max\endcsname{8}
\expandafter\def\csname res@ctrl.ph.w0.n-le1link\endcsname{806}
\expandafter\def\csname res@ctrl.ph.w0.share-ge2links\endcsname{93}
\expandafter\def\csname res@ctrl.ph.w0.share-gt1cm.2to4links.max\endcsname{18}
\expandafter\def\csname res@ctrl.ph.w0.share-gt1cm.2to4links.min\endcsname{4}
\expandafter\def\csname res@ctrl.rainrule.gnn.k\endcsname{0.71}
\expandafter\def\csname res@ctrl.rainrule.gnn.sign\endcsname{80}
\expandafter\def\csname res@ctrl.rainrule.physonly.sign\endcsname{98}
\expandafter\def\csname res@ctrl.rainrule.physres.k\endcsname{0.97}
\expandafter\def\csname res@ctrl.sign.p0.gnn.trained.mean\endcsname{31}
\expandafter\def\csname res@ctrl.sign.p0.hybrid.trained.mean\endcsname{63}
\expandafter\def\csname res@ctrl.sign.p0.physonly.trained.mean\endcsname{30}
\expandafter\def\csname res@ctrl.sign.p0.physres.trained.mean\endcsname{34}
\expandafter\def\csname res@ctrl.sign.r2.gnn.trained.mean\endcsname{47}
\expandafter\def\csname res@ctrl.sign.r2.hybrid.trained.mean\endcsname{49}
\expandafter\def\csname res@ctrl.sign.r2.physonly.reversed.mean\endcsname{11}
\expandafter\def\csname res@ctrl.sign.r2.physonly.trained.mean\endcsname{34}
\expandafter\def\csname res@ctrl.sign.r2.physres.trained.mean\endcsname{31}
\expandafter\def\csname res@ctrl.sign.w0.gnn.trained.mean\endcsname{38}
\expandafter\def\csname res@ctrl.sign.w0.hybrid.trained.mean\endcsname{55}
\expandafter\def\csname res@ctrl.sign.w0.physonly.trained.mean\endcsname{25}
\expandafter\def\csname res@ctrl.sign.w0.physres.reversed.mean\endcsname{1}
\expandafter\def\csname res@ctrl.sign.w0.physres.trained.mean\endcsname{26}
\expandafter\def\csname res@ctrl.sim.p0.change.median.cm\endcsname{2.9}
\expandafter\def\csname res@ctrl.sim.p0.change.p90.cm\endcsname{8.9}
\expandafter\def\csname res@ctrl.sim.p0.dec.avoidable\endcsname{0.6}
\expandafter\def\csname res@ctrl.sim.p0.dec.liftable\endcsname{0.5}
\expandafter\def\csname res@ctrl.sim.p0.int-rain.n-detectable\endcsname{9{,}052}
\expandafter\def\csname res@ctrl.sim.p0.int-stage.n-detectable\endcsname{4{,}072}
\expandafter\def\csname res@ctrl.sim.p0.n-responsive\endcsname{7{,}857}
\expandafter\def\csname res@ctrl.sim.p0.reach.maxhops\endcsname{18}
\expandafter\def\csname res@ctrl.sim.p0.reach.n\endcsname{76}
\expandafter\def\csname res@ctrl.sim.p0.share-negative\endcsname{55}
\expandafter\def\csname res@ctrl.sim.p0.share-responsive\endcsname{12.7}
\expandafter\def\csname res@ctrl.sim.r2.change.median.cm\endcsname{5.3}
\expandafter\def\csname res@ctrl.sim.r2.change.p90.cm\endcsname{23.7}
\expandafter\def\csname res@ctrl.sim.r2.dec.avoidable\endcsname{3.1}
\expandafter\def\csname res@ctrl.sim.r2.dec.liftable\endcsname{7.0}
\expandafter\def\csname res@ctrl.sim.r2.dec.n-warned\endcsname{17{,}436}
\expandafter\def\csname res@ctrl.sim.r2.int-rain.n-detectable\endcsname{18{,}333}
\expandafter\def\csname res@ctrl.sim.r2.int-stage.n-detectable\endcsname{6{,}240}
\expandafter\def\csname res@ctrl.sim.r2.n-responsive\endcsname{12{,}622}
\expandafter\def\csname res@ctrl.sim.r2.reach.maxhops\endcsname{26}
\expandafter\def\csname res@ctrl.sim.r2.reach.n\endcsname{125}
\expandafter\def\csname res@ctrl.sim.r2.share-negative\endcsname{46}
\expandafter\def\csname res@ctrl.sim.r2.share-responsive\endcsname{20.3}
\expandafter\def\csname res@ctrl.sim.w0.change.median.cm\endcsname{7.8}
\expandafter\def\csname res@ctrl.sim.w0.change.p90.cm\endcsname{30.7}
\expandafter\def\csname res@ctrl.sim.w0.dec.avoidable\endcsname{2.3}
\expandafter\def\csname res@ctrl.sim.w0.dec.liftable\endcsname{8.0}
\expandafter\def\csname res@ctrl.sim.w0.int-rain.n-detectable\endcsname{14{,}664}
\expandafter\def\csname res@ctrl.sim.w0.int-stage.n-detectable\endcsname{6{,}899}
\expandafter\def\csname res@ctrl.sim.w0.n-responsive\endcsname{11{,}414}
\expandafter\def\csname res@ctrl.sim.w0.reach.maxhops\endcsname{28}
\expandafter\def\csname res@ctrl.sim.w0.reach.n\endcsname{113}
\expandafter\def\csname res@ctrl.sim.w0.share-negative\endcsname{29}
\expandafter\def\csname res@ctrl.sim.w0.share-responsive\endcsname{18.4}
\expandafter\def\csname res@ctrl.skill.gnn.ctrl-hold.nse.mean\endcsname{0.922}
\expandafter\def\csname res@ctrl.skill.gnn.hold-move.cm\endcsname{0.10}
\expandafter\def\csname res@ctrl.skill.gnn.q0.nse.mean\endcsname{0.423}
\expandafter\def\csname res@ctrl.skill.gnn.real.nse.mean\endcsname{0.922}
\expandafter\def\csname res@ctrl.skill.gnn.real.nse.sd\endcsname{0.002}
\expandafter\def\csname res@ctrl.skill.hybrid.ctrl-hold.nse.mean\endcsname{0.926}
\expandafter\def\csname res@ctrl.skill.hybrid.q0.nse.mean\endcsname{0.479}
\expandafter\def\csname res@ctrl.skill.hybrid.real.nse.mean\endcsname{0.926}
\expandafter\def\csname res@ctrl.skill.hybrid.real.nse.sd\endcsname{0.001}
\expandafter\def\csname res@ctrl.skill.persistence.nse\endcsname{0.298}
\expandafter\def\csname res@ctrl.skill.physonly.ctrl-hold.nse.mean\endcsname{0.683}
\expandafter\def\csname res@ctrl.skill.physonly.q0.nse.mean\endcsname{0.326}
\expandafter\def\csname res@ctrl.skill.physonly.real.nse.mean\endcsname{0.683}
\expandafter\def\csname res@ctrl.skill.physonly.real.nse.sd\endcsname{0.042}
\expandafter\def\csname res@ctrl.skill.physres.ctrl-hold.nse.mean\endcsname{0.809}
\expandafter\def\csname res@ctrl.skill.physres.hold-move.cm\endcsname{0.04}
\expandafter\def\csname res@ctrl.skill.physres.q0.nse.mean\endcsname{0.393}
\expandafter\def\csname res@ctrl.skill.physres.real-minus-hold.maxabs\endcsname{0.002}
\expandafter\def\csname res@ctrl.skill.physres.real.nse.mean\endcsname{0.809}
\expandafter\def\csname res@ctrl.skill.physres.real.nse.sd\endcsname{0.021}
\expandafter\def\csname res@ctrl.valverule.gnn.k\endcsname{0.57}
\expandafter\def\csname res@ctrl.valverule.gnn.sign\endcsname{80}
\expandafter\def\csname res@ctrl.valverule.physonly.k\endcsname{0.87}
\expandafter\def\csname res@ctrl.valverule.physonly.sign\endcsname{97}
\expandafter\def\csname res@data.beta.astro-range-median-cm\endcsname{68.0}
\expandafter\def\csname res@data.beta.catchment-km2\endcsname{1.3}
\expandafter\def\csname res@data.beta.cf-issue-times\endcsname{150}
\expandafter\def\csname res@data.beta.cf-runs-per-issue\endcsname{51}
\expandafter\def\csname res@data.beta.issue-time-max-h\endcsname{33.5}
\expandafter\def\csname res@data.beta.issue-time-min-h\endcsname{16.0}
\expandafter\def\csname res@data.beta.n-conduits\endcsname{205}
\expandafter\def\csname res@data.beta.n-controls\endcsname{3}
\expandafter\def\csname res@data.beta.n-edges\endcsname{211}
\expandafter\def\csname res@data.beta.n-junctions\endcsname{206}
\expandafter\def\csname res@data.beta.n-rain-nodes\endcsname{165}
\expandafter\def\csname res@data.beta.n-scored\endcsname{207}
\expandafter\def\csname res@data.beta.n-state-nodes\endcsname{209}
\expandafter\def\csname res@data.beta.n-storage\endcsname{3}
\expandafter\def\csname res@data.beta.n-storms\endcsname{900}
\expandafter\def\csname res@data.beta.n-storms-test\endcsname{150}
\expandafter\def\csname res@data.beta.n-storms-train\endcsname{600}
\expandafter\def\csname res@data.beta.n-storms-val\endcsname{150}
\expandafter\def\csname res@data.beta.n-subcatchments\endcsname{165}
\expandafter\def\csname res@data.beta.n-threshold-nodes\endcsname{206}
\expandafter\def\csname res@data.beta.rain-add-max-mm\endcsname{76.2}
\expandafter\def\csname res@data.beta.rain-add-min-mm\endcsname{6.35}
\expandafter\def\csname res@data.beta.rain-constant\endcsname{0.0034416}
\expandafter\def\csname res@data.beta.stage-add-max-cm\endcsname{61.0}
\expandafter\def\csname res@data.beta.stage-add-min-cm\endcsname{7.6}
\expandafter\def\csname res@data.beta.step-min\endcsname{15}
\expandafter\def\csname res@data.beta.window-h\endcsname{6}
\expandafter\def\csname res@data.beta.window-steps\endcsname{24}
\expandafter\def\csname res@data.beta.windows-test\endcsname{10{,}950}
\expandafter\def\csname res@data.mia.amax-mm\endcsname{80.7}
\expandafter\def\csname res@data.mia.attrib-tide-range-lwp-cm\endcsname{95.3}
\expandafter\def\csname res@data.mia.attrib-tide-range-vk-cm\endcsname{67.6}
\expandafter\def\csname res@data.mia.datum-canal-match\endcsname{29}
\expandafter\def\csname res@data.mia.datum-canal-offset\endcsname{45}
\expandafter\def\csname res@data.mia.datum-canal-offset-max-cm\endcsname{206.7}
\expandafter\def\csname res@data.mia.datum-canal-offset-min-cm\endcsname{9.9}
\expandafter\def\csname res@data.mia.datum-tol-ft\endcsname{0.05}
\expandafter\def\csname res@data.mia.datum-tol-wells-ft\endcsname{0.01}
\expandafter\def\csname res@data.mia.datum-wells-match\endcsname{35}
\expandafter\def\csname res@data.mia.exposure-canal-test-closes-after\endcsname{2.4}
\expandafter\def\csname res@data.mia.exposure-canal-test-filled\endcsname{7.4}
\expandafter\def\csname res@data.mia.exposure-gates-test-closes-after\endcsname{7.0}
\expandafter\def\csname res@data.mia.exposure-gates-test-filled\endcsname{37.4}
\expandafter\def\csname res@data.mia.exposure-pumps-test-closes-after\endcsname{7.9}
\expandafter\def\csname res@data.mia.exposure-pumps-test-filled\endcsname{34.3}
\expandafter\def\csname res@data.mia.extent-km-ew\endcsname{38}
\expandafter\def\csname res@data.mia.extent-km-ns\endcsname{83}
\expandafter\def\csname res@data.mia.grid-points\endcsname{41}
\expandafter\def\csname res@data.mia.history-h\endcsname{168}
\expandafter\def\csname res@data.mia.horizon-h\endcsname{48}
\expandafter\def\csname res@data.mia.joint-grid-points\endcsname{11}
\expandafter\def\csname res@data.mia.n-canal-gauges\endcsname{74}
\expandafter\def\csname res@data.mia.n-controls\endcsname{36}
\expandafter\def\csname res@data.mia.n-edges\endcsname{242}
\expandafter\def\csname res@data.mia.n-edges-canal\endcsname{105}
\expandafter\def\csname res@data.mia.n-edges-gate\endcsname{24}
\expandafter\def\csname res@data.mia.n-edges-groundwater\endcsname{84}
\expandafter\def\csname res@data.mia.n-edges-gw\endcsname{52}
\expandafter\def\csname res@data.mia.n-edges-pump\endcsname{12}
\expandafter\def\csname res@data.mia.n-edges-seep\endcsname{32}
\expandafter\def\csname res@data.mia.n-edges-structure\endcsname{50}
\expandafter\def\csname res@data.mia.n-edges-tide\endcsname{3}
\expandafter\def\csname res@data.mia.n-edges-ustruct\endcsname{14}
\expandafter\def\csname res@data.mia.n-gates\endcsname{24}
\expandafter\def\csname res@data.mia.n-pump-dir-labelled\endcsname{6}
\expandafter\def\csname res@data.mia.n-pumps\endcsname{12}
\expandafter\def\csname res@data.mia.n-rain-gauges\endcsname{29}
\expandafter\def\csname res@data.mia.n-scored-canal-test\endcsname{71}
\expandafter\def\csname res@data.mia.n-scored-wells-test\endcsname{35}
\expandafter\def\csname res@data.mia.n-series-nodes\endcsname{176}
\expandafter\def\csname res@data.mia.n-state-nodes\endcsname{109}
\expandafter\def\csname res@data.mia.n-tide-gauges\endcsname{2}
\expandafter\def\csname res@data.mia.n-wells\endcsname{35}
\expandafter\def\csname res@data.mia.obs-share-gates-test\endcsname{62.6}
\expandafter\def\csname res@data.mia.obs-share-pumps-test\endcsname{65.7}
\expandafter\def\csname res@data.mia.query-step-h\endcsname{6}
\expandafter\def\csname res@data.mia.rain-annual-share-max\endcsname{11}
\expandafter\def\csname res@data.mia.rain-annual-share-min\endcsname{6}
\expandafter\def\csname res@data.mia.rain-factor-max\endcsname{0.202}
\expandafter\def\csname res@data.mia.rain-factor-median\endcsname{0.0956}
\expandafter\def\csname res@data.mia.rain-factor-min\endcsname{0.064}
\expandafter\def\csname res@data.mia.rain-factor-n\endcsname{28}
\expandafter\def\csname res@data.mia.rain-factor-test-median\endcsname{0.0967}
\expandafter\def\csname res@data.mia.rain-factor-val-median\endcsname{0.0968}
\expandafter\def\csname res@data.mia.rain-hourly-corr-median\endcsname{0.93}
\expandafter\def\csname res@data.mia.rain-hourly-corr-min\endcsname{0.56}
\expandafter\def\csname res@data.mia.rainfc-area-r2-test\endcsname{0.196}
\expandafter\def\csname res@data.mia.rainfc-area-r2-val\endcsname{0.268}
\expandafter\def\csname res@data.mia.rainfc-gauges-r2-positive-test\endcsname{28}
\expandafter\def\csname res@data.mia.rainfc-n-features\endcsname{1{,}958}
\expandafter\def\csname res@data.mia.rainfc-r2-clim-test-median\endcsname{0.028}
\expandafter\def\csname res@data.mia.rainfc-r2-clim-val-median\endcsname{0.008}
\expandafter\def\csname res@data.mia.rainfc-r2-test-median\endcsname{0.106}
\expandafter\def\csname res@data.mia.rainfc-r2-val-median\endcsname{0.137}
\expandafter\def\csname res@data.mia.shuffle-max-days\endcsname{45}
\expandafter\def\csname res@data.mia.shuffle-min-days\endcsname{9}
\expandafter\def\csname res@data.mia.storm-shape-max-share\endcsname{5.8}
\expandafter\def\csname res@data.mia.storm-shape-min-share\endcsname{2.9}
\expandafter\def\csname res@data.mia.surge-max-cm\endcsname{36.5}
\expandafter\def\csname res@data.mia.surge-min-cm\endcsname{$-$49.0}
\expandafter\def\csname res@data.mia.tide-datum-shift-cm\endcsname{45.72}
\expandafter\def\csname res@data.mia.tide-lwp-astro-minus-noaa-pred-cm\endcsname{10.4}
\expandafter\def\csname res@data.mia.tide-lwp-astro-trainfit-rms-test-cm\endcsname{4.2}
\expandafter\def\csname res@data.mia.tide-vk-astro-minus-noaa-pred-cm\endcsname{12.0}
\expandafter\def\csname res@data.mia.tide-vk-astro-trainfit-rms-test-cm\endcsname{4.7}
\expandafter\def\csname res@data.mia.tide-vk-hours-filled\endcsname{277}
\expandafter\def\csname res@data.mia.weekly-blocks-test\endcsname{52}
\expandafter\def\csname res@data.mia.windows-test\endcsname{8{,}545}
\expandafter\def\csname res@data.mia.windows-val-select\endcsname{2{,}849}
\expandafter\def\csname res@gamma.cf.n_issue\endcsname{100}
\expandafter\def\csname res@gamma.cf.n_runs\endcsname{6{,}200}
\expandafter\def\csname res@gamma.cf.noise.median\endcsname{0.00007}
\expandafter\def\csname res@gamma.flood_share\endcsname{83}
\expandafter\def\csname res@gamma.k_rain.gnn.reversed\endcsname{0.39}
\expandafter\def\csname res@gamma.k_rain.gnn.trained\endcsname{0.65}
\expandafter\def\csname res@gamma.k_rain.hybrid.trained\endcsname{0.95}
\expandafter\def\csname res@gamma.k_rain.physonly.trained\endcsname{1.17}
\expandafter\def\csname res@gamma.k_rain.physres.trained\endcsname{1.18}
\expandafter\def\csname res@gamma.k_valve.gnn.half\endcsname{0.13}
\expandafter\def\csname res@gamma.k_valve.gnn.trained\endcsname{0.57}
\expandafter\def\csname res@gamma.k_valve.hybrid.trained\endcsname{0.69}
\expandafter\def\csname res@gamma.k_valve.ncells\endcsname{9{,}320}
\expandafter\def\csname res@gamma.k_valve.physonly.half\endcsname{0.25}
\expandafter\def\csname res@gamma.k_valve.physonly.trained\endcsname{0.87}
\expandafter\def\csname res@gamma.k_valve.physres.trained\endcsname{0.79}
\expandafter\def\csname res@gamma.k_valve.physres.untrained\endcsname{0.01}
\expandafter\def\csname res@gamma.nse.persistence\endcsname{0.365}
\expandafter\def\csname res@gamma.nse.q0.gnn.mean\endcsname{0.536}
\expandafter\def\csname res@gamma.nse.q0.physres.mean\endcsname{0.511}
\expandafter\def\csname res@gamma.nse.real.hybrid.mean\endcsname{0.975}
\expandafter\def\csname res@gamma.nse.real.physonly.mean\endcsname{0.960}
\expandafter\def\csname res@gamma.rain.max_mm\endcsname{127}
\expandafter\def\csname res@gamma.rain.min_mm\endcsname{6.3}
\expandafter\def\csname res@gamma.rf.all.trained.ba.max\endcsname{0.92}
\expandafter\def\csname res@gamma.rf.all.trained.ba.min\endcsname{0.78}
\expandafter\def\csname res@gamma.rf.all.untrained.ba.max\endcsname{0.51}
\expandafter\def\csname res@gamma.rf.n\endcsname{684}
\expandafter\def\csname res@gamma.storms.test\endcsname{100}
\expandafter\def\csname res@gamma.storms.total\endcsname{600}
\expandafter\def\csname res@gamma.storms.train\endcsname{400}
\expandafter\def\csname res@gamma.storms.val\endcsname{100}
\expandafter\def\csname res@gamma.v2.rain.gnn.reversed.slope.max\endcsname{1.21}
\expandafter\def\csname res@gamma.v2.rain.gnn.reversed.slope.min\endcsname{0.92}
\expandafter\def\csname res@gamma.v2.valve.gnn.trained.slope.mean\endcsname{1.04}
\expandafter\def\csname res@gamma.v2.valve.physonly.trained.slope.mean\endcsname{1.01}
\expandafter\def\csname res@gamma.v2.valve.physonly.untrained.slope\endcsname{0.96}
\expandafter\def\csname res@gamma.v2.valve.physres.untrained.slope\endcsname{0.97}
\expandafter\def\csname res@gamma.va.all.trained.ba.max\endcsname{0.90}
\expandafter\def\csname res@gamma.va.all.trained.ba.min\endcsname{0.83}
\expandafter\def\csname res@gamma.va.all.untrained.ba.max\endcsname{0.63}
\expandafter\def\csname res@gamma.va.all.untrained.ba.min\endcsname{0.53}
\expandafter\def\csname res@gamma.va.const_acc\endcsname{0.965}
\expandafter\def\csname res@gamma.va.gnn.trained.acc.max\endcsname{0.903}
\expandafter\def\csname res@gamma.va.gnn.trained.precision.mean\endcsname{0.21}
\expandafter\def\csname res@gamma.va.hybrid.trained.acc.max\endcsname{0.873}
\expandafter\def\csname res@gamma.va.hybrid.trained.precision.mean\endcsname{0.19}
\expandafter\def\csname res@gamma.va.n\endcsname{4{,}576}
\expandafter\def\csname res@gamma.va.physonly.trained.precision.mean\endcsname{0.52}
\expandafter\def\csname res@gamma.va.physres.trained.precision.mean\endcsname{0.54}
\expandafter\def\csname res@gamma.va.pos_rate\endcsname{3.5}
\expandafter\def\csname res@hou.gauges.rain.candidates\endcsname{177}
\expandafter\def\csname res@hou.gauges.rain.selected\endcsname{134}
\expandafter\def\csname res@hou.gauges.stage.candidates\endcsname{172}
\expandafter\def\csname res@hou.gauges.stage.selected\endcsname{84}
\expandafter\def\csname res@hou.groups.surge\endcsname{21}
\expandafter\def\csname res@hou.phys.acc.test.gnn.dnse\endcsname{$-$0.086}
\expandafter\def\csname res@hou.phys.acc.test.gnn.real.nse\endcsname{0.771}
\expandafter\def\csname res@hou.phys.acc.test.persistence.nse\endcsname{$-$0.021}
\expandafter\def\csname res@hou.phys.acc.test.physonly.dnse\endcsname{$-$0.182}
\expandafter\def\csname res@hou.phys.attrib.test.rain.physonly.r\endcsname{13.34}
\expandafter\def\csname res@hou.phys.attrib.test.rain.physres.r.untrained\endcsname{36.51}
\expandafter\def\csname res@hou.phys.cf.a_max_mm\endcsname{274.9}
\expandafter\def\csname res@hou.phys.cf.test.gnn.share_nonmono\endcsname{23.7}
\expandafter\def\csname res@hou.phys.graph.added\endcsname{22}
\expandafter\def\csname res@hou.phys.graph.channel_km_median\endcsname{7.8}
\expandafter\def\csname res@hou.phys.graph.corr_edges_median\endcsname{0.53}
\expandafter\def\csname res@hou.phys.graph.corr_nonedge_median\endcsname{0.41}
\expandafter\def\csname res@hou.phys.graph.dir_nhd_agree\endcsname{61}
\expandafter\def\csname res@hou.phys.graph.dir_stage_n\endcsname{61}
\expandafter\def\csname res@hou.phys.graph.kept\endcsname{41}
\expandafter\def\csname res@hou.phys.graph.n_channel_edges\endcsname{63}
\expandafter\def\csname res@hou.phys.graph.n_channel_edges_inferred\endcsname{64}
\expandafter\def\csname res@hou.phys.graph.n_no_edge\endcsname{2}
\expandafter\def\csname res@hou.phys.graph.n_snapped\endcsname{79}
\expandafter\def\csname res@hou.phys.graph.n_stations\endcsname{84}
\expandafter\def\csname res@hou.phys.graph.n_tide_edges\endcsname{19}
\expandafter\def\csname res@hou.phys.graph.removed\endcsname{23}
\expandafter\def\csname res@hou.phys.groups.n_surge\endcsname{24}
\expandafter\def\csname res@hou.phys.groups.n_surge_tide_only\endcsname{4}
\expandafter\def\csname res@hou.phys.vsinf.test.gnn.sc.q0.dnse\endcsname{$-$0.005}
\expandafter\def\csname res@hou.rainfc.median_r2.test\endcsname{0.00}
\expandafter\def\csname res@hou.rainfc.median_r2.val\endcsname{$-$0.18}
\expandafter\def\csname res@hou.surge.max_cm\endcsname{126.0}
\expandafter\def\csname res@hou.test.cost.csi95\endcsname{0.02}
\expandafter\def\csname res@hou.test.gnn.dnse\endcsname{$-$0.078}
\expandafter\def\csname res@hou.thr.fws\endcsname{79}
\expandafter\def\csname res@hou.tide.gauges\endcsname{4}
\expandafter\def\csname res@mia.attr.eff-resid.max\endcsname{$3\times10^{-5}$}
\expandafter\def\csname res@mia.attr.hybrid-s42-test.n-issue\endcsname{128}
\expandafter\def\csname res@mia.attr.n-issue\endcsname{1{,}425}
\expandafter\def\csname res@mia.attr.test.rain.gnn.trained.phir.p50\endcsname{0.07}
\expandafter\def\csname res@mia.attr.test.rain.gnn.trained.phir.p90\endcsname{0.47}
\expandafter\def\csname res@mia.attr.test.rain.physonly.trained.phir.p50\endcsname{0.42}
\expandafter\def\csname res@mia.attr.test.rain.physonly.trained.phir.p90\endcsname{1.74}
\expandafter\def\csname res@mia.attr.test.rain.physonly.untrained.phir.p90\endcsname{0.15}
\expandafter\def\csname res@mia.attr.test.surge-b.gnn.trained.phis.p90\endcsname{1.21}
\expandafter\def\csname res@mia.attr.test.surge-b.gnn.untrained.phis.p90\endcsname{0.05}
\expandafter\def\csname res@mia.attr.test.surge-b.hybrid.trained.phis.p90\endcsname{6.74}
\expandafter\def\csname res@mia.attr.test.surge-b.hybrid.untrained.phis.p90\endcsname{4.95}
\expandafter\def\csname res@mia.attr.test.surge-b.physonly.trained.phis.p90\endcsname{12.47}
\expandafter\def\csname res@mia.attr.test.surge-b.physonly.untrained.phis.p90\endcsname{6.32}
\expandafter\def\csname res@mia.fc.check.persistence-identical-runs\endcsname{25}
\expandafter\def\csname res@mia.fc.n-canal\endcsname{74}
\expandafter\def\csname res@mia.fc.n-canal-scored\endcsname{71}
\expandafter\def\csname res@mia.fc.n-wells-scored\endcsname{35}
\expandafter\def\csname res@mia.fc.n-windows\endcsname{8{,}545}
\expandafter\def\csname res@mia.fc.test.p95.base-rate\endcsname{6.1}
\expandafter\def\csname res@mia.fc.test.q0.gnn.nse.mean\endcsname{0.811}
\expandafter\def\csname res@mia.fc.test.q0.hybrid.mae.mean\endcsname{2.67}
\expandafter\def\csname res@mia.fc.test.q0.hybrid.nse.mean\endcsname{0.804}
\expandafter\def\csname res@mia.fc.test.q0.physonly.mae-h25-48.mean\endcsname{4.60}
\expandafter\def\csname res@mia.fc.test.q0.ref-physonly.mae-h25-48.mean\endcsname{4.05}
\expandafter\def\csname res@mia.fc.test.q0.untr-gnn.nse\endcsname{0.546}
\expandafter\def\csname res@mia.fc.test.q0.untr-hybrid.nse\endcsname{$-$1.052}
\expandafter\def\csname res@mia.fc.test.q0norain.gnn.nse.mean\endcsname{0.804}
\expandafter\def\csname res@mia.fc.test.q0norain.hybrid.mae.mean\endcsname{2.64}
\expandafter\def\csname res@mia.fc.test.q0norain.hybrid.nse.mean\endcsname{0.811}
\expandafter\def\csname res@mia.fc.val.p95.base-rate\endcsname{3.3}
\expandafter\def\csname res@mia.fc.val.q0norain.hybrid.nse.mean\endcsname{0.885}
\expandafter\def\csname res@mia.obs.group.rain.n\endcsname{57}
\expandafter\def\csname res@mia.obs.group.rain.r-median-all\endcsname{0.55}
\expandafter\def\csname res@mia.obs.group.rain.r-median-selected\endcsname{0.59}
\expandafter\def\csname res@mia.obs.group.rain.rmin\endcsname{0.30}
\expandafter\def\csname res@mia.obs.group.surge.n-ge-015\endcsname{5}
\expandafter\def\csname res@mia.obs.group.surge.r-p90\endcsname{0.09}
\expandafter\def\csname res@mia.obs.group.surge.r.g54t\endcsname{0.17}
\expandafter\def\csname res@mia.obs.group.surge.r.g93t\endcsname{0.22}
\expandafter\def\csname res@mia.obs.test.rain.gnn.fe.untrained.n-big\endcsname{1}
\expandafter\def\csname res@mia.obs.test.rain.interaction.max\endcsname{0.012}
\expandafter\def\csname res@mia.obs.test.rain.physonly.fe.slope.max\endcsname{1.14}
\expandafter\def\csname res@mia.obs.test.rain.physonly.fe.untrained.n-big\endcsname{186}
\expandafter\def\csname res@mia.obs.test.rain.physonly.pooled.slope.max\endcsname{1.10}
\expandafter\def\csname res@mia.obs.test.rain.physres.fe.slope.min\endcsname{0.87}
\expandafter\def\csname res@mia.obs.test.rain.physres.pooled.slope.min\endcsname{0.83}
\expandafter\def\csname res@mia.obs.test.st.bbcw10.gnn.mean-do.max\endcsname{5.1}
\expandafter\def\csname res@mia.obs.test.st.bbcw10.hybrid.mean-do.min\endcsname{5.5}
\expandafter\def\csname res@mia.obs.test.st.bbcw8.gnn.mean-abs-dp.max\endcsname{0.007}
\expandafter\def\csname res@mia.obs.test.st.bbcw8.gnn.mean-do.min\endcsname{3.6}
\expandafter\def\csname res@mia.obs.test.st.bbcw8.hybrid.mean-do.max\endcsname{8.2}
\expandafter\def\csname res@mia.obs.test.st.bbcw8.physonly.mean-do.max\endcsname{24.6}
\expandafter\def\csname res@mia.obs.test.st.bbcw8.physres.mean-do.min\endcsname{18.6}
\expandafter\def\csname res@mia.obs.test.st.s123t.gnn.slope.s42\endcsname{3.70}
\expandafter\def\csname res@mia.obs.test.st.s123t.gnn.slope.s43\endcsname{4.04}
\expandafter\def\csname res@mia.obs.test.st.s123t.gnn.slope.s44\endcsname{4.72}
\expandafter\def\csname res@mia.obs.test.st.s123t.hybrid.slope.s42\endcsname{1.51}
\expandafter\def\csname res@mia.obs.test.st.s123t.hybrid.slope.s43\endcsname{1.46}
\expandafter\def\csname res@mia.obs.test.st.s123t.hybrid.slope.s44\endcsname{1.42}
\expandafter\def\csname res@mia.obs.test.st.s123t.physonly.mean-do.min\endcsname{0.9}
\expandafter\def\csname res@mia.obs.test.st.s123t.physonly.slope.s42\endcsname{1.09}
\expandafter\def\csname res@mia.obs.test.st.s123t.physres.bin1.mean-do\endcsname{$-$5.5}
\expandafter\def\csname res@mia.obs.test.st.s123t.physres.bin2.mean-do\endcsname{$-$1.4}
\expandafter\def\csname res@mia.obs.test.st.s123t.physres.bin3.mean-do\endcsname{1.0}
\expandafter\def\csname res@mia.obs.test.st.s123t.physres.bin4.mean-do\endcsname{3.7}
\expandafter\def\csname res@mia.obs.test.st.s123t.physres.bin5.mean-do\endcsname{9.0}
\expandafter\def\csname res@mia.obs.test.surge-a.gnn.mean-abs-dp\endcsname{0.18}
\expandafter\def\csname res@mia.obs.test.surge-a.gnn.pooled.slope.min\endcsname{3.64}
\expandafter\def\csname res@mia.obs.test.surge-a.hybrid.fe.untrained.slope\endcsname{0.91}
\expandafter\def\csname res@mia.obs.test.surge-a.hybrid.pooled.slope.max\endcsname{1.12}
\expandafter\def\csname res@mia.obs.test.surge-a.hybrid.pooled.slope.min\endcsname{0.99}
\expandafter\def\csname res@mia.obs.test.surge-a.interaction.max\endcsname{0.044}
\expandafter\def\csname res@mia.obs.test.surge-a.physonly.fe.slope.min\endcsname{1.09}
\expandafter\def\csname res@mia.obs.test.surge-a.physonly.fe.untrained.slope\endcsname{1.87}
\expandafter\def\csname res@mia.obs.test.surge-a.physonly.pooled.slope.min\endcsname{$-$0.62}
\expandafter\def\csname res@mia.obs.test.surge-a.physres.fe.slope.max\endcsname{1.12}
\expandafter\def\csname res@mia.obs.test.surge-a.physres.fe.untrained.slope\endcsname{2.34}
\expandafter\def\csname res@mia.obs.test.surge-a.physres.pooled.slope.max\endcsname{$-$0.35}
\expandafter\def\csname res@mia.obs.test.surge-b.gnn.pooled.slope.max\endcsname{4.88}
\expandafter\def\csname res@mia.q2.pooled.max\endcsname{0.00}
\expandafter\def\csname res@mia.q2.pooled.min\endcsname{$-$0.06}
\expandafter\def\csname res@mia.q2.within.min\endcsname{$-$0.08}
\expandafter\def\csname res@mia.robust.grid.agree.min\endcsname{100.0}
\expandafter\def\csname res@mia.robust.grid.medabsdiff.max\endcsname{0.10}
\expandafter\def\csname res@mia.robust.p90.notcrossed.max\endcsname{72}
\expandafter\def\csname res@mia.robust.p90.notcrossed.min\endcsname{28}
\expandafter\def\csname res@mia.robust.p99.notcrossed.max\endcsname{96}
\expandafter\def\csname res@mia.robust.p99.notcrossed.min\endcsname{75}
\expandafter\def\csname res@mia.thr.amax_mm\endcsname{80.7}
\expandafter\def\csname res@mia.thr.grid_step_mm\endcsname{2.02}
\expandafter\def\csname res@mia.thr.joint_step_mm\endcsname{8.07}
\expandafter\def\csname res@mia.thr.n_cells\endcsname{81{,}225}
\expandafter\def\csname res@mia.thr.n_issue\endcsname{1{,}425}
\expandafter\def\csname res@mia.thr.n_rain_stations\endcsname{57}
\expandafter\def\csname res@mia.thr.nonmono.gnn\endcsname{11.6}
\expandafter\def\csname res@mia.thr.nonmono.hybrid\endcsname{10.7}
\expandafter\def\csname res@mia.thr.nonmono.physres\endcsname{0.3}
\expandafter\def\csname res@mia.thr.numerical_failures\endcsname{0}
\expandafter\def\csname res@mia.thr.range.astar-median.max\endcsname{66}
\expandafter\def\csname res@mia.thr.range.astar-median.min\endcsname{55}
\expandafter\def\csname res@mia.thr.range.eps-sensitive.max\endcsname{2{,}700}
\expandafter\def\csname res@mia.thr.range.eps-sensitive.min\endcsname{2{,}097}
\expandafter\def\csname res@mia.thr.range.last-interval.max\endcsname{2{,}258}
\expandafter\def\csname res@mia.thr.range.last-interval.min\endcsname{637}
\expandafter\def\csname res@mia.thr.range.near-threshold.max\endcsname{921}
\expandafter\def\csname res@mia.thr.range.near-threshold.min\endcsname{696}
\expandafter\def\csname res@mia.thr.range.nonmono-visible.max\endcsname{3{,}643}
\expandafter\def\csname res@mia.thr.range.nonmono-visible.min\endcsname{5}
\expandafter\def\csname res@mia.thr.range.notcrossed.max\endcsname{81}
\expandafter\def\csname res@mia.thr.range.notcrossed.min\endcsname{46}
\expandafter\def\csname res@mia.thr.range.notcrossed_all.max\endcsname{81}
\expandafter\def\csname res@mia.thr.range.notcrossed_all.min\endcsname{45}
\expandafter\def\csname res@mia.thr.range.q3-missed.max\endcsname{6.2}
\expandafter\def\csname res@mia.thr.range.q3-missed.min\endcsname{0.0}
\expandafter\def\csname res@mia.thr.range.val.notcrossed.max\endcsname{79}
\expandafter\def\csname res@mia.thr.range.val.notcrossed.min\endcsname{47}
\expandafter\def\csname res@mia.thr.smax_cm\endcsname{36.5}
\expandafter\def\csname res@mia.thr.sstar_found.physonly\endcsname{77}
\expandafter\def\csname res@mia.thr.sstar_found.physres\endcsname{59}
\expandafter\def\csname res@mia.thr.sstar_found_surgegroup.gnn\endcsname{20}
\expandafter\def\csname res@mia.thr.sstar_found_surgegroup.physres\endcsname{880}
\expandafter\def\csname res@mia.thr.sstar_median_surgegroup.gnn\endcsname{25.7}
\expandafter\def\csname res@mia.thr.sstar_median_surgegroup.physonly\endcsname{11.4}
\expandafter\def\csname res@mia.thr.train_max24h_mm\endcsname{113.4}
\expandafter\def\csname res@mia.thr.warned.max\endcsname{5.3}
\expandafter\def\csname res@mia.thr.warned.min\endcsname{5.0}
\expandafter\def\csname res@mia.timing.u1.gainmin.max\endcsname{3.4}
\expandafter\def\csname res@mia.timing.u1.gainmin.min\endcsname{0.3}
\expandafter\def\csname res@mia.timing.u1.gainoracle.max\endcsname{1.8}
\expandafter\def\csname res@mia.timing.u1.gainoracle.min\endcsname{$-$3.0}
\expandafter\def\csname res@mia.timing.u2.gainmin.max\endcsname{0.6}
\expandafter\def\csname res@mia.timing.u2.gainmin.min\endcsname{0.0}
\expandafter\def\csname res@mia.u1.alarm-ratio.margins-vs-astar.max\endcsname{5.75}
\expandafter\def\csname res@mia.u1.alarm-ratio.margins-vs-astar.min\endcsname{1.53}
\expandafter\def\csname res@mia.u1.astar.precision.max\endcsname{0.87}
\expandafter\def\csname res@mia.u1.astar.precision.min\endcsname{0.66}
\expandafter\def\csname res@mia.u1.astar.recall-pct.max\endcsname{29}
\expandafter\def\csname res@mia.u1.astar.recall-pct.min\endcsname{15}
\expandafter\def\csname res@mia.u1.astar.valalarmshare.max\endcsname{0.5}
\expandafter\def\csname res@mia.u1.astar.valalarmshare.min\endcsname{0.2}
\expandafter\def\csname res@mia.u1.cells.max\endcsname{77{,}079}
\expandafter\def\csname res@mia.u1.cells.min\endcsname{66{,}949}
\expandafter\def\csname res@mia.u1.curmargin.precision.max\endcsname{0.22}
\expandafter\def\csname res@mia.u1.curmargin.precision.min\endcsname{0.09}
\expandafter\def\csname res@mia.u1.curmargin.recall.max\endcsname{0.235}
\expandafter\def\csname res@mia.u1.curmargin.recall.min\endcsname{0.054}
\expandafter\def\csname res@mia.u1.linear.precision.max\endcsname{0.92}
\expandafter\def\csname res@mia.u1.linear.precision.min\endcsname{0.74}
\expandafter\def\csname res@mia.u1.linear.recall.max\endcsname{0.220}
\expandafter\def\csname res@mia.u1.linear.recall.min\endcsname{0.079}
\expandafter\def\csname res@mia.u1.nruns.dprauc.astar-vs-linear.ci-above-zero\endcsname{12}
\expandafter\def\csname res@mia.u1.nruns.dprauc.astar-vs-linear.positive\endcsname{12}
\expandafter\def\csname res@mia.u1.nruns.dprauc.astar-vs-logistic.ci-above-zero\endcsname{10}
\expandafter\def\csname res@mia.u1.nruns.dprauc.astar-vs-logistic.positive\endcsname{12}
\expandafter\def\csname res@mia.u1.nruns.dprauc.pred-vs-cur.ci-above-zero\endcsname{11}
\expandafter\def\csname res@mia.u1.nruns.dprauc.pred-vs-cur.positive\endcsname{12}
\expandafter\def\csname res@mia.u1.nruns.precision.linear-gt-astar\endcsname{11}
\expandafter\def\csname res@mia.u1.nruns.recall.astar-gt-linear\endcsname{12}
\expandafter\def\csname res@mia.u1.positives.max\endcsname{741}
\expandafter\def\csname res@mia.u1.positives.min\endcsname{194}
\expandafter\def\csname res@mia.u1.predmargin.precision.max\endcsname{0.34}
\expandafter\def\csname res@mia.u1.predmargin.precision.min\endcsname{0.10}
\expandafter\def\csname res@mia.u1.predmargin.recall.max\endcsname{0.299}
\expandafter\def\csname res@mia.u1.predmargin.recall.min\endcsname{0.084}
\expandafter\def\csname res@mia.u1.prevalence.max\endcsname{1.0}
\expandafter\def\csname res@mia.u1.prevalence.min\endcsname{0.3}
\expandafter\def\csname res@mia.u1eps.nruns.dprauc.astar-vs-linear.positive\endcsname{12}
\expandafter\def\csname res@mia.u1eps.nruns.dprauc.astar-vs-logistic.positive\endcsname{12}
\expandafter\def\csname res@mia.u1eps.nruns.recall.astar-gt-linear\endcsname{12}
\expandafter\def\csname res@mia.u1eps.positives.max\endcsname{502}
\expandafter\def\csname res@mia.u1eps.positives.min\endcsname{94}
\expandafter\def\csname res@mia.u2.alarm-ratio.margins-vs-astar.max\endcsname{5.71}
\expandafter\def\csname res@mia.u2.alarm-ratio.margins-vs-astar.min\endcsname{1.52}
\expandafter\def\csname res@mia.u2.astar.precision.max\endcsname{0.73}
\expandafter\def\csname res@mia.u2.astar.precision.min\endcsname{0.66}
\expandafter\def\csname res@mia.u2.astar.recall-pct.max\endcsname{5}
\expandafter\def\csname res@mia.u2.astar.recall-pct.min\endcsname{3}
\expandafter\def\csname res@mia.u2.astar.recall.max\endcsname{0.048}
\expandafter\def\csname res@mia.u2.astar.recall.min\endcsname{0.030}
\expandafter\def\csname res@mia.u2.astar.testalarmshare.max\endcsname{0.2}
\expandafter\def\csname res@mia.u2.astar.testalarmshare.min\endcsname{0.1}
\expandafter\def\csname res@mia.u2.astar.valalarmshare.max\endcsname{0.5}
\expandafter\def\csname res@mia.u2.astar.valalarmshare.min\endcsname{0.2}
\expandafter\def\csname res@mia.u2.cells.max\endcsname{76{,}382}
\expandafter\def\csname res@mia.u2.cells.min\endcsname{66{,}301}
\expandafter\def\csname res@mia.u2.curmargin.precision.max\endcsname{0.79}
\expandafter\def\csname res@mia.u2.curmargin.precision.min\endcsname{0.47}
\expandafter\def\csname res@mia.u2.curmargin.recall.max\endcsname{0.110}
\expandafter\def\csname res@mia.u2.curmargin.recall.min\endcsname{0.059}
\expandafter\def\csname res@mia.u2.curmargin_zero.precision.max\endcsname{0.95}
\expandafter\def\csname res@mia.u2.curmargin_zero.precision.min\endcsname{0.71}
\expandafter\def\csname res@mia.u2.curmargin_zero.recall.max\endcsname{0.046}
\expandafter\def\csname res@mia.u2.curmargin_zero.recall.min\endcsname{0.006}
\expandafter\def\csname res@mia.u2.dprauc_astar_vs_cur.hybrid\endcsname{0.079}
\expandafter\def\csname res@mia.u2.dprauc_astar_vs_cur.hybrid.hi\endcsname{0.149}
\expandafter\def\csname res@mia.u2.dprauc_astar_vs_cur.hybrid.lo\endcsname{0.001}
\expandafter\def\csname res@mia.u2.events.max\endcsname{2{,}611}
\expandafter\def\csname res@mia.u2.events.min\endcsname{1{,}631}
\expandafter\def\csname res@mia.u2.linear.recall.max\endcsname{0.036}
\expandafter\def\csname res@mia.u2.linear.recall.min\endcsname{0.009}
\expandafter\def\csname res@mia.u2.predmargin.precision.max\endcsname{0.73}
\expandafter\def\csname res@mia.u2.predmargin.precision.min\endcsname{0.48}
\expandafter\def\csname res@mia.u2.predmargin.recall.max\endcsname{0.117}
\expandafter\def\csname res@mia.u2.predmargin.recall.min\endcsname{0.054}
\expandafter\def\csname res@mia.u2.prevalence.max\endcsname{3.4}
\expandafter\def\csname res@mia.u2.prevalence.min\endcsname{2.4}
\expandafter\def\csname res@mia.u2.rerun.precision.max\endcsname{0.73}
\expandafter\def\csname res@mia.u2.rerun.precision.min\endcsname{0.62}
\expandafter\def\csname res@mia.u2.rerun.recall.max\endcsname{0.204}
\expandafter\def\csname res@mia.u2.rerun.recall.min\endcsname{0.076}
\expandafter\def\csname res@mia.u2.rocauc.astar.hybrid\endcsname{0.88}
\expandafter\def\csname res@mia.u2.rocauc.astar.physonly\endcsname{0.72}
\expandafter\def\csname res@mia.u2.rocauc.predmargin.gnn\endcsname{0.91}
\expandafter\def\csname res@mia.u2.rocauc.predmargin.physonly\endcsname{0.88}
\expandafter\def\csname res@mia.u3.auc.astar.max\endcsname{0.94}
\expandafter\def\csname res@mia.u3.auc.astar.min\endcsname{0.76}
\expandafter\def\csname res@mia.u3.auc.predmargin.max\endcsname{0.95}
\expandafter\def\csname res@mia.u3.auc.predmargin.min\endcsname{0.87}
\expandafter\def\csname res@mia.u3.cells.max\endcsname{76{,}967}
\expandafter\def\csname res@mia.u3.cells.min\endcsname{66{,}856}
\expandafter\def\csname res@mia.u3.d.add_astar.max\endcsname{0.005}
\expandafter\def\csname res@mia.u3.d.add_astar.min\endcsname{$-$0.006}
\expandafter\def\csname res@mia.u3.positives.max\endcsname{574}
\expandafter\def\csname res@mia.u3.positives.min\endcsname{358}
\expandafter\def\csname res@mia.u3.prevalence.max\endcsname{0.7}
\expandafter\def\csname res@mia.u3.prevalence.min\endcsname{0.5}
\expandafter\def\csname res@mia.xai.test.rain.gnn.integrated.agree-when-surge-easier.max\endcsname{0}
\expandafter\def\csname res@mia.xai.test.rain.gnn.integrated.constant-answer-agreement.max\endcsname{97}
\expandafter\def\csname res@mia.xai.test.rain.gnn.integrated.constant-answer-agreement.min\endcsname{90}
\expandafter\def\csname res@mia.xai.test.rain.gnn.integrated.ig-driver-agreement.max\endcsname{95}
\expandafter\def\csname res@mia.xai.test.rain.gnn.integrated.ig-driver-agreement.min\endcsname{88}
\expandafter\def\csname res@mia.xai.test.rain.gnn.integrated.ig-share-top-history.max\endcsname{100.0}
\expandafter\def\csname res@mia.xai.test.rain.gnn.linear.median-rel-error-where-both.max\endcsname{0.52}
\expandafter\def\csname res@mia.xai.test.rain.gnn.linear.median-rel-error-where-both.min\endcsname{0.43}
\expandafter\def\csname res@mia.xai.test.rain.gnn.linear.miss-rate.max\endcsname{94}
\expandafter\def\csname res@mia.xai.test.rain.gnn.linear.miss-rate.min\endcsname{91}
\expandafter\def\csname res@mia.xai.test.rain.gnn.unconstrained.cf-mean-share-on-observed-history.max\endcsname{90}
\expandafter\def\csname res@mia.xai.test.rain.gnn.unconstrained.cf-mean-share-on-observed-history.min\endcsname{88}
\expandafter\def\csname res@mia.xai.test.rain.gnn.unconstrained.cf-share-negative-rain.max\endcsname{100.0}
\expandafter\def\csname res@mia.xai.test.rain.physonly.integrated.ig-share-top-history.min\endcsname{99.3}
\expandafter\def\csname res@mia.xai.test.rain.physonly.linear.median-rel-error-where-both.max\endcsname{0.06}
\expandafter\def\csname res@mia.xai.test.rain.physonly.linear.miss-rate.max\endcsname{34}
\expandafter\def\csname res@mia.xai.test.rain.physonly.linear.miss-rate.min\endcsname{32}
\expandafter\def\csname res@mia.xai.test.rain.physonly.single.miss-rate.max\endcsname{6.2}
\expandafter\def\csname res@mia.xai.test.rain.physonly.unconstrained.cf-mean-share-on-observed-history.max\endcsname{47}
\expandafter\def\csname res@mia.xai.test.rain.physonly.unconstrained.cf-mean-share-on-observed-history.min\endcsname{44}
\expandafter\def\csname res@mia.xai.test.rain.physres.single.miss-rate.max\endcsname{1.6}
\expandafter\def\csname res@mia.xai.test.rain.physres.single.miss-rate.min\endcsname{1.4}
\expandafter\def\csname res@mia.xai.test.rain.physres.unconstrained.cf-share-negative-rain.min\endcsname{99.8}
\expandafter\def\csname res@mia.xai.test.surge-a.gnn.single.miss-rate.max\endcsname{27}
\expandafter\def\csname res@mia.xai.test.surge-a.gnn.single.miss-rate.min\endcsname{19}
\expandafter\def\csname res@model.gnn.n-mp-edges\endcsname{1{,}282}
\expandafter\def\csname res@model.gnn.n-params\endcsname{119{,}376}
\expandafter\def\csname res@model.gnn.n-params-beta\endcsname{100{,}920}
\expandafter\def\csname res@model.hybrid.n-params\endcsname{638{,}077}
\expandafter\def\csname res@model.hybrid.n-params-beta\endcsname{610{,}207}
\expandafter\def\csname res@model.physonly.n-params\endcsname{2{,}130}
\expandafter\def\csname res@model.physonly.n-params-beta\endcsname{2{,}313}
\expandafter\def\csname res@model.physres.n-params\endcsname{7{,}891}
\expandafter\def\csname res@model.physres.n-params-beta\endcsname{7{,}370}
\expandafter\def\csname res@s7.best_epoch.physres.min\endcsname{3}
\expandafter\def\csname res@s7.stations.s6p1_water\endcsname{74}
\expandafter\def\csname res@s7.stations.shared\endcsname{72}
\expandafter\def\csname res@s7.stations.water\endcsname{80}
\expandafter\def\csname res@s7.surge.max.test.g1\endcsname{39.8}
\expandafter\def\csname res@s7.surge.max.test.g2\endcsname{43.5}
\expandafter\def\csname res@s7.surge.max.train.g1\endcsname{39.2}
\expandafter\def\csname res@s7.surge.max.train.g2\endcsname{51.9}
\expandafter\def\csname res@s7.surge.windows_gt20.selection\endcsname{0}
\expandafter\def\csname res@s7.surge.windows_gt20.test\endcsname{40}
\expandafter\def\csname res@s7.v2.rain.hybrid.untrained.nbig\endcsname{26}
\expandafter\def\csname res@s7.v2.y2023.rain.gnn.slope.max\endcsname{1.34}
\expandafter\def\csname res@s7.v2.y2023.rain.gnn.slope.min\endcsname{1.10}
\expandafter\def\csname res@s7.wells.shared\endcsname{19}
\expandafter\def\csname res@sf.s6p0.attr.test.gnn.share-r-largest.min\endcsname{100.0}
\expandafter\def\csname res@sf.s6p0.attr.test.hybrid.trained.phir.meanabs.min\endcsname{0.28}
\expandafter\def\csname res@sf.s6p0.attr.test.physonly.trained.phir.meanabs.max\endcsname{0.75}
\expandafter\def\csname res@sf.s6p0.attr.test.physres.untrained.phir.meanabs\endcsname{0.10}
\expandafter\def\csname res@sf.s6p0.fc.test.persistence.nse\endcsname{0.993}
\expandafter\def\csname res@sf.s6p0.fc.test.tcn.q0.nse.mean\endcsname{0.768}
\expandafter\def\csname res@sf.s6p0.groups.n-canal\endcsname{19}
\expandafter\def\csname res@sf.s6p0.groups.n-gates\endcsname{5}
\expandafter\def\csname res@sf.s6p0.groups.n-rain-gauges\endcsname{12}
\expandafter\def\csname res@sf.s6p0.groups.n-tide-edges\endcsname{0}
\expandafter\def\csname res@sf.s6p0.groups.n-wells\endcsname{64}
\expandafter\def\csname res@sf.s6p0.obs.test.rain.gnn.untrained.sign\endcsname{1.00}
\expandafter\def\csname res@sf.s6p0.obs.test.rain.physonly.untrained.sign\endcsname{0.84}
\expandafter\def\csname res@sf.s6p0.obs.test.rain.physres.untrained.sign\endcsname{0.82}
\expandafter\def\csname res@sf.s6p0.u3.test.physres.d-astar-pred.auc.min\endcsname{$-$0.41}
\expandafter\def\csname res@sf.s6p2.fc.test.gnn.q0.nse.mean\endcsname{0.938}
\expandafter\def\csname res@sf.s6p2.fc.test.gnn.real.nse.mean\endcsname{0.931}
\expandafter\def\csname res@sf.s6p2.fc.test.persistence.nse\endcsname{0.941}
\expandafter\def\csname res@sf.s6p2.groups.n-canal\endcsname{12}
\expandafter\def\csname res@sf.s6p2.groups.n-gates\endcsname{3}
\expandafter\def\csname res@sf.s6p2.groups.n-pumps\endcsname{1}
\expandafter\def\csname res@sf.s6p2.groups.n-rain-gauges\endcsname{3}
\expandafter\def\csname res@sf.s6p2.groups.n-wells\endcsname{19}
\expandafter\def\csname res@sf.s6p2.obs.test.rain.hybrid.fe.slope.max\endcsname{0.73}
\expandafter\def\csname res@sf.s6p2.obs.test.rain.physres.fe.slope.min\endcsname{0.50}
\expandafter\def\csname res@sf.s7p0d1.groups.n-canal\endcsname{22}
\expandafter\def\csname res@sf.s7p0d1.groups.surge.n\endcsname{2}
\expandafter\def\csname res@sf.s7p1d1.cost.test.physonly.nse.mean\endcsname{$-$0.050}
\expandafter\def\csname res@sf.s7p1d1.cost.test.physres.nse.mean\endcsname{$-$0.090}
\expandafter\def\csname res@sf.s7p1d1.fc.test.physonly.ref.nse.mean\endcsname{0.703}
\expandafter\def\csname res@sf.s7p1d1.fc.test.physres.ref.nse.mean\endcsname{0.702}
\expandafter\def\csname res@sf.s7p2d1.cost.test.hybrid.nse.mean\endcsname{$-$0.032}
\expandafter\def\csname res@sf.s7p2d1.cost.test.physonly.nse.mean\endcsname{$-$0.039}
\expandafter\def\csname res@sf.s7p2d1.fc.test.mlp.q0.nse.mean\endcsname{0.277}
\expandafter\def\csname res@sf.s7p2d1.groups.n-canal\endcsname{30}
\expandafter\def\csname res@sf.u1chk.s6p0.a-max-mm\endcsname{80.4}
\expandafter\def\csname res@sf.u1chk.s6p0.all.positives.max\endcsname{111}
\expandafter\def\csname res@sf.u1chk.s6p0.all.positives.min\endcsname{10}
\expandafter\def\csname res@sf.u1chk.s6p2.a-max-mm\endcsname{120.2}
\expandafter\def\csname res@uk.data.n_candidates\endcsname{146}
\expandafter\def\csname res@uk.data.n_rain_inputs\endcsname{30}
\expandafter\def\csname res@uk.data.n_stations\endcsname{126}
\expandafter\def\csname res@uk.phys.acc.test.gconv.q0.nse\endcsname{0.623}
\expandafter\def\csname res@uk.phys.acc.test.gconv.q0_rain0.nse\endcsname{0.542}
\expandafter\def\csname res@uk.phys.attr.test.d9phys.phi_rain.mean\endcsname{2.33}
\expandafter\def\csname res@uk.phys.attr.test.d9phys.phi_rain.untrained\endcsname{20.01}
\expandafter\def\csname res@uk.phys.cmp.test.gconv.dnse_phys_minus_inferred\endcsname{0.002}
\expandafter\def\csname res@uk.phys.cost.test.d9phys.dnse\endcsname{$-$0.022}
\expandafter\def\csname res@uk.phys.graph.components\endcsname{13}
\expandafter\def\csname res@uk.phys.graph.corr.inferred\endcsname{0.59}
\expandafter\def\csname res@uk.phys.graph.corr.physical\endcsname{0.65}
\expandafter\def\csname res@uk.phys.graph.corr.random\endcsname{0.31}
\expandafter\def\csname res@uk.phys.graph.dir.downstream_earlier\endcsname{2}
\expandafter\def\csname res@uk.phys.graph.dir.downstream_later\endcsname{56}
\expandafter\def\csname res@uk.phys.graph.dir.n\endcsname{95}
\expandafter\def\csname res@uk.phys.graph.edges\endcsname{113}
\expandafter\def\csname res@uk.phys.graph.isolated\endcsname{4}
\expandafter\def\csname res@uk.phys.graph.off_channel_gauges\endcsname{7}
\expandafter\def\csname res@uk.phys.train.gconv.rt.early_stopped\endcsname{3}
\expandafter\def\csname res@uk.phys.u3.test.gnn.d-astar-pred.auc.max\endcsname{0.07}
\expandafter\def\csname res@uk.phys.u3.test.gnn.d-astar-pred.n-ci-excl0\endcsname{3}
\expandafter\def\csname res@uk.phys.u3.test.physres.d-astar-pred.auc.min\endcsname{0.05}
\expandafter\def\csname res@uk.rainfc.median_test_r2\endcsname{0.12}
\expandafter\def\csname res@uk.ref2.test.dnse\endcsname{0.014}

\expandafter\def\csname res@smb.boot.clusters\endcsname{23}
\expandafter\def\csname res@smb.crb.o3.ba\endcsname{0.78}
\expandafter\def\csname res@smb.crb.o3.ba.hi\endcsname{0.86}
\expandafter\def\csname res@smb.crb.o3.ba.lo\endcsname{0.69}
\expandafter\def\csname res@smb.crb.o4.ba\endcsname{0.83}
\expandafter\def\csname res@smb.crb.o4.ba.hi\endcsname{0.86}
\expandafter\def\csname res@smb.crb.o4.ba.lo\endcsname{0.79}
\expandafter\def\csname res@smb.dev.bank_conduits\endcsname{107}
\expandafter\def\csname res@smb.dev.trunk_conduits\endcsname{52}
\expandafter\def\csname res@smb.groups.near\endcsname{94}
\expandafter\def\csname res@smb.groups.op\endcsname{39}
\expandafter\def\csname res@smb.groups.rain\endcsname{98}
\expandafter\def\csname res@smb.it.gravity\endcsname{86}
\expandafter\def\csname res@smb.it.mixed\endcsname{48}
\expandafter\def\csname res@smb.it.pump\endcsname{66}
\expandafter\def\csname res@smb.it.test\endcsname{200}
\expandafter\def\csname res@smb.k11.mixed_share_train_steps\endcsname{0.95}
\expandafter\def\csname res@smb.k2b.capture.max\endcsname{61}
\expandafter\def\csname res@smb.k2b.capture.min\endcsname{29}
\expandafter\def\csname res@smb.k2b.var.runs\endcsname{3{,}000}
\expandafter\def\csname res@smb.net.conduits\endcsname{165}
\expandafter\def\csname res@smb.net.subcatchments\endcsname{100}
\expandafter\def\csname res@smb.nodes.excluded\endcsname{5}
\expandafter\def\csname res@smb.nodes.scored\endcsname{160}
\expandafter\def\csname res@smb.o4.delta.r0.dd.same.max\endcsname{26}
\expandafter\def\csname res@smb.o4.delta.r0.dd.same.min\endcsname{14}
\expandafter\def\csname res@smb.o4.same.r0.dd.ba.max\endcsname{0.52}
\expandafter\def\csname res@smb.runs.core\endcsname{9{,}800}
\expandafter\def\csname res@smb.runs.primary\endcsname{4{,}000}
\expandafter\def\csname res@smb.storms.test\endcsname{100}
\expandafter\def\csname res@smb.storms.total\endcsname{600}
\expandafter\def\csname res@smb.storms.train\endcsname{400}
\expandafter\def\csname res@smb.storms.val\endcsname{100}
\expandafter\def\csname res@smb.t2.r0.k.gnn\endcsname{1.06}
\expandafter\def\csname res@smb.t2.r0.k.gnn.hi\endcsname{1.09}
\expandafter\def\csname res@smb.t2.r0.k.gnn.lo\endcsname{1.04}
\expandafter\def\csname res@smb.t2.r0.k.hybrid\endcsname{0.98}
\expandafter\def\csname res@smb.t2.r0.k.physonly\endcsname{0.26}
\expandafter\def\csname res@smb.t2.r0.k.physres\endcsname{0.43}
\expandafter\def\csname res@smb.t3.cells\endcsname{18{,}800}
\expandafter\def\csname res@smb.t3.comp\endcsname{16.4}
\expandafter\def\csname res@smb.t3.compshare\endcsname{28.6}
\expandafter\def\csname res@smb.t3.direct\endcsname{57.3}
\expandafter\def\csname res@smb.t3.hold.max_per100\endcsname{73.5}
\expandafter\def\csname res@smb.t3.n_responsive\endcsname{895}
\expandafter\def\csname res@smb.t3.total\endcsname{40.9}
\expandafter\def\csname res@smb.t4.big.min\endcsname{$-$0.16}
\expandafter\def\csname res@smb.t4.maxabs.typical\endcsname{0.07}
\expandafter\def\csname res@smb.t4x.far.maxabs.pump\endcsname{0.11}
\expandafter\def\csname res@smb.t4x.far.simshare\endcsname{66}
\expandafter\def\csname res@smb.t4x.res.gate.r0.pump.untrained.phys.max\endcsname{$-$0.18}
\expandafter\def\csname res@smb.t4x.res.gate.r0.pump.untrained.phys.min\endcsname{$-$0.40}
\expandafter\def\csname res@smb.t4x.res.pump.r0.untrained.phys.max\endcsname{1.27}
\expandafter\def\csname res@smb.t4x.res.pump.r0.untrained.phys.min\endcsname{0.25}
\expandafter\def\csname res@smb.t5a.o3.rate\endcsname{2.28}
\expandafter\def\csname res@smb.t5a.o3.rate.hi\endcsname{3.33}
\expandafter\def\csname res@smb.t5a.o3.rate.lo\endcsname{1.31}
\expandafter\def\csname res@smb.t5b.r0.initmiss.phys.max\endcsname{60}
\expandafter\def\csname res@smb.t5b.r0.initmiss.phys.min\endcsname{37}
\expandafter\def\csname res@smb.t5b.r0.prec.phys.max\endcsname{0.27}
\expandafter\def\csname res@smb.t5b.r0.prec.phys.min\endcsname{0.16}
\expandafter\def\csname res@smb.t6.posrate\endcsname{99.7}
\expandafter\def\csname res@smb.t6.r0.err.gnn\endcsname{4.6}
\expandafter\def\csname res@smb.t6.r0.err.hybrid\endcsname{3.5}
\expandafter\def\csname res@smb.t6.r0.err.physonly\endcsname{14.3}
\expandafter\def\csname res@smb.t6.r0.err.physres\endcsname{17.1}
\expandafter\def\csname res@smb.t7.r0.ctrl.slope.gnn\endcsname{1.24}
\expandafter\def\csname res@smb.t7.r0.ctrl.slope.hybrid\endcsname{1.22}
\expandafter\def\csname res@smb.t7.truth.dsim_med\endcsname{22}
\expandafter\def\csname res@smb.t7.truth.n\endcsname{1{,}232}
\expandafter\def\csname res@smb.t7.truth.r0.none.gnn\endcsname{61}
\expandafter\def\csname res@smb.t7.truth.r0.none.hybrid\endcsname{42}
\expandafter\def\csname res@smb.t7.truth.r0.opp.gnn\endcsname{32}
\expandafter\def\csname res@smb.t7.truth.r0.opp.hybrid\endcsname{47}
\expandafter\def\csname res@smb.t7.truth.r0.opp.phys.max\endcsname{2}
\expandafter\def\csname res@smb.tier1.crb.o3.ba\endcsname{0.80}
\expandafter\def\csname res@smb.tier1.t3.compshare\endcsname{25.3}
\expandafter\def\csname res@smb.tier1.t3.total\endcsname{44.4}
\expandafter\def\csname res@smb.train.capped\endcsname{23}

\newcommand{\smbcolor}{}
\newcommand{\smbadd}[1]{#1}%

\graphicspath{{figures/}}

\begin{document}
\setlength{\textfloatsep}{8pt}\setlength{\dbltextfloatsep}{8pt}
\setlength{\floatsep}{6pt}\setlength{\dblfloatsep}{6pt}\setlength{\intextsep}{6pt}

\title{Interpretable Compound Flood Forecasting with Evidence-Graded Explanations}

\author{Xu Zheng}
\affiliation{\institution{Florida International University}\city{Miami}\state{FL}\country{USA}}
\email{xzhen019@fiu.edu}
\author{Zhuomin Chen}
\affiliation{\institution{Florida International University}\city{Miami}\state{FL}\country{USA}}
\email{zchen051@fiu.edu}
\author{Al Lim}
\affiliation{\institution{Singapore Management University}\country{Singapore}}
\email{allim@smu.edu.sg}
\author{Wei Cheng}
\affiliation{\institution{NEC Laboratories America}\city{Princeton}\state{NJ}\country{USA}}
\email{weicheng@nec-labs.com}
\author{Jayantha Obeysekera}
\affiliation{\institution{Florida International University}\city{Miami}\state{FL}\country{USA}}
\email{jobeysek@fiu.edu}
\author{Mo Sha}
\affiliation{\institution{Florida International University}\city{Miami}\state{FL}\country{USA}}
\email{msha@fiu.edu}
\author{Orlando Woods}
\affiliation{\institution{Singapore Management University}\country{Singapore}}
\email{orlandowoods@smu.edu.sg}
\author{Jason Liu}
\affiliation{\institution{Florida International University}\city{Miami}\state{FL}\country{USA}}
\email{liux@fiu.edu}
\author{Dongsheng Luo}
\authornote{Corresponding author.}
\affiliation{\institution{Singapore Management University}\country{Singapore}}
\email{dsluo@smu.edu.sg}

\begin{abstract}
Learning-based compound flood models achieve high predictive accuracy, yet neither accuracy nor model-faithful explanations guarantee physical validity. Validating explanations against field observations is also difficult because extreme floods cannot be reproduced under controlled interventions. We introduce \emph{evidence grading}, a framework that assesses forecast explanations through model checks, observational diagnostics, and paired simulator runs sharing identical histories alongside untrained model controls, recording an explicit evidence class for each explanation type. We apply this framework to an interpretable forecasting system that predicts water levels from explicit future drivers across four architectures. Through one scenario interface, the system explains each forecast peak via three queries: driver attribution, scenario change, and minimum crossing rainfall (the least added rainfall needed to reach a high-water threshold). Across six flood regions on two continents, explicit scenario inputs largely preserve forecast skill. On a simulated coastal network, evidence grading shows that all four architectures learn the node ranking of rainfall impacts and recover 70\% to 97\% of the simulated rainfall response, a property accuracy alone cannot verify, while missing the inland propagation of coastal surge and control operations. Where observational checks cannot resolve driver responses in real systems, grading classifies explanations as \emph{not checkable}. Graded evidence thus systematically separates partially supported, contradicted, and uncheckable explanations for each architecture.

\end{abstract}



\maketitle

\section{Introduction}\label{sec:intro}

\begin{figure*}[t]
\centering
\includegraphics[width=\textwidth]{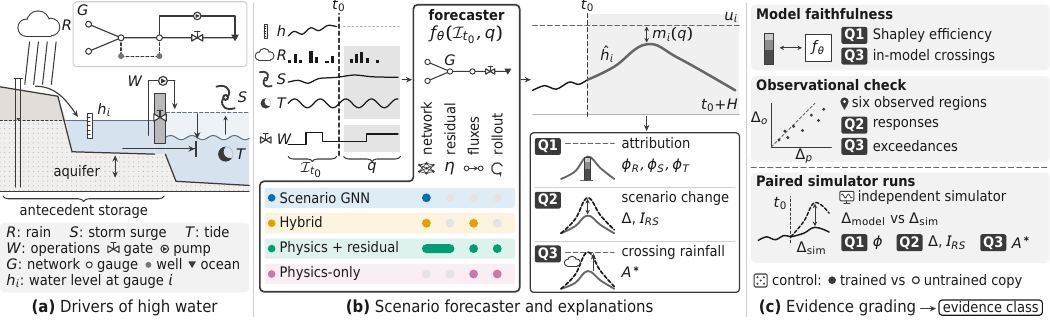}
\caption{Framework overview: (a)~compound flood drivers, (b)~scenario-conditioned peak-margin forecasting ($q=(R,S,T,W)$) for queries Q1--Q3, and (c)~paired-run evidence grading.}
\label{fig:overview}
\end{figure*}

Compound floods pose a growing threat to coastal cities~\cite{wahl2015increasing}, where rising seas weaken gravity drainage~\cite{park2011storm,hughes2016hydrologic}. Forecasting these events is challenging because multiple drivers interact across space and time~\cite{zscheischler2020typology}. Rainfall adds runoff to drainage networks, while elevated coastal water levels obstruct outfall discharge and cause backwater effects~\cite{park2011storm,gold2022inundation}. Antecedent storage in channels and aquifers, together with gate, weir, and pump operations, further shapes the resulting water levels~\cite{hughes2016hydrologic,sukop2018high} (Fig.~\ref{fig:overview}a).

Advances in machine learning have enabled accurate predictions of streamflow, floods, and water levels~\cite{kratzert2018rainfall,nevo2022flood,nearing2024global,shi2025deep,zheng2026uncovering}, and prior work has explored their use in gate and pump scheduling~\cite{shi2025fidlar}. However, flood planning and operations require more than trajectory accuracy under realized drivers. Coastal planning depends on how rainfall and sea level compound~\cite{jane2020multivariate}, while operators must assess whether additional rainfall, a higher tide, or an altered control schedule would cause a threshold crossing. Answering these questions requires understanding how predicted water levels respond to counterfactual changes in individual drivers.

This need motivates explanations that connect forecasts to their physical drivers, and attribution methods have recently been applied to analyze flood formation~\cite{lai2025formation,jiang2024compounding}. Yet because rainfall, sea level, and groundwater are often statistically correlated~\cite{jane2020multivariate}, a model may associate a driver with high water without learning how perturbing that driver would alter water levels. The central question is therefore whether model explanations reliably reflect the physical processes driving floods.

Neither predictive accuracy nor faithfulness alone establishes the physical validity of an explanation. Accuracy evaluates predictions under realized conditions rather than responses to driver perturbations, and autocorrelated water levels can reward shortcuts such as extrapolating recent history~\cite{geirhos2020shortcut}. Faithfulness assesses whether an explanation reflects the model, not whether the model reflects the water system~\cite{jacovi2020towards,haufe2026explainable}. Physical inductive biases can guide learning~\cite{reichstein2019deep,shen2023differentiable}, but explanations still require validation and negative controls, as even untrained networks can produce plausible attributions~\cite{adebayo2018sanity}. Direct validation from field observations is also elusive because the same flood cannot be observed under alternative driver interventions. Despite this gap, existing flood studies present attributions or attention maps without benchmarking them against known responses~\cite{ding2020interpretable,xie2026rapid,xu2026urban,lai2025formation}. Our experiments on a simulated coastal network illustrate the risk: two forecasters with similar accuracy under realized drivers respond very differently to added rainfall (one frequently predicting a falling peak where the simulated peak rises, Section~\ref{sec:results-sim}), yet both pass the observational check available in real systems (Section~\ref{sec:results-observed}).

To address this challenge, we ground forecast explanations in explicit scenario predictions and validate them through \emph{evidence grading} (Fig.~\ref{fig:overview}b,c). Because our scenario forecasters predict water levels from historical states and explicit future trajectories of rainfall, surge, tide, and operations, every explanation queries the forecast peak under well-defined perturbations that can be directly compared against physical system responses. Evidence grading evaluates these explanations against internal model checks, observational diagnostics, and paired simulator runs sharing identical histories but differing in future drivers, alongside untrained model controls, assigning each explanation type an explicit evidence class. Our contributions are:
\begin{itemize}[itemsep=1.5pt,topsep=0pt,parsep=0pt,leftmargin=*]
\item We introduce \emph{evidence grading}, a framework that evaluates forecast explanations via model checks, observational diagnostics, and paired simulator runs with untrained controls, classifying each explanation into one of five evidence classes.
\item We formulate driver attribution, scenario change, and \emph{minimum crossing rainfall} (the least added rainfall lifting a forecast peak to a high-water threshold) as three unified queries of a common scenario interface across four architectures spanning black-box GNNs and differentiable water-balance models.
\item We evaluate the framework across six observed regions on two continents and two simulated networks, showing how empirical support varies across drivers and systems, from partially supported rainfall and valve responses to missing coastal surge and operation responses on the simulated coastal network.
\end{itemize}

\section{Related Work}\label{sec:related}

\par\textbf{Learned and physically structured flood forecasting.}
Deep sequence networks and GNNs have advanced the prediction of streamflow, floods, and water levels~\cite{kratzert2018rainfall,nearing2024global,sun2021explore,nevo2022flood}, as well as the emulation of flood dynamics~\cite{bentivoglio2023rapid}. Hybrid and mass-conserving architectures incorporate physical conservation laws into these models~\cite{karpatne2017theory,reichstein2019deep,shen2023differentiable,hoedt2021mclstm}. Beyond architecture, the input structure determines which operational questions a forecaster can address. Benchmarks such as SF$^2$Bench predict from historical inputs alone and assess driver importance via input ablation~\cite{zheng2026uncovering,you2026multisource}. Conversely, regional water-level surrogates for South Florida~\cite{shi2025deep}, GNN surrogates of SWMM~\cite{zhang2024graph}, and learned controllers for gate and pump scheduling~\cite{shi2025fidlar} explicitly condition predictions on future drivers. While future driver inputs enable scenario forecasting, whether the resulting models learn physically valid responses to driver interventions remains a separate evaluation question.

\par\textbf{Explaining flood forecasts and validating explanations.}
Shapley attributions allocate predictions among input drivers~\cite{shapley1953value,lundberg2017unified}, whereas counterfactual explanations search for input perturbations that alter predictions~\cite{wachter2018counterfactual,wang2023counterfactual}. Flood forecasting studies widely apply attributions and attention weights to interpret driver roles and flood formation~\cite{lai2025formation,jiang2024compounding,ding2020interpretable,hong2025interpretable,xie2026rapid,xu2026urban,basagaoglu2022review}. Operationally, flash flood guidance and tide-conditioned rainfall thresholds express proximity to flooding as the rainfall depth needed to trigger exceedance~\cite{sweeney1992modernized,clark2014conus,lin2025pluvial}. For model explanations, however, faithfulness verifies agreement with the model~\cite{jacovi2020towards,sundararajan2017axiomatic}; agreement with the physical water system requires external validation.

Validating physical explanations requires matching each claim to an appropriate reference response. For attribution, synthetic benchmarks supply known input importance~\cite{mamalakis2022neural,wilming2022scrutinizing}. For interventional responses, references come from physical models or hydrological principles, including comparing neural weather sensitivities with a physical model for one storm~\cite{banomedina2025are}, testing hydrological axioms~\cite{bayati2026evaluating} and expected input responses~\cite{reichert2024metamorphic}, and evaluating behavior under warming~\cite{wi2022assessing}, sensor errors~\cite{kim2026deep}, or lag and zero-rain tests~\cite{duong2025evaluating}. Controlled simulation provides exact counterfactual references: studies on interventional consistency show that surrogates trained without interventions can misestimate intervention effects~\cite{dyer2024interventionally}, while WhatIf-Bench scores trajectories across branching simulated futures~\cite{ashok2026whatif}. Our work brings interventional validation to compound flood explanations, grading rainfall, coastal boundary, and control operation claims into explicit evidence classes.

\section{Background and Problem}\label{sec:problem}

\par\textbf{Forecast target.}
We forecast water levels (stage) across the stations of a drainage network. At issue time $t_0$, a forecaster $f_\theta$ maps the information set $\mathcal{I}_{t_0}$ (history and graph) and a scenario $q=(R,S,T,W)$ of future rainfall, storm surge, tide, and operations to stage forecasts at node $i$ for leads $\ell\le H$, with peak margin $m_i(q)$ relative to a high-water threshold $u_i$:
\begin{align}
\hat{\mathbf h}(q)&=f_\theta(\mathcal{I}_{t_0},q),\label{eq:forecast}\\
m_i(q)&=\max_{1\le\ell\le H}\hat h_{i,t_0+\ell}(q)-u_i ,\label{eq:margin}
\end{align}
where $u_i$ is a training-period percentile $p_u$ and $m_i(q_0)\ge0$ defines a model warning. The issue-time scenario $q_0$ takes rainfall from a statistical forecast fitted on training years (zero on the simulators), holds surge and operations at their last values, and uses the known tide (Fig.~\ref{fig:s-protocol}), while training and diagnostic perfect-driver hindcasts use realized drivers (App.~\ref{app:B2}).

\par\textbf{What an explanation claims.}
An explanation states how the peak margin $m_i(q)$ depends on the drivers: which driver contributed, what an alternative scenario would change, or how much added rainfall reaches the threshold. The problem is to decide, for each explanation type, architecture, and water system, whether evidence supports it, contradicts it, or cannot assess it.

\par\textbf{Role of paired simulations.}
Because each observed flood occurs once, a real watershed's response to a changed driver from the same state is unobservable. Simulated networks supply this reference through paired runs that differ only in future drivers, testing whether an architecture learns physical responses to rainfall, the ocean boundary, or operations. This architecture-level evidence accompanies the model's explanations in observed regions.

\section{The Forecasting System}\label{sec:method}

To assess whether forecast explanations reflect true responses of the water system, we integrate scenario forecasting, peak-margin explanation queries, and evidence grading. The system predicts water levels from historical states and explicit future drivers, explains driver contributions, scenario changes, and minimum crossing rainfall, and grades each explanation type using model checks, observational diagnostics, and paired simulator runs.

\subsection{Scenario Forecasters}\label{sec:method-models}
Evaluating driver effects requires comparing predictions under alternative future trajectories while holding history fixed. We use explicit scenario inputs across four architectures spanning pure data-driven prediction, hybrid physics fusion, and differentiable water-balance evolution with and without residual correction. All four implement the common scenario interface in Eq.~\eqref{eq:forecast} (Fig.~\ref{fig:overview}b).

The \emph{Scenario GNN} represents every time series, including future drivers, as a node with GraphSAGE neighborhood aggregation~\cite{hamilton2017inductive}. The \emph{Hybrid} architecture combines a Transformer encoder~\cite{vaswani2017attention}, spatial message passing, and a physics branch evaluated at issue-time water levels. The \emph{Physics + state residual} (\emph{Physics + residual} in figures) and \emph{Physics-only} models evolve a learned water balance step-by-step, with a bounded neural residual included in the former.

The water-balance models approximate storage continuity on the network graph: rainfall and net edge inflows alter storage, which converts to stage increments through a learned effective storage area~\cite{shen2023differentiable}. In simplified form (App.~\ref{app:C}), the step update is:
\begin{align}
h_i(t{+}1)&=h_i(t)+\frac{\sum_{e^+=i}f_e(t)-\sum_{e^-=i}f_e(t)+p_i(t)}{\Omega_i\big(h_i(t)\big)}+\eta_i(t),\notag\\
f_e(t)&=c_e(t)\,K_e\,\psi_e\big(h_{e^-}(t)-h_{e^+}(t)+b_e\big),\label{eq:balance}
\end{align}
where $h_i$ is stage, $\Omega_i$ effective storage area, $p_i$ rainfall inflow, and $\eta_i$ the bounded residual (Physics + state residual only). Tide nodes impose $T+S$, while flux $f_e$ from $e^-$ to $e^+$ depends on conductance $K_e$, odd monotone head-discharge law $\psi_e$, head offset $b_e$, and control factor $c_e$ (gate opening, learned constant, or unity). Pump fluxes use a separate head-dependent parameterization (App.~\ref{app:C}). Each internal edge flux enters its two end nodes with equal magnitude and opposite sign, while the tide boundary, rainfall, pumps, residual, and stage bounds enter the update directly. This structure gives every driver an explicit pathway to water levels, whose learned responses we evaluate through evidence grading.

\subsection{Explaining Forecast Peaks}\label{sec:method-explanations}
Understanding a forecast peak requires knowing both how drivers contribute to it and how perturbing them would alter it. We formalize attribution, scenario comparison, and counterfactual search over the peak margin $m_i(q)$ with history fixed (Fig.~\ref{fig:overview}b and Algorithm~\ref{alg:explain}).

\emph{Driver attribution (Q1)} partitions $m_i(q)-v(\emptyset)$ among rainfall, surge, and tide relative to a baseline reference. The exact Shapley value~\cite{shapley1953value} averages each driver's marginal contribution across all subsets of the remaining drivers:
\begin{equation}
\phi_D=\sum_{\mathcal{C}\subseteq\mathcal{P}\setminus\{D\}}\frac{|\mathcal{C}|!\,(|\mathcal{P}|-|\mathcal{C}|-1)!}{|\mathcal{P}|!}\big[v(\mathcal{C}\cup\{D\})-v(\mathcal{C})\big],
\label{eq:shapley}
\end{equation}
where $v(\mathcal{C})$ is the peak margin with drivers in $\mathcal{C}\subseteq\mathcal{P}=\{R,S,T\}$ at their scenario values and complement drivers at the reference, keeping operations fixed. Each term compares two scenarios differing only in driver $D$. The allocation depends on the reference~\cite{sundararajan2020many}, and its sum with $v(\emptyset)$ exactly recovers $m_i(q)$ (Shapley efficiency, App.~\ref{app:G}).

\emph{Scenario change (Q2)} evaluates a specified alternative scenario $q'$ via $\Delta=m_i(q')-m_i(q)$. To capture compound effects, we also compare the joint response to rainfall and coastal stage against the sum of their individual responses:
\begin{equation}
I_{RS}=v(R,S)-v(R,0)-v(0,S)+v(0,0),
\label{eq:interaction}
\end{equation}
where $v$ denotes the margin under the specified rain and surge trajectories with tide and operations fixed, and $0$ denotes the query's reference (the realized driver on paired runs, App.~\ref{app:E4}). The Shapley interaction index averages this second difference across contexts~\cite{grabisch1999axiomatic}.

\emph{Minimum crossing rainfall (Q3)} adapts counterfactual search~\cite{wachter2018counterfactual,wang2023counterfactual} to threshold exceedances. Whereas Q2 specifies an input perturbation and measures its response, Q3 fixes the target threshold $u_i+\varepsilon$ and searches for the least rainfall added along a fixed template $g$ to a base scenario $q_b$, holding other drivers fixed:
\begin{equation}
A_i^\ast(g)=\inf\{A\in[0,A_{\max}] \mid m_i(q_b+A\,g)\ge\varepsilon\},
\label{eq:astar}
\end{equation}
where $q_b$ is $q_0$ in observed regions and the realized drivers on the simulators (which lack rain forecasts), $\varepsilon$ is the crossing margin, and $A_{\max}$ is the search bound evaluated by grid search (App.~\ref{app:B3}). Like flash flood guidance~\cite{sweeney1992modernized,clark2014conus} and tide-conditioned thresholds~\cite{lin2025pluvial}, $A^\ast$ expresses proximity to a threshold as rainfall depth, allowing stations to be compared by the additional rainfall needed for a predicted crossing.

\subsection{Evidence Grading}\label{sec:method-grading}
Interpreting attributions and counterfactuals as scientific statements requires evidence about water-system responses. Model checks, field observations, and paired simulator runs test progressively stronger properties; evidence grading links each explanatory claim to its available evidence and records the resulting support class (Fig.~\ref{fig:overview}c).

\par\textbf{Evidence classes and criteria.}
We define five evidence classes based on agreement with reference responses, reference availability, and evaluation status:
\begin{itemize}[itemsep=1.5pt,topsep=0pt,parsep=0pt,leftmargin=*]
\item \emph{Supported}: The explanation agrees with the reference response across all assessed properties and meets control criteria.
\item \emph{Supported in part}: Reference comparisons establish agreement in some assessed properties (e.g., ranking or direction) but not others (e.g., magnitude).
\item \emph{Shown wrong}: Reference comparisons contradict the claim or show no gain over a non-responding baseline.
\item \emph{Not checkable}: The water system lacks the reference response required to assess the explanatory claim.
\item \emph{Not tested}: No reference comparison was run.
\end{itemize}
The assessed properties are driver rankings and attribution magnitudes for Q1, response directions, magnitudes, and interactions for Q2, and threshold crossing decisions for Q3, alongside negative control criteria (Apps.~\ref{app:I6} and~\ref{app:P2}). Each class applies to one explanation type, architecture, and water system.

\par\textbf{Model checks and controls.}
Model checks verify computational consistency, including exact recovery of the forecast margin by the attribution sum (App.~\ref{app:G}). Untrained model copies separate learned responses from architectural priors, while altered-input variants (which halve or reverse the driver perturbation supplied to the forecaster) test whether candidate validation rules successfully reject wrong responses (Apps.~\ref{app:H} and~\ref{app:P}).

\par\textbf{Observational evidence.}
The observational check assesses agreement between predicted driver effects and observed outcomes. Let $Y$ be the observed peak, $F$ the model peak with realized drivers, and $B$ the model peak with driver $D$ held at $q_0$. Regressing $\Delta_o=Y-B$ on $\Delta_p=F-B$ (the model's substitution response between issue-time and realized drivers) yields the Mincer-Zarnowitz slope~\cite{mincer1969evaluation}:
\begin{equation}
\beta=1+\operatorname{Cov}(\Delta_p,\,Y-F)\,/\operatorname{Var}(\Delta_p)
\label{eq:substitution}
\end{equation}
(App.~\ref{app:H1}), measuring how forecast error covaries with the predicted change. The check combines this slope with directional sign agreement and control comparisons (App.~\ref{app:H3}).

\par\textbf{Paired simulator evidence.}
Paired runs share identical history and differ only in future drivers, providing exact reference responses for perturbations $q'$ that add rainfall $\bar A\,g$ to realized drivers, raise the outlet stage by $\bar s$, hold operated assets at fixed settings, or combine rainfall and stage rises. For realized drivers $q$ and window peak $P$, we compute:
\begin{gather}
\Delta_{\mathrm{sim}}=P_{\mathrm{sim}}(q')-P_{\mathrm{sim}}(q),\quad
\Delta_{\mathrm{model}}=P_{\mathrm{model}}(q')-P_{\mathrm{model}}(q),\notag\\
k=\operatorname{median}\{\Delta_{\mathrm{model}}/\Delta_{\mathrm{sim}} \mid |\Delta_{\mathrm{sim}}|>\max(c_1\nu,\delta)\},
\label{eq:paired}
\end{gather}
where $\nu$ is the numerical noise floor, $\delta$ the materiality floor, and $c_1$ a multiplier. The ratio $k$ compares model and simulator response magnitudes across active cells, while sign agreement measures the fraction of active cells where both shift in the same direction. Attributions are compared against simulator attributions under matching references, scenario changes via $k$, sign agreement, and interaction $I_{RS}$, and minimum crossing rainfall via threshold crossing decisions (Apps.~\ref{app:D}, \ref{app:E}, \ref{app:G}, and~\ref{app:K}).

\par\textbf{Class assignment and reporting.}
Under the paired-run grading criteria (Apps.~\ref{app:I6} and~\ref{app:P2}), an explanation is \emph{shown wrong} if $|k|<k_0$ (missing response) or sign agreement falls below $\zeta_0$; \emph{supported} if $k_-\le k\le k_+$, sign agreement is at least $\zeta_1$, and the untrained control has $|k|<k_0$; and \emph{supported in part} otherwise. Across all four architectures, driver and threshold explanations are graded \emph{supported in part} where responses match the simulator in ranking, direction, or crossing decisions but differ in magnitude, and \emph{shown wrong} where the response is missing or fails to improve on a non-responding baseline (App.~\ref{app:I6}). In observed regions, because each flood occurs once and an ideal slope in Eq.~\eqref{eq:substitution} only indicates low error covariance (which a wrong response can also satisfy), observational verdicts serve as diagnostics; every driver explanation there is locally \emph{not checkable} and carries its architecture-level class from the simulated networks.

\begin{algorithm}[t]
\renewcommand{\algorithmicrequire}{\textbf{Input}}%
\caption{Explaining and grading one forecast peak}\label{alg:explain}
\begin{algorithmic}[1]
\Require forecaster $f_\theta$ (one architecture and system), history and graph $\mathcal{I}_{t_0}$, scenarios $q_b$, $q'$, $u_i$, family $g$, evidence ledger
\State $m_i(q_b)$ from $f_\theta(\mathcal{I}_{t_0},q_b)$ by Eqs.~\eqref{eq:forecast} and~\eqref{eq:margin}
\State $v(\mathcal{C})$ for every coalition $\mathcal{C}\subseteq\mathcal{P}$, the others at the reference
\State $\phi_R,\phi_S,\phi_T$ by Eq.~\eqref{eq:shapley}, with the efficiency residual \Comment{Q1}
\State $\Delta=m_i(q')-m_i(q_b)$ for each $q'$, and $I_{RS}$ by Eq.~\eqref{eq:interaction} \Comment{Q2}
\State $A_i^\ast$ from $q_b$ and its status by grid search, Eq.~\eqref{eq:astar} \Comment{Q3}
\State \Return each explanation of $m_i$ with its kind's evidence class
\end{algorithmic}
\end{algorithm}

\section{Experiments}\label{sec:experiments}

\begin{table*}[t]
\caption{Forecasts in nine observed test periods, median station NSE under $q_0$ and $\Delta$NSE against the issue-time reference. Rows marked $\ddagger$ were fixed after the same year's evaluation on inferred graphs (App.~\ref{app:L2}). GNN, P+R and P-only are the Scenario GNN, Physics + state residual and Physics-only model.}
\label{tab:xr}
\centering
\setlength{\tabcolsep}{3pt}%
\resizebox{\textwidth}{!}{%
\begin{tabular}{@{}cllccccccccc@{}}
\toprule
 & & & & \multicolumn{4}{c}{NSE under $q_0$} & \multicolumn{4}{c}{$\Delta$NSE} \\
\cmidrule(lr){5-8}\cmidrule(l){9-12}
 & Region & Year & Persist. & GNN & Hybrid & P+R & P-only & GNN & Hybrid & P+R & P-only \\
\midrule
\multirow{6}{*}{\rotatebox[origin=c]{90}{South Florida}} & Miami-Dade canals & 2019 & 0.755 & 0.811$\pm$0.003 & 0.804$\pm$0.001 & 0.774$\pm$0.002 & 0.752$\pm$0.002 & $-$0.002 & $-$0.005 & $+$0.001 & $-$0.014 \\
 & Miami-Dade canals & 2022 & 0.704 & 0.787$\pm$0.002 & 0.718$\pm$0.006 & 0.612$\pm$0.095 & 0.653$\pm$0.025 & 0.000 & $-$0.021 & $-$0.090 & $-$0.050 \\
 & Kissimmee lakes & 2019 & 0.993 & 0.992$\pm$$<$0.001 & 0.989$\pm$0.001 & 0.984$\pm$0.014 & 0.991$\pm$$<$0.001 & 0.000 & 0.000 & $-$0.008 & $-$0.002 \\
 & Kissimmee lakes & 2022 & 0.980 & 0.976$\pm$$<$0.001 & 0.973$\pm$0.002 & 0.971$\pm$0.001 & 0.977$\pm$$<$0.001 & $-$0.001 & 0.000 & $-$0.006 & $-$0.001 \\
 & Hendry-Collier canals & 2019 & 0.941 & 0.938$\pm$0.003 & 0.922$\pm$0.001 & 0.933$\pm$0.001 & 0.934$\pm$$<$0.001 & $-$0.003 & $-$0.013 & $-$0.004 & $-$0.001 \\
 & Hendry-Collier canals & 2022 & 0.903 & 0.892$\pm$0.001 & 0.870$\pm$0.003 & 0.876$\pm$0.003 & 0.856$\pm$0.001 & $-$0.005 & $-$0.032 & $-$0.020 & $-$0.039 \\
\midrule
\multirow{3}{*}{\rotatebox[origin=c]{90}{\shortstack{River\\networks}}} & Houston bayous & 2019$^{\ddagger}$ & $-$0.021 & 0.326$\pm$0.005 & 0.316$\pm$0.005 & 0.250$\pm$0.006 & 0.111$\pm$0.025 & $-$0.086 & $-$0.092 & $-$0.105 & $-$0.182 \\
 & Louisiana marshes & 2019$^{\ddagger}$ & 0.654 & 0.768$\pm$$<$0.001 & 0.756$\pm$0.006 & 0.714$\pm$0.001 & 0.688$\pm$$<$0.001 & $+$0.004 & $-$0.011 & $-$0.005 & $-$0.002 \\
 & Aire and Calder rivers & 2019$^{\ddagger}$ & 0.523 & 0.623$\pm$0.002 & 0.613$\pm$0.004 & 0.626$\pm$0.002 & 0.546$\pm$0.009 & $+$0.014 & $+$0.007 & $+$0.012 & $-$0.022 \\
\bottomrule
\end{tabular}}
\end{table*}

\subsection{Setup}\label{sec:setup}

\par\textbf{Observed regions.}
We evaluate six observed flood regions across nine test periods (Table~\ref{tab:xr}). Three come from SF$^2$Bench in South Florida~\cite{zheng2026uncovering}: the Miami-Dade coastal canal network (joining canal gauges, aquifer wells, rain gauges, gate and pump series, and tide gauges) and the inland Kissimmee lakes and Hendry-Collier canals (evaluating rainfall). Models are trained on 2015--2017, validated on 2018, and tested on 2019, then retrained on 2020--2021 for a 2022 hold-out period containing Hurricanes Ian and Nicole (App.~\ref{app:L1}). The Houston bayous~\cite{hcfcd2024fws}, coastal Louisiana marshes~\cite{crms2024}, and Aire and Calder rivers in England~\cite{ea2024hydrology} use physical river network graphs~\cite{usgs2024nhdplushr,os2024openrivers}, with coastal tide gauges~\cite{noaa2024coops} in Houston and Louisiana (Apps.~\ref{app:A} and~\ref{app:L}).

\par\textbf{Simulated networks.}
Two pystorms networks~\cite{rimer2023pystorms}, simulated in SWMM~\cite{rossman2022swmm,mcdonnell2020pyswmm} at \res{data.beta.step-min}-min steps, supply the paired counterfactual runs. In the coastal network \emph{beta} (App.~\ref{app:D1}), \res{data.beta.n-state-nodes} depth nodes drain to a single outfall where tide plus stochastic surge sets the boundary stage behind a one-way flap gate. Models are trained, validated, and tested on \res{data.beta.n-storms-train}, \res{data.beta.n-storms-val}, and \res{data.beta.n-storms-test} storms, with one issue time and \res{data.beta.cf-runs-per-issue} paired runs per test storm scored across \res{beta.nodes.rain} runoff-receiving nodes and \res{beta.nodes.near} near-outlet nodes. Three operated assets (an inland orifice gate, a weir, and a pump) keep fixed settings in the baseline runs and follow random schedules when the operation experiment re-simulates the same \res{data.beta.n-storms} storms (App.~\ref{app:P}). The second network, \emph{gamma} (App.~\ref{app:L6}), drains storage basins through dynamically scheduled valves to a free outfall without a coastal boundary. Its \res{gamma.storms.total} storms (split \res{gamma.storms.train}/\res{gamma.storms.val}/\res{gamma.storms.test}) scale a design storm to \res{gamma.rain.min_mm}--\res{gamma.rain.max_mm}~mm totals; across \res{gamma.cf.n_issue} test issue times, \res{gamma.cf.n_runs} paired runs hold individual valves at five fixed settings or add rainfall.

\begin{table}[t]
\caption{Forecast skill (median NSE), driver response ratios $k$ (rain $\bar A$, stage $\bar s$, and operations; App.~\ref{app:P}), and added-rain threshold balanced accuracy on the simulated coastal network.}\label{tab:sim}
\centering
\setlength{\tabcolsep}{3pt}%
\resizebox{\columnwidth}{!}{%
\begin{tabular}{@{}lcccc@{}}
\toprule
 & \begin{tabular}[b]{@{}c@{}}Scenario\\GNN\end{tabular} & Hybrid & \begin{tabular}[b]{@{}c@{}}Physics +\\residual\end{tabular} & \begin{tabular}[b]{@{}c@{}}Physics-\\only\end{tabular} \\
\midrule
NSE, realized drivers & 0.934\,$\pm$\,0.001 & 0.944\,$\pm$\,0.001 & 0.807\,$\pm$\,0.009 & 0.674\,$\pm$\,0.054 \\
NSE, $q_0$ & 0.423\,$\pm$\,0.008 & 0.479\,$\pm$\,0.013 & 0.383\,$\pm$\,0.023 & 0.305\,$\pm$\,0.009 \\
\midrule
Rain $k$ & 0.71\,$\pm$\,0.13 & 0.92\,$\pm$\,0.08 & 0.97\,$\pm$\,0.03 & 0.86\,$\pm$\,0.09 \\
Stage $k$ & 0.00\,$\pm$\,0.00 & 0.00\,$\pm$\,0.00 & 0.00\,$\pm$\,$<$0.01 & 0.00\,$\pm$\,$<$0.01 \\
Operation $k$ & 0.00\,$\pm$\,$<$0.01 & 0.00\,$\pm$\,$<$0.01 & 0.00\,$\pm$\,0.00 & 0.00\,$\pm$\,0.00 \\
\midrule
Balanced accuracy & 0.63\,$\pm$\,0.01 & 0.72\,$\pm$\,0.03 & 0.69\,$\pm$\,0.02 & 0.67\,$\pm$\,0.02 \\
\bottomrule
\end{tabular}}
\end{table}

\par\textbf{Comparisons and metrics.}
We train three seeds per architecture and region and compare $q_0$ forecasts against \emph{persistence} (repeating the last value) and an \emph{issue-time reference} (the same architecture trained without future drivers). Simulator threshold decisions are compared with constant classifiers and a linear rain response, Miami-Dade threshold summaries with issue-time margins (App.~\ref{app:K}), and explanations with Integrated Gradients (IG)~\cite{sundararajan2017axiomatic} and a ForecastCF variant~\cite{wang2023counterfactual} (App.~\ref{app:G}). Forecasts are scored by Nash-Sutcliffe Efficiency (NSE)~\cite{nash1970river} (1 if perfect, 0 for the observed mean), decisions by balanced accuracy~\cite{brodersen2010balanced} (0.5 for chance), and threshold summaries by recall and ROC-AUC. Tables and figures report seed means and standard deviations (SD), while appendix intervals resample storms or 7-day blocks~\cite{kunsch1989jackknife}.

Per-driver station groups and observational check rules, as well as the operation grading thresholds (App.~\ref{app:P2}), are defined from station geography, hydraulic metadata, and validation-split statistics prior to test evaluation (Apps.~\ref{app:H4} and~\ref{app:L2}). Forecasts cover 48~h in observed regions and 6~h on the simulators, thresholds use high training percentiles, and the rain search is bounded (App.~\ref{app:E9}).

\par\textbf{Paired scenarios and scoring.}
Paired runs give a model and SWMM the same history and future driver perturbation: added rain $\bar A=25.4$~mm (1~in) at runoff-receiving nodes, a raised outlet stage $\bar s=30.48$~cm (1~ft) at near-outlet nodes, or holding one or all three operated assets closed or open over the window. A cell (one scored node at one issue time) enters $k$ only when its simulated change exceeds both $2\nu$ and the 1-cm materiality floor, so $k$ and sign agreement evaluate genuine simulator responses. An interaction $I_{RS}$ is detectable above $\max(4\nu,1\text{~cm})$, with grading bounds summarized in Table~\ref{tab:s-settings}.

\par\textbf{Observational check.}
In observed regions, a cell (one gauge at one issue time) is counted when its predicted change exceeds 1~cm. The \emph{pooled rule} fits one slope across all cells with a wide window around one, whereas the \emph{strict rule} fits station fixed-effect slopes within a narrower window and requires minimum active cells, mean response size, and sign agreement above the untrained control. Both rules require sign agreement above 0.5 and a passed shuffled-driver test, and altered-input variants ($a\in\{1/2,-1\}$) test whether they reject wrong responses (Table~\ref{tab:s-settings}, App.~\ref{app:H3}).

\subsection{Forecasting Performance Across Regions}\label{sec:results-forecast}

\begin{figure*}[t]
\centering
\includegraphics[width=\textwidth]{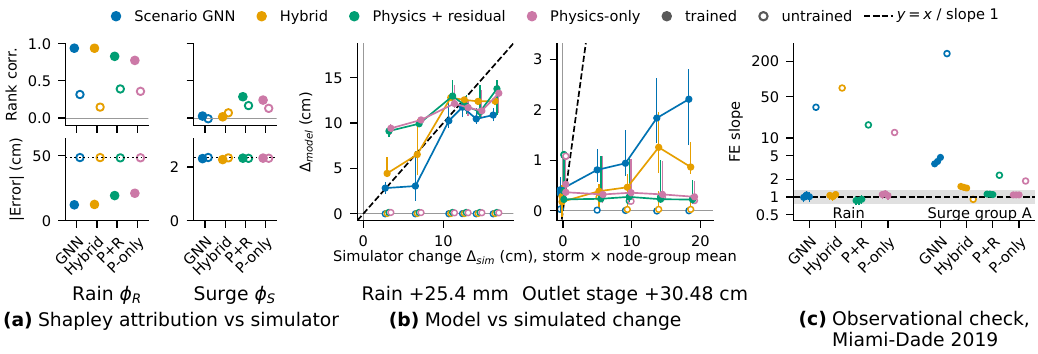}
\vspace{-2em}
\caption{Driver explanations on the simulated coastal network (6-h peak change, cm): (a)~Shapley attribution vs.\ simulator, (b)~predicted vs.\ simulated response to rain $\bar A$ and stage $\bar s$ across storm bins, and (c)~Miami-Dade 2019 station fixed-effect slopes.}
\label{fig:driver-info}
\end{figure*}

To test whether scenario inputs cost forecast skill, Table~\ref{tab:xr} compares each architecture with persistence and its issue-time reference. Outside Houston, the graph networks preserve their skill under $q_0$ (the Scenario GNN with no average loss and the Hybrid within 0.01 NSE) and outperform persistence in every Miami-Dade test year. The water-balance models route every driver through storage continuity, giving each driver an explicit pathway at roughly $1.5\times$ the Hybrid's accuracy cost (App.~\ref{app:L3}). The benefit of future drivers follows the skill of their forecast: in England, the rainfall forecast supplies about four fifths of the Scenario GNN's gain over persistence, whereas in Houston, where it has no skill, omitting it improves NSE for every architecture. Scenario inputs thus preserve forecast skill while making every peak explainable by all three queries.

\subsection{Evidence for Driver Explanations}\label{sec:results-sim}

To test whether driver explanations are accurate where true responses are known, we compare them against paired runs on the simulated coastal network. Trained models' rain attributions $\phi_R$ rank the \res{beta.nodes.scored} scored nodes by simulated rainfall impact with rank correlations at least twice those of untrained copies (Fig.~\ref{fig:driver-info}a), confirming that node rankings are learned while magnitudes are understated (App.~\ref{app:G4}).

For added rain $\bar A$, the Hybrid and water-balance models match the simulated response size more closely ($k=0.86$--$0.97$) than the Scenario GNN ($k=0.71$; Table~\ref{tab:sim}), and every model's response is flatter across storm sizes than SWMM's (Fig.~\ref{fig:driver-info}b). Crucially, similar accuracy conceals divergent interventional responses~\cite{damour2022underspecification}: with realized drivers, the Scenario GNN and Hybrid differ by just 0.01 NSE (and under $q_0$ all models beat persistence NSE \res{beta.fc.persistence.nse}), yet under added rain the Scenario GNN predicts a falling peak where the simulated peak rises in about one cell in five, ten times as often as the Hybrid (Fig.~\ref{fig:s-hexbin}). Rain attribution and added-rain responses are therefore \emph{supported in part} by paired runs for every architecture.

Across the added-rain grid, response ratios also vary with rainfall depth and node set beyond $\bar A$ (Fig.~\ref{fig:rain-response}). On simulator-affected nodes, the Scenario GNN's ratio is lowest at the smallest additions and reaches the others at larger ones. At near-outlet nodes, the Scenario GNN and both water-balance models exceed the simulated rise at the largest additions, whereas the Hybrid stays below it across all amounts.

The ocean boundary gives the opposite verdict. A raised outlet stage $\bar s$ (above the training 99th percentile of surge rises in most storms) leaves the median response of every model, variant, and seed at zero (Table~\ref{tab:sim}), and every surge attribution $\phi_S$ has nearly the error of a non-responding model (Fig.~\ref{fig:driver-info}a, dotted line). Surge attribution and raised-stage responses are therefore \emph{shown wrong} for every architecture. Better forecasts leave this unchanged, cutting outlet-stage response error by less than 5\% of the no-response error (Fig.~\ref{fig:s-forecast-response}), so grading separates a learned rain response from a missing ocean response that accuracy hides.

\subsection{Propagation and Compound Responses}\label{sec:results-compound}

\begin{figure}[t]
\centering
\includegraphics[width=\columnwidth]{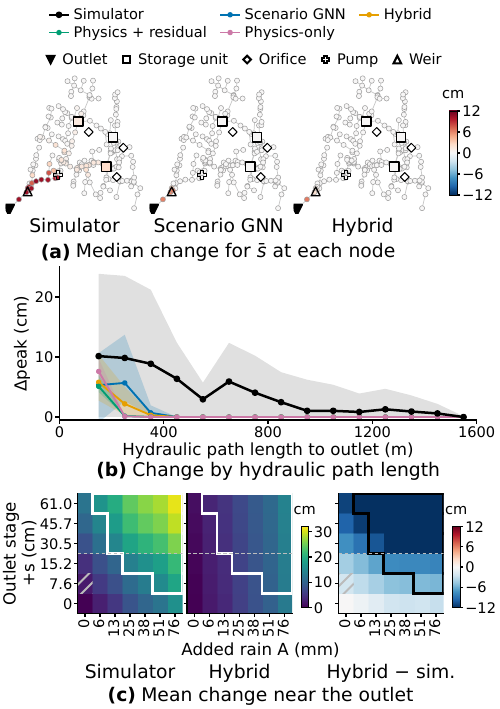}
\caption{Raised outlet stage $\bar s$ and compound rain--stage response on the simulated coastal network (peak change, cm): (a)~spatial median change, (b)~decay with hydraulic distance, and (c)~near-outlet response surface (simulator, Hybrid, and difference).}
\label{fig:propagation}
\end{figure}

To understand why coastal boundary responses fail, we compare spatial backwater propagation and compound rain--stage interactions across the network. Whereas SWMM propagates the raised outlet stage inland, lifting mean peaks by over 1~cm at more than twenty times as many nodes as any learned forecaster, model responses decay immediately past the junctions adjacent to the outfall (Fig.~\ref{fig:propagation}a,b across \res{data.beta.n-storms-test} test storms). For the graph neural networks, this decay partly reflects message-passing depth, which advances the single-edge boundary signal by at most one hop per layer~\cite{xu2018representation,alon2021bottleneck}. Within this reachable receptive field, the Scenario GNN learns a partial response (under half the simulated magnitude versus zero for its untrained control; App.~\ref{app:I-reach}). However, multi-step water-balance rollouts reach nearly every affected node yet still respond only adjacent to the outlet (App.~\ref{app:I-reach}).

In SWMM, coastal stage also interacts with rainfall: the compound interaction $I_{RS}$ at $(\bar A,\bar s)$ is detectable in roughly a quarter of active cells (App.~\ref{app:I4}). Near the outlet, combined rainfall and stage lift more than twice as many initially unwarned cells to the threshold as rainfall alone (Fig.~\ref{fig:propagation}c, where solid outlines mark combinations lifting $\ge 50\%$ of the \res{beta.fig3.grid.n_unwarned_near} unwarned near-outlet cells to exceedance versus at most \res{beta.fig3.grid.wm3fpd.cross_max}\% for the Hybrid; hatching marks changes below $2\nu$ and dashes mark surge rises above the 99th training percentile). Because the mean joint response exceeds the sum of individual responses only slightly (Fig.~\ref{fig:s-compound}), threshold geometry plays a central role in compound flood risk~\cite{zscheischler2020typology}. Paired runs pinpoint the models' deficiency in this compound regime: all architectures recover fewer than 1\% of detectable interactions (App.~\ref{app:I4}), and the Hybrid captures less than half of the simulator's compound threshold crossings (Fig.~\ref{fig:propagation}c).

\begin{table}[t]
\caption{Paired-run evidence classes across all four architectures (valves graded on the river-valve network, other queries on the coastal network; Table~\ref{tab:s-classes}).}\label{tab:classes}
\centering
\begin{tabular}{@{}lll@{}}
\toprule
Query & Explanation & Evidence class \\
\midrule
\multirow{3}{*}{Q1} & Rain $\phi_R$ & supported in part \\
 & Surge $\phi_S$ & shown wrong \\
 & Tide $\phi_T$ & not tested \\
\midrule
\multirow{4}{*}{Q2} & Added rain & supported in part \\
 & Raised outlet stage & shown wrong \\
 & Valves & supported in part \\
 & Gate, weir, pump &\res{ctrl.class.r2.gnn.trained} \\
\midrule
Q3 & $A^\ast$ & supported in part \\
\bottomrule
\end{tabular}
\end{table}

\begin{figure}[t]
\centering
\includegraphics[width=\columnwidth]{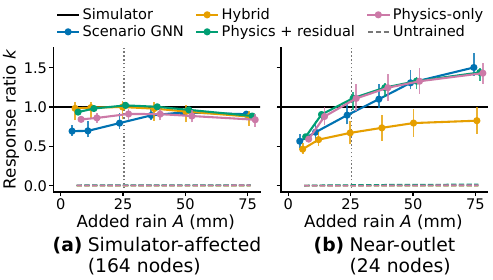}
\vspace{-2em}
\caption{Rain response ratio $k$ across the added-rain grid on the simulated coastal network at (a)~the \res{beta.s12.layer.affected_rain.n} rain-responsive nodes and (b)~the \res{beta.nodes.near} near-outlet nodes (seed means and ranges; dashed: untrained copies; dotted: $\bar A$).}
\label{fig:rain-response}
\end{figure}

\subsection{Operation Responses Across Networks}\label{sec:results-ops}

To test whether models learn hydraulic control effects, the operation experiment (App.~\ref{app:P}) trains all architectures under randomized, state-independent asset schedules and queries the effect of holding individual assets open or closed over the 6-h window. On the coastal network, SWMM shifts mean peaks by over 1~cm across 33\%--60\% of scored nodes, whereas every architecture and seed leaves the median response to the gate, weir, and pump at $k=0.00$ (Table~\ref{tab:sim}), with perturbations confined near each asset (Fig.~\ref{fig:s-ops-reach}). Operation explanations on the coastal network are thus graded \emph{shown wrong} across all twelve architecture--asset pairs. Conversely, on the second network \emph{gamma}, trained identically with random valve schedules, every architecture recovers over half of the simulated valve response, led by the water-balance models, while untrained controls show almost none (\emph{supported in part}, App.~\ref{app:L6}). Under one training design, evidence grading thus separates a missing operation response on the coastal network from a partially learned valve response on the second network.

\subsection{Evidence Grading in Observed Regions}\label{sec:results-observed}

Because observed watersheds lack paired runs, local evaluation relies on the observational check. Crucially, the observational check scores historical driver substitutions (replacing issue-time forecasts with realized drivers), whereas paired runs score causal responses to fixed perturbations (App.~\ref{app:H1}). Applying the observational check to the simulated coastal network reveals what passing it actually establishes. Altered-input variants fed half or reversed driver changes are wrong by construction, yet the pooled rule passes every half-input variant and the reversed-input Hybrid (Fig.~\ref{fig:s-obscheck}), as well as over half of the single-asset operation pairs that paired runs prove wrong (App.~\ref{app:P6}). Under the strict rule, both the Scenario GNN and Hybrid pass for rainfall, even though the Scenario GNN predicts a falling peak under added rain in 20\% of rising cells, while both water-balance models are rejected despite matching the true rainfall response magnitude more closely than the Scenario GNN (App.~\ref{app:H5}). Passing the observational check therefore does not certify a causal response.

Across the nine observed test periods, strict-rule verdicts vary across regions and years (Table~\ref{tab:s-xr-verdicts}). In Miami-Dade 2019, the two rules disagree on rainfall for three of four models even though every station fixed-effect slope lies within 15\% of unity (Fig.~\ref{fig:driver-info}c). Furthermore, untrained copies with a physical branch already produce surge attributions comparable to trained models at Miami-Dade tidal stations (Fig.~\ref{fig:s-miami-shapley}) and respond to surge in Louisiana, where the observational check cannot separate learned dynamics from built-in priors. Grading therefore marks observed-region driver explanations locally \emph{not checkable} and pairs them with their architecture-level class from the simulated networks (Table~\ref{tab:classes}).

\subsection{Evaluating Minimum Crossing Rainfall}\label{sec:results-threshold}

The third query, minimum crossing rainfall (Q3), evaluates proximity to a high-water threshold. On the simulated coastal network, searching from realized drivers, every trained model predicts threshold crossings with higher balanced accuracy than constant classifiers (\res{beta.tab3.always.ba}) and a linear rainfall baseline (\res{beta.tab3.linear.ba}), with the Hybrid achieving nearly three times the Scenario GNN's gain over the linear response (Table~\ref{tab:sim}). Minimum crossing rainfall is therefore graded \emph{supported in part} across all four architectures.

\begin{figure}[t]
\centering
\includegraphics[width=\columnwidth]{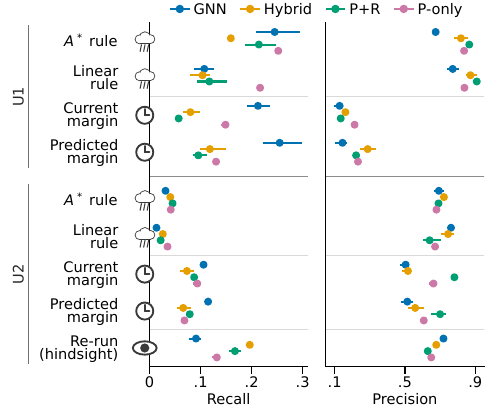}
\vspace{-2em}
\caption{The $A^\ast$ rule in Miami-Dade 2019 against a realized-rain re-run (U1) and observed exceedances (U2), next to
comparison rules.}
\label{fig:state-event}
\end{figure}

In Miami-Dade 2019, $A^\ast$ expresses each unwarned cell's distance to threshold exceedance on a rainfall depth scale (Fig.~\ref{fig:state-event}, showing seed means and ranges across rain-group stations). Flagging a cell when realized rainfall exceeds the forecast by at least $A^\ast$ recovers 14\%--30\% of the crossings found by re-running the same model with realized rain (U1), beating the linear rainfall rule in every run and achieving the highest ROC-AUC among single-number summaries using realized rain totals (App.~\ref{app:K5}). Against observed exceedances with hindsight rain (U2), the $A^\ast$ rule captures more true crossings than the linear rule for every architecture while triggering fewer alarms than issue-time margins matched to its validation alarm rate. In England, $A^\ast$ also ranks future model warnings 12~h ahead more accurately than the raw predicted margin for three architectures (App.~\ref{app:L5}), condensing scenario sensitivity into a single actionable rainfall depth per station and issue time.

\section{Discussion and Conclusion}\label{sec:discussion}

Our experiments show that empirical support for a forecast explanation depends on the driver, the water system, and the response property queried. On the simulated coastal network, all four architectures learn the rainfall response in node ranking and most of its size, and their threshold decisions by minimum crossing rainfall are partly supported. Yet they miss the inland reach of a raised ocean boundary, its joint effect with rain, and gate, weir, and pump responses, while recovering over half of the valve response on a second simulated network. Because forecasters of similar accuracy can differ sharply in these responses, evaluating compound flood models requires pairing predictive performance with interventional evidence on the driver responses a task relies on.

We presented a compound flood forecasting system in which driver attribution, scenario changes, and minimum crossing rainfall operate as three queries of one scenario interface, and evidence grading assigns each explanation type an evidence class from model checks, observational diagnostics, and paired simulator runs. Evaluated across four architectures in six observed regions and two simulated networks, this work provides a framework for testing the driver responses that forecasting models learn and determining when their explanations receive empirical support.

\section{Limitations and Ethical Considerations}\label{sec:limitations}

Our paired-run evidence comes from two synthetic SWMM drainage networks, so the architecture-level classes attached to observed-region explanations reflect the topologies and storm regimes of those networks. Control operations follow randomized schedules rather than closed-loop operating rules\smbadd{, except at the rule-controlled gates and pumps of the synthetic barrage in App.~\ref{app:Lsmb}}, and high-water thresholds correspond to historical stage percentiles at each gauge.

No human participants, user studies, operational deployments, or real-world gate or pump interventions were involved. A threshold crossing is a model output rather than physical inundation, and $A^\ast$ is an analytical diagnostic rather than an operational flood warning. All datasets used in this study are public and documented in Apps.~\ref{app:A1} and~\ref{app:L}.

\section{Generative AI Usage}\label{sec:genai}

We used Anthropic's Claude to assist with code implementation, data visualization and analysis scripts, reference verification, and manuscript editing. All experimental designs, mathematical derivations, and reported empirical results were independently verified by the authors, who take full responsibility for the contents of this work.

\bibliographystyle{ACM-Reference-Format}
\bibliography{references}

\clearpage
\newpage
\appendix
\setcounter{figure}{0}\setcounter{table}{0}
\renewcommand{\thefigure}{S\arabic{figure}}\renewcommand{\theHfigure}{S\arabic{figure}}
\renewcommand{\thetable}{S\arabic{table}}\renewcommand{\theHtable}{S\arabic{table}}
\renewcommand{\topfraction}{0.95}\renewcommand{\bottomfraction}{0.5}\renewcommand{\textfraction}{0.05}
\renewcommand{\floatpagefraction}{0.9}\renewcommand{\dbltopfraction}{0.95}\renewcommand{\dblfloatpagefraction}{0.9}
\setcounter{topnumber}{4}\setcounter{bottomnumber}{2}\setcounter{totalnumber}{6}\setcounter{dbltopnumber}{3}

\section{Study Systems, Data Provenance, and Preprocessing}\label{app:A}
This appendix documents the two study systems, namely what each series measures, how it was prepared before the models saw it,
and which values were filled. The other observed regions are described in App.~\ref{app:L}. The Miami-Dade canals supply observed water levels under real drivers
but no counterfactual labels. The simulated drainage network supplies paired runs that differ only in a specified
driver, from a known controlled system that is not calibrated to Miami-Dade. The two are different hydrological
systems, and no parameter, threshold or result is transferred from one to the other.

\par\textbf{The two systems at a glance.}
The Miami-Dade observational system (SF$^2$Bench S\_6 part~1) predicts the hourly level of \res{data.mia.n-state-nodes}
state nodes, \res{data.mia.n-canal-gauges} canal stages and \res{data.mia.n-wells} well heads, on a graph of
\res{data.mia.n-edges} edges from \res{data.mia.n-rain-gauges} rain gauges, \res{data.mia.n-tide-gauges} tide gauges and
\res{data.mia.n-controls} gate and pump series. Its models read \res{data.mia.history-h}~h and forecast
\res{data.mia.horizon-h}~h, are trained on the years from 2015 to 2017, validated on 2018 and tested on 2019, and are queried every
\res{data.mia.query-step-h}~h. Its labels are observed stages. The simulated drainage network (pystorms
\emph{beta}, SWMM 5.2.4) predicts the depth of \res{data.beta.n-state-nodes} nodes on a graph of
\res{data.beta.n-edges} edges from one rain gauge, the outlet stage and \res{data.beta.n-controls} controls in steps of
\res{data.beta.step-min}~min. Its models read and forecast \res{data.beta.window-h}~h (\res{data.beta.window-steps}
steps), its \res{data.beta.n-storms} storms are split \res{data.beta.n-storms-train} / \res{data.beta.n-storms-val} /
\res{data.beta.n-storms-test}, and its labels are paired runs at \res{data.beta.cf-issue-times} counterfactual issue
times, with added rain from \res{data.beta.rain-add-min-mm} to \res{data.beta.rain-add-max-mm}~mm and added outlet stage from
\res{data.beta.stage-add-min-cm} to \res{data.beta.stage-add-max-cm}~cm. In the issue-time scenario $q_0$, Miami-Dade
rain comes from a ridge forecast and the simulator has none. Both hold surge and operations at their last values and
treat the astronomical tide as known.

\subsection{Study domains and measured quantities}\label{app:A1}

\begin{figure*}[tp]
\centering
\includegraphics[width=\textwidth]{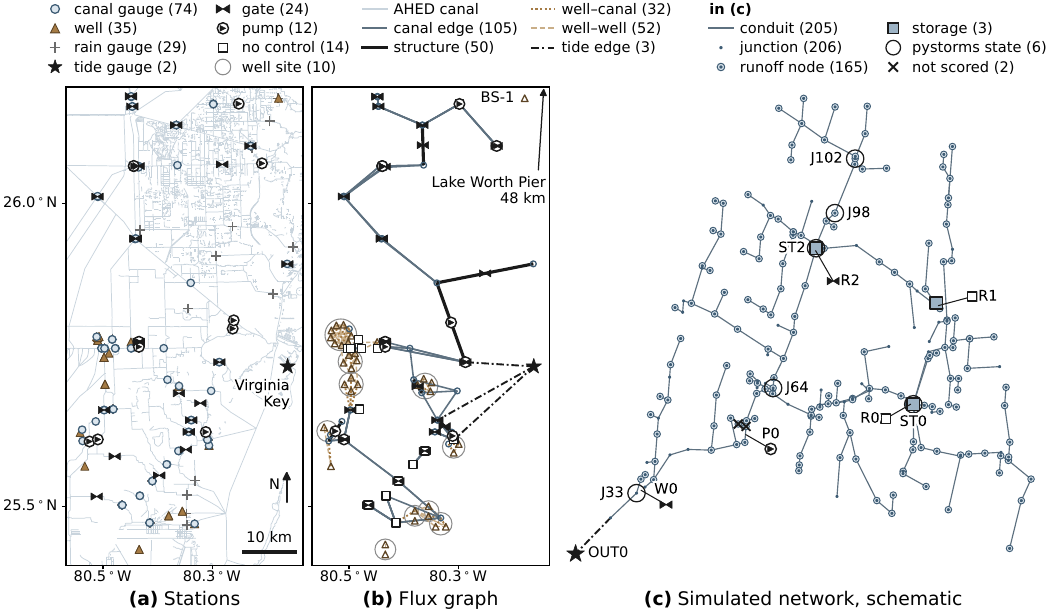}
\caption{Study domains and network structure: (a)~Miami-Dade stations and AHED canal layer, (b)~flux graph, and (c)~simulated coastal drainage network.}
\label{fig:s-domains}
\end{figure*}

\par\textbf{Miami-Dade.}
The Miami-Dade case is a managed canal network, released as SF$^2$Bench S\_6
part~1~\cite{zheng2026uncovering}, whose gauges span about \res{data.mia.extent-km-ew}~km east to west and \res{data.mia.extent-km-ns}~km north to south
(Fig.~\ref{fig:s-domains}a). Its state nodes are \res{data.mia.n-canal-gauges} canal stage gauges, which record the
water level in the canal, and \res{data.mia.n-wells} wells, which record groundwater head. Inputs that are not predicted are \res{data.mia.n-rain-gauges} rain gauges, the
\res{data.mia.n-tide-gauges} tide gauges, \res{data.mia.n-gates} gate-opening series and \res{data.mia.n-pumps} pump
series. The flux graph has \res{data.mia.n-edges} edges (Fig.~\ref{fig:s-domains}b), namely \res{data.mia.n-edges-canal}
canal edges between neighboring gauges along a canal, \res{data.mia.n-edges-structure} structure edges
(\res{data.mia.n-edges-gate} gates, \res{data.mia.n-edges-pump} pumps and \res{data.mia.n-edges-ustruct} structures
without a control series), \res{data.mia.n-edges-groundwater} groundwater edges (\res{data.mia.n-edges-seep} between a
well and a canal gauge, \res{data.mia.n-edges-gw} between wells) and \res{data.mia.n-edges-tide} tide edges. All tide
edges start at Virginia Key and end at G93\_T, S119\_T and S123\_T. The Lake Worth Pier gauge has no edge within the
30-km attachment radius and no message-passing edge, so it influences no prediction of any model; it enters only the
statistical rain forecaster. Well BS-1 has no flux edge, so its physics update is rain alone. In the 2019 test year,
\res{data.mia.n-scored-canal-test} canal gauges and \res{data.mia.n-scored-wells-test} wells are scored. G211\_H, S9\_T
and G58\_H have no observed hour in 2019. Graph nodes, gauges and scored targets are therefore three different counts.

Fig.~\ref{fig:s-domains}a draws every station at its published coordinates (longitude, latitude) over the public canal
and stream layer of the Arc Hydro Enhanced Database (AHED). Of the \res{data.mia.n-tide-gauges} tide gauges, only
Virginia Key lies in this frame, and an arrow points to the second tide gauge, Lake Worth Pier, about 48~km north of the
northernmost station. Panel b draws the flux graph in the same frame, with the state nodes drawn hollow over the edges.
Nodes within 1~km of each other that include a well (10 well sites, gray circles, most of them holding several wells at
one location) are spread on a small ring, each moved by less than 4~km, so that their 51 internal edges show, and all
other nodes are at their coordinates. A structure edge joins the two stage gauges of its structure, and the symbol of
the structure sits at the edge midpoint. The two gauges often share one site and lie under that symbol, as do 9 short
canal edges between gauges of neighboring structures, and symbols overlap in the dense south-western cluster. Well BS-1
is the only node in panel b without an edge.

A level above a gauge's high-water threshold $u_i$ (its training-period p95) is a statistical high-water state of
that gauge rather than a measurement of surface inundation or damage.

\par\textbf{Simulated drainage network.}
The pystorms \emph{beta} network~\cite{rimer2023pystorms} runs in SWMM 5.2.4 through pyswmm
2.1.0~\cite{rossman2022swmm,mcdonnell2020pyswmm}. It has \res{data.beta.n-state-nodes} depth nodes
(\res{data.beta.n-junctions} junctions and \res{data.beta.n-storage} storage units ST0 to ST2) and one outfall, OUT0,
whose prescribed stage acts behind a flap gate. \res{data.beta.n-subcatchments} subcatchments
(\res{data.beta.catchment-km2}~km$^2$) on one rain gauge drain to the \res{data.beta.n-rain-nodes} runoff-receiving
nodes. The structures are the orifice R2 and the weir W0 (both open), the fixed orifices R0 and R1, and the pump P0. The
graph read by the models has \res{data.beta.n-edges} edges, namely \res{data.beta.n-conduits} conduits, the outfall conduit as
a tide edge, and the five structures. The \res{data.beta.n-storms} generated 48-h storms are split by storm into
\res{data.beta.n-storms-train}, \res{data.beta.n-storms-val} and \res{data.beta.n-storms-test}, and the counterfactual
runs use one issue time per test storm. Map coordinates are the schematic coordinates of the model file
(Fig.~\ref{fig:s-domains}c), in which the runoff nodes are the runoff-receiving nodes. Panel c marks R2 and W0 with the
gate symbol, P0, which is held on, with the pump symbol, and R0 and R1 with the open square. The structure links are
too short to show, so their symbols sit beside them with a leader to the link midpoint. Large circles mark the six
depths observed by the pystorms \emph{beta} scenario, and crosses mark J56, the pump's wet well, and its neighbor J57,
which are not scored.
Table~\ref{tab:s-inventory} compares the counts of both systems with those of seven further systems, whose
configurations and results are reported in App.~\ref{app:L}. They are the second simulator (pystorms \emph{gamma} in
SWMM, 19 subcatchments over 0.833~km$^2$ whose storage basins drain through valves on a random, state-independent
schedule to a free outfall), S\_7 part~1 (the same Miami-Dade canal system from 2020 to 2023, with every input before
2022 rebuilt from data before 2022-01-01, the hurricanes Ian and Nicole in the 2022 test year, and 2023 as a second test
year), the inland parts~0 and~2 of both splits, whose graphs have no tide edge, and Houston (Harris
County Flood Warning System stage and rain gauges with tide gauges in Galveston Bay, years in UTC; 134 rain gauges are
selected and 120 of them enter the rain mix). Houston runs all four
structures on the physical river network built from the National Hydrography Dataset Plus High
Resolution~\cite{usgs2024nhdplushr} (App.~\ref{app:L}). Table~\ref{tab:s-xr-baselines} uses
its inferred graph, without physics models.
Coastal Louisiana and the Aire and Calder catchments are described in App.~\ref{app:L} and are not in
Table~\ref{tab:s-inventory}. \smbadd{The synthetic barrage testbed is described in Table~\ref{tab:s-smb-setup},
not in Table~\ref{tab:s-inventory}.}

In Table~\ref{tab:s-inventory}, scored nodes are those with at least 10 observed cells and non-zero variance, which on
the simulator are all nodes except the numerically oscillating J56 and J57.
Counts are for the first test year; in 2023 Miami-Dade scores 79 + 24 nodes. The ocean boundary is the tide gauges, or
the outlet stage of the simulator, and the resampling units are weekly blocks (52 per evaluation year) and storms.
Valves count as gates, and each simulator has one counterfactual issue time per test storm (150 on \emph{beta}, 100 on
\emph{gamma}).

\begin{table*}[t]
\centering
\caption{Study-system inventory. Part~0 is the Kissimmee lakes, part~1 the Miami-Dade canals and part~2 the
Hendry-Collier canals. Level nodes are canal or stream gauges, or simulator depth nodes.}
\label{tab:s-inventory}
\setlength{\tabcolsep}{3pt}%
\begin{tabular*}{\textwidth}{@{\extracolsep{\fill}}lccccccccc@{}}
\toprule
 & \multicolumn{3}{c}{SF\textsuperscript{2}Bench S\_6, observed} & Houston & \multicolumn{3}{c}{SF\textsuperscript{2}Bench S\_7, observed} & \multicolumn{2}{c}{pystorms, simulated} \\
\cmidrule(lr){2-4}\cmidrule(lr){5-5}\cmidrule(lr){6-8}\cmidrule(l){9-10}
 & Part~0 & Part~1 & Part~2 & Bayous & Part~0 & Part~1 & Part~2 & \emph{beta} & \emph{gamma} \\
\midrule
Level nodes + wells & 19 + 64 & 74 + 35 & 12 + 19 & 84 + 0 & 22 + 71 & 80 + 24 & 30 + 53 & 209 + 0 & 11 + 0 \\
Scored nodes & 19 + 64 & 71 + 35 & 12 + 19 & 84 + 0 & 22 + 71 & 76 + 24 & 30 + 53 & 207 + 0 & 11 + 0 \\
Tide edges & 0 & 3 & 0 & 19 & 0 & 6 & 0 & 1 & 0 \\
Extent (km), area (km$^2$) & 46\,$\times$\,59 & 38\,$\times$\,83 & 23\,$\times$\,78 & 96\,$\times$\,75 & 46\,$\times$\,62 & 38\,$\times$\,80 & 23\,$\times$\,78 & 1.315 & 0.833 \\
\midrule
Rain gauges & 12 & 29 & 3 & 120 & 12 & 30 & 9 & 1 & 10 \\
Ocean-boundary series & 2 & 2 & 2 & 4 & 3 & 3 & 3 & 1 & 0 \\
Gates + pumps & 5 + 0 & 24 + 12 & 3 + 1 & 0 + 0 & 7 + 0 & 24 + 13 & 17 + 3 & 2 + 1 & 11 + 0 \\
\midrule
Training & \multicolumn{4}{c}{\cellcolor{black!10}2015--2017} & \multicolumn{3}{c}{\cellcolor{black!10}2020-01 to 2021-07} & 600 storms & 400 storms \\
\addlinespace[1.5pt]
Validation & \multicolumn{4}{c}{\cellcolor{black!10}2018} & \multicolumn{3}{c}{\cellcolor{black!10}2021-08 to 2021-12} & 150 storms & 100 storms \\
\addlinespace[1.5pt]
Test & \multicolumn{4}{c}{\cellcolor{black!10}2019} & \multicolumn{3}{c}{\cellcolor{black!10}2022, 2023} & 150 storms & 100 storms \\
\addlinespace[1.5pt]
Step; history, horizon & \multicolumn{7}{c}{\cellcolor{black!10}1~h; 168~h, 48~h} & \multicolumn{2}{c}{\cellcolor{black!10}15~min; 6~h, 6~h} \\
\bottomrule
\end{tabular*}
\end{table*}

\subsection{Provenance, units, clocks and vertical datum}\label{app:A2}

\par\textbf{Sources and units.}
All Miami-Dade series are the SF$^2$Bench processed hourly files, which derive from DBHYDRO stations of the South
Florida Water Management District and, for the ocean, from tide gauges of the National Oceanic and Atmospheric
Administration (NOAA); the input files are those of the published release~\cite{zheng2026uncovering}. Each
station file holds the columns TIMESTAMP, VALUE, CONFIDENCE and INTERPOLATED\_VALUE; the models read
INTERPOLATED\_VALUE. Raw DBHYDRO records serve to verify units, clocks and datums and to calibrate the
rain units. Stage and well head are converted from feet to centimeters ($\times 30.48$) and tide heights from meters to
centimeters. A gate or pump series is divided by its training-period maximum and clipped at zero but not above, so test
values can exceed one. Rain stays in the release's native units inside the models; its conversion to millimeters is
described at the end of this appendix. Stage, well head and rain are hourly means of the DBHYDRO breakpoint records. Gate
and pump series are hourly, and the release does not state which DBHYDRO gate record, or which pump quantity (speed or
flow), it used. What each driver holds after the issue time is set by the scenarios, and the treatment of missing
values follows below.

\par\textbf{Simulator variables.}
On the simulator the state is the SWMM node depth in cm at the end of each \res{data.beta.step-min}-min step, measured
above the node invert, which is also the reference of the outlet stage. Rain enters in mm per step multiplied by the
constant \res{data.beta.rain-constant}. The outlet stage is a generated astronomical cycle (constituents M2, K1, M4)
plus a random surge pulse (Table~\ref{tab:s-simquality}), and the control settings are read back from SWMM every minute
and averaged over each step. All simulator series are complete.

\par\textbf{Clocks.}
Station files carry hour-beginning labels in Eastern Standard Time (UTC$-$5, no daylight saving). At the gauges that
match DBHYDRO, the hourly means of the raw records, binned to the hour start, agree with the released values, which
confirms the clock at these gauges. The tide files are NOAA
verified hourly heights in meters above NAVD88 (North American Vertical Datum of 1988) and in UTC. They equal the NOAA download at zero lag and are shifted
by $-5$~h to align with Eastern Standard Time. At Virginia Key, \res{data.mia.tide-vk-hours-filled} hours without a NOAA verified value hold filled values
in the released file.

\par\textbf{Vertical datum.}
The release states no vertical datum. The preprocessing adopts NGVD29 (National Geodetic Vertical Datum of 1929) for stage and well head and adds
\res{data.mia.tide-datum-shift-cm}~cm to the NAVD88 tide. Against DBHYDRO records served in feet NAVD88 (median hourly
difference from one record per gauge on observed hours), all \res{data.mia.datum-wells-match} wells agree within
\res{data.mia.datum-tol-wells-ft}~ft and \res{data.mia.datum-canal-match} of the \res{data.mia.n-canal-gauges} canal
gauges within \res{data.mia.datum-tol-ft}~ft, whereas the other \res{data.mia.datum-canal-offset} canal gauges read lower by between
\res{data.mia.datum-canal-offset-min-cm} and \res{data.mia.datum-canal-offset-max-cm}~cm. No single datum conversion
explains these offsets, so the datum is mixed. Thresholds are
station-specific and absorb a constant offset. Neural inputs are standardized per series. Canal, groundwater and tide
edges of the physics branch start with an offset $b_e$ equal to minus the training-mean head difference, so a constant
offset initially cancels on these edges; gate, pump and uncontrolled-structure edges start at $b_e=0$ and learn it, so an
offset between the two gauges of a structure first appears as a head difference.

\par\textbf{Astronomical tide and surge.}
The released astronomical series is a least-squares fit of 15 harmonic constituents (including the annual and
semiannual ones), a constant and a linear trend to the whole record from 2015 to 2019. Surge is the observed total minus this
fit. It is therefore not the NOAA harmonic prediction. It sits \res{data.mia.tide-lwp-astro-minus-noaa-pred-cm}~cm (Lake
Worth Pier) and \res{data.mia.tide-vk-astro-minus-noaa-pred-cm}~cm (Virginia Key) above it on average, because it
carries the record's mean sea level and trend. Because the fit uses the 2018 and 2019 observations, a fit to the years from 2015 to 2017
alone differs in 2019 by \res{data.mia.tide-lwp-astro-trainfit-rms-test-cm} and
\res{data.mia.tide-vk-astro-trainfit-rms-test-cm}~cm root-mean-square at the two gauges. The scenarios treat this series
as known over the horizon.

\par\textbf{Splits and fitted statistics.}
Training covers the years from 2015 to 2017, validation 2018 and test 2019. A window holds 168~h of history and 48~h of horizon
inside one period, and Table~\ref{tab:s-splits} gives the window counts and strides. Stage statistics and thresholds, gate and pump
maxima, rain scales, rain-unit factors, the rain forecaster and the initial edge offsets are fitted on the training
period of the released series, filled hours included. Apart from the release's own gap filling over the full record
(described below), the astronomical fit is the only preprocessing step that uses the validation and test years. On the simulator, storms are assigned to the three splits by a seeded permutation, and
windows of 24 steps of history and 24 steps of horizon never cross a storm.
On the simulator the stage statistics, the thresholds $u_i$ and the rain constant are fitted on the training storms. In
both systems the validation split serves epoch selection only, and nothing is fitted on the test split.

\begin{figure*}[tp]
\centering
\includegraphics[width=\textwidth]{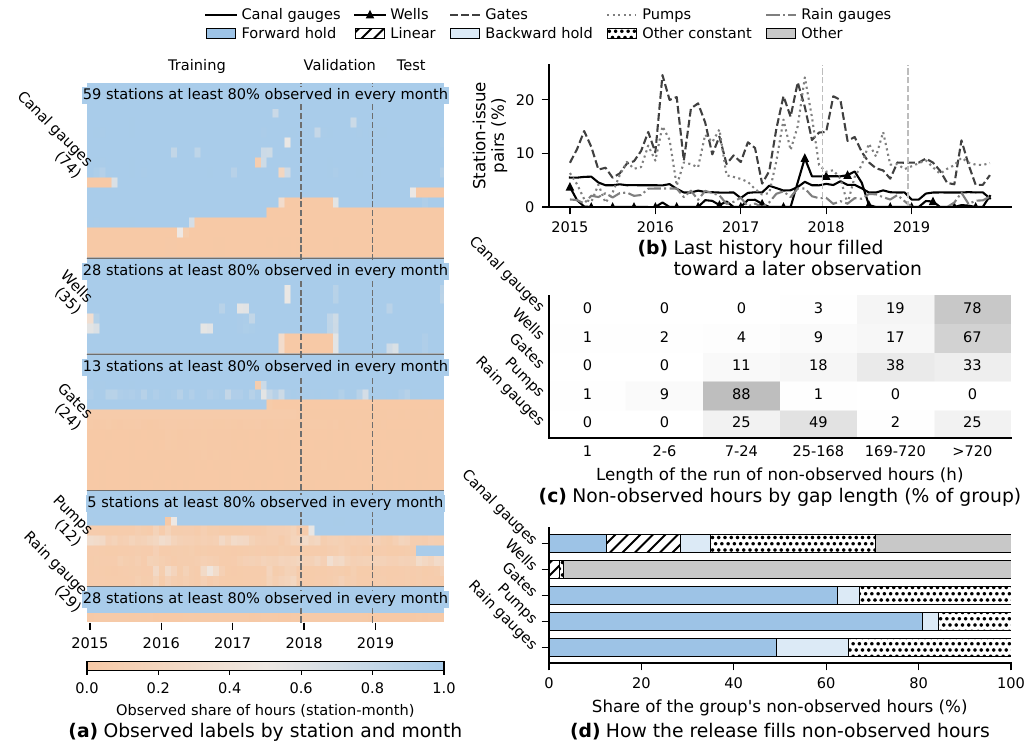}
\caption{Observation availability and imputation in Miami-Dade (2015--2019; station counts in brackets; dashed lines mark the 2018 and 2019 splits).}
\label{fig:s-coverage}
\end{figure*}

\begin{table}[t]
\centering
\caption{Window and issue-time counts per data split (stride in brackets; period definitions in Table~\ref{tab:s-inventory}).}
\label{tab:s-splits}
\resizebox{\columnwidth}{!}{%
\begin{tabular}{@{}lccc@{}}
\toprule
Split & Windows or issue times & Miami-Dade, S\_6 part~1 & Simulator, \emph{beta} \\
\midrule
Training & training windows & 8{,}697 (3~h) & 29{,}400 (3 steps) \\
\midrule
\multirow{3}{*}{Validation} & scored windows & 8{,}545 (1~h) & 10{,}950 (2 steps) \\
 & selection windows & 2{,}849 (3~h) & 7{,}350 (3 steps) \\
 & query issue times & 1{,}425 (6~h) & -- \\
\midrule
\multirow{2}{*}{Test} & scored windows & 8{,}545 (1~h) & 10{,}950 (2 steps) \\
 & query issue times & 1{,}425 (6~h) & 150 \\
\bottomrule
\end{tabular}}
\end{table}

\subsection{Observed labels, filled values and issue-time exposure}\label{app:A3}

\par\textbf{Observed versus filled.}
An hour is an observed label when INTERPOLATED\_VALUE is present and CONFIDENCE is positive, the mask used by the
training loss and by every score, persistence included. CONFIDENCE is the release's count of aggregated raw points; for
rain it exceeds the raw record count of the hour because the release smooths across hours, so an observed hour means
raw support rather than a raw reading in that exact hour. Every station file covers every hour from 2015 to 2019, so every non-observed hour already holds a value filled by
the release (Fig.~\ref{fig:s-coverage}a). Canal gauges and wells are observed in most hours, gates and pumps in about two thirds of them
(\res{data.mia.obs-share-gates-test}\% and \res{data.mia.obs-share-pumps-test}\% of 2019 station-hours;
Table~\ref{tab:s-missingness}). Gate and pump gaps are mostly filled by holding the last observation forward, whereas
canal gaps are filled by a mix of holds, linear segments and other shapes (Fig.~\ref{fig:s-coverage}d). The models read
the filled series as history, the loss and the scores use observed labels only, and the thresholds use all training
hours.

Fig.~\ref{fig:s-coverage}c and d break the non-observed hours of each variable group down by the length of the gap
they belong to and by the fill shape the release used, as shares of that group's non-observed hours. Panel b gives, for
every month from 2015 to 2019, the share of (hourly issue time, station) pairs whose last history hour is filled by a
rule other than a forward hold, inside a gap that closes with an observation after the issue time, which is a lower
bound for history values computed from later measurements.

\par\textbf{Exposure at the issue time.}
Across every 2019 hourly issue time and station, the last history hour is filled in \res{data.mia.exposure-canal-test-filled}\% of canal-gauge pairs, for gates in
\res{data.mia.exposure-gates-test-filled}\% and for pumps in \res{data.mia.exposure-pumps-test-filled}\%. The last
history hour lies inside a gap that is filled by a non-hold rule and closes with an observation after the issue time in
\res{data.mia.exposure-canal-test-closes-after}\% of canal pairs and in between
\res{data.mia.exposure-gates-test-closes-after}\% and \res{data.mia.exposure-pumps-test-closes-after}\% of gate and pump
pairs (Fig.~\ref{fig:s-coverage}b). History inputs are thus the benchmark's preprocessed series, gap-filled over the full record,
and these shares hold for every Miami-Dade result. In Table~\ref{tab:s-missingness}, the observed share of the two tide gauges is the share of hours with a NOAA verified
value.

\begin{table}[t]
\centering
\caption{Observation coverage (\%, lowest-coverage station in brackets) and gap counts in Miami-Dade (S\_6 part~1) by variable group.}
\label{tab:s-missingness}
\setlength{\tabcolsep}{3pt}%
\resizebox{\columnwidth}{!}{%
\begin{tabular}{@{}lrrrrrrr@{}}
\toprule
 &  & \multicolumn{3}{c}{Observed, mean (lowest station)} & \multicolumn{2}{c}{Test year 2019} & Longest \\
\cmidrule(lr){3-5}\cmidrule(lr){6-7}
Group & $n$ & Train & Val. & Test & No obs. & Holds & gap (h) \\
\midrule
Canal gauges & 74 & 94.6 (0.2) & 92.5 (0.0) & 92.6 (0.0) & 3 & 1.6 & 32{,}112 \\
Wells & 35 & 99.0 (89.7) & 97.1 (55.2) & 99.6 (94.4) & 0 & 0.0 & 6{,}628 \\
Gates & 24 & 65.9 (0.3) & 62.5 (0.0) & 62.6 (0.0) & 1 & 26.1 & 21{,}911 \\
Pumps & 12 & 54.6 (6.1) & 61.5 (5.7) & 65.7 (6.2) & 0 & 26.5 & 1{,}048 \\
Rain gauges & 29 & 96.6 (1.7) & 96.6 (2.1) & 96.6 (0.9) & 0 & 2.7 & 3{,}958 \\
\midrule
\multicolumn{8}{@{}c}{\emph{Tide gauges}} \\
Lake Worth Pier & 1 & 100.0 & 100.0 & 100.0 & 0 & 0.0 & 0 \\
Virginia Key & 1 & 98.9 & 100.0 & 100.0 & 0 & 0.0 & 277 \\
\bottomrule
\end{tabular}}
\end{table}

\subsection{Rainfall units and the issue-time rain forecast}\label{app:A4}

\par\textbf{Rain units.}
The released hourly rain is not an hourly depth. At the gauges checked against raw records it follows the mean of the
raw DBHYDRO readings within the hour, smoothed across hours, and at every gauge its totals are a small fraction of the
measured depth (annual totals between \res{data.mia.rain-annual-share-min}\% and \res{data.mia.rain-annual-share-max}\% at the three
gauges checked). For each gauge $r$, one factor $f_r$ is fitted on the years from 2015 to 2017 only, as the ratio of the summed released values to the
summed raw depth in inches, using the raw record with the most training entries. \res{data.mia.rain-factor-n} of the
\res{data.mia.n-rain-gauges} gauges have a factor, between \res{data.mia.rain-factor-min} and
\res{data.mia.rain-factor-max} native units per inch (median \res{data.mia.rain-factor-median}); FT.~LAUD\_R has no raw
rain record and receives the median factor. The same ratio, computed but never used for fitting, has a median of
\res{data.mia.rain-factor-val-median} in 2018 and \res{data.mia.rain-factor-test-median} in 2019. The training-period
hourly correlation between released and raw hourly rain has a median of \res{data.mia.rain-hourly-corr-median} (lowest
\res{data.mia.rain-hourly-corr-min}). The released VALUE and INTERPOLATED\_VALUE training sums coincide at every gauge,
so fitting on VALUE does not bias the factor for the series the models read. An added depth of $A$~mm becomes
$(A/25.4)\,f_r$ native units at gauge $r$. Matching totals fixes a depth scale per gauge and period, so the added
amounts and the minimum crossing rainfall are total depths of a fixed storm shape. Inside the models, rain is
divided by the gauge's training p99 for the neural encoders (floored at
one where the p99 is zero, which happens only at FT.~LAUD\_R) and scaled by a learned gain in the physics branch
(Eq.~\eqref{eq:app-reservoir}).

\par\textbf{Issue-time rain forecast.}
The rain of the issue-time scenario $q_0$ comes from a per-gauge ridge regression of the next-48-h total on
\res{data.mia.rainfc-n-features} past-only features, namely the rain of every gauge in 33 bins over the past 168~h (hourly for
24~h, then 6-h and daily bins), four seasonal terms, the eight 6-h changes over the past 48~h of every state node and of
both tide totals, and the current stages. The penalty is chosen from $\{10,\dots,10^5\}$ on the last fifth of
the years from 2015 to 2017, and the model is refitted on all of these years with rows every 2~h. Predictions are clipped at zero and
spread uniformly over the 48~h; a window uses the forecast made at its last history hour. The per-gauge $R^2$ of the
48-h total has a median of \res{data.mia.rainfc-r2-val-median} in 2018 and \res{data.mia.rainfc-r2-test-median} in
2019, against \res{data.mia.rainfc-r2-clim-val-median} and \res{data.mia.rainfc-r2-clim-test-median} for a monthly
training climatology; it is positive at \res{data.mia.rainfc-gauges-r2-positive-test} of the
\res{data.mia.n-rain-gauges} gauges in 2019, and the area-mean 48-h total reaches $R^2$
\res{data.mia.rainfc-area-r2-val} and \res{data.mia.rainfc-area-r2-test}. Its stage features come from the
same filled series as the model inputs.

\section{Information Sets and Scenario Construction}\label{app:B}
\ifcsname c@appalg\endcsname\else\newcounter{appalg}\renewcommand{\theappalg}{S\arabic{appalg}}\fi

All scenario comparisons are conditional on a specified information set at the forecast origin. This appendix documents
every scenario and query of Eqs.~\eqref{eq:forecast} to \eqref{eq:interaction} as implemented, namely what a model knows at the
issue time $t_0$, which driver a scenario changes, from when, by how much, and what is held fixed.
Algorithm~\ref{alg:explain} of the main text summarizes how these queries explain and grade one forecast peak, and
Algorithm~\ref{alg:s-astar} gives the grid search of its step 5.

\subsection{Forecast-origin information}\label{app:B1}

\begin{figure*}[tp]
\centering
\includegraphics[width=\textwidth]{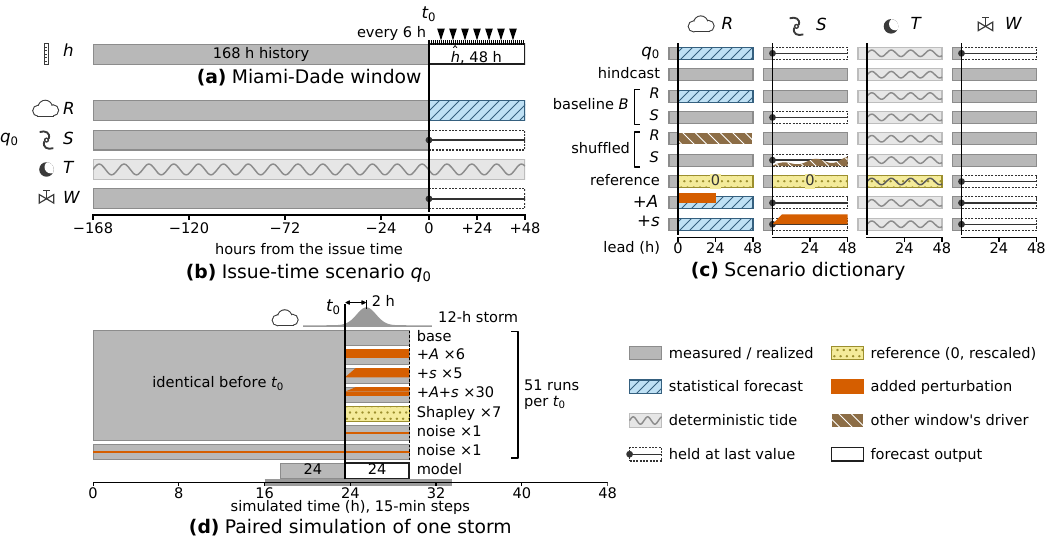}
\caption{Forecast-origin information sets and scenario definitions: (a,b)~Miami-Dade window and $q_0$, (c)~scenario dictionary, and (d)~paired storm simulation.}
\label{fig:s-protocol}
\end{figure*}

The information at the issue time falls into four classes (Fig.~\ref{fig:s-protocol}a,b). (i)~Measured history is the
168~h before $t_0$ of every series, as released by the benchmark, filled values included. No reporting delay is
modeled; every series is available up to the issue time. (ii)~Deterministic drivers consist of the astronomical tide, a harmonic
fit to the record from 2015 to 2019 that is treated as known over the horizon. (iii)~Forecast drivers consist of rain from the ridge
forecaster fitted on the years from 2015 to 2017, issued at the last history hour and spread uniformly over the 48~h. (iv)~Unknown
future drivers are surge and the gate and pump settings, held at their last history values. These four classes define the
issue-time scenario $q_0$. Models are trained with the realized future of rain, surge and operations, the information
of the perfect-driver hindcast, and are scored under $q_0$; the issue-time references are trained and scored with
surge and operations held and without future rain. On the simulator, the history is the 24 steps (6~h) before the issue
time, $q_0$ has no future rain because no rain forecast exists, surge is held at its last value, and the controls are
held at settings equal to the realized ones; training uses the realized drivers with controls held
(Fig.~\ref{fig:s-protocol}d).

In Fig.~\ref{fig:s-protocol}a, triangles mark the next query issue times, every 6~h, and fine ticks mark the hourly
starts of the scored windows. In Fig.~\ref{fig:s-protocol}d, $t_0$ lies 2~h before the applied rain peak, rounded down
to 0.5~h, and the thick gray axis segment shows the range of $t_0$ over the test storms. Every simulated run starts at
$t=0$ and ends at the window end (dashed line), and every run except the whole-storm noise run is identical to the base
run before $t_0$. The model sees the 24 steps before $t_0$ and predicts the next 24.

Every scenario is written into the input batch, and each model's own information-set switches are set to pass every
channel through, so the batch alone defines the information set.

\subsection{Scenario dictionary}\label{app:B2}

Every scenario and query fixes the history and sets the four future drivers, rain $R$, surge $S$, astronomical tide $T$
and operations $W$, over the horizon. The horizon is 48~h in Miami-Dade, where queries are issued every \res{data.mia.query-step-h}~h
and forecasts are scored at every hour, and the 6-h window after each of the \res{data.beta.cf-issue-times} test issue
times on the simulator, with \res{data.beta.cf-runs-per-issue} runs per issue time.
The definitions follow, grouped by class, and templates and timing are detailed in the next subsections.

\par\textbf{Feasible scenarios and queries (Miami-Dade).}
The issue-time scenario $q_0$ takes $R$ from the ridge
forecast, holds $S$ and $W$ at their last values and treats $T$ as known; $q_0$ without rain sets $R=0$. Each query
changes one driver of $q_0$ and reads the margin $m_i$. Added rain $+A\,g$ on \res{data.mia.grid-points} points of
$[0,A_{\max}]$, $A_{\max}=\res{data.mia.amax-mm}$~mm, gives $A^\ast$. Added surge $+s$ on as many points, up to
\res{data.mia.surge-max-cm}~cm and, for cells warned under $q_0$, down to \res{data.mia.surge-min-cm}~cm (the 0.1th and
99.9th percentiles of the training-period 48-h surge change), gives $s^\ast$ and $s^-$. Rain removal scales the forecast
rain of warned cells to zero, and a \res{data.mia.joint-grid-points}\,$\times$\,\res{data.mia.joint-grid-points} joint
grid combines both additions. $A^\ast$ and $s^\ast$ exclude cells already warned under $q_0$.

\par\textbf{Diagnostic reference and observational check.}
The perfect-driver hindcast uses the realized $R$, $S$
and $W$. Two definitions matter for reading the check. The substitution baseline $B$ sets rain to its issue-time
forecast, not to zero, so the rain substitution $F-B$ is realized rain minus forecast rain. For surge it holds the last
value. A third variant substitutes both, and every other driver stays realized. A shuffled control takes the realized
rain of another window between \res{data.mia.shuffle-min-days} and \res{data.mia.shuffle-max-days} days away in the same period
(seed 0), or adds that window's surge change to the current held surge. Both read the forecast peak. The proof that the
check's slope measures residual covariance under this substitution is given in App.~\ref{app:H}.

\par\textbf{Reference scenario.}
The attribution mixes $q_0$ with a reference that sets $R=0$ and $S=0$ and rescales
$T$ to the same-phase tide with the training-median range, over eight coalitions with $W$ held, and returns $\phi_R$,
$\phi_S$, $\phi_T$ and $I_{RS}$.

\par\textbf{Simulator.}
On top of the realized run, added rain from
\res{data.beta.rain-add-min-mm} to \res{data.beta.rain-add-max-mm}~mm and added outlet stage from
\res{data.beta.stage-add-min-cm} to \res{data.beta.stage-add-max-cm}~cm, alone and in 30 pairs, read the window peak and
its crossing of $u_i$; they are feasible but partly beyond the training support. Attribution runs remove $R$ and $S$
after $t_0$ and rescale $T$ about its issue-time value (reference scenario), and noise runs scale rain by $1+10^{-6}$
over the whole storm or after $t_0$ to give the noise floor $\nu$ (numerical check); the controls stay held throughout.

\par\textbf{Stress tests and controls.}
The half- and reversed-input variants feed $q_0+a\,(q_X-q_0)$ with $a=1/2$ or
$a=-1$ and the controls held, and are diagnostic only; the untrained copy of each architecture (initialization seed 0)
runs every scenario as a control.

\subsection{Rainfall families and the minimum crossing rainfall search}\label{app:B3}

\par\textbf{Miami-Dade family.}
The template $g$ is the mean normalized 24-h profile of the 20 largest non-overlapping area-mean 24-h storms of the
training period, in the release's units; its hourly shares lie between \res{data.mia.storm-shape-min-share}\% and
\res{data.mia.storm-shape-max-share}\%, so the shape is nearly uniform. The same depth $A$ (mm) is added at all
\res{data.mia.n-rain-gauges} gauges, converted per gauge with its unit factor, from the issue time over leads 0 to 23~h,
on top of the forecast rain of $q_0$, with surge and operations held. Variants starting 12~h and 24~h later serve only
the timing scans. The range is $[0,A_{\max}]$ with $A_{\max}=\res{data.mia.amax-mm}$~mm, the 99.9th percentile of the
training-period area-mean 24-h depth (training maximum \res{mia.thr.train_max24h_mm}~mm), searched on 41 equally
spaced points (step \res{mia.thr.grid_step_mm}~mm). Rain is only added, never made negative. Adding a template differs
from scaling the rain already in the scenario, which the separate rain-removal query does for cells warned under $q_0$
(forecast rain scaled to zero). Two storms with the same total but different shapes, starts or spatial patterns are
different scenarios, and the search uses this one shape, the same depth at every gauge and amounts up to $A_{\max}$. The same search
with an added surge $s$, ramped linearly over 6~h and then held, gives the minimum crossing surge $s^\ast$ on 41 points
up to \res{mia.thr.smax_cm}~cm, and an 11\,$\times$\,11 joint grid combines both additions.

\par\textbf{Simulator family.}
Rain is added uniformly over the 6-h window at the single rain gauge, and thus at every subcatchment, on top of the
realized drivers, because the simulator has no rain forecast. The grid is
$A\in\{6.35,12.7,25.4,38.1,50.8,76.2\}$~mm, defined in the model file's units as amounts from 0.25 to 3~in. The models receive the same
addition, $A/24$ per 15-min step times the run's rain constant \res{data.beta.rain-constant}.

\par\textbf{Search.}
Algorithm~\ref{alg:s-astar} gives the search exactly as implemented for both systems. All cells of an issue time share
the added amount, so the grid is evaluated by one forward pass per grid point and refined by linear interpolation
between the two grid points that bracket the first crossing (the models are deterministic). Warned, crossed, not
crossed within range, non-monotone and boundary-sensitive cells are counted separately (Table~\ref{tab:s-accounting}
for Miami-Dade and Table~\ref{tab:s-sensitivity} for the simulator). Recomputing Miami-Dade $A^\ast$ on the 11-point rain
axis of the joint grid (step \res{mia.thr.joint_step_mm}~mm) reproduces \res{mia.robust.grid.agree.min}\% of the
crossed and not-crossed calls of the 41-point grid, with a median $|\Delta A^\ast|$ of at most
\res{mia.robust.grid.medabsdiff.max}~mm.

\par\medskip\noindent\rule{\columnwidth}{0.6pt}\par\nopagebreak
\noindent\refstepcounter{appalg}\label{alg:s-astar}\textbf{Algorithm~\theappalg.} The minimum crossing rainfall of one cell
(node $i$, issue time $t_0$) within a scenario family $g$.\par\nopagebreak\vspace{-2pt}
\noindent\rule{\columnwidth}{0.4pt}
\begin{enumerate}
\item \emph{Input.} The grid $0=A^{(0)}<A^{(1)}<\dots<A^{(J)}$ ($J=40$ equal steps to $A_{\max}$ in Miami-Dade, the
six amounts above with $J=6$ on the simulator), the threshold $u_i$, the crossing margin $\varepsilon$ (1~cm Miami-Dade, 0 simulator), the base
scenario $q_b$ ($q_0$ in Miami-Dade, the realized drivers on the simulator) and the peak margin
$m_i(A)=\max_{1\le\ell\le H}\hat h_{i,t_0+\ell}(q_b+A\,g)-u_i$.
\item \emph{Warned at the base.} If $m_i(0)\ge0$, set the status \textsc{warned} and stop. In Miami-Dade the cell
leaves the $A^\ast$ analysis. On the simulator a warned model cell is an initial false alarm, counted as a predicted
crossing. A simulator cell with base peak at or above $u_i$ is not in the common denominator.
\item \emph{Evaluate the grid.} Compute $m_i(A^{(j)})$ for $j=1,\dots,J$.
\item \emph{Monotonicity} (Miami-Dade only). If the margin falls by more than $10^{-3}$~cm between two consecutive grid
points, mark the grid as non-monotone. The simulator side assigns no monotonicity status.
\item \emph{First crossing.} Let $j^\ast$ be the smallest $j$ with $m_i(A^{(j)})\ge\varepsilon$, if any, and
interpolate $A^\ast=A^{(j^\ast-1)}+w\,(A^{(j^\ast)}-A^{(j^\ast-1)})$ with
$w=\big(\varepsilon-m_i(A^{(j^\ast-1)})\big)/\big(m_i(A^{(j^\ast)})-m_i(A^{(j^\ast-1)})\big)$ clipped to $[0,1]$.
\item \emph{Status, in this order of precedence.} \textsc{non-monotone} if step~4 marked the grid; the first observed
crossing of step~5, if one exists, is kept and enters only under the first-observed convention. Otherwise
\textsc{crossed} if $j^\ast$ exists, with $A^\ast$ from step~5. Otherwise \textsc{not crossed within range}, and $A^\ast$ is
censored at $A_{\max}$, neither missing nor equal to $A_{\max}$.
\item \emph{Boundary flags.} In Miami-Dade a cell is flagged if the $q_0$ peak lies within 1~cm below $u_i$, if the crossing falls in the last
grid interval, or if the cell is not crossed with $\varepsilon=1$~cm but would be with $\varepsilon=0$. On the simulator a cell is flagged if a
bracketing grid peak lies within $\nu$ of $u_i$, or, without a crossing, if the highest grid peak lies within $\nu$ below
$u_i$.
\end{enumerate}
\noindent\rule{\columnwidth}{0.6pt}\par\medskip

\subsection{Ocean boundary and operations}\label{app:B4}

\par\textbf{Ocean boundary.}
In Miami-Dade the boundary stage is the astronomical tide plus surge, imposed at the tide node of Virginia Key and
never updated by a model. The astronomical part and the surge residual are separate inputs, so holding or replacing
the surge keeps the astronomical phase. The issue-time scenario holds the surge at its last value, and the added-surge query
adds a ramp on top of it. The attribution reference sets the surge to zero and rescales the astronomical tide about its
window mean to the same-phase tide with the training-median 48-h range (\res{data.mia.attrib-tide-range-lwp-cm}~cm at
Lake Worth Pier, \res{data.mia.attrib-tide-range-vk-cm}~cm at Virginia Key). On the simulator the outlet stage is a
generated astronomical cycle plus a random surge pulse (App.~\ref{app:D1}). An added outlet stage $s$ is
ramped over the first hour after the issue time, held through the window and ramped down after it, with
$s\in\{7.62,15.24,30.48,45.72,60.96\}$~cm (from 0.25 to 2~ft). Behind the flap gate a raised stage blocks outflow and lets no
water flow back in, so the ocean acts only by obstructing drainage~\cite{rossman2022swmm}.

\par\textbf{Operations.}
Gate openings and pump series enter as fractions of their training-period maxima, clipped at zero but not capped
above.
In Miami-Dade they are realized in training and in the perfect-driver hindcast and held at their last history value
under $q_0$ and in every attribution coalition, representing a fixed trajectory rather than a feedback policy.
In the first set of runs on the simulated coastal network, the orifice R2 and the weir W0 stay at setting 1.0 and the pump
P0 is held on in every run, while its flow still follows its pump curve; the models receive these settings as held
controls, which equal the realized ones. Every response and interaction of that set is therefore conditional on one
fixed control path, and operation responses come from other runs.
Gate, weir and pump operations on the same network are evaluated in a second set of runs with random schedules for R2,
W0 and P0 (App.~\ref{app:P}), and valve settings on the second simulated network (App.~\ref{app:L6}). \smbadd{Gate and
pump settings that follow a level rule, and so depend on the water state, are evaluated on the synthetic barrage testbed
of App.~\ref{app:Lsmb}.}

\subsection{Attribution coalitions and common history}\label{app:B5}

The players are rain $R$, surge $S$ and astronomical tide $T$, $\mathcal{P}=\{R,S,T\}$. In Miami-Dade, $R$ is the
issue-time rain forecast, $S$ is the issue-time surge held over the 48~h, and $T$ is the astronomical tide; the
reference sets $R=0$ and $S=0$ and rescales $T$ as above. $\phi_S$ therefore measures the current surge level against no
surge (negative when the held surge is a set-down), and $\phi_T$ measures the window's tidal-range anomaly against the
training-median range, not the tide itself. Operations and the history stay fixed in all eight coalitions. The value
$v(\mathcal{C})$ is the 48-h forecast peak margin at each canal gauge, at the \res{mia.attr.n-issue} issue times of each
year, and the exact Shapley value of player $D\in\mathcal P$ is
\begin{equation}
\phi_D=\sum_{\mathcal{C}\subseteq\mathcal{P}\setminus\{D\}}\frac{|\mathcal{C}|!\,(2-|\mathcal{C}|)!}{3!}
\big[v(\mathcal{C}\cup\{D\})-v(\mathcal{C})\big].
\label{eq:app-shapley}
\end{equation}
The three values sum to $v(\mathcal{P})-v(\emptyset)$ within \res{mia.attr.eff-resid.max}~cm. The reference output
$v(\emptyset)$ contains the history, drainage, the held operations and the reference tide, so the antecedent state
enters through $v(\emptyset)$ rather than as a fourth player. On the simulator, the reference removes the future rain and
the future surge after the issue time and rescales the astronomical part of the outlet stage about its issue-time value
to the training-median 6-h range (\res{data.beta.astro-range-median-cm}~cm); the realized base run is the full
coalition, and seven further runs per issue time complete the eight coalitions. All paired runs of an issue time share
the pre-issue history, the 15-min time grid and the target, the 6-h window peak of node depth.

\subsection{Training support and stress tests}\label{app:B6}

\par\textbf{Support.}
The Miami-Dade added-rain and added-surge ranges come from training-period percentiles, so each searched amount stays
within the training distribution of area-mean 24-h depth or of 48-h surge change; this bounds the added amount, not
its total with the forecast rain or the held surge. On the simulator, a support
analysis of the training windows places the 99th percentile of the 6-h outlet-stage rise at
\res{beta.s4.support.v5.surge_rise.p99}~cm and its maximum at \res{beta.s4.support.v5.surge_rise.max}~cm. The 6-h stage
rise of the test storms with $\bar s$ added, the realized rise plus the ramped addition, exceeds that percentile in
\res{beta.s4.support.test.stage30.above_p99}\% of the test storms and the training maximum in
\res{beta.s4.support.test.stage30.above_max}\% of them; with the largest added stage it exceeds the maximum in
\res{beta.s4.support.test.stage61.above_max}\%. The 6-h rain totals of training windows reach a 99th percentile of
\res{beta.s4.support.v5.rain_total.p99}~mm and a maximum of \res{beta.s4.support.v5.rain_total.max}~mm; realized rain
plus the largest added amount exceeds them in \res{beta.s4.support.test.rain76.above_p99}\% and
\res{beta.s4.support.test.rain76.above_max}\% of the test storms. Amounts beyond the training support are reported
with this label, not removed.

\par\textbf{Stress tests.}
The half-input and reversed-input variants feed the trained model $q_0+a\,(q_X-q_0)$ in place of every scenario $q_X$,
the base included, with $a=1/2$ or $a=-1$ applied to the rain and surge inputs, while the controls stay at their $q_0$
values\smbadd{ except on the synthetic barrage testbed}. On the main simulator $q_0$ is no future rain, surge held and controls held; on the second simulator it is no
rain and valves held\smbadd{; on the synthetic barrage testbed it is no rain with surge, gates and pumps held, and both
variants also scale the gate and pump inputs}. The reversed input feeds negative rain and is physically infeasible. Both variants are diagnostic
constructions that probe what the observational check and the threshold metrics do with a known input distortion; their
response ratios are measured, not implied by $a$, and they are never pooled with feasible scenarios in any application
metric. The untrained copy runs the same scenarios with the initial weights of each architecture (App.~\ref{app:C}).

\section{Model Implementations and Training Protocol}\label{app:C}
\ifcsname c@appalg\endcsname\else\newcounter{appalg}\renewcommand{\theappalg}{S\arabic{appalg}}\fi

\begin{figure*}[tp]
\centering
\includegraphics[width=\textwidth]{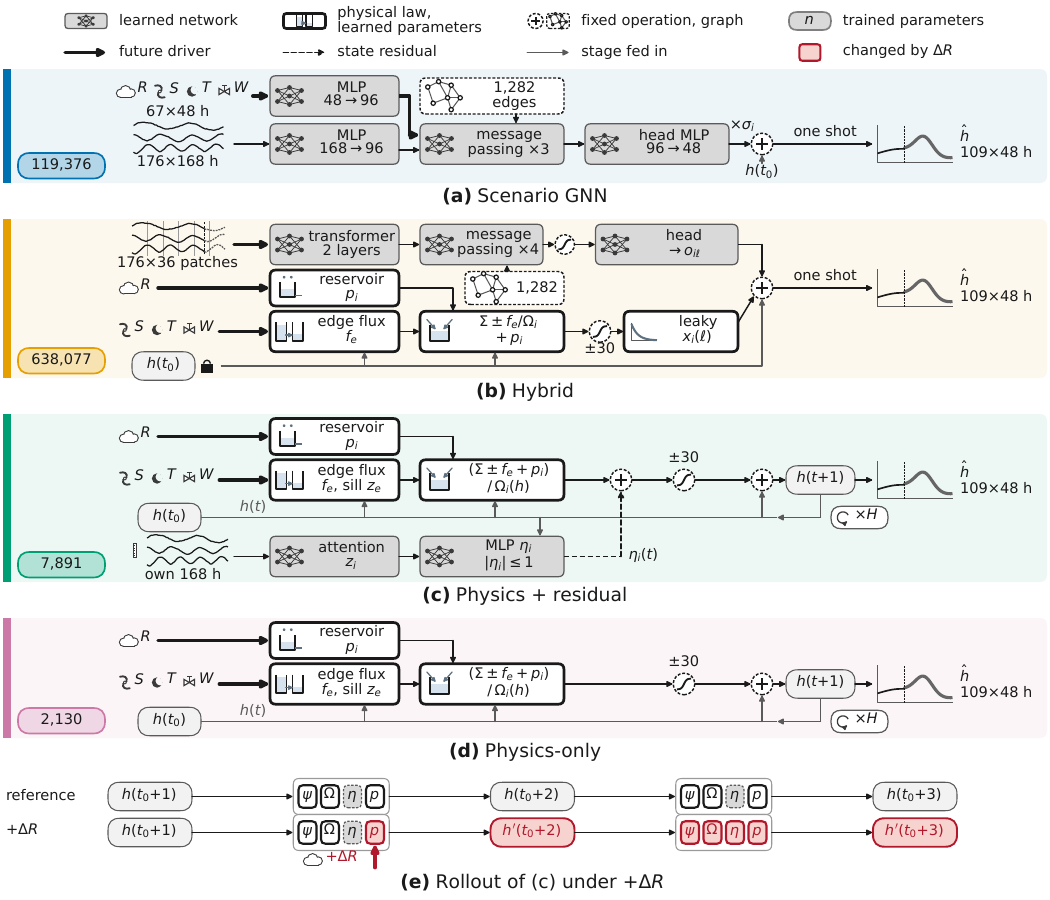}
\caption{Architecture diagrams of the four forecasting models (badges show Miami-Dade parameter counts; locks mark heads frozen at $h(t_0)$; rate bound $\pm30$~cm\,h$^{-1}$).}
\label{fig:s-architectures}
\end{figure*}

This appendix gives the equations, procedures and hyperparameters of the four structures of the main text
(Fig.~\ref{fig:overview}b) as implemented. Fig.~\ref{fig:s-architectures} draws the computational graphs, the text
below states where drivers enter, the constraints and the parameter counts, and Table~\ref{tab:s-training} lists the
training runs. Notation follows Eq.~\eqref{eq:balance}. Edge $e$ runs from node
$e^-$ to node $e^+$, $c_e(t)$ is the gate opening or pump setting acting on it, and $\ell=1,\dots,H$ is the lead.
The multiplier $c_e$ of Eq.~\eqref{eq:balance} is the opening on gate edges, $\operatorname{sigm}(\gamma_e)$ on
uncontrolled structures and 1 on canal, groundwater and tide edges; pump edges replace the flux law by
$C_e\,c_e(t)\,d_e$ (Eq.~\eqref{eq:app-flux}).

In Fig.~\ref{fig:s-architectures}, the future drivers are rain $R$, surge $S$, astronomical tide $T$ and gate and pump
settings $W$. The Scenario GNN encodes the 168-h past of all \res{data.mia.n-series-nodes} series and the 48-h future
of the 67 driver series of Miami-Dade with per-series perceptrons. In panel e, rain added at $t_0{+}1$ first changes the
reservoir outflow $p$, and from $t_0{+}2$ on the changed stage $h'$ alters $\psi$, $\Omega$ and $\eta$ of every later
step.

\subsection{Shared prediction interface}\label{app:C1}

All four structures read the same batch and return the same output, so they differ only inside $f_\theta$ of
Eq.~\eqref{eq:forecast}. For an issue time $t_0$ the batch holds the stage history of the state nodes
(\res{data.mia.n-state-nodes} in Miami-Dade, canal gauges first and wells after, and \res{data.beta.n-state-nodes} depth
nodes on the simulator) over 168~h (24 steps on the simulator) with its observed-label mask, the past and future of the
\res{data.mia.n-controls} control series, of the \res{data.mia.n-rain-gauges} rain gauges and of the astronomical tide
and surge at both tide gauges, and the hour of day and day of year. Units are converted once (App.~\ref{app:A}).
Standardization uses training-period statistics of the released series, namely stage per node (mean $\mu_i$, standard
deviation $\sigma_i$), controls per series, rain divided by its gauge's training p99, and the total tide.
Information-set switches act on the batch. They set the controls to realized, held or removed, the surge to realized, held or zero, and the future
rain to on or off. For scenario evaluation every switch passes the batch through (App.~\ref{app:B}). The output is
$\hat h_{i,t_0+\ell}$ in cm for every state node, wells included; only observed labels are scored. On the simulator one
model time step is one 15-min interval, so every time constant and rate bound below is per step there.

\subsection{The four structures}\label{app:C2}
\begin{table*}[t]
\centering
\caption{Training runs per model and system, mean\,$\pm$\,SD over seeds 42, 43 and 44. Best epochs are 0-indexed, and
wall-clock times are measured on shared GPUs (App.~\ref{app:M}).}
\label{tab:s-training}
\begin{tabular*}{\textwidth}{@{\extracolsep{\fill}}@{}lcccccccc@{}}
\toprule
 & \multicolumn{2}{c}{Learning rate} & \multicolumn{3}{c}{Miami-Dade} & \multicolumn{3}{c}{Simulator} \\
\cmidrule(lr){2-3}\cmidrule(lr){4-6}\cmidrule(lr){7-9}
Model & Network & Physics & Epochs run & Best epoch & Time (min) & Epochs run & Best epoch & Time (min) \\
\midrule
Scenario GNN & $5\times10^{-4}$ & -- & 30.0\,$\pm$\,0.0 & 26.7\,$\pm$\,1.2 & 2.3\,$\pm$\,0.1 & 30.0\,$\pm$\,0.0 & 29.0\,$\pm$\,0.0 & 13.4\,$\pm$\,3.7 \\
Hybrid & $5\times10^{-4}$ & $10^{-2}$ & 28.7\,$\pm$\,1.2 & 20.0\,$\pm$\,1.7 & 25.0\,$\pm$\,1.4 & 30.0\,$\pm$\,0.0 & 28.3\,$\pm$\,0.6 & 50.6\,$\pm$\,12.3 \\
Physics + residual & $3\times10^{-3}$ & $10^{-2}$ & 30.0\,$\pm$\,0.0 & 28.7\,$\pm$\,0.6 & 29.7\,$\pm$\,0.1 & 13.3\,$\pm$\,1.5 & 4.3\,$\pm$\,1.5 & 31.1\,$\pm$\,3.5 \\
Physics-only & -- & $10^{-2}$ & 30.0\,$\pm$\,0.0 & 27.0\,$\pm$\,3.5 & 23.8\,$\pm$\,0.6 & 23.3\,$\pm$\,6.1 & 17.0\,$\pm$\,10.6 & 43.8\,$\pm$\,8.7 \\
\bottomrule
\end{tabular*}
\end{table*}

\par\textbf{Scenario GNN.}
Every series is a graph node, namely the state nodes, the controls, the rain gauges and the tide gauges, which makes
\res{data.mia.n-series-nodes} nodes in Miami-Dade, each with an embedding $\boldsymbol\omega$ of its type. Edges join,
in both directions, the two ends of every flux-graph edge, every control series and both ends of its structure edge,
and every state node and the (up to three) rain gauges of its rain mix, which gives \res{model.gnn.n-mp-edges} directed edges whose
kinds are not distinguished. The Lake Worth Pier node has no edge. With $\mathbf x_n^{\mathrm{past}}$ and
$\mathbf x_n^{\mathrm{fut}}$ the standardized past and future of series $n$ (the future is zero at state nodes),
\begin{align}
\mathbf e_n^{(0)}&=\mathbf W_0\big[\mathcal N_{\mathrm p}(\mathbf x_n^{\mathrm{past}})+\mathcal N_{\mathrm f}(\mathbf x_n^{\mathrm{fut}})+\boldsymbol\omega_{\mathrm{type}(n)}\big],\notag\\
\mathbf e_n^{(l+1)}&=\mathrm{LN}\big(\mathbf e_n^{(l)}+\mathrm{ReLU}\big(\mathbf W_{l+1}[\mathbf e_n^{(l)};\bar{\mathbf e}_n^{(l)}]\big)\big),\notag\\
\hat h_{i,t_0+\ell}&=h_i(t_0)+\sigma_i\,\big[\mathcal N_{\mathrm h}(\mathbf e_i^{(3)})\big]_\ell ,
\label{eq:app-gnn}
\end{align}
where $\bar{\mathbf e}_n^{(l)}$ is the mean of $\mathbf e_m^{(l)}$ over the neighbors $m$ of $n$ in layers $l=0,1,2$,
$h_i(t_0)$ is the last history value, LN is layer normalization, the width is 96, and $\mathcal N_{\mathrm p}$,
$\mathcal N_{\mathrm f}$ and $\mathcal N_{\mathrm h}$ are two-layer perceptrons that map 168 values to 96, 48 values to 96, and 96 values to 48 offsets.
The mean aggregation follows GraphSAGE~\cite{hamilton2017inductive}. Future drivers enter only at their own nodes; no
physical quantity is learned, and nothing is constrained beyond the anchor at the last observation.

\par\textbf{Hybrid forecaster.}
Each of the \res{data.mia.n-series-nodes} series is cut into 6-h patches over its 168-h past and 48-h future, with a flag
for known values, which gives 36 tokens per series. Patches are embedded linearly (width 128) with position and type
embeddings, passed through a two-layer, four-head transformer~\cite{vaswani2017attention} (feed-forward width 512, no dropout)
shared across series, and mean-pooled; four message-passing layers of the form of Eq.~\eqref{eq:app-gnn} on the same edges and a tanh follow. A
direct head (128 to 128 to 48, output layer initialized at zero) returns unbounded level offsets $o_{i\ell}$ in cm. The
physics branch is open loop. At every lead it evaluates the edge fluxes of Eq.~\eqref{eq:app-flux} with all heads
frozen at the last observation $h(t_0)$ and with the scenario's tide, controls and rain at that lead, and integrates the
resulting rate with a node-local leaky filter,
\begin{align}
r_i(\ell)&=\Delta h_{\max}\tanh\!\Big(\frac{1}{\Delta h_{\max}}\Big[\frac{\sum_{e^+=i}f_e(\ell)-\sum_{e^-=i}f_e(\ell)}{\Omega_i\big(h_i(t_0)\big)}+p_i(\ell)\Big]\Big),\notag\\
x_i(\ell)&=e^{-1/\vartheta_i}\,x_i(\ell-1)+r_i(\ell),\qquad x_i(0)=0,\notag\\
\hat h_{i,t_0+\ell}&=h_i(t_0)+o_{i\ell}+x_i(\ell),
\label{eq:app-hybrid}
\end{align}
with time constant $\vartheta_i=1+\operatorname{softplus}(\theta_i)$~h, initialized at 6~h, and the rain source $p_i$ of
Eq.~\eqref{eq:app-reservoir}. Unlike the rollout below, the rain source is added after the division by the storage
area, the downstream head carries no sill, and the flux exponent lies in $(0.5,1.5)$. The
\res{data.mia.n-pump-dir-labelled} pump edges whose direction follows from the station naming have $d_e=+1$; the other
pump edges learn $d_e$, starting at zero. Nothing predicted re-enters the fluxes, and there is no rollout.

\par\textbf{Physics + state residual and Physics-only model.}
Both roll the water balance hourly over the horizon from the last observation; the Physics-only model has no residual.
The edge flux depends on the edge kind,
\begin{equation}
f_e(t)=\begin{cases}
K_e\,\psi_e\big(\Delta_e(t)\big), & e\in\mathcal E_{\mathrm f},\\
c_e(t)\,K_e\,\psi_e\big(\Delta_e(t)\big), & e\in\mathcal E_{\mathrm g},\\
\operatorname{sigm}(\gamma_e)\,K_e\,\psi_e\big(\Delta_e(t)\big), & e\in\mathcal E_{\mathrm u},\\
C_e\,c_e(t)\,d_e, & e\in\mathcal E_{\mathrm p},
\end{cases}
\label{eq:app-flux}
\end{equation}
where $\mathcal E_{\mathrm f}$ holds the canal, groundwater and tide edges, $\mathcal E_{\mathrm g}$ the gate edges,
$\mathcal E_{\mathrm u}$ the structures without a control series and $\mathcal E_{\mathrm p}$ the pump edges. Every
flux is then bounded, $f_e\leftarrow f_{\max}\tanh(f_e/f_{\max})$. Here $\Delta_e(t)=h_{e^-}(t)-\tilde h_{e^+}(t)+b_e$ and
$\psi_e(x)=\operatorname{sign}(x)\big[(|x|+\zeta)^{\alpha_e}-\zeta^{\alpha_e}\big]$, which is odd and monotone. On gate
and uncontrolled-structure edges the downstream stage is replaced by the learned sill,
$\tilde h_{e^+}=\operatorname{relu}(h_{e^+}-z_e)$, or by a measured sill drop where one is known (only the simulator
orifices R0 and R1); elsewhere $\tilde h_{e^+}=h_{e^+}$. Tide nodes $o$ impose $h_o(t)=T_o(t)+S_o(t)$ plus the datum shift
of \res{data.mia.tide-datum-shift-cm}~cm and are never updated. Rain enters through a linear reservoir per node,
\begin{equation}
V_i(t)=\big(1-\tau_i^{-1}\big)\,V_i(t-1)+a_i\sum_r w_{ir}R_r(t),\qquad p_i(t)=V_i(t)/\tau_i ,
\label{eq:app-reservoir}
\end{equation}
which is warmed up on the 168-h rain history and then driven by the scenario's future rain. The fixed weights $w_{ir}$
are inverse-distance weights of the three nearest gauges within 40~km (rows sum to one); $R_r$ is in the release's
native units, so the learned gain $a_i$ absorbs the unit (on the simulator every runoff-receiving node has weight one).
The stage update is Eq.~\eqref{eq:balance} with its soft bounds written out,
\begin{align}
\Delta h_i^{s}(t)&=\frac{\sum_{e^+=i}f_e(t)-\sum_{e^-=i}f_e(t)+p_i(t)}{\Omega_i\big(h_i(t)\big)},\notag\\
h_i(t{+}1)&=h_i(t)+\Delta h_{\max}\tanh\!\Big(\frac{\Delta h_i^{s}(t)+\eta_i(t)}{\Delta h_{\max}}\Big),
\label{eq:app-step}
\end{align}
with storage area $\Omega_i(h)=\operatorname{softplus}\big(\lambda_i+\kappa_i(h-\mu_i)/\sigma_i\big)+10^{-3}$. The state
residual of the Physics + state residual model is
$\eta_i(t)=\eta_{\max}\tanh\!\big(\mathcal N_\eta[\mathbf z_i;\,(h_i(t)-\mu_i)/\sigma_i]\big)$, a two-layer
perceptron of hidden width 32, where $\mathbf z_i$ is a 16-dimensional summary of node $i$'s own standardized 168-h
stage history, computed by a single-head attention over the 168 past hours, with a learned query per node, followed by a linear
map and a tanh. Its rain-mix and calendar inputs are set to zero, so no driver enters it directly; the Physics-only model has
$\eta_i\equiv0$.

\par\textbf{Constants, constraints and initial values.}
The fixed constants are $\zeta=0.1$~cm, $f_{\max}=20$, $\Delta h_{\max}=30$~cm\,h$^{-1}$ and
$\eta_{\max}=1$~cm\,h$^{-1}$. Constraints are imposed by parameterization, so that $K_e$, $C_e$ and $a_i$ are exponentials of
free parameters, $\tau_i=\max(e^{\theta},2\text{ h})$, $\Omega_i>0$, $z_e\ge0$, $\operatorname{sigm}(\gamma_e)\in(0,1)$,
$d_e=\tanh(\theta)\in(-1,1)$, $z_e=\operatorname{softplus}(\theta)$, $\vartheta_i>1$~h, and a sigmoid maps a free
parameter of $\alpha_e$ into $(0.25,1.5)$ for the physics models and $(0.5,1.5)$ for the Hybrid forecaster. Learned are $K_e$, $b_e$, $\alpha_e$, $C_e$, $d_e$, $\gamma_e$, $z_e$,
$\lambda_i$, $\kappa_i$, $a_i$, $\tau_i$, $\vartheta_i$ and the neural parts. Fixed are the edge set and its
orientation, the kind of every edge and the control series it reads, the rain weights, the tide-edge assignment, the
datum shift, the four constants above, the exponent ranges and the pinned pump directions of the Hybrid forecaster.
Initial
values are $K_e=0.05$ on canal, uncontrolled-structure and tide edges, $0.01$ on gate edges and $0.005$ on
groundwater edges; $b_e$ equal to minus the training-mean head difference on canal, groundwater and tide edges, so
that these edges carry almost no flux at mean conditions, and zero on structure edges; $C_e=3$, $\alpha_e=1$, $a_i=5$,
$\tau_i=24$~h, $\lambda_i=\kappa_i=0$ (so $\Omega_i=\ln 2+10^{-3}$), $z_e=10^{-3}$~cm and $\vartheta_i=6$~h. Neural
weights use the PyTorch default initializers, except small random position, type, node and lag embeddings and the
zero-initialized output layer of the Hybrid forecaster's head. In the physics models all
pump directions are learned; the labeled ones start at $\tanh 2$, the others at zero. The trainable parameters number
\res{model.gnn.n-params} for the Scenario GNN, \res{model.hybrid.n-params} for the Hybrid forecaster,
\res{model.physres.n-params} for the Physics + state residual model and \res{model.physonly.n-params} for the
Physics-only model in Miami-Dade, and \res{model.gnn.n-params-beta}, \res{model.hybrid.n-params-beta},
\res{model.physres.n-params-beta} and \res{model.physonly.n-params-beta} on the simulator. Persistence has none.

\par\textbf{Simulator graph.}
On the simulator the state is node depth, the tide node is the outfall OUT0 with the outlet stage, and the
\res{data.beta.n-edges} edges are \res{data.beta.n-conduits} conduits, one tide edge, two gate edges (R2, W0), two
uncontrolled structures (R0, R1) and one pump (P0) of known direction. The outfall conduit is a two-way tide edge. The
models' graph does not represent the flap gate. Simulator rain is multiplied by \res{data.beta.rain-constant} so that
its training p99 per step matches the median training p99 of the Miami-Dade gauges.

\subsection{Loss, optimization and model selection}\label{app:C4}

Every model minimizes
\begin{equation}
\mathcal L=\sum_{G}\frac{1}{|\mathcal O_G|}\sum_{(b,i,\ell)\in\mathcal O_G}
\mathrm{SL}_1\!\Big(\frac{\hat h_{i,t_0+\ell}-h_{i,t_0+\ell}}{\sigma_i}\Big),
\label{eq:app-loss}
\end{equation}
where $\mathcal O_G$ holds the observed cells (window $b$, node $i$, lead $\ell$) of node group $G$ in the batch, the
groups are the canal gauges and the wells in Miami-Dade and all depth nodes as one group on the simulator, and
$\mathrm{SL}_1$ is the smooth-L1 (Huber-type) loss. The
physics models add the same loss under one-step teacher forcing with weight one and use scheduled teacher forcing whose
probability falls from one to zero over the first five epochs. The optimizer is AdamW with weight decay $10^{-6}$. The
learning rate is $5\times10^{-4}$ for the Scenario GNN and the Hybrid forecaster's encoder and head, $3\times10^{-3}$ for the
encoder and residual of the Physics + state residual model, and $10^{-2}$ for physical parameters, annealed by a
cosine schedule to 5\% of its initial value over 30 epochs. Gradients are clipped at norm one and batches hold 64
windows. Training uses the realized future drivers (on the
simulator with controls held). The checkpoint with the lowest mean validation loss on every third validation window
(\res{data.mia.windows-val-select} windows of 2018) is kept; training stops after 8 epochs without improvement once 5
epochs have run, and after 30 epochs at most. Seeds are 42, 43 and 44. Epochs run, best epochs and measured wall-clock
times, as mean\,$\pm$\,SD over the three runs, are in Table~\ref{tab:s-training}. Times come from each finished run's
own time stamps on shared GPUs.
The simulator runs use the same loop and flags with 24-step windows and held controls. On the simulator, the runs of
the Physics + state residual model stopped early, with epoch-mean pre-clip gradient norms up to $8.1\times10^{7}$.

\subsection{Matched comparisons and tuning}\label{app:C5}

Each scenario-input run uses the exact same architecture and hyperparameters as its issue-time reference, differing only in the training information set (future rainfall enabled and surge and controls set to their realized trajectories during training). Hyperparameters were selected on the 2018 validation split prior to scenario evaluation. Across all four architectures, comparisons match the information set, sliding windows, observed-label mask, loss function, optimizer schedule, 30-epoch budget, validation selection rule, and random seeds, while total trainable parameter counts span more than two orders of magnitude from the compact water-balance parameterization to the neural graph architectures. Persistence uses no future drivers, scenario models access future drivers strictly through the queried scenario $q$, and diagnostic perfect-driver hindcasts are reported separately from operational $q_0$ forecasts in all tables. The three sequence baselines evaluated in the multi-region checks are detailed in App.~\ref{app:L}.

\section{Simulator Verification and Numerical Uncertainty}\label{app:D}

The paired-simulator evaluation compares model responses with simulated responses, so the simulated responses must
contain the specified driver change and nothing else. This appendix documents the configuration of the simulated
drainage network, the checks that every paired run changes only its target driver after the issue time, the numerical
noise floor, the continuity errors, the solver-quality rule with its sensitivity, and the floors used for joint effects
and threshold decisions.

\subsection{Configuration and scope}\label{app:D1}

\par\textbf{Engine and numerical settings.}
The pystorms \emph{beta} network~\cite{rimer2023pystorms} runs in SWMM 5.2.4 through pyswmm
2.1.0~\cite{mcdonnell2020pyswmm} with dynamic-wave routing~\cite{rossman2017swmm}. Its input file uses US customary
units, converted here once (1~in${}=25.4$~mm, 1~ft${}=30.48$~cm). Water above a node's rim is lost, because the network
has no surface ponding. The configuration fixes the routing step at 1~s instead of the default variable step of up to
10~s and sets inertial damping to FULL instead of PARTIAL~\cite{rossman2017swmm}; the engine advances in 1-min strides
and the series are sampled every 15~min, reducing numerical noise between paired runs. The pump P0 is held on instead of
switching at the start and stop depths of 30.48 and 15.24~cm (1.0 and 0.5~ft) in the model file, and its flow follows
its pump curve. The outfall OUT0 has a prescribed stage behind a flap gate. A
raised stage blocks outflow, and no water flows back into the network~\cite{rossman2022swmm}.

\par\textbf{Storms, runs and scored nodes.}
Each of the \res{data.beta.n-storms} storms of 48~h is drawn from its own seeded generator, with R2 and W0 fully open;
ranges are drawn uniformly unless stated. With probability \res{beta.gen.antecedent-prob} a storm has antecedent rain
of up to \res{beta.gen.antecedent-max-mm}~mm, spread evenly over the first \res{beta.gen.antecedent-window-h}~h. The
main storm has a log-uniform depth between \res{beta.gen.main-depth-min-mm} and \res{beta.gen.main-depth-max-mm}~mm, the fixed
2-year 12-h pattern of the model file and a start between \res{beta.gen.main-start-min-h} and \res{beta.gen.main-start-max-h}~h.
The outlet stage is a harmonic tide with random phase, its range scaled by a factor between
\res{beta.gen.tide-range-scale-min} and \res{beta.gen.tide-range-scale-max} and offset by between
0 and \res{beta.gen.tide-offset-max-cm}~cm, plus, with probability \res{beta.gen.surge-prob}, a Gaussian surge pulse of
up to \res{beta.gen.surge-amp-max-cm}~cm centered between \res{beta.gen.surge-centre-min-h} and \res{beta.gen.surge-centre-max-h}~h
with a standard deviation between \res{beta.gen.surge-sigma-min-h} and \res{beta.gen.surge-sigma-max-h}~h. Every run starts at $t=0$ of its storm from
the network's initial state, without a hot start. Each test storm has one counterfactual issue time, 2~h before the
applied rain peak and rounded down to 0.5~h, which places the issue times
between \res{data.beta.issue-time-min-h} and \res{data.beta.issue-time-max-h}~h into the run; the window covers the next 24
steps. Of the \res{data.beta.n-state-nodes} depth nodes,
\res{data.beta.n-scored} are scored. J56, the pump's wet well, and J57, its neighbor, oscillate numerically and are
excluded from every scored set. \res{data.beta.n-threshold-nodes} nodes carry a high-water threshold; J65 is also
excluded there, because its training p95 lies within 5~cm of its rim, where a crossing would be indistinguishable
from capping at the rim. The scored node sets are the \res{beta.nodes.rain} runoff-receiving nodes and the
\res{beta.nodes.near} near-outlet nodes within 10 links of OUT0. The network is an independent controlled system, and
the evaluation uses one window length, one issue time per storm and one fixed control path.

\begin{table*}[t]
\caption{Paired runs by run family on the simulated coastal network across 150 issue times (usable runs pass the 5\% primary or 2\% continuity-error threshold).}
\label{tab:s-simquality}
\begin{tabular*}{\textwidth}{@{\extracolsep{\fill}}ll*{13}{c}@{}}
\toprule
 & & Runs/ & & SWMM & & Pre-issue & \multicolumn{3}{c}{Continuity error (\%)} & \multicolumn{3}{c}{Runs with $|e|>$} & \multicolumn{2}{c}{Usable runs} \\
\cmidrule(lr){8-10}\cmidrule(lr){11-13}\cmidrule(l){14-15}
\multicolumn{2}{@{}l}{Run family} & issue & Runs & errors & P0$\ne$1 & diff.\ (cm) & min & median & max & 1\% & 2\% & 5\% & 5\% & 2\% \\
\midrule
\multicolumn{2}{@{}l}{Base run} & 1 & 150 & 0 & 0 & 0.0 & $-$0.20 & +0.43 & +1.40 & 14 & 0 & 0 & 150 & 150 \\
\midrule
\multirow{2}{*}{Noise run} & whole storm & 1 & 150 & 0 & 0 & -- & $-$0.18 & +0.42 & +1.40 & 12 & 0 & 0 & 150 & 150 \\
 & after issue time & 1 & 150 & 0 & 0 & 0.0 & $-$0.04 & +0.41 & +1.40 & 15 & 0 & 0 & 150 & 150 \\
\midrule
\multicolumn{2}{@{}l}{Rain grid} & 6 & 900 & 0 & 0 & 0.0 & $-$1.09 & +0.32 & +1.82 & 40 & 0 & 0 & 900 & 900 \\
\multicolumn{2}{@{}l}{Stage grid} & 5 & 750 & 0 & 0 & 0.0 & $-$0.39 & +0.44 & +1.43 & 78 & 0 & 0 & 750 & 750 \\
\multicolumn{2}{@{}l}{Rain $\times$ stage grid} & 30 & 4{,}500 & 0 & 0 & 0.0 & $-$2.24 & +0.35 & +2.09 & 354 & 6 & 0 & 4{,}500 & 4{,}494 \\
\midrule
\multirow{2}{*}{Shapley} & future rain kept & 3 & 450 & 0 & 0 & 0.0 & $-$0.24 & +0.41 & +1.56 & 47 & 0 & 0 & 450 & 450 \\
 & no future rain & 4 & 600 & 0 & 0 & 0.0 & $-$25.66 & +0.30 & +9.17 & 222 & 84 & 38 & 562 & 516 \\
\bottomrule
\end{tabular*}
\end{table*}

\begin{table*}[t]
\caption{Numerical noise floor $\nu$ per (issue time, node) cell (cm) on the simulated coastal network across 150 issue times ($\le$1\,cm and $=$0 columns in \%).}
\label{tab:s-noisefloor}
\begin{tabular*}{\textwidth}{@{\extracolsep{\fill}}l*{13}{r}@{}}
\toprule
 & & \multicolumn{6}{c}{Whole-storm noise run} & \multicolumn{6}{c}{After-issue noise run} \\
\cmidrule(lr){3-8}\cmidrule(l){9-14}
Node set & Cells & Median & P90 & P99 & Max & $\le$1\,cm & $=$0 & Median & P90 & P99 & Max & $\le$1\,cm & $=$0 \\
\midrule
All scored (207) & 31{,}050 & 0.023 & 1.30 & 8.3 & 65.9 & 86.4 & 4.2 & 0.016 & 1.29 & 8.8 & 69.7 & 86.5 & 4.2 \\
pystorms state nodes (6) & 900 & 0.064 & 1.31 & 6.7 & 10.7 & 84.6 & 0.0 & 0.061 & 1.27 & 6.4 & 10.7 & 85.0 & 0.0 \\
Near-outlet (24) & 3{,}600 & 0.160 & 1.36 & 3.2 & 8.5 & 83.1 & 2.8 & 0.163 & 1.40 & 3.7 & 6.8 & 83.1 & 2.8 \\
Runoff-receiving (163) & 24{,}450 & 0.022 & 1.30 & 10.1 & 65.9 & 86.2 & 0.0 & 0.016 & 1.29 & 10.2 & 69.7 & 86.3 & 0.0 \\
\midrule
J74 & 150 & 10.510 & 55.51 & 63.9 & 65.9 & 44.0 & 0.0 & 12.195 & 58.96 & 67.1 & 69.7 & 44.7 & 0.0 \\
J76 & 150 & 5.250 & 28.35 & 36.4 & 37.7 & 47.3 & 0.0 & 5.524 & 29.05 & 35.2 & 36.3 & 48.7 & 0.0 \\
\midrule
Scored without J74, J76 (205) & 30{,}750 & 0.022 & 1.26 & 6.2 & 39.8 & 86.8 & 4.2 & 0.016 & 1.26 & 6.1 & 57.8 & 86.9 & 4.2 \\
\bottomrule
\end{tabular*}
\end{table*}

\subsection{Paired-run verification}\label{app:D2}

\begin{figure*}[tp]
\centering
\includegraphics[width=\textwidth]{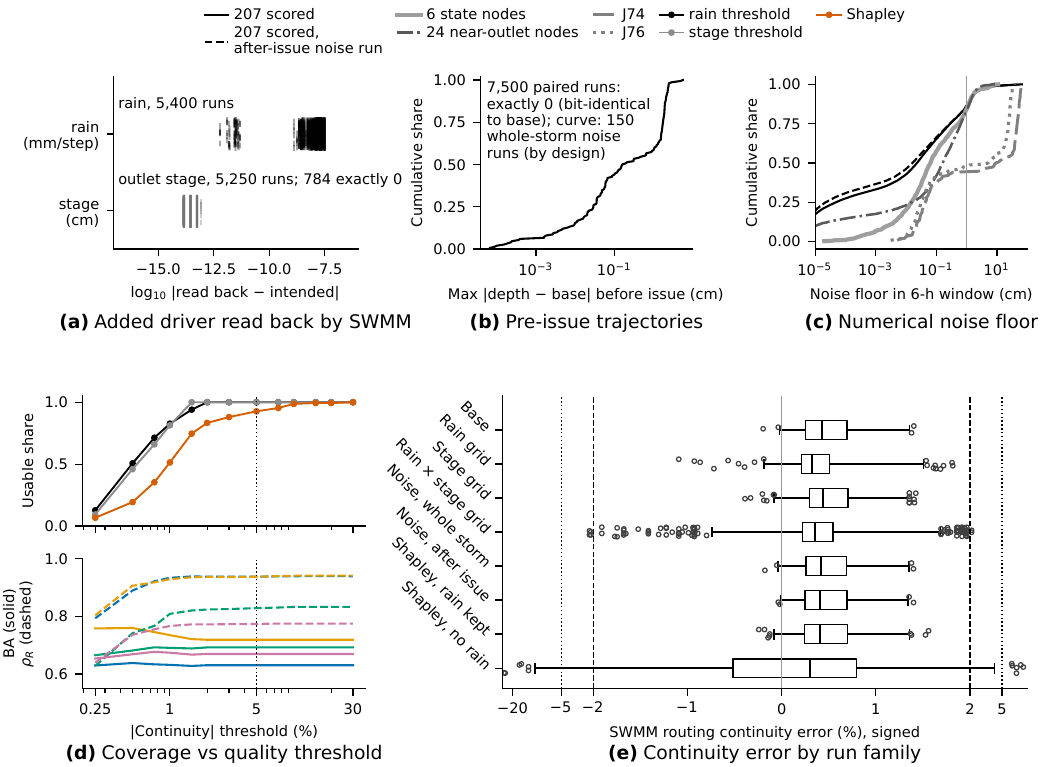}
\caption{Paired-run solver and noise checks on the simulated coastal network across \res{data.beta.cf-issue-times} test issue times (dashed and dotted lines mark 2\% and 5\% bounds).}
\label{fig:s-simchecks}
\end{figure*}

Each issue time has \res{data.beta.cf-runs-per-issue} runs, namely the base run with the realized drivers, two noise runs, six
added-rain and five added-stage runs, the 30 joint pairs and the seven further Shapley coalitions, which makes
\res{beta.cf.n-runs} runs in all. Every run is continuous from $t=0$ and ends at the window end; no run is restored
from a saved state, so a changed future input cannot silently change the initial state. Four checks cover all runs
(Fig.~\ref{fig:s-simchecks}a,b and Table~\ref{tab:s-simquality}). SWMM reported no error, the pump read back at
setting 1.0 at every minute, the \res{beta.quality.preissue-identical-runs} runs other than the whole-storm noise runs
are bit-identical to their base run before the issue time, and the added driver reached the engine, with rain read
back within \res{beta.quality.readback-rain-mm-max}~mm per step and outlet stage within
\res{beta.quality.readback-stage-cm-max}~cm. In every coalition without future rain, the rain after the issue time is
exactly zero. The models receive the same additions as the simulator; their windows end at the issue time, and the last
history depth and the base window peak of the model windows equal those of the simulated runs.

Every attempted run completed. In Table~\ref{tab:s-simquality}, the whole-storm noise run has no pre-issue
difference, because that run differs from its base run before the issue time by design.
Fig.~\ref{fig:s-simchecks} shows every run before the solver-quality rule of App.~\ref{app:D4} is applied. Its six
pystorms state nodes are J33, J64, J98, J102, ST0 and ST2, the depths observed by the pystorms \emph{beta} scenario. The
lower part of its panel d plots the balanced accuracy of Table~\ref{tab:cf-threshold} and the rain Shapley rank
correlation against the continuity threshold, and the dotted line at 5\% marks the primary rule.

\subsection{Numerical noise floor}\label{app:D3}

The noise floor $\nu$ of a cell (issue time, node) is the largest absolute depth difference within the window between
a noise run and the base run, where the noise run multiplies the rain by $1+10^{-6}$ over the whole storm (primary) or
only after the issue time. It
estimates the numerical change of one run's window peak under a negligible input change. Its distribution is strongly
skewed (Fig.~\ref{fig:s-simchecks}c, Table~\ref{tab:s-noisefloor}). Over all scored cells the median is
\res{beta.noise.pert.all.median}~cm, the 90th percentile \res{beta.noise.pert.all.p90}~cm and the maximum
\res{beta.noise.pert.all.max}~cm, with \res{beta.noise.pert.all.share-le-1cm}\% of cells at or below 1~cm. The local
scale differs by node. The median is \res{beta.noise.pert.states6.median}~cm at the six pystorms state nodes and
\res{beta.noise.pert.near.median}~cm at the near-outlet nodes, but \res{beta.noise.pert.j74.median}~cm at J74 (90th
percentile \res{beta.noise.pert.j74.p90}~cm) and \res{beta.noise.pert.j76.median}~cm at J76 (90th percentile
\res{beta.noise.pert.j76.p90}~cm), the two scored nodes next to the pump. The after-issue noise run gives almost the
same distribution (all-cell median \res{beta.noise.pertpost.all.median}~cm). Every detectability test therefore uses
the cell's own $\nu$, one realization per issue time measured on the base run.

\subsection{Continuity errors and solver anomalies}\label{app:D4}

SWMM reports a flow-routing continuity error per run, in percent, which is kept here with its sign. Over all
\res{beta.cf.n-runs} runs it ranges from \res{beta.quality.cont-min}\% to \res{beta.quality.cont-max}\% with a median of
\res{beta.quality.cont-median}\% (Fig.~\ref{fig:s-simchecks}e). \res{beta.quality.n-abs-gt-2pct} runs exceed 2\% in
magnitude; \res{beta.quality.n-abs-gt-2pct-norain-coalitions} of them are Shapley coalitions without future rain and the
rest are joint-grid runs. All \res{beta.quality.n-abs-gt-5pct} runs beyond 5\% are coalitions without future rain. These
coalitions cut the rain at the issue time in the middle of the storm; they produce the most negative error (storm 154)
and every Shapley exclusion below. The base runs and the single-driver rain and stage families stay within the 2\%
envelope of the generated dataset, and the joint-grid runs beyond it stay below 5\% (Table~\ref{tab:s-simquality}).

\par\textbf{Solver-quality rule.}
A run is usable if (1)~it completed without a SWMM error, (2)~the pump read back at 1.0 at every minute, (3)~it is
bit-identical to its base run before the issue time (not applicable to the whole-storm noise run), (4)~the added driver
was read back (rain within $10^{-6}$~in per step, outlet head within $10^{-6}$~ft) and (5)~its routing continuity error is
at most 5\% in magnitude, stricter than the 10\% variant of Table~\ref{tab:s-sensitivity}. A run family is usable if every run it needs is usable,
namely the base run plus the six added-rain runs for rain thresholds, plus the five added-stage runs for stage thresholds, plus
the reference run for each response ratio, plus the seven coalition runs for Shapley values, and the noise runs
themselves for the floor. Every family except the Shapley family is usable at all \res{data.beta.cf-issue-times}
issue times; the Shapley family is usable at \res{beta.quality.shapley-usable-primary} issue times.

\subsection{Sensitivity to the quality rule and floors}\label{app:D5}

The solver-quality rule and its sensitivity variants filter runs by numerical routing continuity and noise-floor
criteria, independent of model rankings. The response ratios
and rank correlations reported here are evaluated under the primary rule, and the response ratios are identical across
continuity thresholds up to 2\%, because every
rain and stage family is usable. Tables~\ref{tab:s-sensitivity} and~\ref{tab:s-sensitivity-shapley} report all sensitivity variants. The continuity variants are no
filter and limits of 10\%, 2\% (the envelope of the dataset runs) and 1\%. The other variants remove boundary-sensitive cells within $\nu$ or $2\nu$ of the threshold,
remove J74 and J76, use noise-floor multipliers $c_1$ of 1, 3 and 5 instead of 2, or use the after-issue floor.

Tables~\ref{tab:s-sensitivity} and~\ref{tab:s-sensitivity-shapley} report the trained models as seed means. Their SD rows give the SD over three seeds under
the primary protocol, and the SDs of the variants lie within 0.01 of it. In the tables, $k$ is the median rain response
ratio, the model change divided by the simulator change for an added 25.4~mm, over the responsive cells of the
runoff-receiving nodes, those with $|\Delta_{\mathrm{sim}}|>\max(c_1\nu,1~\mathrm{cm})$, where $\nu$ is the noise floor
and the primary $c_1=2$. Balanced accuracy is that of Table~\ref{tab:cf-threshold} on $N$ cells, and $\rho_R$ is the
median, over the usable issue times, of the rank correlation of the rain Shapley values across scored nodes. The
column Det.\ gives the share of scored cells whose simulator interaction $|I_{RS}|$ exceeds $c_1\nu$, a
model-independent share in percent, and boundary cells have a bracketing grid peak within $\nu$ or $2\nu$ of $u_i$.

No variant changes the model ranking by response ratio or by balanced accuracy, all four trained response ratios stay
below 1, and the Scenario GNN has the lowest response ratio in every variant. The largest changes occur under the 1\%
limit, which also shrinks the common denominator most. Balanced accuracy moves by at most
\res{beta.s14.cont-1.max-abs-delta-ba} and the response ratio by at most \res{beta.s14.cont-1.max-abs-delta-k}. The
rain Shapley rank correlations of the Scenario GNN and the Hybrid forecaster are tied. Table~\ref{tab:s-sensitivity-shapley} prints equal
values under the primary protocol, and their order flips in the four solver-quality limits (no limit, 10\%, 2\% and 1\%).
The threshold families stay complete for
limits down to about the 2\% envelope and shrink quickly below 1\% (Fig.~\ref{fig:s-simchecks}d), whereas the Shapley
family keeps \res{beta.quality.shapley-usable-primary} issue times at 5\% and \res{beta.quality.shapley-usable-2pct} at
2\%. The
interaction column of Table~\ref{tab:s-sensitivity-shapley} uses the coalition definition (Shapley coalition runs,
realized tide, $|I_{RS}|$ above a multiple of $\nu$) on the \res{beta.quality.shapley-usable-primary} usable issue
times; this model-independent share moves from \res{beta.s14.m5.det-irs}\% (multiple 5) to
\res{beta.s14.m1.det-irs}\% (multiple 1) around \res{beta.s14.primary.det-irs}\% for the primary multiple 2.

\begin{table*}[t]
\caption{Rain response ratio $k$ and threshold balanced accuracy across simulator solver-quality and noise-floor variants (mean and SD over three seeds; P+R: Physics + residual).}
\label{tab:s-sensitivity}
\centering
\begin{tabular*}{\textwidth}{@{\extracolsep{\fill}}@{}cl*{10}{r}@{}}
\toprule
 & & \multicolumn{5}{c}{Rain response ratio $k$} & \multicolumn{5}{c}{Balanced accuracy} \\
\cmidrule(lr){3-7}\cmidrule(lr){8-12}
\multicolumn{2}{@{}l}{Variant} & Cells & GNN & Hybrid & P+R & P-only & $N$ & GNN & Hybrid & P+R & P-only \\
\midrule
\multicolumn{2}{@{}l}{Primary ($\le$5\%, $c_1=2$)} & 17{,}543 & 0.71 & 0.92 & 0.97 & 0.86 & 12{,}778 & 0.631 & 0.719 & 0.692 & 0.670 \\
\multicolumn{2}{@{}l}{\quad SD} & & 0.13 & 0.08 & 0.03 & 0.09 & & 0.007 & 0.029 & 0.020 & 0.016 \\
\midrule
\multirow{4}{*}{\rotatebox[origin=c]{90}{Continuity}} & no limit & 17{,}543 & 0.71 & 0.92 & 0.97 & 0.86 & 12{,}778 & 0.631 & 0.719 & 0.692 & 0.670 \\
 & $\le$10\% (SWMM manual) & 17{,}543 & 0.71 & 0.92 & 0.97 & 0.86 & 12{,}778 & 0.631 & 0.719 & 0.692 & 0.670 \\
 & $\le$2\% (dataset envelope) & 17{,}543 & 0.71 & 0.92 & 0.97 & 0.86 & 12{,}778 & 0.631 & 0.719 & 0.692 & 0.670 \\
 & $\le$1\% & 14{,}728 & 0.75 & 0.91 & 0.94 & 0.84 & 9{,}168 & 0.632 & 0.735 & 0.691 & 0.674 \\
\midrule
\multirow{3}{*}{\rotatebox[origin=c]{90}{Removed}} & boundary cells, $\nu$ & 17{,}543 & 0.71 & 0.92 & 0.97 & 0.86 & 12{,}195 & 0.632 & 0.717 & 0.691 & 0.667 \\
 & boundary cells, $2\nu$ & 17{,}543 & 0.71 & 0.92 & 0.97 & 0.86 & 11{,}829 & 0.633 & 0.715 & 0.692 & 0.667 \\
 & nodes J74, J76 & 17{,}395 & 0.70 & 0.92 & 0.96 & 0.85 & 12{,}608 & 0.634 & 0.720 & 0.691 & 0.670 \\
\midrule
\multirow{4}{*}{\rotatebox[origin=c]{90}{Noise floor}} & $c_1=1$ & 18{,}002 & 0.71 & 0.92 & 0.99 & 0.88 & 12{,}778 & 0.631 & 0.719 & 0.692 & 0.670 \\
 & $c_1=3$ & 17{,}146 & 0.70 & 0.91 & 0.95 & 0.84 & 12{,}778 & 0.631 & 0.719 & 0.692 & 0.670 \\
 & $c_1=5$ & 16{,}573 & 0.70 & 0.91 & 0.93 & 0.81 & 12{,}778 & 0.631 & 0.719 & 0.692 & 0.670 \\
 & after-issue $\nu$ & 17{,}524 & 0.71 & 0.92 & 0.97 & 0.85 & 12{,}778 & 0.631 & 0.719 & 0.692 & 0.670 \\
\bottomrule
\end{tabular*}
\end{table*}

\begin{table}[t]
\caption{Rain Shapley rank correlation $\rho_R$ and detectable interaction share (Det., $|I_{RS}|>c_1\nu$, \%) across the sensitivity variants of Table~\ref{tab:s-sensitivity} (mean and SD over three seeds).}
\label{tab:s-sensitivity-shapley}
\centering
\setlength{\tabcolsep}{3pt}%
\resizebox{\columnwidth}{!}{%
\begin{tabular}{@{}cl*{6}{r}@{}}
\toprule
 & & \multicolumn{5}{c}{Rain Shapley $\rho_R$} & \\
\cmidrule(lr){3-7}
\multicolumn{2}{@{}l}{Variant} & Issues & GNN & Hybrid & P+R & P-only & Det. \\
\midrule
\multicolumn{2}{@{}l}{Primary ($\le$5\%, $c_1=2$)} & 139 & 0.937 & 0.937 & 0.828 & 0.774 & 10.5 \\
\multicolumn{2}{@{}l}{\quad SD} & & $<$0.001 & 0.001 & 0.017 & 0.019 & \\
\midrule
\multirow{4}{*}{\rotatebox[origin=c]{90}{Continuity}} & no limit & 150 & 0.939 & 0.941 & 0.833 & 0.775 & 10.6 \\
 & $\le$10\% (SWMM manual) & 148 & 0.939 & 0.941 & 0.832 & 0.775 & 10.5 \\
 & $\le$2\% (dataset envelope) & 125 & 0.937 & 0.936 & 0.824 & 0.772 & 10.5 \\
 & $\le$1\% & 77 & 0.933 & 0.929 & 0.809 & 0.767 & 9.5 \\
\midrule
\multirow{3}{*}{\rotatebox[origin=c]{90}{Removed}} & boundary cells, $\nu$ & 139 & 0.937 & 0.937 & 0.828 & 0.774 & 10.5 \\
 & boundary cells, $2\nu$ & 139 & 0.937 & 0.937 & 0.828 & 0.774 & 10.5 \\
 & nodes J74, J76 & 139 & 0.937 & 0.937 & 0.825 & 0.769 & 10.5 \\
\midrule
\multirow{4}{*}{\rotatebox[origin=c]{90}{Noise floor}} & $c_1=1$ & 139 & 0.937 & 0.937 & 0.828 & 0.774 & 13.5 \\
 & $c_1=3$ & 139 & 0.937 & 0.937 & 0.828 & 0.774 & 9.1 \\
 & $c_1=5$ & 139 & 0.937 & 0.937 & 0.828 & 0.774 & 7.7 \\
 & after-issue $\nu$ & 139 & 0.937 & 0.937 & 0.828 & 0.774 & 11.4 \\
\bottomrule
\end{tabular}}
\end{table}

\subsection{Floors for joint effects and threshold resolution}\label{app:D6}

A paired difference $\Delta_{\mathrm{sim}}$ of one run and its base is detectable when it exceeds $\max(c_1\nu,\delta)$
with $c_1=2$, the sum of the two runs' floors, and the materiality floor $\delta=1$~cm. The interaction
$I_{RS}=v(R,S)-v(R,0)-v(0,S)+v(0,0)$ of Eq.~\eqref{eq:interaction} combines four runs, so a single-run floor does not
apply. It is detectable when $|I_{RS}|>\max(c_2\nu,\delta)$ with $c_2=4$, the triangle-inequality sum of four per-run
floors, plus the same materiality floor, which binds in most cells. The detectable share of the simulator's interaction
at $(\bar A,\bar s)$ depends on this choice. It is \res{beta.fig3.irs.sim.share_det}\% with the primary floor,
\res{beta.fig3.irs.sim.share_det_2nf}\% with the two-run floor $2\nu$ and \res{beta.fig3.irs.sim.share_det_4nf_nofloor}\%
with $4\nu$ alone. App.~\ref{app:I} reports the interaction results with both this floor and the coalition
definition. Because $\nu$ is measured on the base run only, a difference near its floor is treated as undetectable rather
than estimated.

Threshold decisions on the simulator rest on grids of six added rain amounts and five added outlet stages, refined by
linear interpolation
(Algorithm~\ref{alg:s-astar}). Cells whose bracketing grid peak lies within $\nu$ or $2\nu$ of the threshold are kept
in the primary protocol, and removing them changes balanced accuracy by at most
\res{beta.s14.no-bsens-2nf.max-abs-delta-ba} (Table~\ref{tab:s-sensitivity}). For the outlet stage, few cells are
crossed by both the simulator and a model without an initial false alarm, so stage crossing errors are reported
together with their coverage (App.~\ref{app:K2}).

\section{Estimands, Metrics, and Statistical Analysis}\label{app:E}

This appendix is the formula dictionary of the paper. For every metric it states the target, unit, denominator,
aggregation, treatment of undefined cases, reference baseline and interval method. The protocol of the statistical
analysis and multiplicity conventions close it. In tables, -- marks a cell that does not apply to
its row and UD a mathematically undefined value.

\subsection{Targets}\label{app:E1}

\emph{Forecast target.} The stage of every state node at every lead $\ell=1,\dots,H$ (cm), scored only on observed
labels. \emph{Query target.} The peak margin $m_i(q)$ of Eq.~\eqref{eq:forecast}, which is the largest forecast over all leads,
without a mask, minus $u_i$, at the canal gauges and at query issue times every 6~h. \emph{Observed peak.} The largest
observed stage over the 48-h horizon, defined only for cells with at least 40 of the 48 hours observed. It is the
outcome of the observational check and of U2. \emph{Simulator target.} The window peak $P$ of node depth over the 24
steps after the issue time (cm). A difference of window peaks (maximum first, then difference) is not the window
maximum of time-wise differences. The fixed-time variant, the difference at the realized run's peak step, is named
separately wherever it is used. \emph{Threshold states.} A threshold crossing is $m_i\ge\varepsilon$. An observed
exceedance is an observed 48-h peak at or above $u_i$. The forecast event of the critical success index is an observed
hourly stage above $u_i$, so it counts hours, not independent events. Durations of exceedance and peak timing are not
queried, and Fig.~\ref{fig:s-compound} shows the shift of the simulated window-peak time descriptively.

\subsection{Forecast metrics}\label{app:E2}

For station $i$ with observed cells $c$ (window, lead), $\mathrm{NSE}_i=1-\sum_c(\hat h_c-h_c)^2/\sum_c(h_c-\bar h_i)^2$,
where $\bar h_i$ is the mean over the same cells~\cite{nash1970river}. Canal NSE is the median of $\mathrm{NSE}_i$ over
scored canal gauges. A gauge is scored if it has at least 10 observed cells and an observed variance above
$10^{-3}$~cm$^2$, which leaves \res{mia.fc.n-canal-scored} of \res{mia.fc.n-canal} canal gauges in both years.
Groundwater NSE is the median over the \res{mia.fc.n-wells-scored} wells. MAE and the root-mean-square error (RMSE, cm)
pool all observed canal cells of all gauges, windows and leads, so gauges with more observed hours weigh more. The lead
bands from 1 to 6, from 7 to 24 and from 25 to 48~h restrict the same pooling to those leads. CSI is hits / (hits + misses + false
alarms)~\cite{schaefer1990critical} from one contingency table pooled over observed (issue time, canal gauge, lead)
cells, with an event when the observed stage exceeds $u_i$ and a warning when the forecast does. The event base rate
is \res{mia.fc.test.p95.base-rate}\% of observed canal cells in 2019 and \res{mia.fc.val.p95.base-rate}\% in 2018.
Because windows overlap, one observed hour enters up to 48 windows. Persistence repeats the last history value, the
filled series at the issue hour, over the horizon. It is accumulated inside every evaluation on the same batches,
windows and mask, and its scores are identical in all \res{mia.fc.check.persistence-identical-runs} evaluation
summaries. $\Delta$NSE is the scenario model minus its issue-time reference under $q_0$, reported as mean\,$\pm$\,SD of
seed-matched differences. On the simulator, NSE is the median over scored nodes with
non-constant depth, MAE, RMSE and CSI (at the node's training p95) pool all scored cells, and the same scoring code is
used over the \res{data.beta.windows-test} test windows.

\subsection{Attribution and response metrics}\label{app:E3}

\par\textbf{Shapley recovery on the simulator.}
For each usable issue time, the rank correlation is Spearman's correlation between model and simulator values of one
player across the \res{beta.nodes.scored} scored nodes. It is undefined for a constant vector. The reported value
is the median over the \res{beta.quality.shapley-usable-primary} usable issue times, computed per seed and then
averaged over seeds. The absolute error is the mean $|\phi_{\mathrm{model}}-\phi_{\mathrm{sim}}|$ over discernible
cells, where $|\phi_{\mathrm{sim}}|>2\nu$ (\res{beta.fig2a.share-disc.r}\% of cells for rain,
\res{beta.fig2a.share-disc.s}\% for surge). The signed error averages $\phi_{\mathrm{model}}-\phi_{\mathrm{sim}}$ over the
same cells, so a negative value is under-attribution. The no-response reference is the mean $|\phi_{\mathrm{sim}}|$ over
those cells (\res{beta.fig2a.true-abs-phi-disc.r}~cm for rain, \res{beta.fig2a.true-abs-phi-disc.s}~cm for surge), which is the
error of a model whose attribution is zero.

\par\textbf{Direct responses.}
The response ratio $k$ of Eq.~\eqref{eq:paired} is the median of $\Delta_{\mathrm{model}}/\Delta_{\mathrm{sim}}$ over
responsive cells with $|\Delta_{\mathrm{sim}}|>\max(c_1\nu,\delta)$, for $\bar A$ at the \res{beta.nodes.rain} scored
runoff-receiving nodes (\res{beta.k.rain.n-cells} responsive cells of the \res{data.beta.cf-issue-times} issue times) and for $\bar s$ at the
\res{beta.nodes.near} near-outlet nodes (\res{beta.k.stage.n-cells} responsive cells). It is computed per seed and then
averaged. The response MAE is the mean $|\Delta_{\mathrm{model}}-\Delta_{\mathrm{sim}}|$ over all cells of the node
group, without the responsiveness filter. Its no-response reference is the mean $|\Delta_{\mathrm{sim}}|$ over the same
cells (\res{beta.k.rain.zero-response.mae}~cm for rain, \res{beta.k.stage.zero-response.mae}~cm for the outlet stage),
and the relative reduction is one minus their ratio. The wrong-direction share is the share of
responsive rain cells in which the simulated peak rises but the model's peak falls. The non-monotone share
is the share of scored cells whose change drops by more than $\max(2\nu,0.1~\text{cm})$ between consecutive
amounts of the rain grid. In Fig.~\ref{fig:driver-info}b the unit is a storm and node group, namely the mean over the group's
nodes for one storm, with storms binned into six quantile bins of the simulated change.

\par\textbf{Node sets and distances.}
Simulator-affected nodes are the scored nodes whose simulated window-peak change under the reference change ($\bar A$ or
$\bar s$) exceeds $\max(c_1\nu,\delta)$ in at least half of the issue times (\res{beta.s12.layer.affected_rain.n} nodes for
rain and \res{beta.s12.layer.affected_stage.n} for the outlet stage). Hydraulic path length is the shortest path to the
outfall along links, using conduit lengths and counting structures as zero length. The hop distance counts links.
Neither is a travel time.

\par\textbf{Observational check.}
The cells are (issue time, gauge) pairs with at least 40 of 48 hours observed, at the \res{data.mia.windows-test} hourly
issue times of a year. With $B$, $F$ and $Y$ as in Eq.~\eqref{eq:substitution} (model peaks over all leads, observed
peak over observed hours), the slope $\beta$ is the least-squares slope of $\Delta_o$ on $\Delta_p$ with intercept. The
station fixed-effect version first demeans $\Delta_p$, $\Delta_o$ and the shuffled response within each station. Sign
agreement is the share of cells with $|\Delta_p|>\delta_o$ (on the demeaned $\Delta_p$ in the fixed-effect version)
whose $\Delta_p$ and $\Delta_o$ have the same sign. Its chance level is not one half when signs are imbalanced. The
shuffle slope repeats the regression with the driver taken from another window, the untrained copy repeats the whole
computation, and the mean $|\Delta_p|$ is reported on raw cells. The pooled rule passes a driver if, for all three
seeds, the slope lies in $[0.5,2]$ with a positive interval lower bound, the agreement interval lies above one half, the
shuffle slope is smaller than half the slope in magnitude, and the untrained copy does not meet the first two criteria.
The strict rule counts a driver as informative only if at least 100 cells have $|\Delta_p|>1$~cm and the mean
$|\Delta_p|$ is at least 0.1~cm, both on demeaned values in Miami-Dade and on pooled values on the simulators. It uses the fixed-effect slope with the window
$[0.75,1.33]$, requires the agreement to exceed the untrained copy's by 0.10, tightens the shuffle bound to a quarter of
the slope and restates the clause on untrained copies. The proof of Eq.~\eqref{eq:substitution}, the complete verdict
criteria, and the station groups are in App.~\ref{app:H}. On the
simulator the same code runs on the \res{data.beta.windows-test} test windows, with the runoff-receiving nodes as the
rain group and the near-outlet nodes as the surge group.

\subsection{Interactions}\label{app:E4}

The finite-difference interaction is $I_{RS}=v(R,S)-v(R,0)-v(0,S)+v(0,0)$ (Eq.~\eqref{eq:interaction}), where 0 is a
defined reference, not always zero rain or zero surge. In Miami-Dade, 0 is the attribution reference for rain and
surge, the tide stays at its issue-time value, and operations and history are fixed, so
$I_{RS}=v(\{R,S,T\})-v(\{S,T\})-v(\{R,T\})+v(\{T\})$. The trained models' near-additivity is summarized as the mean
$|$joint response $-$ sum of single responses$|$ over cells. On the simulator's paired grid, 0 is the realized driver
(the base run), and the four runs share the history, the astronomical tide and the fixed control path. A simulator
interaction is detectable when $|I_{\mathrm{sim}}|>\max(c_2\nu,\delta)$. The primary evaluation point is $(\bar A,\bar s)$, and all 30 grid points are reported in Fig.~\ref{fig:s-compound}. Signed
recall is the share of simulator-detectable cells in which $|I_{\mathrm{model}}|$ exceeds the same cell threshold with
the simulator's sign. Sign agreement is the share of detectable cells in which the model has the simulator's sign and
$|I_{\mathrm{model}}|\ge0.1$~cm, smaller model values counting as no interaction. The error is the mean
$|I_{\mathrm{model}}-I_{\mathrm{sim}}|$ over detectable cells and, without the filter, over all
\res{beta.nodes.scored}\,$\times$\,\res{data.beta.cf-issue-times} node and storm cells. The no-interaction baseline
$I\equiv0$ has error \res{beta.fig3.irs.nointeraction.mae_det}~cm on detectable cells and
\res{beta.fig3.irs.nointeraction.mae_all}~cm on all cells. Model detections where the simulator's interaction is not
detectable enter the all-cell error. The coalition definition forms
$I_{RS}$ from the Shapley coalition runs (realized tide, with 0 meaning no future rain or no future surge) and counts
$|I_{RS}|>2\nu$, which gives \res{beta.fig3.irs.coalition.share_gt_2nf}\% of cells over all
\res{data.beta.cf-issue-times} issue times, before the quality rule. With $\max(c_2\nu,\delta)$ it gives
\res{beta.fig3.irs.coalition.share_det}\%, and \res{beta.fig3.irs.coalition.share_det_quality}\% after removing issue
times with a coalition run beyond 2\% continuity error. The fixed-time interaction is the same difference of depths at
the realized run's peak step. It is computed for the simulator, while the models' interaction is measured on window
peaks, the quantity they save. The complete interaction results are in App.~\ref{app:I}.

\subsection{Counterfactual threshold classification}\label{app:E5}

The common denominator $N$ holds the (issue time, threshold node) cells whose simulated base window peak is below
$u_i$ and whose rain-threshold run family is usable, which gives \res{beta.tab3.n} cells, the same for every model and baseline.
The target is $y=1$ if the simulated window peak reaches $u_i$ for some grid amount, which happens in
\res{beta.tab3.positives} cells (\res{beta.tab3.pos-rate}\%). The prediction is $\hat y=1$ if the model's own window
peak reaches $u_i$ for some grid amount including $A=0$. A model already at or above $u_i$ at the base scenario counts
as a predicted crossing and is tallied separately as an initial false alarm, whose share of $N$ is reported. This is a
classification of crossing within the scenario range; a crossing predicted from an initial false alarm is not a
successful prediction of a crossing caused by the added driver. From the confusion matrix, crossing recall is
TP/(TP+FN), crossing precision TP/(TP+FP) (UD if TP+FP${}=0$), non-crossing recall TN/(TN+FP), balanced accuracy the
mean of the two recalls~\cite{brodersen2010balanced}, the Matthews correlation coefficient (MCC) the correlation of the
two binary labels~\cite{chicco2020advantages}, and the positive rate (TP+FN)/$N$. Raw agreement is shown only next to the
constant classifiers, because it rewards always predicting a crossing when most cells cross. The constant classifiers
always or never predict a crossing and have no crossing rainfall to compare. The training-fit linear response fits, for
each threshold node, a least-squares intercept and slope of the window-peak rise on the window rain over the training
storms and predicts the window peak at added rain $A$ as the issue-time depth plus the intercept plus the slope times
the realized window rain plus $A$. Its stage version regresses on the window-mean outlet-stage change. The strict diagnostic subset holds the
\res{beta.tab3.strict.n} cells of $N$ in which no trained configuration (four models, three seeds) is initially warned,
with the same target and prediction. Each metric is computed per seed and reported
as mean\,$\pm$\,SD. The outlet-stage version uses the stage grid and the stage family in the same way, and MCC is given
in Table~\ref{tab:s-confusion}. Raw accuracy of the same decision, with model base warnings counted as crossings, is
reported separately in App.~\ref{app:K}.

\subsection{Threshold error and censoring}\label{app:E6}

The simulated minimum crossing rainfall is defined on cells with $y=1$, and the model's on cells where the model is not
initially warned and crosses at some $A>0$, both by Algorithm~\ref{alg:s-astar}. The conditional median $|A^\ast|$ error is the
median of $|A^\ast_{\mathrm{model}}-A^\ast_{\mathrm{sim}}|$ over cells where the simulator crosses and the model crosses
without an initial false alarm, reported together with its coverage $n/N$. The signed error is the model's $A^\ast$ minus the simulator's, so a positive value
means the model needs more added rain to cross. On the subset of \res{beta.tab3.common-both.n} cells where the simulator
and all twelve trained configurations cross without an initial false alarm, the errors are also compared on one common
set of cells.

In Miami-Dade, \emph{not crossed within range} means that a monotone grid reaches no crossing up to $A_{\max}$. Such a
cell is censored at the end of the range, so it is neither a missing label nor a cell with $A^\ast=A_{\max}$. Two
conventions handle non-monotone cells. The monotone-grid convention removes them from the denominator and defines the
monotone-grid sample used by Fig.~\ref{fig:state-event} and by U1 to U3. The first-observed
convention keeps every initially unwarned cell and enters non-monotone cells at their first observed crossing. It gives
the not-crossed shares of Fig.~\ref{fig:s-censoring}a and~c. Each panel names its convention, and Table~\ref{tab:s-accounting}
reports both. Medians of $A^\ast$ over crossers describe crossers only. Boundary-sensitive cells (Algorithm~\ref{alg:s-astar},
step~7) are counted, not removed. The U1 to U3 protocols, with their reference outcomes, denominators and the information
used by each rule, are in App.~\ref{app:K}.

\subsection{Sampling uncertainty}\label{app:E7}

Windows issued every hour overlap, and node and storm cells share storms, so neither is an independent sample. In
Miami-Dade the resampling unit is a 7-day block of issue times (168~h, \res{data.mia.weekly-blocks-test} blocks in the
test year), drawn with replacement~\cite{kunsch1989jackknife}. Blocks keep the dependence between overlapping windows
and between the stations of one issue time. The observational check uses 500 block resamples per run (seed 0) and the
threshold-free U2 and U3 differences 500, all as percentile intervals computed
per seed. The tables show the per-seed intervals or their envelope over seeds. On the simulator each test storm
contributes one issue time, so the \res{data.beta.n-storms-test} storms are resampled with replacement and all nodes
of a storm move together, with 2,000 resamples (seed 0) for the Shapley, propagation and interaction intervals and 5,000 for
the paired balanced-accuracy differences between the models of Table~\ref{tab:cf-threshold}. The same resample indices serve every model,
so paired differences keep their correlation. For the balanced-accuracy differences and the propagation and
interaction measures, seeds are averaged within each resample, so these intervals describe storm sampling variation of
the seed-mean metric, not seed variation. The SD over the three seeds is a
run-to-run spread and is reported separately, while untrained copies use one initialization each. Forecast scores under $q_0$ are compared through their
means and SDs over seeds.

\subsection{Multiplicity and reporting conventions}\label{app:E8}

No correction for multiple comparisons is applied, and no null-hypothesis significance test is reported. Instead, the
paper reports effect sizes, seed means and standard deviations (or seed ranges), and block or storm bootstrap percentile
intervals across all evaluated station groups, rule variants, thresholds and sample conventions.

\subsection{Settings}\label{app:E9}

Table~\ref{tab:s-settings} lists the forecast, query and checking settings that Section~\ref{sec:setup} summarizes.
The threshold percentile $p_u$ is taken over each gauge's training stage, and the observed $A_{\max}$ is the
Miami-Dade value per 24~h, with the values of the other regions in App.~\ref{app:L5}. The operation classes are
defined in App.~\ref{app:P2} and the rules of the observational check in App.~\ref{app:H3}.

\begin{table}[h]
\caption{Forecast, query and checking settings of the observed and simulated systems.}
\label{tab:s-settings}
\centering
\small
\begin{tabular}{@{}lcc@{}}
\toprule
Setting & Observed & Simulated \\
\midrule
Horizon $H$ & \res{data.mia.horizon-h}~h & \res{data.beta.window-h}~h \\
Threshold percentile $p_u$ & 95th & 95th \\
Crossing margin $\varepsilon$ & 1~cm & 0 \\
Search bound $A_{\max}$ & \res{data.mia.amax-mm}~mm & 76.2~mm \\
Materiality floor $\delta$ & 1~cm & 1~cm \\
\midrule
\multicolumn{3}{@{}l}{\emph{Scoring paired runs}} \\
Noise multiple for a cell in $k$, $c_1$ & -- & 2 \\
Noise multiple for a detectable $I_{RS}$, $c_2$ & -- & 4 \\
Missing response, $|k|<k_0$ & -- & 0.2 \\
Right-sized response, $[k_-,k_+]$ & -- & $[0.75,1.33]$ \\
Sign agreement, $\zeta_0$ and $\zeta_1$ & -- & 0.5, 0.8 \\
Altered-input factor $a$ & -- & $1/2$, $-1$ \\
\midrule
\multicolumn{3}{@{}l}{\emph{Observational check}} \\
Counted predicted change, $\delta_o$ & 1~cm & 1~cm \\
Pooled slope window & $[0.5,2]$ & $[0.5,2]$ \\
Strict station FE slope window & $[0.75,1.33]$ & $[0.75,1.33]$ \\
Strict, counted cells & $\ge100$ & $\ge100$ \\
Strict, mean $|\Delta_p|$ & $\ge0.1$~cm & $\ge0.1$~cm \\
Strict, sign margin over untrained copy & 0.10 & 0.10 \\
\bottomrule
\end{tabular}
\end{table}

\section{Complete Forecasting Results}\label{app:F}

This appendix reports the complete forecast scores behind Table~\ref{tab:xr} and the forecast axis of
Fig.~\ref{fig:s-forecast-response}, organized by system, information set at forecast time, state
variable, and lead band. Every score follows the metric definitions of App.~\ref{app:E}.

Table~\ref{tab:s-forecast} gives the complete Miami-Dade 2019 test-year evaluation across all information sets, and
Table~\ref{tab:s-forecast-lead} breaks canal MAE down by lead band across both the 2019 test and 2018 validation years. In these tables CSI counts
exceedances of the training p95 on observed (issue time, station, lead) cells.

\subsection{Miami-Dade}\label{app:F1}

\par\textbf{Protocol.}
Table~\ref{tab:s-forecast} gives the test-year scores and Table~\ref{tab:s-forecast-lead} the canal MAE by lead band.
The lead bands use the same runs and seeds as Table~\ref{tab:s-forecast}, which reports the all-lead MAE. Rows are grouped
by the scenario given at forecast time, namely the issue-time scenario $q_0$, $q_0$ without the rain forecast (future
rain set to zero, rain${}=0$ in the table), and the perfect-driver hindcast with realized future rain, surge, and
operations. Within each block, each architecture lists its issue-time reference, its scenario model, and its untrained copy,
and the Training column gives the information set each was trained with. Scores use all
\res{mia.fc.n-windows} hourly windows of the year, the same observed-label mask in every row, the median over
\res{mia.fc.n-canal-scored} scored canal gauges and \res{mia.fc.n-wells-scored} wells for NSE, and pooled observed
cells for MAE, RMSE, and CSI. Entries are mean\,$\pm$\,SD over seeds 42, 43, and 44. Untrained rows use one
initialization each. The $q_0$ scores of the Physics-only model's issue-time reference come from its training-run
evaluation on the same windows and mask, and its hindcast rows come from the scenario
evaluation of the same three runs. Issue-time
references do not read future rain, so their
rows without the rain forecast equal their $q_0$ rows to every printed digit and are not repeated.

\par\textbf{What the complete table adds.}
Under $q_0$, the differences between each scenario model and its reference match Table~\ref{tab:xr}. In
the 2018 validation year, the seed ranges of the Scenario GNN and the Hybrid forecaster lie slightly below their references, and the
Physics-only model has the largest cost in both years. Its cost grows with lead: at leads from 25 to 48~h its canal MAE is
\res{mia.fc.test.q0.physonly.mae-h25-48.mean}~cm against \res{mia.fc.test.q0.ref-physonly.mae-h25-48.mean}~cm for its
reference in 2019, whereas the Scenario GNN matches its reference in every band (Table~\ref{tab:s-forecast-lead}). Without the rain forecast, the
Scenario GNN loses canal NSE in 2019 (\res{mia.fc.test.q0.gnn.nse.mean} to \res{mia.fc.test.q0norain.gnn.nse.mean}),
whereas the Hybrid forecaster gains slightly (\res{mia.fc.test.q0.hybrid.nse.mean} to \res{mia.fc.test.q0norain.hybrid.nse.mean}, with MAE from
\res{mia.fc.test.q0.hybrid.mae.mean} to \res{mia.fc.test.q0norain.hybrid.mae.mean}~cm) and scores
\res{mia.fc.val.q0norain.hybrid.nse.mean} both with and without it in 2018. The perfect-driver hindcast rows show how much each
model utilizes realized drivers as a diagnostic reference. The large test RMSE spread of the Physics + state
residual reference in Table~\ref{tab:s-forecast} reflects one of its three seeds. The untrained copies score far below persistence (2019 canal NSE \res{mia.fc.test.q0.untr-hybrid.nse} to
\res{mia.fc.test.q0.untr-gnn.nse}) and serve strictly as structural controls.

\par\textbf{Lead time and high water.}
Lead-resolved skill is the canal MAE in the three lead bands of Table~\ref{tab:s-forecast-lead}, and high-water skill is the CSI column. The
forecast scores of the later Miami-Dade period and of Houston belong to the transfer checks (App.~\ref{app:L}).

\begin{table*}[t]
\caption{Complete forecast scores in Miami-Dade (2019, mean\,$\pm$\,SD over three seeds; $^{\dagger}$own training-run evaluation).}
\label{tab:s-forecast}
\begin{tabular*}{\textwidth}{@{\extracolsep{\fill}}cllr@{\extracolsep{0pt}}l@{\extracolsep{\fill}}r@{\extracolsep{0pt}}l@{\extracolsep{\fill}}r@{\extracolsep{0pt}}l@{\extracolsep{\fill}}r@{\extracolsep{0pt}}l@{\extracolsep{\fill}}r@{\extracolsep{0pt}}l@{}}
\toprule
 & & & \multicolumn{8}{c}{Canal} & \multicolumn{2}{c}{Groundwater} \\
\cmidrule(lr){4-11}\cmidrule(l){12-13}
 & Model & Training & \multicolumn{2}{c}{NSE} & \multicolumn{2}{c}{MAE (cm)} & \multicolumn{2}{c}{RMSE (cm)} & \multicolumn{2}{c}{CSI} & \multicolumn{2}{c}{NSE} \\
\midrule
\multirow{13}{*}{\rotatebox[origin=c]{90}{Issue-time scenario $q_0$}} & Persistence & -- & 0.755 &  & 3.98 &  & 9.04 &  & 0.683 &  & 0.829 &  \\
\cmidrule(l){2-13}
 & \multirow{3}{*}{Scenario GNN} & issue-time & 0.814 & \,$\pm$\,0.001 & 2.53 & \,$\pm$\,$<$0.01 & 4.93 & \,$\pm$\,0.01 & 0.735 & \,$\pm$\,0.003 & 0.914 & \,$\pm$\,0.002 \\
 & & realized & 0.811 & \,$\pm$\,0.003 & 2.53 & \,$\pm$\,$<$0.01 & 4.97 & \,$\pm$\,0.01 & 0.732 & \,$\pm$\,0.002 & 0.916 & \,$\pm$\,0.001 \\
 & & untrained & 0.546 &  & 6.09 &  & 10.16 &  & 0.571 &  & 0.642 &  \\
\cmidrule(l){2-13}
 & \multirow{3}{*}{Hybrid} & issue-time & 0.809 & \,$\pm$\,0.006 & 2.63 & \,$\pm$\,0.02 & 5.09 & \,$\pm$\,0.01 & 0.716 & \,$\pm$\,0.005 & 0.901 & \,$\pm$\,0.002 \\
 & & realized & 0.804 & \,$\pm$\,0.001 & 2.67 & \,$\pm$\,0.02 & 5.05 & \,$\pm$\,0.02 & 0.721 & \,$\pm$\,0.006 & 0.901 & \,$\pm$\,0.003 \\
 & & untrained & $-$1.052 &  & 13.27 &  & 18.23 &  & 0.045 &  & 0.782 &  \\
\cmidrule(l){2-13}
 & \multirow{3}{*}{Physics + residual} & issue-time & 0.773 & \,$\pm$\,0.002 & 3.37 & \,$\pm$\,0.07 & 8.02 & \,$\pm$\,1.70 & 0.696 & \,$\pm$\,0.008 & 0.846 & \,$\pm$\,0.001 \\
 & & realized & 0.774 & \,$\pm$\,0.002 & 3.47 & \,$\pm$\,0.02 & 7.37 & \,$\pm$\,0.14 & 0.698 & \,$\pm$\,0.006 & 0.854 & \,$\pm$\,0.001 \\
 & & untrained & 0.031 &  & 9.19 &  & 12.50 &  & 0.048 &  & 0.744 &  \\
\cmidrule(l){2-13}
 & \multirow{3}{*}{Physics-only} & issue-time$^{\dagger}$ & 0.766 & \,$\pm$\,$<$0.001 & 3.43 & \,$\pm$\,0.01 & 7.17 & \,$\pm$\,0.01 & 0.694 & \,$\pm$\,0.001 & 0.830 & \,$\pm$\,$<$0.001 \\
 & & realized & 0.752 & \,$\pm$\,0.002 & 3.76 & \,$\pm$\,0.03 & 8.43 & \,$\pm$\,0.52 & 0.670 & \,$\pm$\,0.011 & 0.816 & \,$\pm$\,$<$0.001 \\
 & & untrained & 0.027 &  & 9.02 &  & 12.30 &  & 0.056 &  & 0.758 &  \\
\midrule
\multirow{4}{*}{\rotatebox[origin=c]{90}{\begin{tabular}{@{}c@{}}$q_0$,\\rain${}=0$\end{tabular}}} & Scenario GNN & \multirow{4}{*}{realized} & 0.804 & \,$\pm$\,0.005 & 2.57 & \,$\pm$\,0.02 & 5.03 & \,$\pm$\,0.02 & 0.722 & \,$\pm$\,0.003 & 0.904 & \,$\pm$\,0.001 \\
 & Hybrid &  & 0.811 & \,$\pm$\,0.005 & 2.64 & \,$\pm$\,0.01 & 5.09 & \,$\pm$\,0.02 & 0.710 & \,$\pm$\,0.007 & 0.898 & \,$\pm$\,0.002 \\
 & Physics + residual &  & 0.772 & \,$\pm$\,0.002 & 3.48 & \,$\pm$\,0.03 & 7.49 & \,$\pm$\,0.18 & 0.683 & \,$\pm$\,0.007 & 0.812 & \,$\pm$\,0.005 \\
 & Physics-only &  & 0.746 & \,$\pm$\,$<$0.001 & 3.70 & \,$\pm$\,0.03 & 8.49 & \,$\pm$\,0.55 & 0.656 & \,$\pm$\,0.008 & 0.789 & \,$\pm$\,$<$0.001 \\
\midrule
\multirow{12}{*}{\rotatebox[origin=c]{90}{Perfect-driver hindcast}} & \multirow{3}{*}{Scenario GNN} & issue-time & 0.813 & \,$\pm$\,0.002 & 2.53 & \,$\pm$\,$<$0.01 & 4.93 & \,$\pm$\,0.01 & 0.736 & \,$\pm$\,0.003 & 0.914 & \,$\pm$\,0.002 \\
 & & realized & 0.846 & \,$\pm$\,0.002 & 2.41 & \,$\pm$\,0.01 & 4.74 & \,$\pm$\,0.02 & 0.731 & \,$\pm$\,0.005 & 0.922 & \,$\pm$\,0.002 \\
 & & untrained & 0.546 &  & 6.08 &  & 10.16 &  & 0.571 &  & 0.642 &  \\
\cmidrule(l){2-13}
 & \multirow{3}{*}{Hybrid} & issue-time & 0.809 & \,$\pm$\,0.006 & 2.60 & \,$\pm$\,0.02 & 5.04 & \,$\pm$\,$<$0.01 & 0.719 & \,$\pm$\,0.005 & 0.901 & \,$\pm$\,0.002 \\
 & & realized & 0.844 & \,$\pm$\,0.001 & 2.44 & \,$\pm$\,$<$0.01 & 4.63 & \,$\pm$\,0.03 & 0.715 & \,$\pm$\,0.007 & 0.931 & \,$\pm$\,0.001 \\
 & & untrained & $-$1.052 &  & 13.24 &  & 18.20 &  & 0.045 &  & 0.783 &  \\
\cmidrule(l){2-13}
 & \multirow{3}{*}{Physics + residual} & issue-time & 0.773 & \,$\pm$\,0.002 & 3.24 & \,$\pm$\,0.02 & 6.88 & \,$\pm$\,0.56 & 0.702 & \,$\pm$\,0.008 & 0.846 & \,$\pm$\,0.001 \\
 & & realized & 0.813 & \,$\pm$\,0.004 & 3.13 & \,$\pm$\,0.02 & 6.48 & \,$\pm$\,0.06 & 0.692 & \,$\pm$\,0.005 & 0.911 & \,$\pm$\,0.001 \\
 & & untrained & 0.029 &  & 9.16 &  & 12.46 &  & 0.048 &  & 0.745 &  \\
\cmidrule(l){2-13}
 & \multirow{3}{*}{Physics-only} & issue-time & 0.764 & \,$\pm$\,0.001 & 3.36 & \,$\pm$\,0.01 & 7.09 & \,$\pm$\,0.01 & 0.700 & \,$\pm$\,0.001 & 0.830 & \,$\pm$\,$<$0.001 \\
 & & realized & 0.776 & \,$\pm$\,0.018 & 3.38 & \,$\pm$\,0.01 & 7.18 & \,$\pm$\,0.05 & 0.672 & \,$\pm$\,0.007 & 0.891 & \,$\pm$\,$<$0.001 \\
 & & untrained & 0.027 &  & 8.98 &  & 12.26 &  & 0.057 &  & 0.758 &  \\
\bottomrule
\end{tabular*}
\end{table*}

\begin{table}[tp]
\caption{Canal MAE (cm) by lead-time band in Miami-Dade (mean\,$\pm$\,SD over three seeds; Ref.: issue-time reference; P+R: Physics + residual; $^{\dagger}$own training-run evaluation).}
\label{tab:s-forecast-lead}
\setlength{\tabcolsep}{4pt}%
\resizebox{\columnwidth}{!}{%
\begin{tabular}{@{}cclr@{}lr@{}lr@{}l@{}}
\toprule
 & & Model & \multicolumn{2}{c}{1--6 h} & \multicolumn{2}{c}{7--24 h} & \multicolumn{2}{c}{25--48 h} \\
\midrule
\multirow{13}{*}{\rotatebox[origin=c]{90}{2019 (test)}} & \multirow{9}{*}{\rotatebox[origin=c]{90}{$q_0$}} & Persistence & 2.77 &  & 3.59 &  & 4.56 &  \\
\cmidrule(l){3-9}
 &  & Ref. GNN & 1.09 & \,$\pm$\,0.01 & 2.13 & \,$\pm$\,0.01 & 3.19 & \,$\pm$\,0.01 \\
 &  & Scenario GNN & 1.10 & \,$\pm$\,$<$0.01 & 2.13 & \,$\pm$\,$<$0.01 & 3.18 & \,$\pm$\,0.01 \\
\cmidrule(l){3-9}
 &  & Ref. Hybrid & 1.23 & \,$\pm$\,0.02 & 2.29 & \,$\pm$\,0.02 & 3.23 & \,$\pm$\,0.02 \\
 &  & Hybrid & 1.26 & \,$\pm$\,0.01 & 2.34 & \,$\pm$\,0.02 & 3.27 & \,$\pm$\,0.01 \\
\cmidrule(l){3-9}
 &  & Ref. P+R & 2.09 & \,$\pm$\,$<$0.01 & 2.96 & \,$\pm$\,0.02 & 3.99 & \,$\pm$\,0.15 \\
 &  & Physics + residual & 2.09 & \,$\pm$\,0.01 & 3.04 & \,$\pm$\,0.02 & 4.13 & \,$\pm$\,0.03 \\
\cmidrule(l){3-9}
 &  & Ref. P-only$^{\dagger}$ & 2.07 & \,$\pm$\,$<$0.01 & 3.05 & \,$\pm$\,0.01 & 4.05 & \,$\pm$\,0.01 \\
 &  & Physics-only & 2.11 & \,$\pm$\,0.01 & 3.20 & \,$\pm$\,0.01 & 4.60 & \,$\pm$\,0.06 \\
\cmidrule(l){2-9}
 & \multirow{4}{*}{\rotatebox[origin=c]{90}{Hindcast}} & Scenario GNN & 1.11 & \,$\pm$\,$<$0.01 & 2.07 & \,$\pm$\,0.01 & 2.99 & \,$\pm$\,0.01 \\
 &  & Hybrid & 1.22 & \,$\pm$\,0.01 & 2.17 & \,$\pm$\,0.01 & 2.95 & \,$\pm$\,$<$0.01 \\
 &  & Physics + residual & 2.03 & \,$\pm$\,0.01 & 2.81 & \,$\pm$\,0.02 & 3.64 & \,$\pm$\,0.01 \\
 &  & Physics-only & 2.05 & \,$\pm$\,0.01 & 2.97 & \,$\pm$\,0.01 & 4.02 & \,$\pm$\,0.02 \\
\midrule
\multirow{13}{*}{\rotatebox[origin=c]{90}{2018 (validation)}} & \multirow{9}{*}{\rotatebox[origin=c]{90}{$q_0$}} & Persistence & 2.51 &  & 3.25 &  & 4.16 &  \\
\cmidrule(l){3-9}
 &  & Ref. GNN & 1.03 & \,$\pm$\,0.02 & 1.93 & \,$\pm$\,0.01 & 2.90 & \,$\pm$\,0.01 \\
 &  & Scenario GNN & 1.04 & \,$\pm$\,0.01 & 1.94 & \,$\pm$\,0.01 & 2.89 & \,$\pm$\,0.01 \\
\cmidrule(l){3-9}
 &  & Ref. Hybrid & 1.14 & \,$\pm$\,0.02 & 2.06 & \,$\pm$\,0.02 & 2.93 & \,$\pm$\,0.03 \\
 &  & Hybrid & 1.19 & \,$\pm$\,0.01 & 2.14 & \,$\pm$\,0.01 & 2.99 & \,$\pm$\,0.02 \\
\cmidrule(l){3-9}
 &  & Ref. P+R & 1.83 & \,$\pm$\,$<$0.01 & 2.64 & \,$\pm$\,0.02 & 3.48 & \,$\pm$\,0.01 \\
 &  & Physics + residual & 1.87 & \,$\pm$\,0.01 & 2.77 & \,$\pm$\,0.01 & 3.66 & \,$\pm$\,0.02 \\
\cmidrule(l){3-9}
 &  & Ref. P-only$^{\dagger}$ & 1.81 & \,$\pm$\,$<$0.01 & 2.70 & \,$\pm$\,0.01 & 3.65 & \,$\pm$\,0.01 \\
 &  & Physics-only & 1.89 & \,$\pm$\,0.01 & 2.96 & \,$\pm$\,0.01 & 4.09 & \,$\pm$\,0.01 \\
\cmidrule(l){2-9}
 & \multirow{4}{*}{\rotatebox[origin=c]{90}{Hindcast}} & Scenario GNN & 1.05 & \,$\pm$\,0.01 & 1.88 & \,$\pm$\,0.01 & 2.73 & \,$\pm$\,$<$0.01 \\
 &  & Hybrid & 1.16 & \,$\pm$\,$<$0.01 & 2.01 & \,$\pm$\,0.01 & 2.69 & \,$\pm$\,$<$0.01 \\
 &  & Physics + residual & 1.83 & \,$\pm$\,0.01 & 2.58 & \,$\pm$\,0.01 & 3.28 & \,$\pm$\,0.03 \\
 &  & Physics-only & 1.83 & \,$\pm$\,0.01 & 2.75 & \,$\pm$\,0.01 & 3.69 & \,$\pm$\,0.01 \\
\bottomrule
\end{tabular}}
\end{table}

\subsection{Simulated drainage network}\label{app:F2}

\par\textbf{Protocol.}
Table~\ref{tab:s-betafc} scores the four structures on the \res{data.beta.n-storms-test} test storms
(\res{data.beta.windows-test} windows every 2 steps, that is, issue times every 30~min, \res{data.beta.n-scored} scored
nodes and a horizon of 24 steps or 6~h) under two information sets with controls held, namely the perfect-driver
hindcast (realized rain and outlet stage, a diagnostic reference) and the simulator's
issue-time scenario (no future rain, because no rain forecast exists, and surge held at its last value). It differs
from the Miami-Dade $q_0$, which carries a statistical rain forecast. Persistence repeats the last depth. NSE is the
median over non-flat nodes, and CSI is taken at each node's training-period p95.

\par
\begin{figure}[tbp]
\centering
\includegraphics[width=\columnwidth]{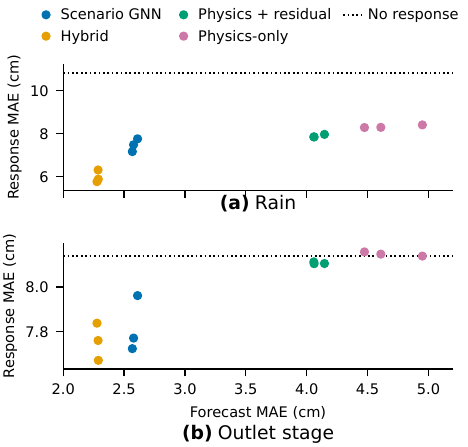}
\caption{Forecast error vs.\ response error under added rain $\bar A$ and raised outlet stage $\bar s$ on the simulated coastal network (one marker per model and seed; dotted: no-response reference).}
\label{fig:s-forecast-response}
\end{figure}

\textbf{Results.}
With realized drivers the Hybrid forecaster and the Scenario GNN reach NSE \res{beta.fc.hybrid.real.nse.mean} and
\res{beta.fc.gnn.real.nse.mean}, ahead of the two physics models. Without future rain all four structures keep about half
of their NSE or less (NSE from \res{beta.fc.physonly.q0.nse.mean} to \res{beta.fc.hybrid.q0.nse.mean}), and persistence has the
highest CSI under this scenario. The forecast axis of Fig.~\ref{fig:s-forecast-response} is the perfect-driver hindcast MAE of this table, per model and
seed, plotted against the response MAE of the paired runs on the same test storms (Eq.~\eqref{eq:paired}); the
window-peak MAE at the \res{data.beta.cf-issue-times} counterfactual issue times, an alternative forecast axis, gives
the same model order.

The response MAE of that figure is the mean $|\Delta_{\mathrm{model}}-\Delta_{\mathrm{sim}}|$ of the 6-h window-peak
change over the storm\,$\times$\,node cells of Fig.~\ref{fig:driver-info}b, with $\bar A$ at the
\res{beta.nodes.rain} runoff-receiving nodes and $\bar s$ at the \res{beta.nodes.near} near-outlet nodes. Its
no-response line is the error of a model whose peak never changes. The outlet-stage
response error follows the forecast-error order but stays within \res{beta.k.stage.mae.reduction-vs-zero.max}\% of the
no-response error, unlike the rain-response error.

\par\textbf{Training stability.}
All three Physics + state residual runs on the simulator stopped early, after very large epoch-mean gradient norms
before clipping, after 12, 13 and 15 of 30 epochs (Table~\ref{tab:s-training} gives the mean and SD of epochs run and
best epochs, and App.~\ref{app:C} the largest norm). Two of the three Physics-only runs also stopped before the epoch
limit, after 22 and 18 epochs. Their
scores are those of the checkpoints selected by validation loss under the common rule.

\begin{table}[t]
\caption{Forecast scores on the simulated coastal network (test storms, 6-h horizon, mean\,$\pm$\,SD over three seeds).}
\label{tab:s-betafc}
\setlength{\tabcolsep}{4pt}%
\resizebox{\linewidth}{!}{%
\begin{tabular}{@{}clcccc@{}}
\toprule
\multicolumn{2}{@{}l}{Model} & NSE & MAE (cm) & RMSE (cm) & CSI \\
\midrule
\multirow{4}{*}{\rotatebox[origin=c]{90}{Hindcast}} & Scenario GNN & 0.934\,$\pm$\,0.001 & 2.59\,$\pm$\,0.02 & 6.91\,$\pm$\,0.08 & 0.601\,$\pm$\,0.013 \\
 & Hybrid & 0.944\,$\pm$\,0.001 & 2.28\,$\pm$\,0.01 & 6.14\,$\pm$\,0.01 & 0.625\,$\pm$\,0.008 \\
 & Physics + residual & 0.807\,$\pm$\,0.009 & 4.09\,$\pm$\,0.05 & 10.40\,$\pm$\,0.15 & 0.491\,$\pm$\,0.049 \\
 & Physics-only & 0.674\,$\pm$\,0.054 & 4.68\,$\pm$\,0.25 & 11.33\,$\pm$\,0.41 & 0.361\,$\pm$\,0.021 \\
\midrule
\multirow{4}{*}{\rotatebox[origin=c]{90}{Issue time}} & Scenario GNN & 0.423\,$\pm$\,0.008 & 5.19\,$\pm$\,$<$0.01 & 16.45\,$\pm$\,0.04 & 0.250\,$\pm$\,0.002 \\
 & Hybrid & 0.479\,$\pm$\,0.013 & 4.70\,$\pm$\,0.08 & 15.44\,$\pm$\,0.28 & 0.274\,$\pm$\,0.010 \\
 & Physics + residual & 0.383\,$\pm$\,0.023 & 6.08\,$\pm$\,0.05 & 17.31\,$\pm$\,0.06 & 0.259\,$\pm$\,0.026 \\
 & Physics-only & 0.305\,$\pm$\,0.009 & 6.36\,$\pm$\,0.14 & 17.56\,$\pm$\,0.08 & 0.253\,$\pm$\,0.002 \\
\midrule
\multicolumn{2}{@{}l}{Persistence} & 0.267 & 6.32 & 18.47 & 0.302 \\
\bottomrule
\end{tabular}}
\end{table}

\section{Attribution References, Faithfulness, and Structural Priors}\label{app:G}

This appendix gives the complete attribution evidence behind the single-driver results, namely the attribution distributions
on Miami-Dade, the meaning of the reference scenario, history attribution and the explanation baselines, Shapley
recovery on the simulator and the untrained controls. It treats three questions separately and does not merge them
into one explanation score. The three are whether an attribution is faithful to the model it explains, whether it agrees with the
simulator under the same query, and whether its features come from training or from the architecture alone.

\subsection{Future-driver attribution on Miami-Dade}\label{app:G1}

\par\textbf{Computation.}
For every canal gauge and every query issue time (every 6~h, \res{mia.attr.n-issue} per year), the coalition value
$v(\mathcal C)$ is the 48-h forecast peak margin with the players in $\mathcal C\subseteq\{R,S,T\}$ at their
issue-time values and the others at the reference scenario, and $\phi_R,\phi_S,\phi_T$ follow exactly from the eight
coalitions of Eq.~\eqref{eq:app-shapley}. The three values sum to $v(\{R,S,T\})-v(\emptyset)$ within
\res{mia.attr.eff-resid.max}~cm over all cells, models, seeds and years. This efficiency is a property of the exact
Shapley computation~\cite{shapley1953value} and confirms faithfulness to the model, not physical correctness.
Untrained copies are evaluated with the same coalitions, cells and reference.

\par\textbf{Coverage.}
Every run evaluates all \res{mia.attr.n-issue} issue times of both years (for the Hybrid forecaster's seed-42 test-year
attribution, \res{mia.attr.hybrid-s42-test.n-issue} issue times were evaluated, so pooled Hybrid test-year
distributions use the full-year seeds 43 and 44).

\begin{figure*}[tp]
\centering
\includegraphics[width=\textwidth]{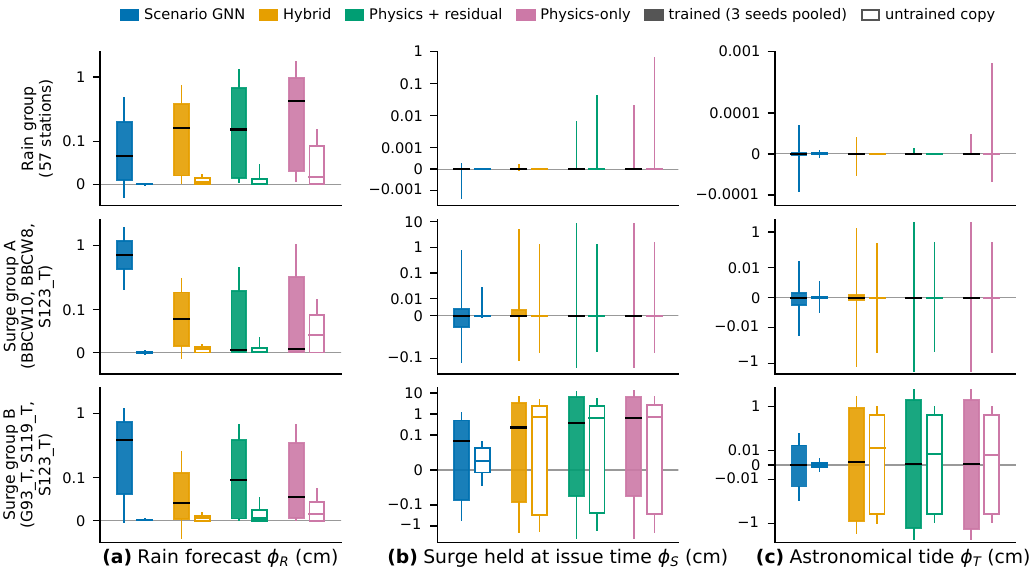}
\caption{Exact Shapley distributions (cm, symmetric-log scale) for the 48-h peak margin across Miami-Dade canal gauges in 2019 (median, IQR, and 10th--90th percentiles).}
\label{fig:s-miami-shapley}
\end{figure*}

\par\textbf{Results.}
Fig.~\ref{fig:s-miami-shapley} shows the distributions over (issue time, gauge) cells, with operations and history held
in every coalition. Its rows are the rain group, surge group~A (selected from 2018 validation stations), and surge
group~B (the tidal-edge gauges). The values
describe how each model uses the issue-time drivers relative to the reference scenario of App.~\ref{app:G2}, not driver
effects in the canal system. In the rain group the forecast rain adds little to most peaks.
The median $\phi_R$ runs from \res{mia.attr.test.rain.gnn.trained.phir.p50}~cm (Scenario GNN) to
\res{mia.attr.test.rain.physonly.trained.phir.p50}~cm (Physics-only model), and the 90th percentile from
\res{mia.attr.test.rain.gnn.trained.phir.p90} to \res{mia.attr.test.rain.physonly.trained.phir.p90}~cm. The untrained
copies attribute at most \res{mia.attr.test.rain.physonly.untrained.phir.p90}~cm at their 90th percentile, so the rain
attribution comes from training. For the trained models, surge and tide contributions at rain-group gauges stay within
a few hundredths of a centimeter. At the tidal-edge gauges of surge group~B the picture reverses. The 90th percentile of $\phi_S$ is
\res{mia.attr.test.surge-b.gnn.trained.phis.p90}~cm for the Scenario GNN and
from \res{mia.attr.test.surge-b.hybrid.trained.phis.p90} to \res{mia.attr.test.surge-b.physonly.trained.phis.p90}~cm for the
three physics-branch models, while their untrained copies already reach between
\res{mia.attr.test.surge-b.hybrid.untrained.phis.p90} and \res{mia.attr.test.surge-b.physonly.untrained.phis.p90}~cm and
the untrained Scenario GNN \res{mia.attr.test.surge-b.gnn.untrained.phis.p90}~cm. Physics-branch structure therefore
supplies surge attributions of the same order before any training. Surge group~A mixes two gauges with essentially no
modeled surge response (BBCW10, BBCW8) with S123\_T, so its $\phi_S$ and $\phi_T$ distributions collapse near zero with
long whiskers. The tide contribution is centered on zero at the tidal-edge gauges. It measures the window's tidal-range
anomaly against the training-median range, not the tide itself.

The distributions of Fig.~\ref{fig:s-miami-shapley} take the place of dominant-player shares, since an arg max over
players defaults to rain wherever all three contributions are near zero. Taken together, the Miami-Dade attributions
describe how each model uses the issue-time drivers. They establish that the forecast rain moves most rain-group peaks
little and that physics-branch structure produces surge contributions before training.

\subsection{What the reference scenario means}\label{app:G2}

A Shapley value answers a question relative to its reference~\cite{sundararajan2020many}. Here the reference sets
future rain and surge to zero and rescales the astronomical tide to the same-phase tide with the training-median 48-h
range, while operations and history stay fixed. Three consequences follow. First, $v(\emptyset)$ is the model's
forecast under this reference, not a baseline water level of the canal. It contains the history, drainage, the held
operations and the reference tide. The antecedent state is not a player, so the decomposition does not attribute
anything to antecedent storage. Second, $\phi_S$ compares the surge held at the issue time with no surge. It is
negative when the held surge is a set-down and says nothing about future surge changes; $\phi_T$ compares the window's
tidal range with the training-median range.

Third, a different reference changes the question, not only the numbers. Taking $q_0$ itself as the reference would
make every Miami-Dade value zero, because the players are queried at their $q_0$ values. A seasonally matched reference
would measure departures from typical conditions instead of departures from no rain and no surge. On the simulator the
reference has the same definition for the model and the simulator, namely no future rain, no future surge and the
astronomical part rescaled about its issue-time value to the training-median 6-h range of
\res{data.beta.astro-range-median-cm}~cm. The recovery test below therefore asks one question of both under this
reference.

\subsection{History attribution and explanation baselines}\label{app:G3}

Three explanation baselines are compared in Table~\ref{tab:s-xai} with the grid answer of the same model for the
minimum added rain or surge that makes the 48-h peak cross the p95 threshold (the in-model reference), as
mean\,$\pm$\,SD over seeds. Ranges
quoted below run over seeds. They
describe the models, not the water system.

\begin{table}[t]
\caption{Explanation baselines against the in-model grid-search reference in Miami-Dade (2019, \%; relative error as a fraction; mean\,$\pm$\,SD over three seeds; $^{\ddagger}$seeds 43--44).}
\label{tab:s-xai}
\centering
\setlength{\tabcolsep}{3pt}%
\resizebox{\columnwidth}{!}{%
\begin{tabular}{@{}lr@{}lr@{}lr@{}lr@{}l@{}}
\toprule
Measure & \multicolumn{2}{c}{\begin{tabular}[b]{@{}c@{}}Scenario\\GNN\end{tabular}} & \multicolumn{2}{c}{Hybrid} & \multicolumn{2}{c}{\begin{tabular}[b]{@{}c@{}}Physics +\\residual\end{tabular}} & \multicolumn{2}{c}{\begin{tabular}[b]{@{}c@{}}Physics-\\only\end{tabular}} \\
\midrule
\multicolumn{9}{@{}c}{\emph{Linear extrapolation}} \\
Miss & 92 & \,$\pm$\,2 & 56 & \,$\pm$\,15 & 53 & \,$\pm$\,4 & 33 & \,$\pm$\,1 \\
Relative error & 0.48 & \,$\pm$\,0.05 & 0.15 & \,$\pm$\,0.08 & 0.20 & \,$\pm$\,0.01 & 0.06 & \,$\pm$\,$<$0.01 \\
\midrule
\multicolumn{9}{@{}c}{\emph{Single-factor analysis}} \\
Miss, rain & 2 & \,$\pm$\,$<$1 & 0 & \,$\pm$\,$<$1 & 2 & \,$\pm$\,$<$1 & 6 & \,$\pm$\,$<$1 \\
Miss, surge A & 24 & \,$\pm$\,5 & 14 & \,$\pm$\,9 & 0 & \,$\pm$\,0 & 0 & \,$\pm$\,0 \\
\midrule
\multicolumn{9}{@{}c}{\emph{Integrated gradients (IG)}} \\
Top = history & 100 & \,$\pm$\,0 & 100 & \,$\pm$\,0 & 100 & \,$\pm$\,$<$1 & 100 & \,$\pm$\,$<$1 \\
Agreement, rain & 92 & \,$\pm$\,4 & 77 & \,$\pm$\,3 & 100 & \,$\pm$\,0 & 100 & \,$\pm$\,$<$1 \\
\quad constant & 94 & \,$\pm$\,4 & 77 & \,$\pm$\,3 & 100 & \,$\pm$\,0 & 100 & \,$\pm$\,0 \\
Agreement, surge A & 77 & \,$\pm$\,$<$1$^{\ddagger}$ & 84 & \,$\pm$\,1 & 66 & \,$\pm$\,1 & 64 & \,$\pm$\,$<$1 \\
\quad constant & 96 & \,$\pm$\,3$^{\ddagger}$ & 70 & \,$\pm$\,8 & 59 & \,$\pm$\,1 & 58 & \,$\pm$\,$<$1 \\
\midrule
\multicolumn{9}{@{}c}{\emph{Unconstrained counterfactual}} \\
Success & 100 & \,$\pm$\,0 & 100 & \,$\pm$\,0 & 100 & \,$\pm$\,0 & 100 & \,$\pm$\,0 \\
On history & 89 & \,$\pm$\,1 & 89 & \,$\pm$\,1 & 80 & \,$\pm$\,2 & 46 & \,$\pm$\,1 \\
Negative rain & 100 & \,$\pm$\,0 & 100 & \,$\pm$\,0 & 100 & \,$\pm$\,$<$1 & 100 & \,$\pm$\,$<$1 \\
\bottomrule
\end{tabular}}
\end{table}

\par\textbf{Integrated gradients.}
Integrated gradients~\cite{sundararajan2017axiomatic} of a gauge's $q_0$ peak, with a zero baseline and 32 steps, are
summed in absolute value over each input group. The observed water-level history is the largest group in between
\res{mia.xai.test.rain.physonly.integrated.ig-share-top-history.min} and \res{mia.xai.test.rain.gnn.integrated.ig-share-top-history.max}\%
of rain-group station-windows. IG thus answers what the model looked at, which here is mainly the canal's own recent
level, not which driver change would produce a crossing. Its two-driver ranking (future rain against future surge)
agrees with the in-model reference in between
\res{mia.xai.test.rain.gnn.integrated.ig-driver-agreement.min} and \res{mia.xai.test.rain.gnn.integrated.ig-driver-agreement.max}\%
of rain-group cells for the Scenario GNN, but always naming the more frequent easier driver, rain, reaches between
\res{mia.xai.test.rain.gnn.integrated.constant-answer-agreement.min} and \res{mia.xai.test.rain.gnn.integrated.constant-answer-agreement.max}\%.
Where surge is the easier driver in the rain group, IG agrees in
\res{mia.xai.test.rain.gnn.integrated.agree-when-surge-easier.max}\% of cells for both data-driven models (for the
physics models the grid never finds surge easier there). In surge group~A, IG
exceeds the constant answer only for the Hybrid forecaster and the two physics models. The agreement rates are
therefore not evidence that IG ranks drivers correctly.

\par\textbf{Unconstrained counterfactual.}
This baseline, our variant of the ForecastCF counterfactual~\cite{wang2023counterfactual}, searches by gradient descent
for an additive change to every input, history included, until the margin reaches 1~cm; unlike the original method it
imposes no plausibility bound. It succeeds in every rain-group case. The share of the change placed on the observed history is between
\res{mia.xai.test.rain.gnn.unconstrained.cf-mean-share-on-observed-history.min} and \res{mia.xai.test.rain.gnn.unconstrained.cf-mean-share-on-observed-history.max}\%
for the Scenario GNN but between
\res{mia.xai.test.rain.physonly.unconstrained.cf-mean-share-on-observed-history.min} and \res{mia.xai.test.rain.physonly.unconstrained.cf-mean-share-on-observed-history.max}\%
for the Physics-only model, and between
\res{mia.xai.test.rain.physres.unconstrained.cf-share-negative-rain.min} and \res{mia.xai.test.rain.gnn.unconstrained.cf-share-negative-rain.max}\%
of the counterfactuals contain negative rain somewhere. An unconstrained input counterfactual therefore mostly rewrites
the past and leaves the space of physical scenarios, which is why the scenario queries specify the change instead of
searching for it.

\par\textbf{Linear extrapolation and single-factor analysis.}
Extrapolating the $q_0$ margin linearly from the first joint-grid step misses from
\res{mia.xai.test.rain.gnn.linear.miss-rate.min} to \res{mia.xai.test.rain.gnn.linear.miss-rate.max}\% of the
grid-found rain crossings for the Scenario GNN and from
\res{mia.xai.test.rain.physonly.linear.miss-rate.min} to \res{mia.xai.test.rain.physonly.linear.miss-rate.max}\% for
the Physics-only model. Where both find a crossing, the median relative error of the extrapolated amount lies between
\res{mia.xai.test.rain.gnn.linear.median-rel-error-where-both.min} and \res{mia.xai.test.rain.gnn.linear.median-rel-error-where-both.max}
for the Scenario GNN and is \res{mia.xai.test.rain.physonly.linear.median-rel-error-where-both.max} for the Physics-only
model. The Scenario GNN's rain response is thus far from linear in the added amount. Joint-grid crossings that neither
the largest rain nor the largest surge alone produces make up at most
\res{mia.xai.test.rain.physonly.single.miss-rate.max}\% of rain-group crossings, but from
\res{mia.xai.test.surge-a.gnn.single.miss-rate.min} to \res{mia.xai.test.surge-a.gnn.single.miss-rate.max}\% for the
Scenario GNN in surge group~A. The missed crossings of the linear extrapolation are counted on unwarned, monotone
cells. Per seed, the rain group of these baselines has 120 issue times $\times$ 10 stations and surge group~A 240
issue times $\times$ 3 stations (for the Scenario GNN in surge group~A, where seed 42 evaluated 8 issue times, the
agreement values marked with a double dagger in Table~\ref{tab:s-xai} use the full-sample seeds 43 and 44). The gradient
baselines cover the test year 2019.

\subsection{Shapley recovery on the simulator}\label{app:G4}

\par\textbf{Protocol.}
The simulator's Shapley values come from the eight coalition runs of each issue time, and the model's from the same
eight scenarios. Issue times whose coalition runs all pass the solver-quality rule are used
(\res{beta.quality.shapley-usable-primary} of \res{data.beta.cf-issue-times}). The rank correlation is computed across
the \res{beta.nodes.scored} scored nodes per issue time and summarized by its median. Errors use the discernible cells
where the simulated value exceeds twice the noise floor, \res{beta.fig2a.share-disc.r}\% of cells for rain and
\res{beta.fig2a.share-disc.s}\% for surge. Intervals are 95\% percentile intervals from 2,000 storm resamples of the
seed-mean median.

\par\textbf{Rain.}
Median rank correlations and storm intervals are \res{beta.fig2a.r.gnn.rho.mean}
[\res{beta.fig2a.r.gnn.rho.boot-lo}, \res{beta.fig2a.r.gnn.rho.boot-hi}] for the Scenario GNN,
\res{beta.fig2a.r.hybrid.rho.mean} [\res{beta.fig2a.r.hybrid.rho.boot-lo}, \res{beta.fig2a.r.hybrid.rho.boot-hi}] for
the Hybrid forecaster, \res{beta.fig2a.r.physres.rho.mean} [\res{beta.fig2a.r.physres.rho.boot-lo},
\res{beta.fig2a.r.physres.rho.boot-hi}] for the Physics + state residual model and \res{beta.fig2a.r.physonly.rho.mean}
[\res{beta.fig2a.r.physonly.rho.boot-lo}, \res{beta.fig2a.r.physonly.rho.boot-hi}] for the Physics-only model, against
values from \res{beta.fig2a.r.hybrid-untrained.rho} to \res{beta.fig2a.r.physres-untrained.rho} for the untrained copies. The
Scenario GNN and the Hybrid forecaster are tied, and their order reverses under several of the sensitivity variants of
Table~\ref{tab:s-sensitivity-shapley}. Every signed error is negative. The mean of model
minus simulator over discernible cells is \res{beta.fig2a.r.gnn.signed-err.mean} (Scenario GNN),
\res{beta.fig2a.r.hybrid.signed-err.mean} (Hybrid forecaster), \res{beta.fig2a.r.physres.signed-err.mean} (Physics +
state residual) and \res{beta.fig2a.r.physonly.signed-err.mean}~cm (Physics-only model), and
\res{beta.fig2a.r.gnn-untrained.signed-err} to \res{beta.fig2a.r.physres-untrained.signed-err}~cm for the untrained
copies, close to minus the no-response reference of \res{beta.fig2a.true-abs-phi-disc.r}~cm. The trained models thus
order nodes by their simulated rain effect and under-attribute its size.

\par\textbf{Surge.}
For the outlet stage the rank correlations range from \res{beta.fig2a.s.hybrid.rho.mean} to \res{beta.fig2a.s.gnn.rho.mean} for
the Hybrid forecaster and the Scenario GNN and from \res{beta.fig2a.s.physonly.rho.mean} to \res{beta.fig2a.s.physres.rho.mean}
for the physics models, whose storm intervals stay above zero (lower bounds \res{beta.fig2a.s.physonly.rho.boot-lo} and
\res{beta.fig2a.s.physres.rho.boot-lo}). Their untrained copies reach \res{beta.fig2a.s.physonly-untrained.rho} and
\res{beta.fig2a.s.physres-untrained.rho}. Absolute errors of the trained models, from
\res{beta.fig2a.s.hybrid.abs-err.mean} to \res{beta.fig2a.s.physres.abs-err.mean}~cm, equal the no-response reference of
\res{beta.fig2a.true-abs-phi-disc.s}~cm within a few hundredths of a centimeter, and the signed errors are close to its
negative. The small positive surge ranking of the physics models therefore comes with no recovered size, and part of
it is present before training. No model recovers the simulated surge contribution.

\subsection{Untrained copies as structural controls}\label{app:G5}

\par\textbf{What the control tests.}
The untrained copy of each architecture is one random initialization with the same standardization, graph, windows,
scenarios and reference as the trained runs. It separates what training adds from what the fixed parts of an
architecture produce. For the physics-branch models these fixed parts are substantial, namely the graph, the flux signs,
the initial head offsets, the storage areas and the rain reservoirs. An untrained copy is therefore a control for
structure, not a model without prior~\cite{adebayo2018sanity}.

\par\textbf{What it shows.}
The control reads differently by driver and system. On the simulator the rain ranking requires training. Untrained rank
correlations stay far below the trained ones. On Miami-Dade, physics-branch copies attribute surge contributions at the
tidal-edge gauges of the same order as the trained models before any training
(Fig.~\ref{fig:s-miami-shapley}). On surge group~A they reach station fixed-effect slopes of the observational check
between \res{mia.obs.test.surge-a.hybrid.fe.untrained.slope} and \res{mia.obs.test.surge-a.physres.fe.untrained.slope}
(Table~\ref{tab:s-obscheck}). On the simulator their sign agreement for rain is already high (Table~\ref{tab:s-simrows}),
which decides the strict-rule verdicts of the physics models. Agreement of an untrained physics copy with the
observational check thus shows structure, not learned physics.

\par\textbf{Scope.}
Each architecture has one initialization, so the control carries no spread across initializations. The randomization
is applied to the whole network at once, and the faithfulness of the history attribution rests on the axioms of
integrated gradients.

\section{Observational Diagnostics and Identification Limits}\label{app:H}

This appendix documents the observational check of Eq.~\eqref{eq:substitution} and its results.
It gives the proof of the substitution identity with its assumptions, the within-station version, the conditions under which
the check carries information, the station groups and decision rules, and the counterexamples on
the simulators. The check needs no counterfactual labels, which is its value for an observed system, and the results
below define what its verdicts establish.

\subsection{The substitution identity}\label{app:H1}

\par\textbf{Setting.}
Fix a driver $D$ and a set of $n$ cells $c$, each an (issue time, gauge) pair with at least 40 of 48 horizon hours
observed. Let $B_c$ be the model's forecast peak with $D$ at its issue-time value and every other driver realized,
$F_c$ the peak with all drivers realized and $Y_c$ the observed peak (App.~\ref{app:B2} defines the
substitution baselines). $B_c$ and $F_c$ are maxima over all 48 leads, whereas $Y_c$ is the maximum over the observed
hours only. Define the predicted change $\Delta_{p,c}=F_c-B_c$, the observed discrepancy
$\Delta_{o,c}=Y_c-B_c$ and the model's peak error under realized drivers $e_c=Y_c-F_c$, which also contains the
difference between the two peak windows. With sample means $\bar x$,
$\widehat{\operatorname{Cov}}(x,y)=n^{-1}\sum_c(x_c-\bar x)(y_c-\bar y)$ and
$\widehat{\operatorname{Var}}(x)=\widehat{\operatorname{Cov}}(x,x)$, the check fits ordinary least squares (OLS) with
intercept,
\begin{equation}
\hat\beta=\frac{\widehat{\operatorname{Cov}}(\Delta_p,\Delta_o)}{\widehat{\operatorname{Var}}(\Delta_p)},\qquad
\hat\alpha=\bar\Delta_o-\hat\beta\,\bar\Delta_p .
\label{eq:app-ols}
\end{equation}
This is the Mincer-Zarnowitz regression of outcomes on forecasts~\cite{mincer1969evaluation} applied to changes, with
the observed peak $Y_c$ as the outcome, the realized-driver peak $F_c$ as the forecast and the shared baseline $B_c$
subtracted from both.

\par\textbf{Proposition 1 (substitution identity of Eq.~\eqref{eq:substitution}).}
(i)~If $\widehat{\operatorname{Var}}(\Delta_p)>0$, then
\begin{equation}
\hat\beta=1+\frac{\widehat{\operatorname{Cov}}(\Delta_p,e)}{\widehat{\operatorname{Var}}(\Delta_p)},\qquad
\hat\alpha=\bar e-(\hat\beta-1)\,\bar\Delta_p .
\label{eq:app-identity}
\end{equation}
(ii)~If the cells are draws of random variables $(\Delta_p,Y-F)$ with finite variances and
$\operatorname{Var}(\Delta_p)>0$, the population slope is $\beta=1+\operatorname{Cov}(\Delta_p,Y-F)/\operatorname{Var}(\Delta_p)$,
which is Eq.~\eqref{eq:substitution} of the main text.

\par\textbf{Proof.}
Cell by cell, $\Delta_{o,c}=Y_c-B_c=(F_c-B_c)+(Y_c-F_c)=\Delta_{p,c}+e_c$. This step uses no property of the model,
the drivers or the water system. Covariance is bilinear, so
$\widehat{\operatorname{Cov}}(\Delta_p,\Delta_o)=\widehat{\operatorname{Var}}(\Delta_p)+\widehat{\operatorname{Cov}}(\Delta_p,e)$,
and dividing by $\widehat{\operatorname{Var}}(\Delta_p)>0$ gives the slope. For the intercept,
$\bar\Delta_o=\bar\Delta_p+\bar e$, so $\hat\alpha=\bar\Delta_p+\bar e-\hat\beta\bar\Delta_p$. Part~(ii) repeats the
argument with expectations. Finite variances make every covariance finite by the Cauchy-Schwarz inequality, and
$\operatorname{Var}(\Delta_p)>0$ makes the slope defined. $\square$

If $\widehat{\operatorname{Var}}(\Delta_p)=0$, because the model does not respond or responds equally in every cell,
the slope is UD. A slope of 1 therefore means exactly that the model's error under realized drivers does
not covary with its predicted substitution response in these cells. Whether the predicted response is right needs a
further assumption, made explicit next.

\par\textbf{Corollary 1 (what a calibration reading would require).}
Let $Y^B_c$ be the peak that would have been observed had $D$ followed its issue-time value while every other driver,
operations included, followed its realized path, and $Y^B_c$ is never observed. Write
$\Delta^{\mathrm{true}}_c=Y_c-Y^B_c$ for the true substitution response and $b_c=Y^B_c-B_c$ for the model's error in
that unobserved state. Then $e_c=(\Delta^{\mathrm{true}}_c-\Delta_{p,c})+b_c$, and under the assumptions of
Proposition~1(ii), extended to these variables,
\begin{equation}
\beta=\frac{\operatorname{Cov}(\Delta_p,\Delta^{\mathrm{true}})+\operatorname{Cov}(\Delta_p,b)}{\operatorname{Var}(\Delta_p)} .
\label{eq:app-calibration}
\end{equation}
\emph{Proof.} Substitute the decomposition of $e$ into Proposition~1(ii). $\square$

The slope equals the regression slope of the true substitution response on the predicted one only if
$\operatorname{Cov}(\Delta_p,b)=0$, an assumption about an error in an unobserved state that the data cannot check.
Because $B$ enters both $\Delta_p=F-B$ and $b=Y^B-B$, a model error at the baseline can covary with $\Delta_p$. This is
the shared baseline named in the main text. A slope near 1 is thus consistent with a predicted substitution response
that is calibrated on average (a slope of 1 of the true on the predicted response, which still allows cell-by-cell
errors) together with an uncorrelated baseline error, and equally with a miscalibrated response whose error is offset
by $\operatorname{Cov}(\Delta_p,b)$. Two scope conditions hold regardless of this assumption. The realized operations
are held on both sides, so $Y^B$ describes a world in which operators would not have acted differently. And the
substitution compares realized with issue-time drivers, whereas the scenario queries add a fixed amount of a driver to
a scenario, so even a correct substitution response would not establish the response to an added driver.

\subsection{Station fixed effects}\label{app:H2}

\par\textbf{Proposition 1 (continued), within-station version.}
For station $s$ with cell set $\mathcal C_s$, let $\tilde x_c=x_c-\bar x_s$ with $\bar x_s$ the station mean. The
station FE slope is $\hat\beta_{\mathrm{FE}}=\sum_c\tilde\Delta_{p,c}\tilde\Delta_{o,c}\big/\sum_c\tilde\Delta_{p,c}^2$.
(iii)~If $\sum_c\tilde\Delta_{p,c}^2>0$, that is, if $\Delta_p$ varies within at least one station, then
\begin{equation}
\hat\beta_{\mathrm{FE}}=1+\frac{\sum_c\tilde\Delta_{p,c}\,\tilde e_c}{\sum_c\tilde\Delta_{p,c}^2}.
\label{eq:app-fe}
\end{equation}
\emph{Proof.} Demeaning is linear, so $\tilde\Delta_{o,c}=\tilde\Delta_{p,c}+\tilde e_c$ cell by cell. Substituting
into the slope gives the result. $\square$

\par\textbf{Pooled, within-station and per-station slopes.}
With the within-station sums $W_{xy}=\sum_s\sum_{c\in\mathcal C_s}(x_c-\bar x_s)(y_c-\bar y_s)$ and the between-station
sums $Q_{xy}=\sum_s n_s(\bar x_s-\bar x)(\bar y_s-\bar y)$, the pooled slope is
$(W_{\Delta_p\Delta_o}+Q_{\Delta_p\Delta_o})/(W_{\Delta_p\Delta_p}+Q_{\Delta_p\Delta_p})$, whereas the FE slope keeps
only the within parts. Because $W_{xy}$ is the sum of the stations' own sums, the FE slope is a weighted mean of the
per-station slopes $\hat\beta_s$ with weights proportional to each station's $\sum_{c\in\mathcal C_s}\tilde\Delta_{p,c}^2$.
A station whose model does not respond gets weight zero. The pooled slope can therefore be governed by differences in
station means that have nothing to do with the response. Surge group~A shows this directly. BBCW10 and BBCW8 have essentially no modeled surge response in any trained model (mean $|\Delta_p|$ at
most \res{mia.obs.test.st.bbcw8.gnn.mean-abs-dp.max}~cm), yet for the two physics models their observed peaks exceed
the baseline forecast by between \res{mia.obs.test.st.bbcw8.physres.mean-do.min} and \res{mia.obs.test.st.bbcw8.physonly.mean-do.max}~cm
on average, whereas S123\_T has larger responses and a mean discrepancy of only
\res{mia.obs.test.st.s123t.physonly.mean-do.min}~cm for the Physics-only model. The between-station covariance is
negative, the pooled slopes are negative
(\res{mia.obs.test.surge-a.physonly.pooled.slope.min} to \res{mia.obs.test.surge-a.physres.pooled.slope.max}), and the
FE slope of the Physics-only model, \res{mia.obs.test.surge-a.physonly.fe.slope.min}, is essentially the per-station
slope of S123\_T (\res{mia.obs.test.st.s123t.physonly.slope.s42} in every seed).

\par\textbf{What FE removes and what it does not.}
Demeaning removes station-constant parts of $\Delta_p$ and $e$, such as datum offsets and a systematic bias of the
baseline at a gauge. It leaves every within-station covariance between $\Delta_p$ and $e$, including covariance from
time-varying confounding (for example realized operations that respond to the same storms) and from the shared baseline
of Corollary~1. FE is a correction for offsets, not an adjustment that identifies a response.

\subsection{When the check is informative}\label{app:H3}

\begin{figure*}[tp]
\centering
\includegraphics[width=\textwidth]{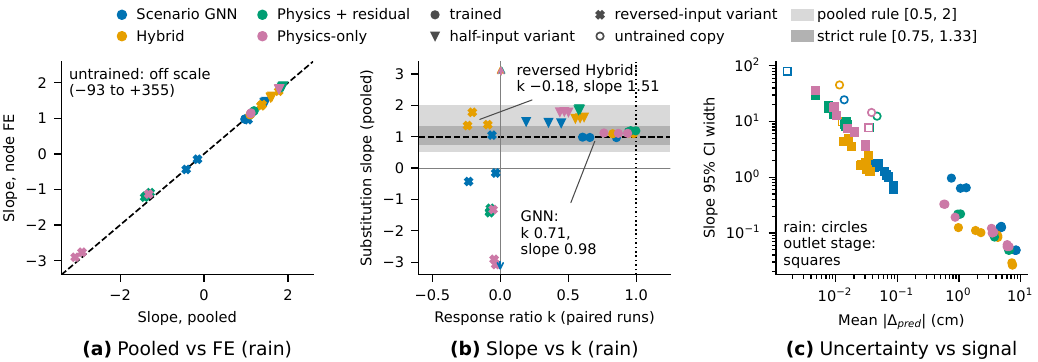}
\caption{Observational-check slopes vs.\ simulated response ratios $k$ on the simulated coastal network across models, variants, and seeds (95\% storm-block intervals in (c)).}
\label{fig:s-obscheck}
\end{figure*}

\par\textbf{Signal size.}
For OLS, the sampling variance of $\hat\beta$ is approximately the residual variance divided by
$n\,\widehat{\operatorname{Var}}(\Delta_p)$, and overlapping windows reduce the effective $n$. The intervals therefore
resample 7-day blocks or storms. When predicted responses are hundredths of a centimeter and residuals are
centimeters, the interval widens without bound and the point slope can take almost any value. Fig.~\ref{fig:s-obscheck}c
shows the interval width growing as the mean $|\Delta_p|$ shrinks on the simulator, and untrained copies with barely
any response produce slopes in the tens or hundreds (Tables~\ref{tab:s-obscheck} and~\ref{tab:s-simrows}).

\par\textbf{Driver variation and station mix.}
The check sees only the difference between the realized driver and its issue-time value. For surge this difference is
small outside surge events, and the Miami-Dade test year 2019 contains no large surge episode. For rain it is realized
minus forecast rain. Groups that mix stations with different offsets make the pooled slope reflect those offsets,
as the within-station decomposition above shows.

\par\textbf{Sign agreement.}
Agreement is computed on cells with $|\Delta_p|>\delta_o$. If the signs of $\Delta_p$ and $\Delta_o$ were independent,
the expected agreement would be $\pi_p\pi_o+(1-\pi_p)(1-\pi_o)$, where $\pi_p$ and $\pi_o$ are the shares of positive
values among those cells. It equals one half only when one of the two shares is one half. The agreement clauses (2)
and (2a) below compare with one half, not with this chance level, so with imbalanced signs a passed agreement
clause does not by itself show agreement above chance.

\par\textbf{Complete verdict criteria.}
Under the \emph{pooled rule} a driver passes if, for every seed, (1)~the pooled slope lies in $[0.5,2]$ with the
lower bound of its 95\% block interval above 0, (2)~the agreement interval lies above one half, (3)~the shuffled-driver
slope is smaller than half the slope in magnitude, and the untrained copy does not meet (1) and (2). If it does, the
verdict is ``not separable''. The \emph{station fixed-effect correction} applies the same criteria to within-station
demeaned quantities. The \emph{strict rule} first requires (0)~at least 100 cells with $|\Delta_p|>1$~cm and a
mean $|\Delta_p|$ of at least 0.1~cm (on demeaned values on Miami-Dade, on pooled values on the simulators), otherwise
the verdict is ``no information'' (``uninformative''). It then requires (1)~an FE slope in $[0.75,1.33]$ with
interval above 0, (2a)~an agreement interval above one half, (2b)~an agreement at least 0.10 above the untrained copy's
agreement and (3)~a shuffled-driver FE slope below a quarter of the slope in magnitude. An untrained copy that meets
(1) and (2a) gives ``cannot separate''. An untrained copy with no cell above 1~cm sets no bar in (2b) on Miami-Dade and
a bar of 0.6 on the simulators.

\par\textbf{Behavior of clause (2b).}
Clause~(2b) compares against the untrained copy's sign agreement even when that copy has fewer than 100 active cells. When an untrained physics-branch copy inherits valid flux signs on its few active cells, its sign agreement can approach unity and set a strict threshold; each verdict is therefore reported alongside the specific clause that decided it.

\par\textbf{What the check can establish.}
Within these conditions, the check can show that a model's predicted changes have the observed sign more often than
one half of the time and, under the strict rule, more often than its untrained copy, and that the residual
covariance is small within the observed range of $\Delta_p$.
It does not measure the size of the response to an added driver. Its slope and the response ratio $k$ refer to
different perturbations, and no relation between them is assumed or tested.

\subsection{Station groups and decision rules}\label{app:H4}

\par\textbf{Station groups.}
The rain group and surge group~A were formed on 2018 validation data without any model, and surge group~B is defined by the
graph topology. The \emph{rain group} holds the \res{mia.obs.group.rain.n}
canal gauges whose 48-h observed rise correlates with the 48-h realized rain of their rain mix at
$r\ge\res{mia.obs.group.rain.rmin}$. The median correlation is
\res{mia.obs.group.rain.r-median-selected} over the selected gauges and \res{mia.obs.group.rain.r-median-all} over all
gauges with a defined correlation. For surge, no gauge reached $r\ge0.3$ (90th percentile of the per-gauge correlation
\res{mia.obs.group.surge.r-p90}). \emph{Surge group~A} (BBCW10, BBCW8, S123\_T) comprises three coastal gauges selected from
the 2018 distribution (overall, \res{mia.obs.group.surge.n-ge-015} gauges had correlations above 0.15 in 2018, including
G93\_T at \res{mia.obs.group.surge.r.g93t} and G54\_T at \res{mia.obs.group.surge.r.g54t}). \emph{Surge group~B}
(G93\_T, S119\_T, S123\_T) holds the three gauges directly connected to a tidal edge in the flux graph. Both surge groups
have three gauges and share S123\_T.

\par\textbf{Decision rules.}
We evaluate three rule variants to test sensitivity to between-station offsets and slope window width, and a pass is never read
as a correct response size.
\begin{itemize}[itemsep=1.5pt,topsep=0pt,parsep=0pt,leftmargin=*]
\item \emph{Pooled rule}, defined on the 2018 validation split with slope window $[0.5,2]$ and applied unchanged to the
2019 test year and the simulators.
\item \emph{Station fixed-effect (FE) correction}, which applies the pooled-rule criteria after demeaning $\Delta_p$ and
$\Delta_o$ within each station to remove station-constant offsets such as those in surge group~A.
\item \emph{Strict rule}, which adds the informative clause~(0) and the untrained-copy margin clause~(2b) and tightens
the station-FE slope window to $[0.75,1.33]$. On the main simulator, this window separates the four trained models (FE slopes from
\res{beta.s9.rain.gnn.slope-fe.mean} to \res{beta.s9.rain.physres.slope-fe.mean}) from the four half-input variants (FE slopes from
\res{beta.s9.rain.gnn-half.slope-fe.mean} to \res{beta.s9.rain.physres-half.slope-fe.mean}).
\item \emph{Pooled check on surge group~B}, which evaluates the pooled and FE rules on the graph-defined tidal-edge gauges.
\end{itemize}

\begin{table*}[t]
\caption{Observational response-consistency check in Miami-Dade (2019): pooled and station fixed-effect (FE) slopes of $\Delta_o$ on $\Delta_p$ (mean\,$\pm$\,SD over three seeds; UD: undefined).}
\label{tab:s-obscheck}
\setlength{\tabcolsep}{4pt}%
\resizebox{\textwidth}{!}{%
\begin{tabular}{@{}clrr@{}lr@{}lr@{}lr@{}lr@{}lr@{}lrlr@{}}
\toprule
 & & & \multicolumn{12}{c}{Trained models} & \multicolumn{3}{c}{Untrained copy} \\
\cmidrule(lr){4-15}\cmidrule(l){16-18}
 & Model & Cells & \multicolumn{2}{c}{\begin{tabular}[b]{@{}c@{}}Pooled\\slope\end{tabular}} & \multicolumn{2}{c}{\begin{tabular}[b]{@{}c@{}}FE\\slope\end{tabular}} & \multicolumn{2}{c}{\begin{tabular}[b]{@{}c@{}}Shuffle\\FE slope\end{tabular}} & \multicolumn{2}{c}{\begin{tabular}[b]{@{}c@{}}FE\\agreement\end{tabular}} & \multicolumn{2}{c}{\begin{tabular}[b]{@{}c@{}}$|\Delta_p|{>}1$~cm\\($10^3$ cells)\end{tabular}} & \multicolumn{2}{c}{\begin{tabular}[b]{@{}c@{}}Mean\\$|\Delta_p|$ (cm)\end{tabular}} & \begin{tabular}[b]{@{}c@{}}FE\\slope\end{tabular} & Agreement & \begin{tabular}[b]{@{}c@{}}$|\Delta_p|{>}1$~cm\\(cells)\end{tabular} \\
\midrule
\multirow{4}{*}{\rotatebox[origin=c]{90}{\begin{tabular}{@{}c@{}}Rain\\group\end{tabular}}} & Scenario GNN & \multirow{4}{*}{482{,}891} & 0.97 & \,$\pm$\,0.05 & 1.03 & \,$\pm$\,0.05 & $-$0.02 & \,$\pm$\,$<$0.01 & 0.83 & \,$\pm$\,$<$0.01 & 108.1 & \,$\pm$\,8.3 & 0.73 & \,$\pm$\,0.04 & 33.27 & 1.00 & 1 \\
 & Hybrid &  & 1.02 & \,$\pm$\,0.05 & 1.06 & \,$\pm$\,0.05 & $-$0.02 & \,$\pm$\,$<$0.01 & 0.82 & \,$\pm$\,$<$0.01 & 153.2 & \,$\pm$\,6.8 & 0.89 & \,$\pm$\,0.04 & 70.49 & UD & 0 \\
 & Physics + residual &  & 0.84 & \,$\pm$\,0.02 & 0.89 & \,$\pm$\,0.02 & $-$0.02 & \,$\pm$\,$<$0.01 & 0.81 & \,$\pm$\,$<$0.01 & 214.0 & \,$\pm$\,13.5 & 1.12 & \,$\pm$\,0.05 & 16.70 & 1.00 & 1 \\
 & Physics-only &  & 1.07 & \,$\pm$\,0.03 & 1.10 & \,$\pm$\,0.04 & $-$0.03 & \,$\pm$\,0.01 & 0.80 & \,$\pm$\,$<$0.01 & 145.9 & \,$\pm$\,3.5 & 0.94 & \,$\pm$\,0.02 & 12.43 & 0.81 & 186 \\
\midrule
\multirow{4}{*}{\rotatebox[origin=c]{90}{\begin{tabular}{@{}c@{}}Surge\\group A\end{tabular}}} & Scenario GNN & \multirow{4}{*}{25{,}569} & 4.11 & \,$\pm$\,0.53 & 4.12 & \,$\pm$\,0.53 & 0.00 & \,$\pm$\,0.13 & 0.90 & \,$\pm$\,0.02 & 1.4 & \,$\pm$\,0.2 & 0.18 & \,$\pm$\,0.01 & 270 & UD & 0 \\
 & Hybrid &  & 1.05 & \,$\pm$\,0.07 & 1.47 & \,$\pm$\,0.05 & $-$0.03 & \,$\pm$\,0.02 & 0.84 & \,$\pm$\,0.01 & 5.7 & \,$\pm$\,0.1 & 0.75 & \,$\pm$\,0.02 & 0.91 & 0.55 & 2{,}984 \\
 & Physics + residual &  & $-$0.37 & \,$\pm$\,0.02 & 1.12 & \,$\pm$\,0.01 & 0.00 & \,$\pm$\,$<$0.01 & 0.91 & \,$\pm$\,$<$0.01 & 7.1 & \,$\pm$\,$<$0.1 & 1.39 & \,$\pm$\,0.02 & 2.34 & 0.67 & 2{,}770 \\
 & Physics-only &  & $-$0.59 & \,$\pm$\,0.03 & 1.09 & \,$\pm$\,$<$0.01 & 0.00 & \,$\pm$\,$<$0.01 & 0.90 & \,$\pm$\,$<$0.01 & 7.1 & \,$\pm$\,$<$0.1 & 1.43 & \,$\pm$\,$<$0.01 & 1.87 & 0.66 & 3{,}340 \\
\midrule
\multirow{4}{*}{\rotatebox[origin=c]{90}{\begin{tabular}{@{}c@{}}Surge\\group B\end{tabular}}} & Scenario GNN & \multirow{4}{*}{25{,}635} & 4.30 & \,$\pm$\,0.55 & 4.27 & \,$\pm$\,0.55 & $-$0.03 & \,$\pm$\,0.12 & 0.91 & \,$\pm$\,0.02 & 2.4 & \,$\pm$\,0.4 & 0.35 & \,$\pm$\,0.03 & 252 & UD & 0 \\
 & Hybrid &  & 1.67 & \,$\pm$\,0.06 & 1.58 & \,$\pm$\,0.06 & $-$0.02 & \,$\pm$\,0.02 & 0.85 & \,$\pm$\,0.01 & 10.4 & \,$\pm$\,0.3 & 1.33 & \,$\pm$\,0.06 & 1.01 & 0.58 & 10{,}356 \\
 & Physics + residual &  & 0.94 & \,$\pm$\,0.02 & 1.05 & \,$\pm$\,0.01 & 0.01 & \,$\pm$\,$<$0.01 & 0.90 & \,$\pm$\,$<$0.01 & 14.1 & \,$\pm$\,$<$0.1 & 2.75 & \,$\pm$\,0.02 & 1.51 & 0.63 & 9{,}123 \\
 & Physics-only &  & 0.90 & \,$\pm$\,0.02 & 1.02 & \,$\pm$\,$<$0.01 & 0.01 & \,$\pm$\,$<$0.01 & 0.90 & \,$\pm$\,$<$0.01 & 14.4 & \,$\pm$\,0.1 & 2.86 & \,$\pm$\,0.02 & 1.32 & 0.63 & 10{,}727 \\
\bottomrule
\end{tabular}}
\end{table*}

\par\textbf{Verdicts.}
The verdicts of the three rules are the same in 2019 and
2018 except for one. For rain, all four structures pass the pooled rule and its station-FE correction. Under the
strict rule the Hybrid forecaster passes and both physics models fail clause~(2b), as does the Scenario GNN in
2019 (it passes in 2018). The Scenario GNN fails clause~(1) in both surge groups under every version. On surge
group~A the Hybrid forecaster passes the pooled rule, is not separable after the FE correction and fails~(1) under
the strict rule. Both physics models fail~(1) and~(3) under the pooled rule. After the FE correction the Physics +
state residual model passes and the Physics-only model is not separable, and both pass the strict rule. On surge
group~B the Hybrid forecaster is not separable under the pooled rule and its correction and fails~(1) under the
strict rule, the Physics + state residual model is not separable until it passes the strict rule, and the
Physics-only model is not separable under all three. ``Not separable'' means that the trained model passes but its
untrained copy also meets the slope and agreement criteria.

\par\textbf{Results on Miami-Dade.}
Table~\ref{tab:s-obscheck} gives the 2019 test-year diagnostics across station groups and estimators (trained models as mean\,$\pm$\,SD over seeds), with
the definitions of App.~\ref{app:E3} and the groups of App.~\ref{app:H4}. Their pooled slopes use all cells of a group.
Their shuffle column is the FE slope with the driver of another window. Their agreement is the share of the cells with
demeaned $|\Delta_p|>1$~cm, counted in the next column, where $\Delta_p$ and $\Delta_o$ share sign, and their mean
$|\Delta_p|$ is computed on raw cells. In addition,
Fig.~\ref{fig:driver-info}c shows the station fixed-effect slopes of 2019 for rain and surge
group~A. Ranges quoted below run over seeds and models. For rain, all four structures are
consistent, with pooled slopes from \res{mia.obs.test.rain.physres.pooled.slope.min} to \res{mia.obs.test.rain.physonly.pooled.slope.max},
FE slopes from \res{mia.obs.test.rain.physres.fe.slope.min} to \res{mia.obs.test.rain.physonly.fe.slope.max}, shuffled-driver
slopes near zero, and untrained copies with almost no response (at most \res{mia.obs.test.rain.physonly.fe.untrained.n-big}
cells above 1~cm). Under the strict rule the Scenario GNN and both physics models fail rain only through
clause~(2b). The untrained Scenario GNN and Physics + state residual copies have
\res{mia.obs.test.rain.gnn.fe.untrained.n-big} cell above 1~cm each, and the Physics-only copy
\res{mia.obs.test.rain.physonly.fe.untrained.n-big}. The Hybrid forecaster passes because its untrained copy has no
such cell, so no bar applies. In 2018 the Scenario GNN passes as well.

For surge, the Scenario GNN's slopes range from \res{mia.obs.test.surge-a.gnn.pooled.slope.min} to \res{mia.obs.test.surge-b.gnn.pooled.slope.max}
under every version and in both groups, resting on mean responses of \res{mia.obs.test.surge-a.gnn.mean-abs-dp}~cm on
group~A. Reading such a slope as a response that is several times too small would need the assumption of Corollary~1.
The Hybrid forecaster passes the pooled rule on group~A (pooled from \res{mia.obs.test.surge-a.hybrid.pooled.slope.min} to \res{mia.obs.test.surge-a.hybrid.pooled.slope.max})
but is not separable after the FE correction, because its untrained copy reaches an FE slope of
\res{mia.obs.test.surge-a.hybrid.fe.untrained.slope}. The physics models' pooled slopes on group~A are negative, and
the FE correction brings them to values from \res{mia.obs.test.surge-a.physonly.fe.slope.min} to \res{mia.obs.test.surge-a.physres.fe.slope.max},
with untrained copies at \res{mia.obs.test.surge-a.physonly.fe.untrained.slope} and
\res{mia.obs.test.surge-a.physres.fe.untrained.slope}. Under the FE correction and the strict rule on
Miami-Dade, the Physics + state residual model passes the FE-corrected pooled rule on group~A and the strict rule on
both groups, whereas the Physics-only model is not separable under the FE correction and passes the strict rule on
group~A only. In a test year without a large surge episode (App.~\ref{app:H3}), these surge diagnostics need known
scenario responses before they can be read.

The offsets behind the pooled values are model-dependent. The mean discrepancy at BBCW10 and BBCW8 ranges from
\res{mia.obs.test.st.bbcw8.gnn.mean-do.min} to \res{mia.obs.test.st.bbcw10.gnn.mean-do.max}~cm for the Scenario GNN and from
\res{mia.obs.test.st.bbcw10.hybrid.mean-do.min} to \res{mia.obs.test.st.bbcw8.hybrid.mean-do.max}~cm for the Hybrid
forecaster, against \res{mia.obs.test.st.bbcw8.physres.mean-do.min} to \res{mia.obs.test.st.bbcw8.physonly.mean-do.max}~cm
for the two physics models, so the large offset holds for the physics models only. At S123\_T the
per-station slopes in seeds 42, 43 and 44 are \res{mia.obs.test.st.s123t.gnn.slope.s42}, \res{mia.obs.test.st.s123t.gnn.slope.s43}
and \res{mia.obs.test.st.s123t.gnn.slope.s44} for the Scenario GNN and \res{mia.obs.test.st.s123t.hybrid.slope.s42},
\res{mia.obs.test.st.s123t.hybrid.slope.s43} and \res{mia.obs.test.st.s123t.hybrid.slope.s44} for the Hybrid forecaster.
For the physics models the mean discrepancy at S123\_T rises steadily over five equal-count bins of $\Delta_p$
(Physics + state residual, seed~42, with \res{mia.obs.test.st.s123t.physres.bin1.mean-do}, \res{mia.obs.test.st.s123t.physres.bin2.mean-do},
\res{mia.obs.test.st.s123t.physres.bin3.mean-do}, \res{mia.obs.test.st.s123t.physres.bin4.mean-do} and
\res{mia.obs.test.st.s123t.physres.bin5.mean-do}~cm).

Finally, the trained Miami-Dade models act almost additively in rain and surge. The mean absolute difference between
the joint response and the sum of the single responses is at most \res{mia.obs.test.rain.interaction.max}~cm on the
rain group and \res{mia.obs.test.surge-a.interaction.max}~cm on surge group~A in 2019. This additivity is a property of
the models, not a finding about the canal system.

\subsection{Counterexamples on the simulators}\label{app:H5}

\par\textbf{Setting.}
The same analysis code runs on the \res{data.beta.windows-test} test windows of the simulated drainage network, with
the \res{beta.nodes.rain} runoff-receiving nodes as the rain group and the near-outlet nodes as the surge group. There
the substitution compares realized rain with no future rain, the issue-time scenario of the simulator, and paired
simulations at \res{data.beta.cf-issue-times} issue times provide the response ratio $k$ to an added amount $\bar A$.
The ratio $k$ is the median of the model change over the simulator change across responsive cells, for +25.4~mm of
added rain and +30.48~cm of added outlet stage. Each model is checked trained, as half-input and reversed-input variants
and as an untrained copy. The variants feed $q_0+a(q_X-q_0)$ in place of every scenario $q_X$ with $a=1/2$ or $a=-1$.
The reversed input contains negative rain and is a non-physical diagnostic stress test. Table~\ref{tab:s-simrows} lists
every row, and Fig.~\ref{fig:s-obscheck} places slopes and response ratios side by side, each with its own reference.

In Table~\ref{tab:s-simrows} the pooled slope uses all cells of the node group and the FE slope is demeaned within
each node. The verdicts agree in all seeds. The untrained rows state whether the untrained copy meets criteria (1)
and (2) of App.~\ref{app:H3}, or (1) and (2a) under the strict rule. UD marks a sign agreement that is undefined
because no cell has $|\Delta_p|>1$~cm.

\par\textbf{Counterexample 1. A trained model with a wrong response size passes.}
The Scenario GNN passes the pooled rule with a pooled substitution slope of \res{beta.s9.rain.gnn.slope-pooled.mean}
(FE \res{beta.s9.rain.gnn.slope-fe.mean}), although its median response to the added rain $\bar A$ is
\res{beta.k.rain.gnn.k-median.mean} of the simulated change and its peak falls in
\res{beta.k.rain.gnn.share-negative.mean}\% of the responsive cells where the simulated peak rises. Its response ratio
also varies across seeds (SD \res{beta.k.rain.gnn.k-median.sd}) while its slope stays near 1 in every seed
(Fig.~\ref{fig:s-obscheck}b).

\par\textbf{Counterexample 2. Variants fed half of every input change pass.}
All four half-input variants obtain pooled slopes from \res{beta.s9.rain.half.slope-pooled.min} to \res{beta.s9.rain.half.slope-pooled.max}
and pass the pooled rule. Their measured response ratios to $\bar A$ are \res{beta.k.rain.gnn-half.k-median.mean}
(Scenario GNN), \res{beta.k.rain.physonly-half.k-median.mean} (Physics-only model),
\res{beta.k.rain.hybrid-half.k-median.mean} (Hybrid forecaster) and \res{beta.k.rain.physres-half.k-median.mean}
(Physics + state residual). The slope and $k$ again refer to different perturbations. What holds exactly is the
construction, half of every input change.

\par\textbf{Counterexample 3. A variant fed the negated input change passes.}
The reversed-input Hybrid forecaster passes the pooled rule with a pooled slope of
\res{beta.s9.rain.hybrid-reversed.slope-pooled.mean} (FE \res{beta.s9.rain.hybrid-reversed.slope-fe.mean}), although
its response ratio to $\bar A$ is \res{beta.k.rain.hybrid-reversed.k-median.mean}. The other three reversed-input variants
fail, with pooled slopes of \res{beta.s9.rain.gnn-reversed.slope-pooled.mean} (Scenario GNN),
\res{beta.s9.rain.physres-reversed.slope-pooled.mean} (Physics + state residual) and
\res{beta.s9.rain.physonly-reversed.slope-pooled.mean} (Physics-only model).

\par\textbf{The strict rule on the simulator.}
The strict rule passes the Scenario GNN and the Hybrid forecaster for rain and rejects every half-input and
reversed-input variant, as their fixed-effect slopes fall outside its $[0.75,1.33]$ window. It rejects both physics models through clause~(2b). Their untrained copies already
agree in sign almost everywhere (Table~\ref{tab:s-simrows}), so the bar exceeds what the trained models reach. These
verdicts follow from the bar of clause~(2b) described above and are not evidence about the physics models' responses.

\begin{table*}[t]
\caption{Observational-check slopes and simulated response ratios $k$ on the simulated coastal network (mean\,$\pm$\,SD over three seeds; untrained rows indicate whether criteria (1)--(2a) hold; uninf.: no information; UD: undefined).}
\label{tab:s-simrows}
\setlength{\tabcolsep}{4pt}%
\resizebox{\textwidth}{!}{%
\begin{tabular}{@{}cc*{5}{c}ccc@{}}
\toprule
 & & & \multicolumn{2}{c}{Slope} & Sign agr. & Mean $|\Delta_p|$ & \multicolumn{2}{c}{Pooled rule} & Strict \\
\cmidrule(lr){4-5}\cmidrule(lr){8-9}
\multicolumn{2}{@{}l}{Model} & $k$ & pooled & FE & FE & (cm) & pooled & FE & rule \\
\midrule
\multicolumn{10}{@{}c}{\emph{Rain, 163 runoff-receiving nodes}} \\
\midrule
\multirow{4}{*}{\rotatebox[origin=c]{90}{Trained}} & Scenario GNN & +0.71\,$\pm$\,0.13 & +0.98\,$\pm$\,0.01 & +0.98\,$\pm$\,0.01 & 0.97\,$\pm$\,$<$0.01 & 8.315\,$\pm$\,0.166 & pass & pass & pass \\
 & Hybrid & +0.92\,$\pm$\,0.08 & +1.11\,$\pm$\,0.01 & +1.11\,$\pm$\,0.01 & 0.97\,$\pm$\,$<$0.01 & 7.305\,$\pm$\,0.103 & pass & pass & pass \\
 & Physics + residual & +0.97\,$\pm$\,0.03 & +1.18\,$\pm$\,0.02 & +1.20\,$\pm$\,0.02 & 0.95\,$\pm$\,$<$0.01 & 6.475\,$\pm$\,0.109 & pass & pass & fail \\
 & Physics-only & +0.86\,$\pm$\,0.09 & +1.11\,$\pm$\,0.01 & +1.14\,$\pm$\,0.01 & 0.94\,$\pm$\,$<$0.01 & 6.196\,$\pm$\,0.267 & pass & pass & fail \\
\cmidrule(l){1-10}
\multirow{4}{*}{\rotatebox[origin=c]{90}{\begin{tabular}{@{}c@{}}Half-\\input\end{tabular}}} & Scenario GNN & +0.33\,$\pm$\,0.13 & +1.46\,$\pm$\,0.02 & +1.45\,$\pm$\,0.02 & 0.97\,$\pm$\,$<$0.01 & 4.782\,$\pm$\,0.163 & pass & pass & fail \\
 & Hybrid & +0.58\,$\pm$\,0.03 & +1.60\,$\pm$\,0.02 & +1.61\,$\pm$\,0.02 & 0.97\,$\pm$\,$<$0.01 & 4.228\,$\pm$\,0.050 & pass & pass & fail \\
 & Physics + residual & +0.58\,$\pm$\,0.01 & +1.87\,$\pm$\,0.02 & +1.90\,$\pm$\,0.02 & 0.95\,$\pm$\,$<$0.01 & 3.693\,$\pm$\,0.065 & pass & pass & fail \\
 & Physics-only & +0.47\,$\pm$\,0.03 & +1.79\,$\pm$\,0.01 & +1.83\,$\pm$\,0.01 & 0.95\,$\pm$\,$<$0.01 & 3.499\,$\pm$\,0.156 & pass & pass & fail \\
\cmidrule(l){1-10}
\multirow{4}{*}{\rotatebox[origin=c]{90}{\begin{tabular}{@{}c@{}}Reversed-\\input\end{tabular}}} & Scenario GNN & $-$0.11\,$\pm$\,0.11 & +0.16\,$\pm$\,0.79 & +0.13\,$\pm$\,0.74 & 0.57\,$\pm$\,0.17 & 1.042\,$\pm$\,0.279 & fail & fail & fail \\
 & Hybrid & $-$0.18\,$\pm$\,0.08 & +1.51\,$\pm$\,0.24 & +1.50\,$\pm$\,0.23 & 0.87\,$\pm$\,0.03 & 1.703\,$\pm$\,0.659 & pass & pass & fail \\
 & Physics + residual & $-$0.08\,$\pm$\,$<$0.01 & $-$1.36\,$\pm$\,0.07 & $-$1.16\,$\pm$\,0.07 & 0.57\,$\pm$\,0.01 & 1.013\,$\pm$\,0.052 & fail & fail & fail \\
 & Physics-only & $-$0.04\,$\pm$\,0.01 & $-$2.44\,$\pm$\,0.97 & $-$2.27\,$\pm$\,0.99 & 0.37\,$\pm$\,0.14 & 0.677\,$\pm$\,0.166 & fail & fail & fail \\
\cmidrule(l){1-10}
 & & & & & & & \emph{(1)/(2) met} & \emph{(1)/(2) met} & \emph{(1)/(2a) met} \\
\multirow{4}{*}{\rotatebox[origin=c]{90}{Untrained}} & Scenario GNN & $-$0.00 & $-$91.72 & $-$92.97 & 0.49 & 0.014 & no/no & no/no & no/no \\
 & Hybrid & +0.00 & +351 & +355 & UD & 0.012 & no/no & no/no & no/no \\
 & Physics + residual & +0.01 & +110 & +112 & 0.99 & 0.047 & no/yes & no/yes & no/yes \\
 & Physics-only & +0.01 & +92.92 & +95.00 & 0.99 & 0.038 & no/yes & no/yes & no/yes \\
\midrule
\multicolumn{10}{@{}c}{\emph{Outlet stage, 24 near-outlet nodes}} \\
\midrule
\multirow{4}{*}{\rotatebox[origin=c]{90}{Trained}} & Scenario GNN & +0.00\,$\pm$\,0.00 & +0.75\,$\pm$\,0.33 & +0.73\,$\pm$\,0.33 & 0.55\,$\pm$\,0.01 & 0.082\,$\pm$\,0.008 & fail & fail & uninf. \\
 & Hybrid & +0.00\,$\pm$\,0.00 & +0.90\,$\pm$\,0.08 & +0.89\,$\pm$\,0.09 & 0.56\,$\pm$\,0.03 & 0.032\,$\pm$\,0.005 & fail & fail & uninf. \\
 & Physics + residual & +0.00\,$\pm$\,$<$0.01 & $-$1.67\,$\pm$\,0.73 & $-$1.51\,$\pm$\,0.91 & 0.49\,$\pm$\,0.02 & 0.013\,$\pm$\,0.003 & fail & fail & uninf. \\
 & Physics-only & +0.00\,$\pm$\,$<$0.01 & $-$1.45\,$\pm$\,2.23 & $-$1.66\,$\pm$\,2.77 & 0.49\,$\pm$\,0.02 & 0.020\,$\pm$\,0.011 & fail & fail & uninf. \\
\cmidrule(l){1-10}
\multirow{4}{*}{\rotatebox[origin=c]{90}{\begin{tabular}{@{}c@{}}Half-\\input\end{tabular}}} & Scenario GNN & +0.00\,$\pm$\,0.00 & +0.87\,$\pm$\,0.50 & +0.85\,$\pm$\,0.51 & 0.55\,$\pm$\,0.01 & 0.041\,$\pm$\,0.004 & fail & fail & uninf. \\
 & Hybrid & +0.00\,$\pm$\,$<$0.01 & +1.56\,$\pm$\,0.08 & +1.60\,$\pm$\,0.08 & 0.57\,$\pm$\,0.03 & 0.016\,$\pm$\,0.002 & fail & fail & uninf. \\
 & Physics + residual & +0.00\,$\pm$\,$<$0.01 & $-$3.41\,$\pm$\,1.67 & $-$2.81\,$\pm$\,1.84 & 0.48\,$\pm$\,0.05 & 0.007\,$\pm$\,0.002 & fail & fail & uninf. \\
 & Physics-only & +0.00\,$\pm$\,$<$0.01 & $-$3.07\,$\pm$\,4.75 & $-$3.25\,$\pm$\,5.58 & 0.48\,$\pm$\,0.05 & 0.010\,$\pm$\,0.005 & fail & fail & uninf. \\
\cmidrule(l){1-10}
\multirow{4}{*}{\rotatebox[origin=c]{90}{\begin{tabular}{@{}c@{}}Reversed-\\input\end{tabular}}} & Scenario GNN & +0.00\,$\pm$\,0.00 & +0.48\,$\pm$\,0.49 & +0.49\,$\pm$\,0.50 & 0.49\,$\pm$\,0.03 & 0.060\,$\pm$\,0.009 & fail & fail & uninf. \\
 & Hybrid & $-$0.00\,$\pm$\,$<$0.01 & $-$0.94\,$\pm$\,0.09 & $-$0.91\,$\pm$\,0.14 & 0.44\,$\pm$\,0.03 & 0.030\,$\pm$\,0.004 & fail & fail & uninf. \\
 & Physics + residual & +0.00\,$\pm$\,0.00 & +1.43\,$\pm$\,1.05 & +1.09\,$\pm$\,0.96 & 0.49\,$\pm$\,0.01 & 0.013\,$\pm$\,0.003 & fail & fail & uninf. \\
 & Physics-only & +0.00\,$\pm$\,0.00 & +1.48\,$\pm$\,2.59 & +1.45\,$\pm$\,2.75 & 0.50\,$\pm$\,0.01 & 0.020\,$\pm$\,0.011 & fail & fail & uninf. \\
\cmidrule(l){1-10}
 & & & & & & & \emph{(1)/(2) met} & \emph{(1)/(2) met} & \emph{(1)/(2a) met} \\
\multirow{4}{*}{\rotatebox[origin=c]{90}{Untrained}} & Scenario GNN & +0.00 & $-$4.57 & $-$6.11 & 1.00 & 0.002 & no/yes & no/yes & no/yes \\
 & Hybrid & +0.00 & $-$2.71 & $-$2.83 & 0.48 & 0.013 & no/no & no/no & no/no \\
 & Physics + residual & +0.00 & $-$2.69 & $-$2.70 & 0.46 & 0.035 & no/no & no/no & no/no \\
 & Physics-only & +0.00 & $-$2.69 & $-$2.69 & 0.45 & 0.035 & no/no & no/no & no/no \\
\bottomrule
\end{tabular}}
\end{table*}

\par\textbf{Outlet stage.}
For the outlet-stage driver the pooled rule fails all four trained models, and the strict rule calls every
trained, half-input and reversed-input row uninformative. The pooled surge slopes of the trained models range from
\res{beta.s9.surge.physres.slope-pooled.mean} to \res{beta.s9.surge.hybrid.slope-pooled.mean}, and their mean
$|\Delta_p|$ ranges from \res{beta.s9.surge.mean-abs-dpred.min} to \res{beta.s9.surge.mean-abs-dpred.max}~cm, while the response
ratio to $\bar s$ is zero for every model and variant. The check therefore treats the two drivers differently, but for the
outlet stage it carries no size information and cannot distinguish a missing response from a weak signal. Among the
configurations modified to restore the outlet-stage response (Table~\ref{tab:s-stage4}), the Scenario GNN trained with
more surge storms and the one trained with paired twin storms pass the outlet-stage driver under the pooled rule
with the station fixed-effect correction, although their response ratio stays at \res{beta.s4.h1.gconv.k_stage}.

\par\textbf{Second simulator.}
The same pattern appears on the second simulator, where valve settings are the scenario channel. The FE slopes for
valves range from \res{gamma.v2.valve.physonly.trained.slope.mean} to \res{gamma.v2.valve.gnn.trained.slope.mean} for every
trained model, while the valve response ratios range only from \res{gamma.k_valve.gnn.trained} to \res{gamma.k_valve.physonly.trained}.
The untrained physics copies already reach valve slopes of \res{gamma.v2.valve.physonly.untrained.slope} and
\res{gamma.v2.valve.physres.untrained.slope}, and the reversed-input Scenario GNN passes the rain check with slopes from
\res{gamma.v2.rain.gnn.reversed.slope.min} to \res{gamma.v2.rain.gnn.reversed.slope.max} (Fig.~\ref{fig:s-transfer}e).
Across \smbadd{the main and second} simulators a slope near 1 thus coexists with response ratios well below 1, with a negated input and with
responses supplied by structure alone. The slope measures residual covariance, as Proposition~1 states, and not a
correct response size.

\section{Driver Responses, Propagation, and Compound Effects}\label{app:I}

This appendix gives the complete paired-simulator response results behind the single-driver responses, the propagation
of the outlet stage and its interaction with rainfall. It reports per-cell responses, response curves by node set,
spatial profiles, the reach of the outlet-stage input, joint surfaces with their training support, responses under data and boundary modifications, and the interaction estimands, including the coalition definition. These results concern the first set of runs on the simulated coastal network under one fixed
control path. The last subsection collects the evidence class of every kind of explanation, from these and the other
paired-run results (Table~\ref{tab:s-classes}).

\subsection{Single-driver responses by node set}\label{app:I1}

\par
\begin{figure*}[tp]
\centering
\includegraphics[width=\textwidth]{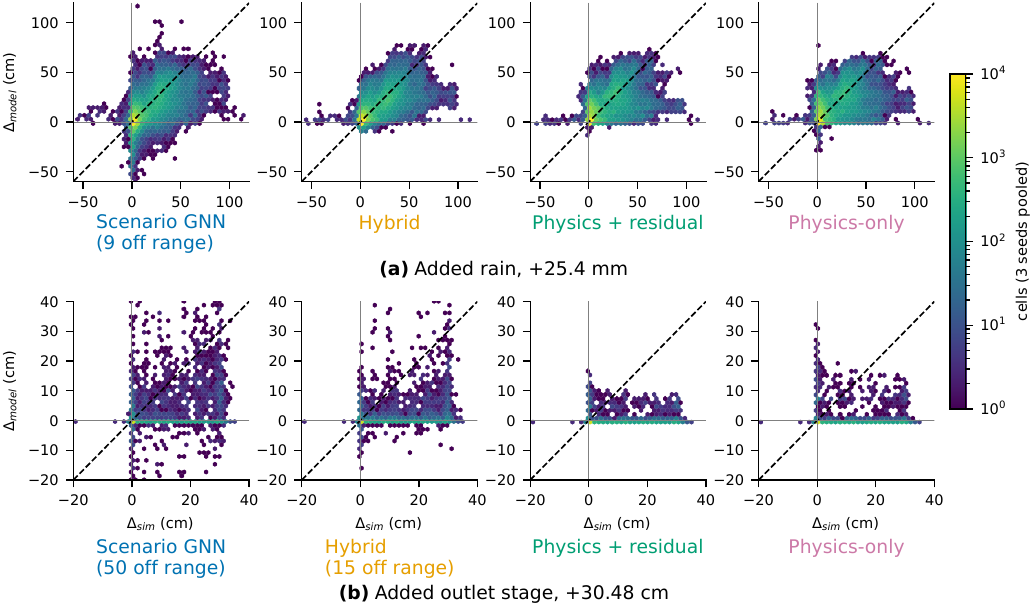}
\caption{Per-cell paired 6-h peak depth changes on the simulated coastal network across \res{data.beta.cf-issue-times} test issue times (three seeds pooled) for added rain $\bar A$ and raised outlet stage $\bar s$ (dashed: $1{:}1$ line).}
\label{fig:s-hexbin}
\end{figure*}

\textbf{Per-cell responses.}
Fig.~\ref{fig:s-hexbin} shows every cell behind the storm-level means of the main results. For the added rain $\bar A$ the
trained models follow the simulated changes with wide scatter, and the Scenario GNN is the only model with a large
share of falling peaks. Its peak falls in \res{beta.k.rain.gnn.share-negative.mean}\% of the responsive cells in which
the simulated peak rises, against \res{beta.k.rain.hybrid.share-negative.mean}\% for the Hybrid forecaster and
\res{beta.k.rain.physres.share-negative.mean}\% and \res{beta.k.rain.physonly.share-negative.mean}\% for the two
physics models. The 10th percentile of its per-cell ratio is \res{beta.k.rain.gnn.k-p10.mean}. For the added outlet
stage $\bar s$, most model changes are zero while the simulated changes at the same near-outlet nodes reach tens of
centimeters. The profiles below locate the few non-zero model changes next to the outlet.
A cell of Fig.~\ref{fig:s-hexbin} is one (issue time, node) pair, and the panel titles count the cells outside the
axis range. Cells of one storm share its state and are not independent samples.

\par
\begin{figure*}[tp]
\centering
\includegraphics[width=\textwidth]{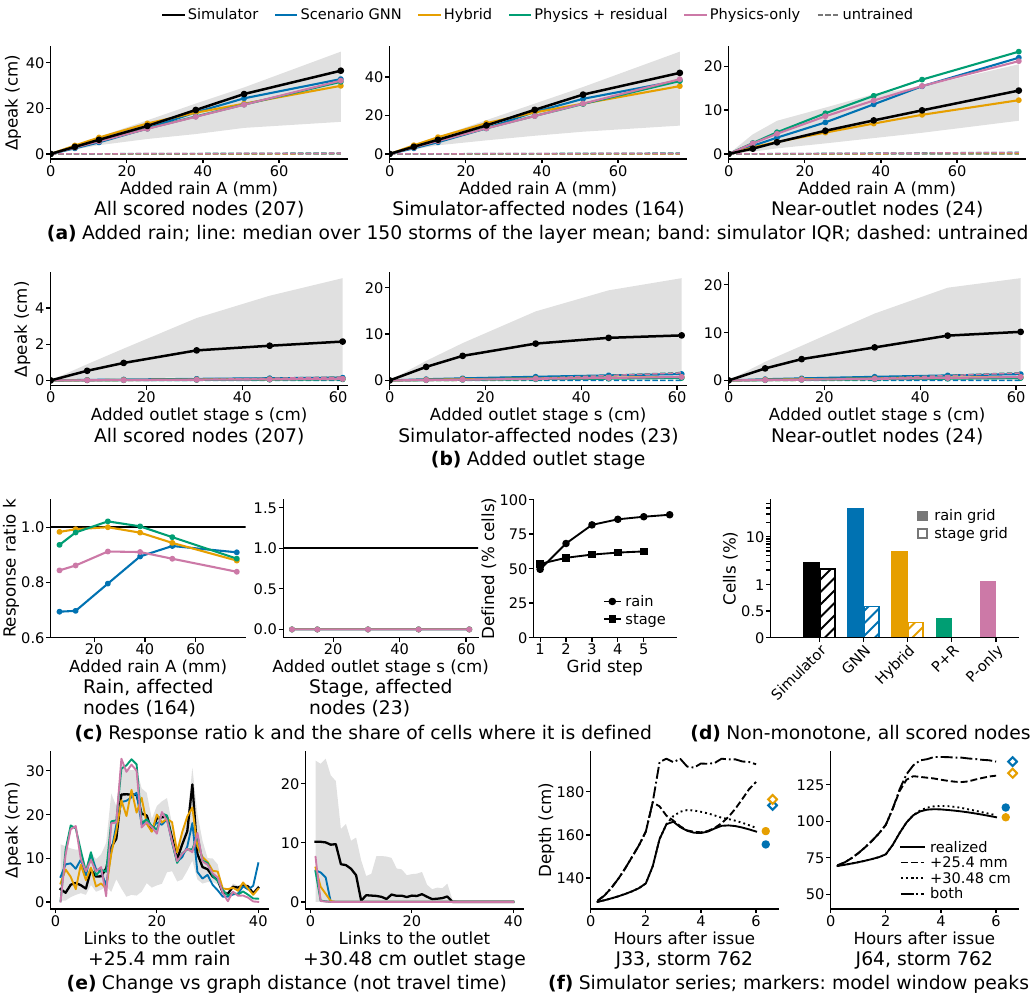}
\caption{Driver responses on the simulated coastal network (\res{data.beta.cf-issue-times} test storms, seed means): (a,b)~response curves, (c)~ratio $k$, (d)~non-monotone cell shares, (e)~spatial propagation, and (f)~time series.}
\label{fig:s-response}
\end{figure*}

\textbf{Node sets.}
Responses are reported for three node sets. The first holds all \res{beta.nodes.scored} scored nodes. The second holds
the simulator-affected nodes, \res{beta.s12.layer.affected_rain.n} for rain and \res{beta.s12.layer.affected_stage.n}
for the outlet stage, defined from the simulator alone as the nodes whose simulated change exceeds
$\max(c_1\nu,\delta)$ in at least half of the issue times. The third holds the \res{beta.nodes.near} near-outlet nodes,
defined by topology. On the simulator-affected nodes, the median response ratio to $\bar A$ is
\res{beta.s12.k.rain.affected.gconv} for the Scenario GNN, \res{beta.s12.k.rain.affected.wm3fpd} for the Hybrid
forecaster, \res{beta.s12.k.rain.affected.d9wm4} for the Physics + state residual model and
\res{beta.s12.k.rain.affected.d9phys} for the Physics-only model, computed on the
\res{beta.s12.k.rain.affected.coverage}\% of cells where it is defined. Across the rain grid (Fig.~\ref{fig:s-response}a,c)
the Scenario GNN's ratio is lowest at small amounts. At the near-outlet nodes the Scenario GNN and both physics models
exceed the simulated rise for the largest amounts, whereas the Hybrid forecaster stays below it.
Fig.~\ref{fig:s-response}c gives $k$ on the simulator-affected nodes at every grid amount, together with the share of
cells where it is defined, the cells with $|\Delta_{\mathrm{sim}}|>\max(c_1\nu,\delta)$.

\par\textbf{Outlet stage.}
For the added outlet stage, the median response ratio is zero at every amount, in every node set and for every model.
The largest of these medians is \res{beta.s12.k.stage.max-over-amounts} (Fig.~\ref{fig:s-response}b,c). At the
reference amount $\bar s$, $k$ is defined on \res{beta.s12.k.stage.near.coverage}\% of near-outlet and
\res{beta.s12.k.stage.affected.coverage}\% of simulator-affected cells. The same computation returns non-zero changes at
the junctions next to the outlet, stage-response shares (mean $|$model change$|$ over mean $|$simulated change$|$) from
\res{beta.s4.3a.d9wm4.share_stage} to \res{beta.s4.3a.gconv.share_stage} and response ratios to rain that differ by
model, so the zero median reflects responses confined to a few near-outlet nodes.

\par\textbf{Non-monotone responses.}
A cell is non-monotone over a grid when its change drops by more than $\max(2\nu,0.1~\text{cm})$ between consecutive
amounts. Over the rain grid this holds for \res{beta.s12.nonmono.rain.allscored.simulator}\% of
scored cells in the simulator, \res{beta.s12.nonmono.rain.allscored.gconv}\% for the Scenario GNN,
\res{beta.s12.nonmono.rain.allscored.wm3fpd}\% for the Hybrid forecaster and
\res{beta.s12.nonmono.rain.allscored.d9wm4}\% and \res{beta.s12.nonmono.rain.allscored.d9phys}\% for the two physics
models (Fig.~\ref{fig:s-response}d). In the simulator the share is higher at the near-outlet nodes
(\res{beta.s12.nonmono.rain.nearoutlet.simulator}\%). Over the stage grid
no model exceeds \res{beta.s12.nonmono.stage.nearoutlet.gconv}\% (the Scenario GNN at near-outlet nodes). The Scenario
GNN's rain response is therefore not only too small in the median but also often non-monotone in the added amount, a
property its median response ratio does not show.

\subsection{Spatial propagation}\label{app:I2}

\par\textbf{Simulator.}
An added outlet stage $\bar s$ changes window peaks detectably in at least half of the storms at
\res{beta.fig3.sim.n_nodes_affected_s1} scored nodes, including the storage units ST0 and ST2, up to
\res{beta.fig3.sim.maxpath_affected_m}~m of hydraulic path from the outlet, and raises the mean peak by more than 1~cm at
\res{beta.fig3.sim.n_nodes_mean_gt1cm_s1} nodes. The response is skewed across storms. At the near-outlet nodes the
median change is \res{beta.fig3.sim.near_median_s1}~cm and the mean \res{beta.fig3.sim.near_mean_s1}~cm. The four
junctions next to the outlet rise by \res{beta.s4.sim.adj4.mean_range}~cm on average and the other near-outlet nodes
by \res{beta.s4.sim.other20.mean_range}~cm. Only \res{beta.s4.sim.n_near_median_gt1cm} of the \res{beta.nodes.near}
have a median change above 1~cm. Behind the flap gate the raised stage obstructs gravity drainage at the outfall, and
the simulated network carries the resulting backwater inland, into the storage units ST0 and ST2.

\par\textbf{Models.}
Every model raises the mean peak by more than 1~cm at no more than \res{beta.fig3.gconv.n_nodes_mean_gt1cm_s1} junctions
(Scenario GNN), \res{beta.fig3.wm3fpd.n_nodes_mean_gt1cm_s1} (Hybrid forecaster) and
\res{beta.fig3.d9wm4.n_nodes_mean_gt1cm_s1} (each physics model), all next to the outlet. At the four outlet-adjacent
junctions the Scenario GNN reaches \res{beta.s4.3a.gconv.adj4.mean_range}~cm, \res{beta.s4.3a.gconv.adj4.share_range}\%
of the simulated change at the same node, and the Hybrid forecaster \res{beta.s4.3a.wm3fpd.adj4.mean_range}~cm
(\res{beta.s4.3a.wm3fpd.adj4.share_range}\%). For added rain, by contrast, the models follow the simulated profile over
the link distance from the outlet (Fig.~\ref{fig:s-response}e). The models therefore reproduce part of the response at
the \res{beta.fig3.d9wm4.n_nodes_mean_gt1cm_s1} to \res{beta.fig3.gconv.n_nodes_mean_gt1cm_s1} junctions next to the
outlet and miss it at every other node, including nodes up to \res{beta.fig3.sim.maxpath_affected_m}~m inland
where the simulator responds detectably in at least half of the storms.

\par\textbf{Distances and timing.}
Hop distance and hydraulic path length are distances, not travel times. Responses are compared through window peaks,
because the saved 24-step simulator series sample a 1-h ramp every 15~min, which resolves few lags, and the models
save window peaks.
Fig.~\ref{fig:s-response}f shows, for one illustrative test storm, how the realized, added-rain,
added-stage and joint series differ at a node two links and at a node sixteen links from the outlet,
with markers showing the models' window peaks.

\subsection{Reach of the outlet-stage input}\label{app:I-reach}

\begin{figure*}[tp]
\centering
\includegraphics[width=\textwidth]{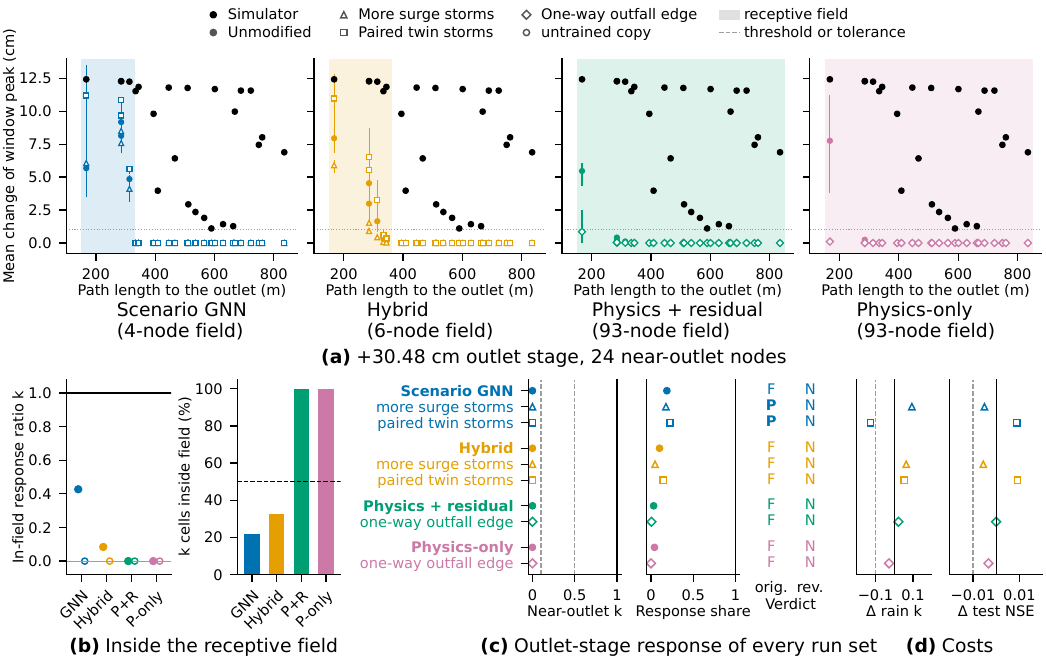}
\caption{Outlet-stage receptive-field reach and ablation responses across \res{data.beta.cf-issue-times} test storms on the simulated coastal network (seed mean with range in (a) and SD in (b); P: pass, F: fail, N: no information).}
\label{fig:s-receptive}
\end{figure*}

From the network topology and the architecture layer counts of the simulator checkpoints, every architecture
receives the outlet stage (astronomical tide plus surge) as one series at the outfall node, which touches the network
only through the outfall edge to junction J79. The reach of that input, the receptive field of the main text, is the
set of scored nodes whose forecast it can change. For the Scenario GNN it holds the nodes within
\res{beta.rf.gnn.layers} hops of the tide node on its encoder graph (one hop per mean-aggregation layer). For the Hybrid
forecaster it holds the nodes within \res{beta.rf.hybrid.enc_layers} encoder hops, plus J79 through its open-loop
physics branch (\res{beta.rf.hybrid.phys_reach_links} link). For the two physics models it holds the nodes within
\res{beta.rf.phys.rollout_steps} links of the outfall, one link per rollout step. The reach holds
\res{beta.rf.gnn.field_size}, \res{beta.rf.hybrid.field_size} and \res{beta.rf.physres.field_size} scored nodes,
respectively. Of the \res{beta.s12.layer.affected_stage.n} simulator-affected nodes, \res{beta.rf.gnn.beyond_aff23},
\res{beta.rf.hybrid.beyond_aff23} and \res{beta.rf.physres.beyond_aff23} lie outside it. Of the
\res{beta.fig3.sim.n_nodes_mean_gt1cm_s1} nodes whose mean simulated peak rises by more than 1~cm,
\res{beta.rf.gnn.beyond_aff86}, \res{beta.rf.hybrid.beyond_aff86} and \res{beta.rf.physres.beyond_aff86} lie outside
it. Only \res{beta.rf.gnn.kstage_cells_in_reach_pct}\% and \res{beta.rf.hybrid.kstage_cells_in_reach_pct}\% of the
\res{beta.k.stage.n-cells} near-outlet cells that define $k_{\mathrm{stage}}$ lie inside the reach of the Scenario GNN
and the Hybrid forecaster, fewer than half, so their near-outlet median $k_{\mathrm{stage}}$ is zero by construction
(Fig.~\ref{fig:s-receptive}b).

The saved outputs confirm the derived reach. Outside it no trained model changes any window peak by more than
\res{beta.rf.max_change_outside_reach}~cm. Inside it the Scenario GNN raises the mean peak by more than 1~cm at all
\res{beta.rf.gnn.n_responding} reachable nodes in all three seeds, the Hybrid forecaster at
\res{beta.rf.hybrid.n_responding} of its \res{beta.rf.hybrid.field_size}, and each physics model only at J79
(Fig.~\ref{fig:s-receptive}a). Restricted to the responsive outlet-stage cells inside each reach, the median response
ratio (mean over three seeds) is \res{beta.rf.gnn.k_stage_in_reach.mean} for the Scenario GNN
(\res{beta.rf.gnn.k_stage_in_reach.n_cells} cells), against \res{beta.rf.gnn.k_stage_in_reach.untrained} for its
untrained copy on the same cells, and \res{beta.rf.hybrid.k_stage_in_reach.mean} for the Hybrid forecaster
(\res{beta.rf.hybrid.k_stage_in_reach.n_cells} cells). For the Physics + state residual model it is
\res{beta.rf.physres.k_stage_in_reach.mean} on \res{beta.rf.physres.k_stage_in_reach.n_cells} cells, although its
reach holds \res{beta.rf.stage.physres.affected_in_reach_share}\% of the simulator-affected nodes. Inside its reach the
Scenario GNN's outlet-stage response is thus learned, since its untrained copy does not respond, but smaller than the
simulated one. The modified runs of App.~\ref{app:I-mods} use the same layer counts and rollout length. The missing inland response is
therefore fixed by architecture for the Scenario GNN and, for most of its missed affected nodes, for the Hybrid
forecaster, which also leaves two to three reachable affected nodes without a response. For the physics models it is
not a reach limit.

Fig.~\ref{fig:s-receptive} uses the added outlet stage of $+30.48$~cm.
In panel~b the in-reach $k$ of each architecture is taken on that architecture's own cells, so the cell sets differ by
architecture.

\subsection{Joint surfaces and training support}\label{app:I3}

\par
\begin{figure*}[tp]
\centering
\includegraphics[width=\textwidth]{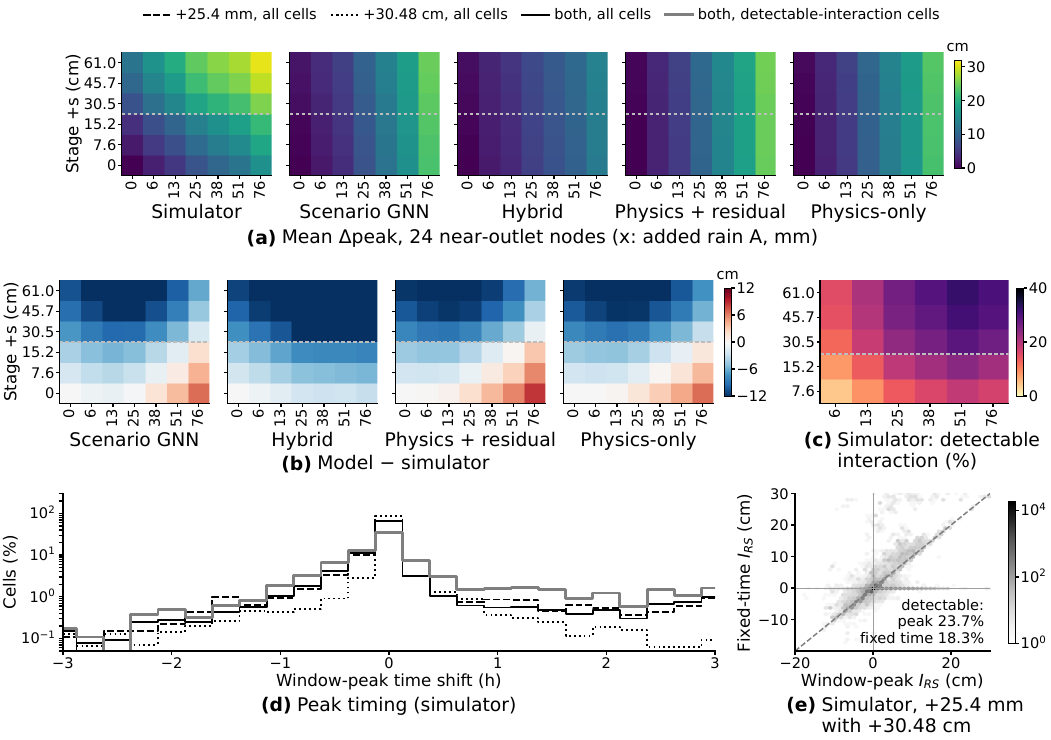}
\caption{Compound rain--stage responses and interaction targets $I_{RS}$ on the simulated coastal network (\res{data.beta.cf-issue-times} test storms, seed means; dashed line: training 99th-percentile stage rise).}
\label{fig:s-compound}
\end{figure*}

\textbf{Mean surfaces.}
History, astronomical tide and control path are shared by all runs of an issue time.
Fig.~\ref{fig:s-compound}a shows the mean window-peak change over the near-outlet cells on the sampled rain $\times$
outlet-stage grid, without smoothing. Means are shown because the mean surface's interaction equals the mean of the
cell interactions, whereas medians of this bimodal cell distribution create a strong super-additivity that the cells
do not have. The median changes are \res{beta.fig3.grid.sim_median_a6_s0}~cm for the smallest rain step alone,
\res{beta.fig3.grid.sim_median_a0_s8}~cm for the smallest stage step alone and \res{beta.fig3.grid.sim_median_a6_s8}~cm
for both together. On the mean surface the simulator responds to both drivers. At $(\bar A,\bar s)$ the mean change is
\res{beta.fig3.grid.sim_mean_a25}~cm for rain alone, \res{beta.fig3.grid.sim_mean_s30}~cm for the stage alone and
\res{beta.fig3.grid.sim_mean_a25_s30}~cm together, an interaction of the mean surface of
\res{beta.fig3.grid.sim_mean_interaction_a25_s30}~cm.

\par\textbf{Threshold crossings on the grid.}
Of the \res{beta.fig3.grid.n_unwarned_near} near-outlet cells below the high-water threshold at the simulator base, rain
alone brings at most \res{beta.fig3.grid.sim_cross_rain_only_max}\% to the threshold within the grid and the outlet
stage alone at most \res{beta.fig3.grid.sim_cross_stage_only_max}\%, whereas combinations bring more than half. The
models' surfaces vary mainly with rain. For $\bar s$ alone the mean change is \res{beta.fig3.grid.wm3fpd.mean_s30}~cm for
the Hybrid forecaster and \res{beta.fig3.grid.gconv.mean_s30}~cm for the Scenario GNN, against
\res{beta.fig3.grid.sim_mean_s30}~cm in the simulator. At $(\bar A,\bar s)$ the Scenario GNN predicts a crossing for
\res{beta.fig3.grid.gconv.cross_a25_s30}\% of these cells, against \res{beta.fig3.grid.sim_cross_a25_s30}\% in the
simulator, and the Hybrid forecaster reaches at most \res{beta.fig3.grid.wm3fpd.cross_max}\% at any grid point. The
model-minus-simulator differences are most negative where the added stage is large (Fig.~\ref{fig:s-compound}b).

\par\textbf{Training support.}
The tested outlet-stage amounts lie partly beyond the training windows. The reference amount $\bar s$ exceeds the 99th
percentile of the 6-h stage rise seen in training in \res{beta.s4.support.test.stage30.above_p99}\% of the test storms
but the training maximum in \res{beta.s4.support.test.stage30.above_max}\%, and the largest amount exceeds the maximum
in \res{beta.s4.support.test.stage61.above_max}\%. The dashed line of Fig.~\ref{fig:s-compound} marks this boundary.
Model recall of the interaction stays near zero at every grid point, the smallest stage steps included. For the physics
models, whose reach is not limiting, the missing joint response therefore does not rest on extrapolation alone. For
the Scenario GNN and the Hybrid forecaster most interaction cells lie beyond the reach of their outlet-stage input
(App.~\ref{app:I-reach}), so their recall reflects the reach rather than extrapolation.

\par\textbf{Outlet-stage signal in the training data.}
In the \res{beta.s4.train_windows} training windows of the simulated drainage network, which start every 2 steps of
each training storm, the surge part of the outlet stage changes little over 6~h. Its median absolute change is
\res{beta.s4.dsurge.median}~cm and its 99th percentile \res{beta.s4.dsurge.p99}~cm, and \res{beta.s4.dsurge.share_gt10}\%
of the windows change by more than 10~cm, whereas the astronomical part changes by a median of
\res{beta.s4.dastro.median}~cm. A least-squares fit of the window rise at each near-outlet node on the window rain, surge
change and astronomical change gives a median partial $R^2$ of \res{beta.s4.pr2.rain.median} for rain,
\res{beta.s4.pr2.surge.median} for surge and \res{beta.s4.pr2.astro.median} for the astronomical tide (largest
\res{beta.s4.pr2.rain.max}, \res{beta.s4.pr2.surge.max} and \res{beta.s4.pr2.astro.max}). The near-outlet rise in
training is thus driven by rain inflow, and a model fitted to these windows gains almost nothing by representing the
backwater response. App.~\ref{app:I-mods} tests this candidate cause.

\subsection{Responses under data and boundary modifications}\label{app:I-mods}

\begin{table*}[t]
\caption{Outlet-stage modifications on the simulated coastal network across 150 test storms (mean\,$\pm$\,SD over three seeds; node counts: seed range; $^{\dagger}k_{\mathrm{rain}}$ fell by 0.13, beyond tolerance; uninf.: uninformative).}
\label{tab:s-stage4}
\begin{tabular*}{\textwidth}{@{\extracolsep{\fill}}clcccccccccc@{}}
\toprule
\multicolumn{2}{@{}l}{\multirow{2}{*}{Run set, model}} & \multicolumn{3}{c}{Outlet-stage response} & \multicolumn{2}{c}{Tolerance checks} & \multirow{2}{*}{Verdict} & \multicolumn{2}{c}{Pooled + FE} & \multicolumn{2}{c}{Strict} \\
\cmidrule(lr){3-5}\cmidrule(lr){6-7}\cmidrule(lr){9-10}\cmidrule(l){11-12}
\multicolumn{2}{@{}l}{} & $k_{\mathrm{stage}}$ & Stage share & Nodes $>$1~cm & $k_{\mathrm{rain}}$ & Test NSE & & Rain & Surge & Rain & Surge \\
\midrule
\multirow{4}{*}{\rotatebox[origin=c]{90}{Main}} & Scenario GNN & 0.000 & 0.188\,$\pm$\,0.025 & 4 & 0.71\,$\pm$\,0.13 & 0.934\,$\pm$\,0.001 & -- & pass & fail & pass & uninf. \\
 & Hybrid & 0.000 & 0.102\,$\pm$\,0.028 & 3--4 & 0.92\,$\pm$\,0.08 & 0.944\,$\pm$\,0.001 & -- & pass & fail & pass & uninf. \\
 & Physics + residual & 0.000 & 0.032\,$\pm$\,0.006 & 1 & 0.97\,$\pm$\,0.03 & 0.807\,$\pm$\,0.009 & -- & pass & fail & fail & uninf. \\
 & Physics-only & 0.000 & 0.042\,$\pm$\,0.020 & 1 & 0.86\,$\pm$\,0.09 & 0.674\,$\pm$\,0.054 & -- & pass & fail & fail & uninf. \\
\midrule
\multirow{2}{*}{\rotatebox[origin=c]{90}{H1}} & Scenario GNN & 0.000 & 0.178\,$\pm$\,0.017 & 4 & 0.80\,$\pm$\,0.04 & 0.929\,$\pm$\,0.003 & no effect & pass & pass & pass & uninf. \\
 & Hybrid & 0.000 & 0.049\,$\pm$\,0.001 & 2--3 & 0.98\,$\pm$\,0.09 & 0.939\,$\pm$\,0.002 & no effect & pass & fail & pass & uninf. \\
\midrule
\multirow{2}{*}{\rotatebox[origin=c]{90}{H2}} & Scenario GNN & 0.000 & 0.226\,$\pm$\,0.002 & 4 & 0.58\,$\pm$\,0.05 & 0.943\,$\pm$\,0.002 & no effect$^{\dagger}$ & pass & pass & pass & uninf. \\
 & Hybrid & 0.000 & 0.145\,$\pm$\,0.061 & 3--5 & 0.97\,$\pm$\,0.08 & 0.954\,$\pm$\,$<$0.001 & no effect & pass & fail & pass & uninf. \\
\midrule
\multirow{2}{*}{\rotatebox[origin=c]{90}{H3}} & Physics + residual & 0.000 & 0.006\,$\pm$\,0.010 & 0--1 & 0.99\,$\pm$\,0.02 & 0.807\,$\pm$\,0.020 & no effect & pass & fail & fail & uninf. \\
 & Physics-only & 0.000 & 0.001\,$\pm$\,0.001 & 0 & 0.83\,$\pm$\,0.13 & 0.670\,$\pm$\,0.099 & no effect & pass & fail & fail & uninf. \\
\bottomrule
\end{tabular*}
\end{table*}

\par\textbf{Design.}
Three modifications each test one candidate cause of the missing outlet-stage response. The verdict rests on the
median outlet-stage response ratio $k_{\mathrm{stage}}$, the response ratio $k$ of Eq.~\eqref{eq:paired} for $\bar s$
at the \res{beta.nodes.near} near-outlet nodes. A value $k_{\mathrm{stage}}\ge0.5$ counts as fixed,
$0.1\le k_{\mathrm{stage}}<0.5$ as partial and $k_{\mathrm{stage}}<0.1$ as no effect, and a fix counts only if the rain
response ratio $k_{\mathrm{rain}}$ drops by no more than 0.10 and the test NSE with realized drivers by no more than
0.010 against the main-study run of the same architecture. Each modification is retrained with three seeds and scored
on the same \res{data.beta.n-storms-test} test storms. \emph{H1, more surge storms} (Scenario GNN, Hybrid forecaster)
trains on new storms of the same number, each with its own surge pulse, which lifts the 99th percentile of the 6-h
stage rise in training to \res{beta.s4.support.h1.surge_rise.p99}~cm and places $\bar s$ inside the training range.
\emph{H2, paired twin storms} (same two architectures) adds to the \res{data.beta.n-storms-train} training storms twins
that differ only in the surge, so that the models see outlet-stage contrasts at equal rain (stage-rise 99th percentile
\res{beta.s4.support.h2.surge_rise.p99}~cm). \emph{H3, one-way outfall edge} (Physics + state residual, Physics-only)
makes the two-way outfall edge of the models' graph one-way, like the simulator's flap gate. The run set Main holds
the unmodified runs, and the label Physics + residual denotes the Physics + state residual model. The rain share of
the main-study runs, the mean $|$model change$|$ over the mean $|$simulated change$|$ for $+25.4$~mm at the
\res{beta.nodes.rain} runoff-receiving nodes, lies from 0.98 to 1.10, and the stage share of
Table~\ref{tab:s-stage4} is the same ratio for $+30.48$~cm of added outlet stage at the \res{beta.nodes.near}
near-outlet nodes.

\par\textbf{Results.}
All six modified configurations keep $k_{\mathrm{stage}}=\res{beta.s4.h1.gconv.k_stage}$, so every verdict is no
effect (Table~\ref{tab:s-stage4}). For the Scenario GNN and the Hybrid forecaster this follows from the reach
(App.~\ref{app:I-reach}), which holds \res{beta.rf.gnn.field_size} and \res{beta.rf.hybrid.field_size} of the
\res{beta.nodes.near} near-outlet nodes, so H1 and H2 test these two architectures at their reachable junctions. There
the response does change. H2 moves the Scenario GNN's response close to the simulated change at three of its four
reachable junctions and the Hybrid forecaster's at the junction next to the outlet, whereas H1 leaves the Scenario
GNN's response about unchanged and lowers the Hybrid forecaster's (Fig.~\ref{fig:s-receptive}a). The stage-response
share over the 24 nodes, capped by the reach, moves from \res{beta.s4.3a.gconv.share_stage} and
\res{beta.s4.3a.wm3fpd.share_stage} in the main study to \res{beta.s4.h1.gconv.share_stage} and
\res{beta.s4.h1.wm3fpd.share_stage} under H1 and to \res{beta.s4.h2.gconv.share_stage} and
\res{beta.s4.h2.wm3fpd.share_stage} under H2. H3 lowers the share of the Physics + state residual and Physics-only
models from \res{beta.s4.3a.d9wm4.share_stage} and \res{beta.s4.3a.d9phys.share_stage} to
\res{beta.s4.h3.d9wm4.share_stage} and \res{beta.s4.h3.d9phys.share_stage}. The Physics-only model, which has no
residual, responds at a single near-outlet junction with $k_{\mathrm{stage}}=\res{beta.s4.3a.d9phys.k_stage}$, like
the Physics + state residual model, so removing the residual does not restore the response, although the two models
also differ in capacity. H2 lowered the Scenario GNN's rain response ratio from \res{beta.s4.3a.gconv.k_rain} to
\res{beta.s4.h2.gconv.k_rain}, beyond the tolerance, while the test NSE of every modified configuration stayed within
it.

On the same runs the observational check passes rain everywhere under the pooled rule with the station fixed-effect
correction. For the outlet stage it passes the Scenario GNN under H1 and H2 although $k_{\mathrm{stage}}$ is zero, and
the strict rule calls the outlet stage uninformative everywhere. In Table~\ref{tab:s-stage4} the columns Pooled + FE
and Strict give these two rules, uninf.\ is the verdict N, no information, of Table~\ref{tab:s-xr-verdicts}, and the
column of nodes above 1~cm compares with the simulator, in which the mean change exceeds 1~cm at all
\res{beta.nodes.near} near-outlet nodes. In Fig.~\ref{fig:s-receptive}, panels~c and~d give, for the unmodified runs and the six modified configurations with three seeds each,
the near-outlet $k$, the response share (ratio of mean changes) and the changes of rain $k$ and test NSE against the
unmodified runs, with the thresholds and tolerances of the verdict rules. The letters in
panel~c give the outlet-stage verdicts of the observational check, as in Table~\ref{tab:s-xr-verdicts}, in the column
orig.\ under the pooled rule with the station fixed-effect correction and in the column rev.\ under the strict rule.

None of the three modifications restores the outlet-stage response under the verdict criteria, so each is
ruled out as a sufficient remedy. For the Scenario GNN and the Hybrid forecaster the reach bounds the response near the
outlet by construction, and for the physics models the response stops at the adjacent junction although their reach
extends further. Two further candidate explanations, the timing of the fixed storm shape and the current distance to
the threshold, concern the event-level analyses in Miami-Dade and are tested with those analyses (App.~\ref{app:K}).

\subsection{Interaction estimands}\label{app:I4}

\par\textbf{Window peak and fixed time.}
An interaction counts as detectable where $|I_{RS}|>\max(c_2\nu,\delta)$.
The main interaction measure is the finite difference of window peaks at $(\bar A,\bar s)$. It is detectable in
\res{beta.s13.share_det.peak}\% of node and storm cells. The same difference of depths at the realized run's peak step
is detectable in \res{beta.s13.share_det.fixed}\%, and \res{beta.s13.peak_det_also_fixed}\% of the peak-detectable
cells are also detectable at that fixed time, with \res{beta.s13.sign_peak_vs_fixed}\% sign agreement
(Fig.~\ref{fig:s-compound}e). A detectable fixed-time interaction occurs at some step of the window in more cells
(\res{beta.s13.share_det.maxt}\%) than at the realized peak step. The interaction therefore also appears at fixed times
and does not come from the maximum operator alone.

\par\textbf{Peak timing.}
The joint change moves the window-peak time by at least one step in \res{beta.s13.peakshift.joint.any}\% of all cells
and in \res{beta.s13.peakshift.joint.det}\% of the cells with a detectable interaction. In
\res{beta.s13.joint_peak_time_new}\% of the latter the joint peak time differs from both single-driver peak times
(Fig.~\ref{fig:s-compound}d). In \res{beta.s13.base_peak_boundary}\% of cells the realized window maximum lies on the
first or last step of the window.
One step is 0.25~h.

\par\textbf{Sign and grid.}
Of the detectable interactions at $(\bar A,\bar s)$, \res{beta.fig3.irs.sim.share_positive}\% are positive (joint rise above
the sum of the single rises), and at the largest rain amount fewer are positive. The sign names no mechanism, because
storage, obstruction of outflow and capping at node rims all enter a window-peak difference. Over
the 30 grid points the detectable share ranges from \res{beta.s13.grid.share_det.min}\% to
\res{beta.s13.grid.share_det.max}\% (Fig.~\ref{fig:s-compound}c), while the largest signed recall of any model at any
point is \res{beta.s13.grid.recall_max.gconv}\% (Scenario GNN), \res{beta.s13.grid.recall_max.wm3fpd}\% (Hybrid
forecaster), \res{beta.s13.grid.recall_max.d9wm4}\% (Physics + state residual) and
\res{beta.s13.grid.recall_max.d9phys}\% (Physics-only model).
Fig.~\ref{fig:s-compound}c takes the simulator's detectable share over all
\res{beta.nodes.scored}\,$\times$\,\res{data.beta.cf-issue-times} node and storm cells at each grid point.

\par\textbf{Errors.}
On detectable cells every model's interaction error equals that of a model with no interaction,
\res{beta.fig3.irs.nointeraction.mae_det}~cm. Over all node and storm cells, without the detectability filter, the
errors are \res{beta.fig3.irs.gconv.trained.mae_all}~cm for the Scenario GNN and the Hybrid forecaster and
\res{beta.fig3.irs.d9wm4.trained.mae_all}~cm for the physics models, against
\res{beta.fig3.irs.nointeraction.mae_all}~cm without interaction. The all-cell error is small because most cells have
almost no interaction, and no trained model is closer to the simulator than the no-interaction baseline.

\par\textbf{Coalition definition.}
The coalition definition takes the interaction from the
Shapley coalition runs (realized tide, with $0$ meaning no future rain or no future surge) and counts it where
$|I_{RS}|>2\nu$, in \res{beta.fig3.irs.coalition.share_gt_2nf}\% of cells. With the four-run floor $\max(c_2\nu,\delta)$ it gives
\res{beta.fig3.irs.coalition.share_det}\%, and \res{beta.fig3.irs.coalition.share_det_quality}\% after removing the
\res{beta.fig3.irs.coalition.n_issue_flagged} issue times in which one of the four realized-tide runs used here exceeds
2\% continuity error (the 2\% Shapley sensitivity rule of App.~\ref{app:D} checks all eight coalitions and leaves
\res{beta.quality.shapley-usable-2pct} issue times). The Scenario GNN's signed recall under this definition is
\res{beta.fig3.irs.coalition.gconv.recall}\%. Both definitions lead to the same reading. The simulator's rain $\times$
outlet-stage interaction is present and no model reproduces it.

\subsection{Evidence classes by kind of explanation}\label{app:I6}

\begin{table}[tp]
\caption{Evidence classes and seed-mean metrics across queries (valves graded on the river-valve network, others on the coastal network).}
\label{tab:s-classes}
\centering
\setlength{\tabcolsep}{3pt}%
\resizebox{\columnwidth}{!}{%
\begin{tabular}{@{}llccccll@{}}
\toprule
Explanation & Metric & \begin{tabular}[b]{@{}c@{}}Scenario\\GNN\end{tabular} & Hybrid & \begin{tabular}[b]{@{}c@{}}Physics +\\residual\end{tabular} & \begin{tabular}[b]{@{}c@{}}Physics-\\only\end{tabular} & Ref. & Class \\
\midrule
\multicolumn{8}{@{}c}{\emph{Q1}} \\
Rain $\phi_R$ & $\rho$ & \res{beta.fig2a.r.gnn.rho.mean} & \res{beta.fig2a.r.hybrid.rho.mean} & \res{beta.fig2a.r.physres.rho.mean} & \res{beta.fig2a.r.physonly.rho.mean} & \res{beta.fig2a.r.physres-untrained.rho}$^{u}$ & in part \\
Surge $\phi_S$ & $|e|$ & \res{beta.fig2a.s.gnn.abs-err.mean} & \res{beta.fig2a.s.hybrid.abs-err.mean} & \res{beta.fig2a.s.physres.abs-err.mean} & \res{beta.fig2a.s.physonly.abs-err.mean} & \res{beta.fig2a.true-abs-phi-disc.s}$^{n}$ & wrong \\
Tide $\phi_T$ & -- & -- & -- & -- & -- & -- & untested \\
\midrule
\multicolumn{8}{@{}c}{\emph{Q2}} \\
Added rain & $k$ & \res{beta.k.rain.gnn.k-median.mean} & \res{beta.k.rain.hybrid.k-median.mean} & \res{beta.k.rain.physres.k-median.mean} & \res{beta.k.rain.physonly.k-median.mean} & \res{beta.k.rain.physres-untrained.k-median}$^{u}$ & in part \\
Raised stage & $k$ & \res{beta.k.stage.gnn.k-median.mean} & \res{beta.k.stage.hybrid.k-median.mean} & \res{beta.k.stage.physres.k-median.mean} & \res{beta.k.stage.physonly.k-median.mean} & \res{beta.k.stage.physres-untrained.k-median}$^{u}$ & wrong \\
Valves & $k$ & \res{gamma.k_valve.gnn.trained} & \res{gamma.k_valve.hybrid.trained} & \res{gamma.k_valve.physres.trained} & \res{gamma.k_valve.physonly.trained} & \res{gamma.k_valve.physres.untrained}$^{u}$ & in part \\
Gate R2 & $k$ & \res{ctrl.k.r2.gnn.trained.mean} & \res{ctrl.k.r2.hybrid.trained.mean} & \res{ctrl.k.r2.physres.trained.mean} & \res{ctrl.k.r2.physonly.trained.mean} & \res{ctrl.k.r2.physres.untrained}$^{u}$ & wrong \\
Weir W0 & $k$ & \res{ctrl.k.w0.gnn.trained.mean} & \res{ctrl.k.w0.hybrid.trained.mean} & \res{ctrl.k.w0.physres.trained.mean} & \res{ctrl.k.w0.physonly.trained.mean} & \res{ctrl.k.w0.physres.untrained}$^{u}$ & wrong \\
Pump P0 & $k$ & \res{ctrl.k.p0.gnn.trained.mean} & \res{ctrl.k.p0.hybrid.trained.mean} & \res{ctrl.k.p0.physres.trained.mean} & \res{ctrl.k.p0.physonly.trained.mean} & \res{ctrl.k.p0.physres.untrained}$^{u}$ & wrong \\
\midrule
\multicolumn{8}{@{}c}{\emph{Q3}} \\
$A^\ast$ & BA & \res{beta.tab3.gnn.ba.mean} & \res{beta.tab3.hybrid.ba.mean} & \res{beta.tab3.physres.ba.mean} & \res{beta.tab3.physonly.ba.mean} & \begin{tabular}[t]{@{}l@{}}\res{beta.tab3.always.ba}$^{c}$\\\res{beta.tab3.linear.ba}$^{l}$\end{tabular} & in part \\
\bottomrule
\end{tabular}}
\end{table}

\par\textbf{Local and architecture-level classes.}
Evidence grading assigns each explanation type one evidence class per architecture and water system (Section~\ref{sec:method}), and
Table~\ref{tab:s-classes} summarizes these classes alongside the underlying paired-run metrics. In an observed region, the forecast itself carries the class \emph{forecast} (evaluated retrospectively against gauge observations in Table~\ref{tab:xr}), whereas every driver explanation is locally \emph{not checkable} due to the absence of counterfactual ground truth, and is paired with its architecture-level class from the simulated drainage networks. The simulated
coastal network (\emph{beta}) supplies the grades for rainfall and surge attributions ($\phi_R,\phi_S$), scenario responses to added rainfall, raised outlet stage, and gate, weir, and pump operations (App.~\ref{app:P}), and minimum crossing rainfall ($A^\ast$), while the second simulated
network (\emph{gamma}, App.~\ref{app:L6}) supplies the grade for valve settings. Because astronomical tide is deterministic and not perturbed in paired runs, tide attribution ($\phi_T$) is graded \emph{not tested}. Compound interaction $I_{RS}$ is evaluated as part of Q2 scenario changes in App.~\ref{app:I4}.

\par\textbf{Metrics and grading criteria.}
In Table~\ref{tab:s-classes}, rows are grouped by queries Q1--Q3, and the four architecture columns report the primary metric behind each grade: Spearman node rank correlation $\rho$, mean absolute attribution error $|e|$ (cm), median response ratio $k$, or threshold balanced accuracy (BA). The reference column (Ref.) reports the highest untrained control ($u$), a non-responding baseline ($n$), a constant classifier ($c$), or the training-fit linear rainfall baseline ($l$), while the Class column abbreviates \emph{supported in part} as \emph{in part}, \emph{shown wrong} as \emph{wrong}, and \emph{not tested} as \emph{untested}. Under the paired-run grading criteria (App.~\ref{app:P2}), all twelve gate, weir, and pump pairs on the coastal network ($k=0.00$) as well as the raised-stage response ($k=0.00$) and surge attribution ($|e|$ equal to the non-responding baseline) are graded \emph{shown wrong}. For added rainfall (where $k$ reaches \res{ctrl.rainrule.gnn.k}--\res{ctrl.rainrule.physres.k} with sign agreement \res{ctrl.rainrule.gnn.sign}\%--\res{ctrl.rainrule.physonly.sign}\%) and valve operations on the second network ($k$ from \res{ctrl.valverule.gnn.k} to \res{ctrl.valverule.physonly.k} with sign agreement \res{ctrl.valverule.gnn.sign}\%--\res{ctrl.valverule.physonly.sign}\%), although the Hybrid and water-balance models satisfy the single-point magnitude window $[0.75,1.33]$, their response curves flatten across storm intensities or understate multi-step magnitude variations (Apps.~\ref{app:I1} and~\ref{app:L6}), so Table~\ref{tab:s-classes} assigns the conservative grade \emph{supported in part} across all four architectures.

\section{Threshold Decisions and Usage Tasks}\label{app:K}

This appendix gives the complete results behind the analyses of threshold decisions. It covers the accounting of every
threshold record, the full counterfactual classification against the simulator, the censoring and robustness checks,
the protocols of the three usage tasks with their operating points and comparisons without thresholds, and the timing
scans.
Fig.~\ref{fig:state-event} in Section~\ref{sec:results-threshold} shows U1 and U2 in Miami-Dade, and
Fig.~\ref{fig:s-censoring} the censoring and the threshold robustness.

\subsection{Threshold-record accounting}\label{app:K1}

\par\textbf{Records.}
The Miami-Dade records cover the \res{mia.thr.n_rain_stations} rain-group gauges at \res{mia.thr.n_issue} issue times
per year, \res{mia.thr.n_cells} cells per year. The high-water threshold $u_i$ is the training-period 95th percentile of
each gauge, a crossing is a forecast peak of at least $u_i+\varepsilon$ with $\varepsilon=1$~cm, and the added-rain
range is $[0,A_{\max}]$ with $A_{\max}=\res{mia.thr.amax_mm}$~mm per 24~h (training maximum
\res{mia.thr.train_max24h_mm}~mm), searched on 41 points \res{mia.thr.grid_step_mm}~mm apart. Every cell receives one
status (Table~\ref{tab:s-accounting}), and Fig.~\ref{fig:s-censoring}b shows where every rain-group cell ends up, per
seed and year. No cell has a non-finite margin (\res{mia.thr.numerical_failures} numerical failures).

\begin{table*}[t]
\caption{Threshold-record accounting for the Miami-Dade rain group in 2019 (Test) and 2018 (Val.), mean\,$\pm$\,SD over three seeds (\% of unwarned cells unless noted).}
\label{tab:s-accounting}
\resizebox{\textwidth}{!}{%
\begin{tabular}{llcccccccc}
\toprule
 &  & \multicolumn{2}{c}{Scenario GNN} & \multicolumn{2}{c}{Hybrid} & \multicolumn{2}{c}{Physics + residual} & \multicolumn{2}{c}{Physics-only} \\
\cmidrule(lr){3-4}\cmidrule(lr){5-6}\cmidrule(lr){7-8}\cmidrule(lr){9-10}
 &  & Test & Val. & Test & Val. & Test & Val. & Test & Val. \\
\midrule
\multirow{5}{*}{\rotatebox[origin=c]{90}{(a) Shares (\%)}} & Warned at issue & 5.3\,$\pm$\,0.0 & 3.2\,$\pm$\,0.0 & 5.3\,$\pm$\,0.1 & 3.5\,$\pm$\,0.0 & 5.1\,$\pm$\,0.0 & 2.9\,$\pm$\,0.0 & 5.2\,$\pm$\,0.0 & 3.4\,$\pm$\,0.0 \\
\cmidrule(l){2-10}
 & Not crossed, monotone-grid & 46.7\,$\pm$\,0.5 & 47.6\,$\pm$\,0.7 & 69.2\,$\pm$\,9.9 & 67.5\,$\pm$\,10.2 & 68.0\,$\pm$\,1.0 & 66.8\,$\pm$\,0.8 & 80.5\,$\pm$\,0.6 & 77.3\,$\pm$\,1.4 \\
 & Not crossed, first-observed & 46.2\,$\pm$\,1.5 & 47.0\,$\pm$\,1.6 & 69.6\,$\pm$\,10.7 & 68.7\,$\pm$\,10.5 & 67.8\,$\pm$\,0.8 & 66.5\,$\pm$\,0.6 & 80.3\,$\pm$\,0.9 & 77.2\,$\pm$\,1.5 \\
 & Non-monotone & 11.6\,$\pm$\,1.2 & 11.2\,$\pm$\,1.3 & 10.7\,$\pm$\,1.7 & 12.6\,$\pm$\,2.3 & 0.3\,$\pm$\,0.3 & 0.6\,$\pm$\,0.4 & 0.3\,$\pm$\,0.4 & 0.2\,$\pm$\,0.3 \\
\cmidrule(l){2-10}
 & Q3 missed & 1.58\,$\pm$\,0.31 & 1.97\,$\pm$\,0.13 & 0.14\,$\pm$\,0.15 & 0.21\,$\pm$\,0.19 & 1.53\,$\pm$\,0.10 & 0.97\,$\pm$\,0.05 & 5.92\,$\pm$\,0.38 & 1.80\,$\pm$\,0.22 \\
\midrule
\multirow{7}{*}{\rotatebox[origin=c]{90}{(b) Cells, $A^\ast$}} & Visible dip & 1{,}925\,$\pm$\,68 & 1{,}421\,$\pm$\,101 & 3{,}027\,$\pm$\,661 & 2{,}366\,$\pm$\,537 & 31\,$\pm$\,41 & 77\,$\pm$\,118 & 16\,$\pm$\,14 & 13\,$\pm$\,9 \\
 & Near threshold & 850\,$\pm$\,6 & 672\,$\pm$\,7 & 868\,$\pm$\,47 & 689\,$\pm$\,14 & 708\,$\pm$\,10 & 623\,$\pm$\,23 & 854\,$\pm$\,0 & 644\,$\pm$\,25 \\
 & Last interval & 2{,}105\,$\pm$\,232 & 2{,}055\,$\pm$\,240 & 1{,}254\,$\pm$\,282 & 1{,}169\,$\pm$\,338 & 1{,}096\,$\pm$\,15 & 978\,$\pm$\,26 & 646\,$\pm$\,10 & 526\,$\pm$\,17 \\
 & $\varepsilon$-sensitive & 2{,}499\,$\pm$\,201 & 2{,}917\,$\pm$\,86 & 2{,}252\,$\pm$\,225 & 2{,}279\,$\pm$\,288 & 2{,}315\,$\pm$\,103 & 2{,}035\,$\pm$\,84 & 2{,}163\,$\pm$\,67 & 1{,}726\,$\pm$\,84 \\
 & $s^\ast$ found & 0\,$\pm$\,0 & 0\,$\pm$\,0 & 0\,$\pm$\,0 & 0\,$\pm$\,0 & 59\,$\pm$\,3 & 2\,$\pm$\,0 & 77\,$\pm$\,3 & 55\,$\pm$\,5 \\
\cmidrule(l){2-10}
 & Median $A^\ast$ (mm) & 64.6\,$\pm$\,1.6 & 65.6\,$\pm$\,1.6 & 62.7\,$\pm$\,0.8 & 61.1\,$\pm$\,1.2 & 58.2\,$\pm$\,0.6 & 54.5\,$\pm$\,0.7 & 56.0\,$\pm$\,1.0 & 53.8\,$\pm$\,0.5 \\
 & IQR of $A^\ast$ (mm) & 54.4--72.2 & 54.6--72.5 & 48.7--72.2 & 47.4--71.1 & 43.4--70.0 & 41.7--66.8 & 38.7--68.7 & 38.5--65.9 \\
\bottomrule
\end{tabular}}
\end{table*}

\par\textbf{Statuses.}
In the test year from \res{mia.thr.warned.min}\% to \res{mia.thr.warned.max}\% of the cells are warned under $q_0$ and
leave the analysis. Of the remaining cells, from \res{mia.thr.range.notcrossed.min}\% to
\res{mia.thr.range.notcrossed.max}\% are not crossed within range under the record convention, which removes
non-monotone cells from the denominator (the monotone-grid rows of Table~\ref{tab:s-accounting}), and from
\res{mia.thr.range.notcrossed_all.min}\% to \res{mia.thr.range.notcrossed_all.max}\% under the first-observed
convention, which keeps every cell and enters non-monotone cells at their first observed crossing. Under this
convention the crossed share is one minus the not-crossed share. The validation year gives from
\res{mia.thr.range.val.notcrossed.min}\% to \res{mia.thr.range.val.notcrossed.max}\% under the record convention.
Non-monotone grids, where the margin decreases by more than $10^{-3}$~cm somewhere, occur in
\res{mia.thr.nonmono.gnn}\% of the Scenario GNN's and \res{mia.thr.nonmono.hybrid}\% of the Hybrid forecaster's unwarned
cells and in at most \res{mia.thr.nonmono.physres}\% for the physics models. On the 11-point subgrid between
\res{mia.thr.range.nonmono-visible.min} and \res{mia.thr.range.nonmono-visible.max} cells per run show a decrease
larger than 0.1~cm, the visible dips of Table~\ref{tab:s-accounting}. The median $A^\ast$ among crossers lies between
\res{mia.thr.range.astar-median.min} and
\res{mia.thr.range.astar-median.max}~mm. With a large share of the cells not crossed, it describes crossers, not the
stations' risk.

\par\textbf{Boundary-sensitive cells.}
Per run, between \res{mia.thr.range.near-threshold.min} and \res{mia.thr.range.near-threshold.max} unwarned cells have
a $q_0$ peak within 1~cm below the threshold, between \res{mia.thr.range.last-interval.min} and
\res{mia.thr.range.last-interval.max} crossings fall in the last grid interval, where $A^\ast$ depends most on
$A_{\max}$, and between \res{mia.thr.range.eps-sensitive.min} and \res{mia.thr.range.eps-sensitive.max} non-crossers
would cross with $\varepsilon=0$, because their margin at $A_{\max}$ is at least zero. Table~\ref{tab:s-accounting}
lists them as near-threshold, last-interval and $\varepsilon$-sensitive cells. These cells are counted, not removed.

\par\textbf{Surge, joint and tidal-range records.}
The analogous minimum crossing surge $s^\ast$ (added surge up to \res{mia.thr.smax_cm}~cm, ramped over 6~h) is found in
no rain-group cell of the Scenario GNN or the Hybrid forecaster and in \res{mia.thr.sstar_found.physres} and
\res{mia.thr.sstar_found.physonly} cells of the two physics models. In surge group~A it is found in
\res{mia.thr.sstar_found_surgegroup.gnn} (Scenario GNN) to \res{mia.thr.sstar_found_surgegroup.physres} cells (Physics +
state residual), with median $s^\ast$ between \res{mia.thr.sstar_median_surgegroup.physonly} and
\res{mia.thr.sstar_median_surgegroup.gnn}~cm. Joint-grid crossings that neither single-factor end point produces, the
Q3 missed row of Table~\ref{tab:s-accounting}, make up from \res{mia.thr.range.q3-missed.min}\% to
\res{mia.thr.range.q3-missed.max}\% of the joint rain and surge crossings over the twelve runs. For the Physics + state
residual model the per-seed range is from
\res{mia.xai.test.rain.physres.single.miss-rate.min}\% to \res{mia.xai.test.rain.physres.single.miss-rate.max}\%. At
the rain-group gauges the astronomical tidal range of the horizon is almost unrelated to $A^\ast$ (pooled rank
correlation from \res{mia.q2.pooled.min} to \res{mia.q2.pooled.max}, within-station from \res{mia.q2.within.min} to
about zero).

\subsection{Counterfactual threshold classification in full}\label{app:K2}

\par\textbf{Protocol.}
The common denominator holds the \res{beta.tab3.n} usable (issue time, threshold node) cells whose simulated base peak
lies below the threshold, the node's training p95, and \res{beta.tab3.pos-rate}\% of them cross within the added-rain
grid in the simulator. The cells come from \res{data.beta.n-storms-test} test-storm issue times and
\res{data.beta.n-threshold-nodes} threshold nodes. Rain is added uniformly over the 6-h window, from 0 to
\res{data.beta.rain-add-max-mm}~mm, and the outlet stage from 0 to 60.96~cm. A model predicts a crossing if its own
window peak reaches the threshold at some grid amount including zero, so an initial false alarm counts as a predicted
crossing and is tallied separately. Table~\ref{tab:cf-threshold} summarizes the added-rain decisions of the four trained models against
the constant and linear baselines, and Table~\ref{tab:s-confusion} gives the full confusion matrices for both drivers,
every variant, and the constant and linear baselines. In these tables, GNN, P+R and P-only stand for the Scenario GNN, the
Physics + state residual model and the Physics-only model, and the linear fit is the training-fit linear response. BA
is the balanced accuracy, Rec.+ and Rec.$-$ are the crossing and non-crossing recall, Prec.\ is the crossing precision
and IFA counts the initial false alarms, the cells whose model base peak lies above the threshold. The untrained rows
use one initialization.

\begin{table*}[t]
\caption{Counterfactual threshold classification under added rain on the simulated coastal network ($N=12{,}778$ cells; mean\,$\pm$\,SD over three seeds; UD: undefined).}
\label{tab:cf-threshold}
\setlength{\tabcolsep}{4pt}%
\begin{tabular*}{\textwidth}{@{\extracolsep{\fill}}lcccccc@{}}
\toprule
Predictor & \begin{tabular}[b]{@{}c@{}}Balanced\\acc.\end{tabular} & \begin{tabular}[b]{@{}c@{}}Crossing\\recall\end{tabular}
 & \begin{tabular}[b]{@{}c@{}}Crossing\\precision\end{tabular} & \begin{tabular}[b]{@{}c@{}}Median $|A^\ast|$\\error (mm)\end{tabular}
 & \begin{tabular}[b]{@{}c@{}}Error\\coverage $n$\end{tabular} & \begin{tabular}[b]{@{}c@{}}Initial false\\alarms (\%)\end{tabular} \\
\midrule
Always-cross & 0.50 & 1.00 & 0.89 & -- & -- & -- \\
Never-cross & 0.50 & 0.00 & UD & -- & -- & -- \\
Linear response & 0.58 & 0.87 & 0.91 & 13.5 & 8{,}459 & 13.1 \\
\midrule
Scenario GNN & 0.63\,$\pm$\,0.01 & 0.88\,$\pm$\,0.02 & 0.92\,$\pm$\,$<$0.01 & 8.8\,$\pm$\,0.9 & 9{,}256\,$\pm$\,108 & 8.0\,$\pm$\,1.3 \\
Hybrid & 0.72\,$\pm$\,0.03 & 0.72\,$\pm$\,0.12 & 0.95\,$\pm$\,0.01 & 11.6\,$\pm$\,1.9 & 7{,}453\,$\pm$\,1{,}402 & 6.6\,$\pm$\,0.2 \\
Physics + residual & 0.69\,$\pm$\,0.02 & 0.80\,$\pm$\,0.03 & 0.94\,$\pm$\,$<$0.01 & 11.1\,$\pm$\,0.6 & 8{,}494\,$\pm$\,278 & 6.5\,$\pm$\,0.1 \\
Physics-only & 0.67\,$\pm$\,0.02 & 0.71\,$\pm$\,0.03 & 0.94\,$\pm$\,$<$0.01 & 11.3\,$\pm$\,0.3 & 7{,}511\,$\pm$\,331 & 5.9\,$\pm$\,0.4 \\
\bottomrule
\end{tabular*}
\end{table*}

\begin{table*}[tp]
\caption{Counterfactual threshold confusion on the simulated coastal network ($N=12{,}778$; rates: mean\,$\pm$\,SD over three seeds; counts: seed means; UD: undefined).}
\label{tab:s-confusion}
\centering
\setlength{\tabcolsep}{4pt}%
\begin{adjustbox}{max width=\textwidth}
\begin{tabular}{@{}cc*{4}{c}*{5}{c}c@{}}
\toprule
\multicolumn{2}{@{}l}{\multirow{2}{*}{Predictor}} & \multicolumn{4}{c}{Counts} & \multicolumn{5}{c}{Classification} & \multirow{2}{*}{IFA} \\
\cmidrule(lr){3-6}\cmidrule(lr){7-11}
\multicolumn{2}{@{}l}{} & TP & FP & TN & FN & BA & MCC & Rec.+ & Rec.$-$ & Prec. & \\
\midrule
\multicolumn{12}{@{}c}{\emph{Rain}} \\
\multicolumn{2}{l}{Always-cross} & 11{,}341 & 1{,}437 & 0 & 0 & 0.50 & UD & 1.00 & 0.00 & 0.89 & -- \\
\multicolumn{2}{l}{Never-cross} & 0 & 0 & 1{,}437 & 11{,}341 & 0.50 & UD & 0.00 & 1.00 & UD & -- \\
\multicolumn{2}{l}{Linear fit} & 9{,}822 & 1{,}019 & 418 & 1{,}519 & 0.58 & 0.14 & 0.87 & 0.29 & 0.91 & 1{,}673 \\
\cmidrule(l){1-12}
\multirow{4}{*}{\rotatebox[origin=c]{90}{Trained}} & GNN & 10{,}014 & 893 & 544 & 1{,}327 & 0.63\,$\pm$\,0.01 & 0.23\,$\pm$\,0.02 & 0.88\,$\pm$\,0.02 & 0.38\,$\pm$\,$<$0.01 & 0.92\,$\pm$\,$<$0.01 & 1{,}026 \\
 & Hybrid & 8{,}169 & 406 & 1{,}031 & 3{,}172 & 0.72\,$\pm$\,0.03 & 0.30\,$\pm$\,0.06 & 0.72\,$\pm$\,0.12 & 0.72\,$\pm$\,0.08 & 0.95\,$\pm$\,0.01 & 842 \\
 & P+R & 9{,}094 & 599 & 838 & 2{,}247 & 0.69\,$\pm$\,0.02 & 0.28\,$\pm$\,0.03 & 0.80\,$\pm$\,0.03 & 0.58\,$\pm$\,0.03 & 0.94\,$\pm$\,$<$0.01 & 836 \\
 & P-only & 8{,}053 & 533 & 904 & 3{,}288 & 0.67\,$\pm$\,0.02 & 0.23\,$\pm$\,0.02 & 0.71\,$\pm$\,0.03 & 0.63\,$\pm$\,0.03 & 0.94\,$\pm$\,$<$0.01 & 753 \\
\cmidrule(l){1-12}
\multirow{4}{*}{\rotatebox[origin=c]{90}{\begin{tabular}{@{}c@{}}Half-\\input\end{tabular}}} & GNN & 1{,}282 & 276 & 1{,}161 & 10{,}059 & 0.46\,$\pm$\,0.01 & $-$0.08\,$\pm$\,0.01 & 0.11\,$\pm$\,0.02 & 0.81\,$\pm$\,0.03 & 0.82\,$\pm$\,$<$0.01 & 498 \\
 & Hybrid & 2{,}035 & 213 & 1{,}224 & 9{,}306 & 0.52\,$\pm$\,0.02 & 0.02\,$\pm$\,0.03 & 0.18\,$\pm$\,0.06 & 0.85\,$\pm$\,0.03 & 0.90\,$\pm$\,0.02 & 346 \\
 & P+R & 1{,}340 & 250 & 1{,}187 & 10{,}001 & 0.47\,$\pm$\,$<$0.01 & $-$0.05\,$\pm$\,$<$0.01 & 0.12\,$\pm$\,$<$0.01 & 0.83\,$\pm$\,0.01 & 0.84\,$\pm$\,$<$0.01 & 428 \\
 & P-only & 1{,}025 & 230 & 1{,}207 & 10{,}316 & 0.47\,$\pm$\,0.01 & $-$0.07\,$\pm$\,0.01 & 0.09\,$\pm$\,0.02 & 0.84\,$\pm$\,0.02 & 0.82\,$\pm$\,0.02 & 375 \\
\cmidrule(l){1-12}
\multirow{4}{*}{\rotatebox[origin=c]{90}{\begin{tabular}{@{}c@{}}Reversed-\\input\end{tabular}}} & GNN & 166 & 79 & 1{,}358 & 11{,}175 & 0.48\,$\pm$\,0.02 & $-$0.08\,$\pm$\,0.09 & 0.01\,$\pm$\,$<$0.01 & 0.94\,$\pm$\,0.05 & 0.73\,$\pm$\,0.19 & 229 \\
 & Hybrid & 692 & 176 & 1{,}261 & 10{,}649 & 0.47\,$\pm$\,$<$0.01 & $-$0.09\,$\pm$\,0.03 & 0.06\,$\pm$\,0.05 & 0.88\,$\pm$\,0.05 & 0.76\,$\pm$\,0.08 & 350 \\
 & P+R & 375 & 151 & 1{,}286 & 10{,}966 & 0.46\,$\pm$\,0.01 & $-$0.11\,$\pm$\,0.04 & 0.03\,$\pm$\,$<$0.01 & 0.90\,$\pm$\,0.02 & 0.71\,$\pm$\,0.05 & 341 \\
 & P-only & 297 & 59 & 1{,}378 & 11{,}044 & 0.49\,$\pm$\,0.02 & $-$0.03\,$\pm$\,0.08 & 0.03\,$\pm$\,0.01 & 0.96\,$\pm$\,0.03 & 0.83\,$\pm$\,0.15 & 257 \\
\cmidrule(l){1-12}
\multirow{4}{*}{\rotatebox[origin=c]{90}{Untrained}} & GNN & 272 & 204 & 1{,}233 & 11{,}069 & 0.44 & $-$0.20 & 0.02 & 0.86 & 0.57 & 475 \\
 & Hybrid & 140 & 88 & 1{,}349 & 11{,}201 & 0.48 & $-$0.12 & 0.01 & 0.94 & 0.61 & 227 \\
 & P+R & 181 & 167 & 1{,}270 & 11{,}160 & 0.45 & $-$0.19 & 0.02 & 0.88 & 0.52 & 343 \\
 & P-only & 173 & 113 & 1{,}324 & 11{,}168 & 0.47 & $-$0.14 & 0.02 & 0.92 & 0.60 & 276 \\
\midrule
\multicolumn{12}{@{}c}{\emph{Outlet stage}} \\
\multicolumn{2}{l}{Always-cross} & 1{,}002 & 11{,}776 & 0 & 0 & 0.50 & UD & 1.00 & 0.00 & 0.08 & -- \\
\multicolumn{2}{l}{Never-cross} & 0 & 0 & 11{,}776 & 1{,}002 & 0.50 & UD & 0.00 & 1.00 & UD & -- \\
\multicolumn{2}{l}{Linear fit} & 48 & 621 & 11{,}155 & 954 & 0.50 & $-$0.01 & 0.05 & 0.95 & 0.07 & 654 \\
\cmidrule(l){1-12}
\multirow{4}{*}{\rotatebox[origin=c]{90}{Trained}} & GNN & 356 & 730 & 11{,}046 & 646 & 0.65\,$\pm$\,0.02 & 0.28\,$\pm$\,0.03 & 0.35\,$\pm$\,0.05 & 0.94\,$\pm$\,0.01 & 0.33\,$\pm$\,0.03 & 1{,}026 \\
 & Hybrid & 137 & 714 & 11{,}062 & 865 & 0.54\,$\pm$\,$<$0.01 & 0.08\,$\pm$\,0.01 & 0.14\,$\pm$\,$<$0.01 & 0.94\,$\pm$\,$<$0.01 & 0.16\,$\pm$\,0.01 & 842 \\
 & P+R & 121 & 717 & 11{,}059 & 881 & 0.53\,$\pm$\,$<$0.01 & 0.06\,$\pm$\,0.01 & 0.12\,$\pm$\,0.01 & 0.94\,$\pm$\,$<$0.01 & 0.14\,$\pm$\,$<$0.01 & 836 \\
 & P-only & 116 & 638 & 11{,}138 & 886 & 0.53\,$\pm$\,$<$0.01 & 0.07\,$\pm$\,$<$0.01 & 0.12\,$\pm$\,0.01 & 0.95\,$\pm$\,$<$0.01 & 0.15\,$\pm$\,$<$0.01 & 753 \\
\cmidrule(l){1-12}
\multirow{4}{*}{\rotatebox[origin=c]{90}{\begin{tabular}{@{}c@{}}Half-\\input\end{tabular}}} & GNN & 102 & 408 & 11{,}368 & 900 & 0.53\,$\pm$\,0.01 & 0.09\,$\pm$\,0.02 & 0.10\,$\pm$\,0.02 & 0.97\,$\pm$\,$<$0.01 & 0.20\,$\pm$\,0.03 & 498 \\
 & Hybrid & 57 & 289 & 11{,}487 & 945 & 0.52\,$\pm$\,$<$0.01 & 0.05\,$\pm$\,0.01 & 0.06\,$\pm$\,$<$0.01 & 0.98\,$\pm$\,$<$0.01 & 0.17\,$\pm$\,0.02 & 346 \\
 & P+R & 49 & 379 & 11{,}397 & 953 & 0.51\,$\pm$\,$<$0.01 & 0.03\,$\pm$\,$<$0.01 & 0.05\,$\pm$\,$<$0.01 & 0.97\,$\pm$\,$<$0.01 & 0.11\,$\pm$\,0.01 & 428 \\
 & P-only & 52 & 323 & 11{,}453 & 950 & 0.51\,$\pm$\,$<$0.01 & 0.04\,$\pm$\,0.01 & 0.05\,$\pm$\,$<$0.01 & 0.97\,$\pm$\,$<$0.01 & 0.14\,$\pm$\,0.02 & 375 \\
\cmidrule(l){1-12}
\multirow{4}{*}{\rotatebox[origin=c]{90}{\begin{tabular}{@{}c@{}}Reversed-\\input\end{tabular}}} & GNN & 12 & 217 & 11{,}559 & 990 & 0.50\,$\pm$\,$<$0.01 & $-$0.02\,$\pm$\,0.02 & 0.01\,$\pm$\,0.02 & 0.98\,$\pm$\,0.01 & 0.04\,$\pm$\,0.04 & 229 \\
 & Hybrid & 61 & 289 & 11{,}487 & 941 & 0.52\,$\pm$\,0.02 & 0.06\,$\pm$\,0.05 & 0.06\,$\pm$\,0.03 & 0.98\,$\pm$\,$<$0.01 & 0.17\,$\pm$\,0.07 & 350 \\
 & P+R & 27 & 314 & 11{,}462 & 975 & 0.50\,$\pm$\,0.01 & 0.00\,$\pm$\,0.04 & 0.03\,$\pm$\,0.02 & 0.97\,$\pm$\,0.01 & 0.08\,$\pm$\,0.06 & 341 \\
 & P-only & 14 & 243 & 11{,}533 & 988 & 0.50\,$\pm$\,0.01 & $-$0.01\,$\pm$\,0.04 & 0.01\,$\pm$\,0.02 & 0.98\,$\pm$\,$<$0.01 & 0.05\,$\pm$\,0.07 & 257 \\
\cmidrule(l){1-12}
\multirow{4}{*}{\rotatebox[origin=c]{90}{Untrained}} & GNN & 42 & 433 & 11{,}343 & 960 & 0.50 & 0.01 & 0.04 & 0.96 & 0.09 & 475 \\
 & Hybrid & 4 & 223 & 11{,}553 & 998 & 0.49 & $-$0.03 & 0.00 & 0.98 & 0.02 & 227 \\
 & P+R & 26 & 317 & 11{,}459 & 976 & 0.50 & $-$0.00 & 0.03 & 0.97 & 0.08 & 343 \\
 & P-only & 22 & 254 & 11{,}522 & 980 & 0.50 & 0.00 & 0.02 & 0.98 & 0.08 & 276 \\
\bottomrule
\end{tabular}
\end{adjustbox}
\end{table*}

\par\textbf{Rain.}
Paired storm-bootstrap differences of balanced accuracy (5,000 resamples, seeds averaged within each resample) place
every trained model above the training-fit linear response. The difference to the linear response is
\res{beta.tab3.boot.gnn-minus-linear.ba.point} [\res{beta.tab3.boot.gnn-minus-linear.ba.lo},
\res{beta.tab3.boot.gnn-minus-linear.ba.hi}] for the Scenario GNN, \res{beta.tab3.boot.hybrid-minus-linear.ba.point}
[\res{beta.tab3.boot.hybrid-minus-linear.ba.lo}, \res{beta.tab3.boot.hybrid-minus-linear.ba.hi}] for the Hybrid
forecaster, \res{beta.tab3.boot.physres-minus-linear.ba.point} [\res{beta.tab3.boot.physres-minus-linear.ba.lo},
\res{beta.tab3.boot.physres-minus-linear.ba.hi}] for the Physics + state residual model and
\res{beta.tab3.boot.physonly-minus-linear.ba.point} [\res{beta.tab3.boot.physonly-minus-linear.ba.lo},
\res{beta.tab3.boot.physonly-minus-linear.ba.hi}] for the Physics-only model. The Hybrid forecaster exceeds the
Scenario GNN by \res{beta.tab3.boot.hybrid-minus-gnn.ba.point} [\res{beta.tab3.boot.hybrid-minus-gnn.ba.lo},
\res{beta.tab3.boot.hybrid-minus-gnn.ba.hi}]. Initial false alarms make up from
\res{beta.tab3.physonly.init-fa-share.mean}\% to \res{beta.tab3.gnn.init-fa-share.mean}\% of the denominator for the
trained models and supply from \res{beta.tab3.physres.tp-from-ifa.mean}\% to \res{beta.tab3.hybrid.tp-from-ifa.mean}\%
of their rain hits. The half-input and reversed-input variants feed $q_0+a(q_X-q_0)$ in place of every scenario $q_X$,
with $a=0.5$ and $a=-1$, respectively, and the reversed input is a non-physical stress test with negative rain.
Half-input, reversed-input and untrained configurations reach balanced accuracies from
\res{beta.tab3.gnn-untrained.ba} to \res{beta.tab3.hybrid-half.ba.mean}, close to or below the constant classifiers'
\res{beta.tab3.always.ba}.

\par\textbf{Strict subset.}
On the \res{beta.tab3.strict.n} cells where no trained configuration (four models, three seeds) is initially warned, a
diagnostic subset that removes inherited warnings (in which 90.2\% of cells cross under added rain and 4.8\% under added outlet stage), balanced accuracy is \res{beta.tab3.strict.gnn.ba.mean} for the
Scenario GNN, \res{beta.tab3.strict.hybrid.ba.mean} for the Hybrid forecaster, \res{beta.tab3.strict.physres.ba.mean}
for the Physics + state residual model and \res{beta.tab3.strict.physonly.ba.mean} for the Physics-only model, against
\res{beta.tab3.strict.linear.ba} for the linear response. The models keep the same order as on the full denominator, so
the rain ranking does not come from inherited warnings.

\par\textbf{Conditional errors.}
Median absolute $A^\ast$ errors are conditional on the cells where both the simulator and the model cross without an
initial false alarm, so they come with their coverage (Table~\ref{tab:cf-threshold}). On the \res{beta.tab3.common-both.n}
cells where the simulator and all twelve trained configurations cross in this way, the errors are
\res{beta.tab3.common-both.gnn.err-mm.mean}~mm (Scenario GNN), \res{beta.tab3.common-both.hybrid.err-mm.mean}~mm
(Hybrid forecaster), \res{beta.tab3.common-both.physres.err-mm.mean}~mm (Physics + state residual) and
\res{beta.tab3.common-both.physonly.err-mm.mean}~mm (Physics-only model). The Scenario GNN thus has the lowest
conditional error and the lowest balanced accuracy, as it predicts crossings in most cells.

\par\textbf{Outlet stage.}
For the added outlet stage \res{beta.s10.stage.pos-rate}\% of the cells cross. The Scenario GNN reaches a
balanced accuracy of \res{beta.s10.stage.gnn.ba.mean}, but \res{beta.s10.stage.gnn.tp-from-ifa.mean}\% of its hits are
warnings already present at the base scenario, against \res{beta.s10.stage.hybrid.tp-from-ifa.mean}\%,
\res{beta.s10.stage.physres.tp-from-ifa.mean}\% and \res{beta.s10.stage.physonly.tp-from-ifa.mean}\% for the other
three models. On the strict subset its stage balanced accuracy falls to \res{beta.s10.stage.strict.gnn.ba.mean}, and the
other models score \res{beta.s10.stage.strict.hybrid.ba.mean}. Cells where both the simulator and a model cross number
from \res{beta.s10.stage.physonly.n-both.mean} to \res{beta.s10.stage.gnn.n-both.mean} per model on average, and stage
threshold errors (in cm) are reported with this coverage. The apparent stage skill
is inherited from base warnings, in line with a zero outlet-stage response ratio.

\par\textbf{Raw agreement.}
Raw agreement of the same decision, the primary simulator metric with model base warnings counted as crossings, ranges
from \res{beta.hist.agree.physonly.mean} to \res{beta.hist.agree.gnn.mean} for the trained models. Always predicting a
crossing scores \res{beta.tab3.pos-rate}\% on the same cells, so this metric rewards
predicting crossings and ranks the Scenario GNN first.

\subsection{Censoring and robustness to thresholds and grid}\label{app:K4}

\begin{figure*}[tp]
\centering
\includegraphics[width=\textwidth]{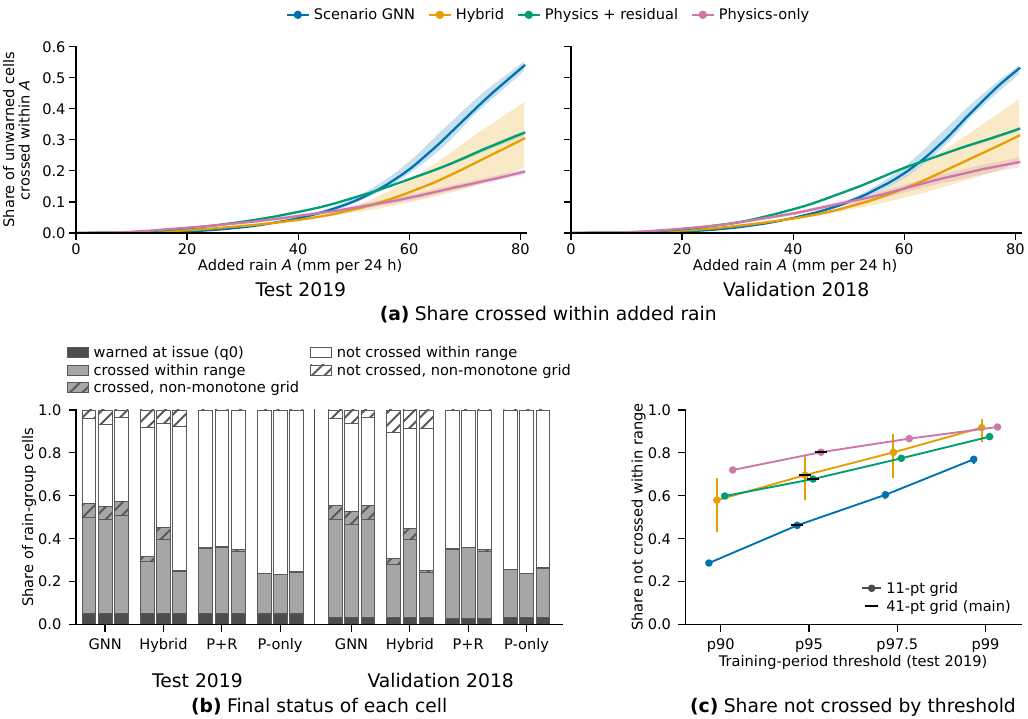}
\caption{Minimum crossing rainfall $A^\ast$ for unwarned Miami-Dade rain-group cells: (a)~cumulative crossing share (seed range shaded), (b)~cell status breakdown, and (c)~non-crossing share by threshold.}
\label{fig:s-censoring}
\end{figure*}

\par\textbf{Censoring.}
Cells not crossed within range are censored at $A_{\max}$. They are neither missing nor equal to $A_{\max}$.
Fig.~\ref{fig:s-censoring}a keeps them in the denominator under the first-observed convention and shows empirical
cumulative crossing shares within added rain $A$ on the deterministic grid, without a survival model and without
imputing non-crossers. Over all unwarned cells the curves end at the not-crossed shares of this convention, from
\res{mia.thr.range.notcrossed_all.min}\% to \res{mia.thr.range.notcrossed_all.max}\% per run in the test year
(App.~\ref{app:K1}).

\par\textbf{Thresholds and grid.}
Recomputed from the saved 11-point rain axis for training-period thresholds from p90 to p99, the share not crossed
within range rises with the threshold, from between \res{mia.robust.p90.notcrossed.min}\% and
\res{mia.robust.p90.notcrossed.max}\% at p90 to between \res{mia.robust.p99.notcrossed.min}\% and
\res{mia.robust.p99.notcrossed.max}\% at p99 (Fig.~\ref{fig:s-censoring}c). The grid step barely matters
(Algorithm~\ref{alg:s-astar}). In every run the 11-point and 41-point grids agree on almost every crossed or not-crossed
call (the lowest agreement is just below 100\% and prints as \res{mia.robust.grid.agree.min}\% at one decimal), with a
median $|\Delta A^\ast|$ of at most \res{mia.robust.grid.medabsdiff.max}~mm.

\subsection{Protocols of U1, U2 and U3}\label{app:K5}

\begin{figure*}[tp]
\centering
\includegraphics[width=\textwidth]{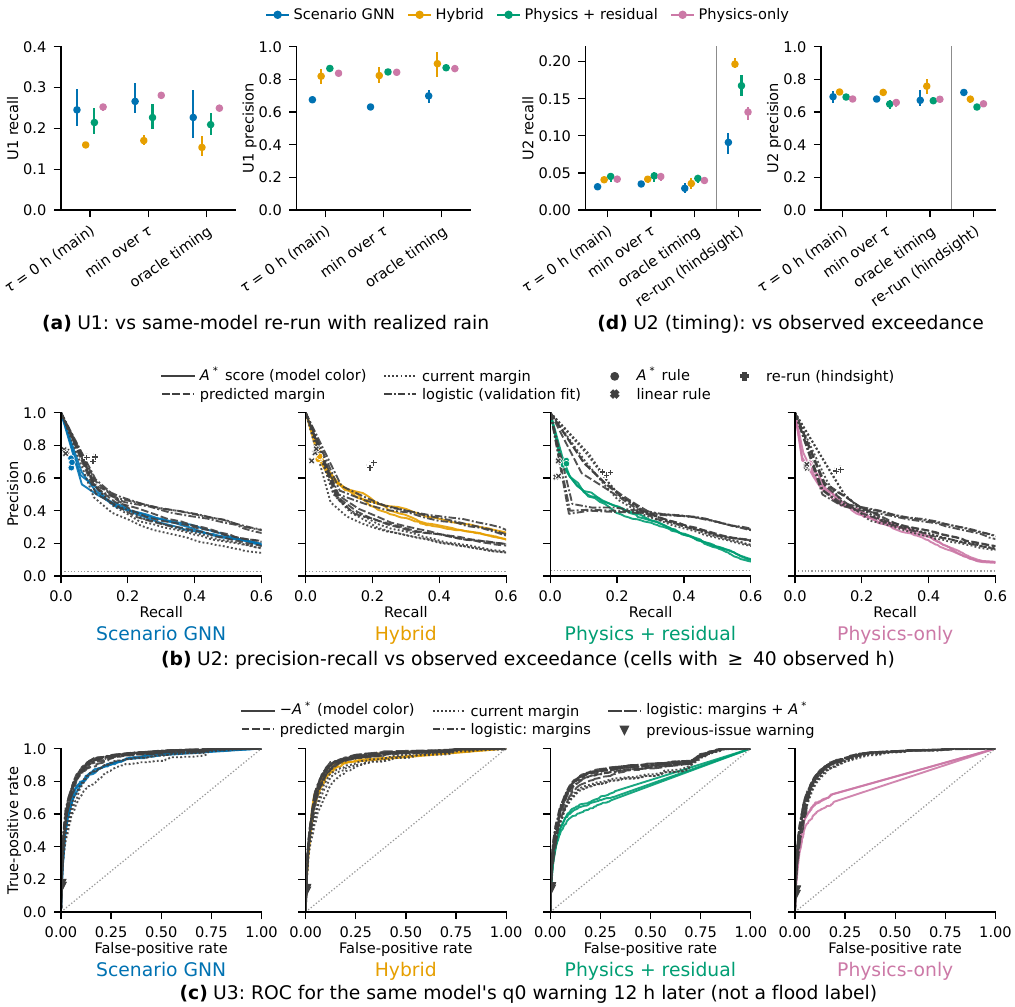}
\caption{Offline evaluation of $A^\ast$ in Miami-Dade (2019): (a,d)~performance by storm timing (seed mean and range); (b,c)~PR and ROC curves per seed.}
\label{fig:s-usage}
\end{figure*}

\begin{table*}[t]
\caption{Operating points for tasks U1 and U2 (Scenario GNN and Hybrid, Miami-Dade 2019, mean\,$\pm$\,SD over three seeds; $^{h}$realized rain; rate-m.: matched to $A^\ast$ validation alarm rate).}
\label{tab:s-usage}
\begin{tabular*}{\textwidth}{@{\extracolsep{\fill}}lllcccccc@{}}
\toprule
 & & & \multicolumn{3}{c}{Scenario GNN} & \multicolumn{3}{c}{Hybrid} \\
\cmidrule(lr){4-6}\cmidrule(lr){7-9}
 & Method & Cutoff & Recall & Precision & Alarms & Recall & Precision & Alarms \\
\midrule
\multirow{9}{*}{\rotatebox[origin=c]{90}{U1}} & $A^\ast$ rule$^{h}$ & binary & 0.245\,$\pm$\,0.044 & 0.675\,$\pm$\,0.011 & 82\,$\pm$\,12 & 0.159\,$\pm$\,0.004 & 0.819\,$\pm$\,0.040 & 98\,$\pm$\,6 \\
\cmidrule(l){2-9}
 & Linear rain rule$^{h}$ & binary & 0.107\,$\pm$\,0.020 & 0.772\,$\pm$\,0.034 & 32\,$\pm$\,9 & 0.104\,$\pm$\,0.022 & 0.873\,$\pm$\,0.032 & 60\,$\pm$\,15 \\
\cmidrule(l){2-9}
 & \multirow{2}{*}{Logistic$^{h}$} & rate-m. & 0.366\,$\pm$\,0.051 & 0.324\,$\pm$\,0.032 & 255\,$\pm$\,18 & 0.097\,$\pm$\,0.038 & 0.676\,$\pm$\,0.050 & 73\,$\pm$\,34 \\
 &  & max-F1 & 0.345\,$\pm$\,0.046 & 0.339\,$\pm$\,0.023 & 230\,$\pm$\,24 & 0.674\,$\pm$\,0.038 & 0.278\,$\pm$\,0.012 & 1{,}223\,$\pm$\,107 \\
\cmidrule(l){2-9}
 & \multirow{3}{*}{Current margin} & $\geq$\,0\,cm & 0.017\,$\pm$\,0.002 & 0.208\,$\pm$\,0.096 & 22\,$\pm$\,8 & 0.009\,$\pm$\,0.003 & 0.094\,$\pm$\,0.029 & 46\,$\pm$\,2 \\
 &  & rate-m. & 0.212\,$\pm$\,0.022 & 0.129\,$\pm$\,0.030 & 380\,$\pm$\,18 & 0.080\,$\pm$\,0.017 & 0.162\,$\pm$\,0.006 & 251\,$\pm$\,59 \\
 &  & max-F1 & 0.325\,$\pm$\,0.044 & 0.103\,$\pm$\,0.028 & 743\,$\pm$\,152 & 0.290\,$\pm$\,0.076 & 0.143\,$\pm$\,0.006 & 1{,}033\,$\pm$\,295 \\
\cmidrule(l){2-9}
 & \multirow{2}{*}{Predicted margin} & rate-m. & 0.255\,$\pm$\,0.040 & 0.145\,$\pm$\,0.036 & 405\,$\pm$\,14 & 0.118\,$\pm$\,0.028 & 0.288\,$\pm$\,0.046 & 210\,$\pm$\,56 \\
 &  & max-F1 & 0.300\,$\pm$\,0.022 & 0.141\,$\pm$\,0.043 & 505\,$\pm$\,112 & 0.337\,$\pm$\,0.031 & 0.171\,$\pm$\,0.008 & 1{,}000\,$\pm$\,129 \\
\midrule
\multirow{10}{*}{\rotatebox[origin=c]{90}{U2}} & $A^\ast$ rule$^{h}$ & binary & 0.031\,$\pm$\,0.002 & 0.693\,$\pm$\,0.029 & 82\,$\pm$\,12 & 0.041\,$\pm$\,0.003 & 0.722\,$\pm$\,0.010 & 98\,$\pm$\,6 \\
\cmidrule(l){2-9}
 & Linear rain rule$^{h}$ & binary & 0.014\,$\pm$\,0.004 & 0.762\,$\pm$\,0.011 & 32\,$\pm$\,9 & 0.026\,$\pm$\,0.007 & 0.745\,$\pm$\,0.037 & 60\,$\pm$\,15 \\
\cmidrule(l){2-9}
 & \multirow{2}{*}{Logistic$^{h}$} & rate-m. & 0.077\,$\pm$\,0.004 & 0.581\,$\pm$\,0.036 & 240\,$\pm$\,13 & 0.027\,$\pm$\,0.012 & 0.653\,$\pm$\,0.027 & 73\,$\pm$\,32 \\
 &  & max-F1 & 0.483\,$\pm$\,0.018 & 0.340\,$\pm$\,0.018 & 2{,}563\,$\pm$\,69 & 0.440\,$\pm$\,0.015 & 0.340\,$\pm$\,0.018 & 2{,}266\,$\pm$\,197 \\
\cmidrule(l){2-9}
 & \multirow{3}{*}{Current margin} & $\geq$\,0\,cm & 0.010\,$\pm$\,0.004 & 0.833\,$\pm$\,0.119 & 22\,$\pm$\,8 & 0.021\,$\pm$\,0.002 & 0.795\,$\pm$\,0.054 & 46\,$\pm$\,2 \\
 &  & rate-m. & 0.106\,$\pm$\,0.004 & 0.504\,$\pm$\,0.029 & 378\,$\pm$\,17 & 0.073\,$\pm$\,0.014 & 0.517\,$\pm$\,0.029 & 248\,$\pm$\,60 \\
 &  & max-F1 & 0.284\,$\pm$\,0.042 & 0.304\,$\pm$\,0.007 & 1{,}701\,$\pm$\,397 & 0.242\,$\pm$\,0.021 & 0.286\,$\pm$\,0.027 & 1{,}493\,$\pm$\,264 \\
\cmidrule(l){2-9}
 & \multirow{2}{*}{Predicted margin} & rate-m. & 0.115\,$\pm$\,0.002 & 0.514\,$\pm$\,0.032 & 403\,$\pm$\,15 & 0.066\,$\pm$\,0.013 & 0.558\,$\pm$\,0.046 & 208\,$\pm$\,56 \\
 &  & max-F1 & 0.320\,$\pm$\,0.020 & 0.333\,$\pm$\,0.012 & 1{,}733\,$\pm$\,178 & 0.289\,$\pm$\,0.014 & 0.298\,$\pm$\,0.018 & 1{,}699\,$\pm$\,192 \\
\cmidrule(l){2-9}
 & Re-run, realized rain$^{h}$ & reference & 0.091\,$\pm$\,0.013 & 0.719\,$\pm$\,0.016 & 228\,$\pm$\,39 & 0.196\,$\pm$\,0.006 & 0.678\,$\pm$\,0.015 & 504\,$\pm$\,8 \\
\bottomrule
\end{tabular*}
\end{table*}

\begin{table*}[t]
\caption{Operating points of U1 and U2 for the physics models, Miami-Dade test year 2019, with the rows and notation of Table~\ref{tab:s-usage}.}
\label{tab:s-usage-phys}
\begin{tabular*}{\textwidth}{@{\extracolsep{\fill}}lllcccccc@{}}
\toprule
 & & & \multicolumn{3}{c}{Physics + residual} & \multicolumn{3}{c}{Physics-only} \\
\cmidrule(lr){4-6}\cmidrule(lr){7-9}
 & Method & Cutoff & Recall & Precision & Alarms & Recall & Precision & Alarms \\
\midrule
\multirow{9}{*}{\rotatebox[origin=c]{90}{U1}} & $A^\ast$ rule$^{h}$ & binary & 0.214\,$\pm$\,0.030 & 0.866\,$\pm$\,0.010 & 169\,$\pm$\,16 & 0.252\,$\pm$\,0.006 & 0.837\,$\pm$\,0.007 & 147\,$\pm$\,10 \\
\cmidrule(l){2-9}
 & Linear rain rule$^{h}$ & binary & 0.117\,$\pm$\,0.031 & 0.909\,$\pm$\,0.010 & 87\,$\pm$\,16 & 0.217\,$\pm$\,0.005 & 0.839\,$\pm$\,0.010 & 126\,$\pm$\,8 \\
\cmidrule(l){2-9}
 & \multirow{2}{*}{Logistic$^{h}$} & rate-m. & 0.118\,$\pm$\,0.007 & 0.533\,$\pm$\,0.051 & 153\,$\pm$\,21 & 0.098\,$\pm$\,0.021 & 0.431\,$\pm$\,0.035 & 110\,$\pm$\,15 \\
 &  & max-F1 & 0.535\,$\pm$\,0.132 & 0.288\,$\pm$\,0.038 & 1{,}341\,$\pm$\,540 & 0.569\,$\pm$\,0.018 & 0.294\,$\pm$\,0.016 & 943\,$\pm$\,49 \\
\cmidrule(l){2-9}
 & \multirow{3}{*}{Current margin} & $\geq$\,0\,cm & 0.009\,$\pm$\,0.002 & 0.053\,$\pm$\,0.017 & 116\,$\pm$\,18 & 0.014\,$\pm$\,0.003 & 0.136\,$\pm$\,0.020 & 50\,$\pm$\,16 \\
 &  & rate-m. & 0.057\,$\pm$\,0.004 & 0.135\,$\pm$\,0.005 & 290\,$\pm$\,16 & 0.149\,$\pm$\,0.008 & 0.214\,$\pm$\,0.010 & 339\,$\pm$\,27 \\
 &  & max-F1 & 0.251\,$\pm$\,0.016 & 0.123\,$\pm$\,0.001 & 1{,}407\,$\pm$\,189 & 0.401\,$\pm$\,0.017 & 0.109\,$\pm$\,0.002 & 1{,}782\,$\pm$\,72 \\
\cmidrule(l){2-9}
 & \multirow{2}{*}{Predicted margin} & rate-m. & 0.096\,$\pm$\,0.015 & 0.222\,$\pm$\,0.013 & 292\,$\pm$\,13 & 0.130\,$\pm$\,0.004 & 0.233\,$\pm$\,0.010 & 272\,$\pm$\,9 \\
 &  & max-F1 & 0.301\,$\pm$\,0.016 & 0.155\,$\pm$\,0.006 & 1{,}324\,$\pm$\,39 & 0.380\,$\pm$\,0.009 & 0.152\,$\pm$\,0.002 & 1{,}215\,$\pm$\,65 \\
\midrule
\multirow{10}{*}{\rotatebox[origin=c]{90}{U2}} & $A^\ast$ rule$^{h}$ & binary & 0.045\,$\pm$\,0.005 & 0.691\,$\pm$\,0.014 & 169\,$\pm$\,16 & 0.042\,$\pm$\,0.003 & 0.679\,$\pm$\,0.012 & 147\,$\pm$\,10 \\
\cmidrule(l){2-9}
 & Linear rain rule$^{h}$ & binary & 0.022\,$\pm$\,0.004 & 0.641\,$\pm$\,0.055 & 87\,$\pm$\,16 & 0.035\,$\pm$\,0.001 & 0.672\,$\pm$\,0.012 & 126\,$\pm$\,8 \\
\cmidrule(l){2-9}
 & \multirow{2}{*}{Logistic$^{h}$} & rate-m. & 0.032\,$\pm$\,0.002 & 0.375\,$\pm$\,0.035 & 220\,$\pm$\,28 & 0.021\,$\pm$\,0.001 & 0.602\,$\pm$\,0.022 & 85\,$\pm$\,5 \\
 &  & max-F1 & 0.457\,$\pm$\,0.005 & 0.349\,$\pm$\,0.004 & 3{,}376\,$\pm$\,35 & 0.369\,$\pm$\,0.006 & 0.346\,$\pm$\,0.004 & 2{,}560\,$\pm$\,103 \\
\cmidrule(l){2-9}
 & \multirow{3}{*}{Current margin} & $\geq$\,0\,cm & 0.040\,$\pm$\,0.006 & 0.892\,$\pm$\,0.014 & 116\,$\pm$\,18 & 0.019\,$\pm$\,0.006 & 0.914\,$\pm$\,0.033 & 50\,$\pm$\,16 \\
 &  & rate-m. & 0.087\,$\pm$\,0.005 & 0.781\,$\pm$\,0.012 & 289\,$\pm$\,16 & 0.094\,$\pm$\,0.007 & 0.662\,$\pm$\,0.020 & 339\,$\pm$\,27 \\
 &  & max-F1 & 0.291\,$\pm$\,0.007 & 0.376\,$\pm$\,0.006 & 2{,}004\,$\pm$\,49 & 0.295\,$\pm$\,0.009 & 0.307\,$\pm$\,0.008 & 2{,}304\,$\pm$\,89 \\
\cmidrule(l){2-9}
 & \multirow{2}{*}{Predicted margin} & rate-m. & 0.079\,$\pm$\,0.001 & 0.700\,$\pm$\,0.044 & 291\,$\pm$\,13 & 0.068\,$\pm$\,0.002 & 0.607\,$\pm$\,0.006 & 270\,$\pm$\,9 \\
 &  & max-F1 & 0.306\,$\pm$\,0.003 & 0.372\,$\pm$\,0.013 & 2{,}122\,$\pm$\,63 & 0.308\,$\pm$\,0.020 & 0.311\,$\pm$\,0.010 & 2{,}382\,$\pm$\,250 \\
\cmidrule(l){2-9}
 & Re-run, realized rain$^{h}$ & reference & 0.167\,$\pm$\,0.012 & 0.630\,$\pm$\,0.011 & 686\,$\pm$\,62 & 0.132\,$\pm$\,0.008 & 0.650\,$\pm$\,0.005 & 486\,$\pm$\,21 \\
\bottomrule
\end{tabular*}
\end{table*}

\begin{table*}[t]
\caption{Threshold-free AUC comparisons in Miami-Dade (2019, mean\,$\pm$\,SD and 95\% block-bootstrap envelope over three seeds; $^{h}$hindsight).}
\label{tab:s-usage-ranking}
\resizebox{\textwidth}{!}{%
\begin{tabular}{llcccccccc}
\toprule
 &  & \multicolumn{2}{c}{Scenario GNN} & \multicolumn{2}{c}{Hybrid} & \multicolumn{2}{c}{Physics + residual} & \multicolumn{2}{c}{Physics-only} \\
\cmidrule(lr){3-4}\cmidrule(lr){5-6}\cmidrule(lr){7-8}\cmidrule(lr){9-10}
 & Score & PR-AUC & ROC-AUC & PR-AUC & ROC-AUC & PR-AUC & ROC-AUC & PR-AUC & ROC-AUC \\
\midrule
\multirow{10}{*}{\rotatebox[origin=c]{90}{U1}} & $A^\ast$ score$^{h}$ & 0.461\,$\pm$\,0.036 & 0.992\,$\pm$\,0.001 & 0.493\,$\pm$\,0.016 & 0.974\,$\pm$\,0.012 & 0.582\,$\pm$\,0.022 & 0.982\,$\pm$\,0.004 & 0.618\,$\pm$\,0.005 & 0.991\,$\pm$\,0.000 \\
 & Linear-rule score$^{h}$ & 0.330\,$\pm$\,0.033 & 0.867\,$\pm$\,0.041 & 0.380\,$\pm$\,0.030 & 0.887\,$\pm$\,0.014 & 0.438\,$\pm$\,0.031 & 0.879\,$\pm$\,0.014 & 0.560\,$\pm$\,0.008 & 0.955\,$\pm$\,0.004 \\
 & Current margin & 0.091\,$\pm$\,0.023 & 0.938\,$\pm$\,0.003 & 0.111\,$\pm$\,0.003 & 0.941\,$\pm$\,0.007 & 0.095\,$\pm$\,0.003 & 0.913\,$\pm$\,0.007 & 0.114\,$\pm$\,0.002 & 0.946\,$\pm$\,0.007 \\
 & Predicted margin & 0.117\,$\pm$\,0.027 & 0.950\,$\pm$\,0.007 & 0.165\,$\pm$\,0.009 & 0.966\,$\pm$\,0.002 & 0.127\,$\pm$\,0.009 & 0.939\,$\pm$\,0.007 & 0.143\,$\pm$\,0.002 & 0.961\,$\pm$\,0.007 \\
 & Logistic$^{h}$ (margins + total) & 0.300\,$\pm$\,0.012 & 0.972\,$\pm$\,0.002 & 0.390\,$\pm$\,0.001 & 0.986\,$\pm$\,0.000 & 0.319\,$\pm$\,0.017 & 0.979\,$\pm$\,0.003 & 0.304\,$\pm$\,0.019 & 0.987\,$\pm$\,0.001 \\
\cmidrule(l){2-10}
 & Prevalence & \multicolumn{2}{c}{0.0034\,$\pm$\,0.0005} & \multicolumn{2}{c}{0.0073\,$\pm$\,0.0001} & \multicolumn{2}{c}{0.0089\,$\pm$\,0.0008} & \multicolumn{2}{c}{0.0063\,$\pm$\,0.0003} \\
\cmidrule(l){2-10}
 & & \emph{mean}\,$\pm$\,\emph{SD} & \emph{interval} & \emph{mean}\,$\pm$\,\emph{SD} & \emph{interval} & \emph{mean}\,$\pm$\,\emph{SD} & \emph{interval} & \emph{mean}\,$\pm$\,\emph{SD} & \emph{interval} \\
 & $\Delta$PR ($A^\ast$ score $-$ linear) & 0.132\,$\pm$\,0.069 & [0.012, 0.271] & 0.114\,$\pm$\,0.031 & [0.046, 0.217] & 0.144\,$\pm$\,0.023 & [0.073, 0.214] & 0.058\,$\pm$\,0.008 & [0.021, 0.105] \\
 & $\Delta$PR ($A^\ast$ score $-$ logistic) & 0.161\,$\pm$\,0.033 & [0.041, 0.275] & 0.103\,$\pm$\,0.015 & [$-$0.047, 0.187] & 0.263\,$\pm$\,0.007 & [0.188, 0.315] & 0.314\,$\pm$\,0.016 & [0.223, 0.393] \\
 & $\Delta$PR (predicted $-$ current) & 0.026\,$\pm$\,0.009 & [$-$0.012, 0.065] & 0.054\,$\pm$\,0.010 & [0.020, 0.102] & 0.033\,$\pm$\,0.006 & [0.017, 0.061] & 0.029\,$\pm$\,0.003 & [0.008, 0.053] \\
\midrule
\multirow{8}{*}{\rotatebox[origin=c]{90}{U2}} & $A^\ast$ score$^{h}$ & 0.269\,$\pm$\,0.006 & 0.852\,$\pm$\,0.003 & 0.305\,$\pm$\,0.018 & 0.885\,$\pm$\,0.022 & 0.229\,$\pm$\,0.006 & 0.732\,$\pm$\,0.005 & 0.207\,$\pm$\,0.004 & 0.718\,$\pm$\,0.003 \\
 & Linear-rule score$^{h}$ & 0.169\,$\pm$\,0.020 & 0.668\,$\pm$\,0.026 & 0.200\,$\pm$\,0.001 & 0.730\,$\pm$\,0.001 & 0.165\,$\pm$\,0.011 & 0.652\,$\pm$\,0.010 & 0.165\,$\pm$\,0.006 & 0.669\,$\pm$\,0.005 \\
 & Current margin & 0.252\,$\pm$\,0.020 & 0.884\,$\pm$\,0.002 & 0.226\,$\pm$\,0.007 & 0.875\,$\pm$\,0.003 & 0.318\,$\pm$\,0.009 & 0.876\,$\pm$\,0.001 & 0.274\,$\pm$\,0.014 & 0.869\,$\pm$\,0.002 \\
 & Predicted margin & 0.277\,$\pm$\,0.015 & 0.906\,$\pm$\,0.002 & 0.253\,$\pm$\,0.005 & 0.905\,$\pm$\,0.002 & 0.319\,$\pm$\,0.010 & 0.896\,$\pm$\,0.001 & 0.265\,$\pm$\,0.010 & 0.884\,$\pm$\,0.002 \\
 & Logistic$^{h}$ (margins + total) & 0.332\,$\pm$\,0.010 & 0.929\,$\pm$\,0.003 & 0.318\,$\pm$\,0.013 & 0.931\,$\pm$\,0.003 & 0.293\,$\pm$\,0.004 & 0.919\,$\pm$\,0.000 & 0.289\,$\pm$\,0.008 & 0.907\,$\pm$\,0.001 \\
\cmidrule(l){2-10}
 & Prevalence & \multicolumn{2}{c}{0.027\,$\pm$\,0.002} & \multicolumn{2}{c}{0.026\,$\pm$\,0.000} & \multicolumn{2}{c}{0.034\,$\pm$\,0.000} & \multicolumn{2}{c}{0.032\,$\pm$\,0.001} \\
\cmidrule(l){2-10}
 & & \emph{mean}\,$\pm$\,\emph{SD} & \emph{interval} & \emph{mean}\,$\pm$\,\emph{SD} & \emph{interval} & \emph{mean}\,$\pm$\,\emph{SD} & \emph{interval} & \emph{mean}\,$\pm$\,\emph{SD} & \emph{interval} \\
 & $\Delta$PR: $A^\ast$ score $-$ current & 0.017\,$\pm$\,0.015 & [$-$0.045, 0.073] & 0.079\,$\pm$\,0.012 & [0.001, 0.149] & $-$0.089\,$\pm$\,0.003 & [$-$0.152, $-$0.031] & $-$0.067\,$\pm$\,0.013 & [$-$0.129, 0.004] \\
\midrule
\multirow{11}{*}{\rotatebox[origin=c]{90}{U3}} & $-A^\ast$ & 0.130\,$\pm$\,0.033 & 0.916\,$\pm$\,0.003 & 0.126\,$\pm$\,0.012 & 0.924\,$\pm$\,0.013 & 0.103\,$\pm$\,0.009 & 0.773\,$\pm$\,0.014 & 0.111\,$\pm$\,0.026 & 0.800\,$\pm$\,0.017 \\
 & Predicted margin & 0.174\,$\pm$\,0.025 & 0.945\,$\pm$\,0.005 & 0.147\,$\pm$\,0.020 & 0.936\,$\pm$\,0.005 & 0.152\,$\pm$\,0.010 & 0.876\,$\pm$\,0.008 & 0.161\,$\pm$\,0.015 & 0.935\,$\pm$\,0.003 \\
 & Current margin & 0.132\,$\pm$\,0.027 & 0.904\,$\pm$\,0.016 & 0.108\,$\pm$\,0.022 & 0.905\,$\pm$\,0.010 & 0.127\,$\pm$\,0.006 & 0.833\,$\pm$\,0.005 & 0.150\,$\pm$\,0.008 & 0.924\,$\pm$\,0.002 \\
 & Previous warning ($t-6$\,h) & 0.040\,$\pm$\,0.001 & 0.580\,$\pm$\,0.005 & 0.031\,$\pm$\,0.004 & 0.568\,$\pm$\,0.005 & 0.034\,$\pm$\,0.003 & 0.573\,$\pm$\,0.004 & 0.024\,$\pm$\,0.004 & 0.560\,$\pm$\,0.007 \\
 & Persistence & 0.006\,$\pm$\,0.001 & 0.500\,$\pm$\,0.000 & 0.006\,$\pm$\,0.000 & 0.500\,$\pm$\,0.000 & 0.007\,$\pm$\,0.000 & 0.500\,$\pm$\,0.000 & 0.007\,$\pm$\,0.001 & 0.500\,$\pm$\,0.000 \\
 & Logistic (margins) & 0.150\,$\pm$\,0.005 & 0.945\,$\pm$\,0.004 & 0.113\,$\pm$\,0.007 & 0.938\,$\pm$\,0.004 & 0.129\,$\pm$\,0.013 & 0.878\,$\pm$\,0.009 & 0.144\,$\pm$\,0.020 & 0.935\,$\pm$\,0.003 \\
 & Logistic (margins + $A^\ast$) & 0.150\,$\pm$\,0.019 & 0.943\,$\pm$\,0.004 & 0.126\,$\pm$\,0.010 & 0.943\,$\pm$\,0.004 & 0.123\,$\pm$\,0.014 & 0.872\,$\pm$\,0.015 & 0.124\,$\pm$\,0.024 & 0.931\,$\pm$\,0.004 \\
\cmidrule(l){2-10}
 & Prevalence & \multicolumn{2}{c}{0.0061\,$\pm$\,0.0008} & \multicolumn{2}{c}{0.0058\,$\pm$\,0.0001} & \multicolumn{2}{c}{0.0071\,$\pm$\,0.0001} & \multicolumn{2}{c}{0.0071\,$\pm$\,0.0006} \\
\cmidrule(l){2-10}
 & & \emph{mean}\,$\pm$\,\emph{SD} & \emph{interval} & \emph{mean}\,$\pm$\,\emph{SD} & \emph{interval} & \emph{mean}\,$\pm$\,\emph{SD} & \emph{interval} & \emph{mean}\,$\pm$\,\emph{SD} & \emph{interval} \\
 & $\Delta$ROC: $-A^\ast$ $-$ predicted & $-$0.029\,$\pm$\,0.003 & [$-$0.045, $-$0.008] & $-$0.012\,$\pm$\,0.009 & [$-$0.039, 0.004] & $-$0.103\,$\pm$\,0.006 & [$-$0.146, $-$0.066] & $-$0.135\,$\pm$\,0.014 & [$-$0.196, $-$0.094] \\
 & $\Delta$ROC: adding $A^\ast$ & $-$0.003\,$\pm$\,0.000 & [$-$0.009, 0.003] & 0.005\,$\pm$\,0.002 & [$-$0.004, 0.014] & $-$0.006\,$\pm$\,0.007 & [$-$0.022, 0.002] & $-$0.004\,$\pm$\,0.001 & [$-$0.010, 0.001] \\
\bottomrule
\end{tabular}}
\end{table*}

\par\textbf{Common start set and references.}
All three tasks start from the monotone-grid sample of rain-group cells not warned under $q_0$ in the test year 2019, and
each has its own reference outcome and denominator. \emph{U1} takes as positives the cells
whose margin reaches zero when the same model is re-run with the realized rain field, gates, pumps and surge held. This
gives between \res{mia.u1.positives.min} and \res{mia.u1.positives.max} positives per run out of between
\res{mia.u1.cells.min} and \res{mia.u1.cells.max} cells. \emph{U2} takes as positives the observed exceedances, an
observed 48-h peak at or above $u_i$, in cells with at least 40 observed hours. This gives between
\res{mia.u2.events.min} and \res{mia.u2.events.max} positives per run out of between \res{mia.u2.cells.min} and
\res{mia.u2.cells.max} cells, a prevalence from \res{mia.u2.prevalence.min}\% to \res{mia.u2.prevalence.max}\%.
\emph{U3} takes as positives the same model's $q_0$ warning at the issue time $\Delta t_{\mathrm{U3}}=12$~h later, a
future model output, not a flood. This gives between \res{mia.u3.positives.min} and \res{mia.u3.positives.max}
positives per run out of between \res{mia.u3.cells.min} and \res{mia.u3.cells.max} cells (from
\res{mia.u3.prevalence.min}\% to \res{mia.u3.prevalence.max}\%). U3 uses only the issue-time $A^\ast$
(Table~\ref{tab:s-usage-ranking}).

\par\textbf{Rules and information.}
The $A^\ast$ rule flags a cell when the realized area-mean 48-h rain in excess of the forecast reaches $A^\ast$, so it
uses a hindsight scalar. The linear rain rule flags a cell when the $q_0$ margin plus the first joint-grid step's slope
times the same excess rain reaches $\varepsilon$. It involves no fitting and uses the same hindsight scalar. The margin
baselines use issue-time information only, namely the current water-level margin $m^{\mathrm{cur}}_i$, the last
observed level minus $u_i$, and the predicted-peak margin $m_i(q_0)$. The fixed rule ``currently at or above the
threshold'' flags $m^{\mathrm{cur}}_i\ge0$. Logistic baselines are fitted on the validation year of the same model and
seed. The $A^\ast$ rule and the linear rain rule are binary and have no cutoff. The cutoffs of the score baselines are
fixed on the validation year. There the rate-matched baselines (rate-m.\ in Tables~\ref{tab:s-usage} and~\ref{tab:s-usage-phys}) receive the
$A^\ast$ rule's validation alarm share (from \res{mia.u2.astar.valalarmshare.min}\% to
\res{mia.u2.astar.valalarmshare.max}\% of cells), and the max-F1 baselines take the cutoff of maximal F1. The $A^\ast$
rule then alarms on only between \res{mia.u2.astar.testalarmshare.min}\% and \res{mia.u2.astar.testalarmshare.max}\% of
test cells, so the rate-matched current and predicted margins raise between
\res{mia.u2.alarm-ratio.margins-vs-astar.min} and \res{mia.u2.alarm-ratio.margins-vs-astar.max} times as many test
alarms as the $A^\ast$ rule in the same run (24 per-run ratios). For the comparisons without thresholds, the $A^\ast$
rule of U1 and U2 becomes a score, the realized excess rain minus $A^\ast$, in which not-crossed cells rank lowest.
Ranking quality is measured by PR-AUC, computed as average precision, and by ROC-AUC. The intervals of
Table~\ref{tab:s-usage-ranking} are 95\% block-bootstrap intervals from 500 resamples of 7-day blocks per seed, and the table
gives their envelope over seeds.

\par\textbf{U1.}
The $A^\ast$ rule recovers between \res{mia.u1.astar.recall-pct.min}\% and \res{mia.u1.astar.recall-pct.max}\% of the
U1 positives with precision from \res{mia.u1.astar.precision.min} to \res{mia.u1.astar.precision.max}. Baselines for
U1 are scored against the same re-run, whose saved per-cell margins reproduce every U1 count of
the $A^\ast$ rule exactly. U1 positives arise from
the realized rain, which the margins do not see, so rules are compared only within an information set. For U1 the
rate-matched cutoffs use the $A^\ast$ rule's alarm share on the validation U1 cells (from
\res{mia.u1.astar.valalarmshare.min}\% to \res{mia.u1.astar.valalarmshare.max}\%), and the max-F1 cutoffs and the
logistic models use the validation U1 labels of the same model and seed.

Among the rules that use the realized rain total, the $A^\ast$ rule recovers more U1 positives than the linear rain
rule in \res{mia.u1.nruns.recall.astar-gt-linear} of 12 runs, the linear rule reaching recall from
\res{mia.u1.linear.recall.min} to \res{mia.u1.linear.recall.max}, while the linear rule is the more precise in
\res{mia.u1.nruns.precision.linear-gt-astar} of 12 runs (precision from \res{mia.u1.linear.precision.min} to
\res{mia.u1.linear.precision.max}). The $A^\ast$ score has a higher PR-AUC than the linear-rule score in
\res{mia.u1.nruns.dprauc.astar-vs-linear.positive} of 12 runs, with
\res{mia.u1.nruns.dprauc.astar-vs-linear.ci-above-zero} block-bootstrap intervals above zero, and a higher PR-AUC
than the validation-fitted logistic model in \res{mia.u1.nruns.dprauc.astar-vs-logistic.positive} of 12 runs, with
\res{mia.u1.nruns.dprauc.astar-vs-logistic.ci-above-zero} intervals above zero (Table~\ref{tab:s-usage-ranking}). Among the
issue-time rules, the rate-matched current margin recovers from \res{mia.u1.curmargin.recall.min} to
\res{mia.u1.curmargin.recall.max} of the U1 positives with precision from \res{mia.u1.curmargin.precision.min} to
\res{mia.u1.curmargin.precision.max}, and the rate-matched predicted-peak margin from
\res{mia.u1.predmargin.recall.min} to \res{mia.u1.predmargin.recall.max} with precision from
\res{mia.u1.predmargin.precision.min} to \res{mia.u1.predmargin.precision.max}. They raise between
\res{mia.u1.alarm-ratio.margins-vs-astar.min} and \res{mia.u1.alarm-ratio.margins-vs-astar.max} times as many test
alarms as the $A^\ast$ rule in the same run. The predicted-peak margin has a higher PR-AUC than the current margin in
\res{mia.u1.nruns.dprauc.pred-vs-cur.positive} of 12 runs, with \res{mia.u1.nruns.dprauc.pred-vs-cur.ci-above-zero}
intervals above zero. Positives make up only from \res{mia.u1.prevalence.min}\% to \res{mia.u1.prevalence.max}\% of
the U1 cells (from \res{mia.u1.positives.min} to \res{mia.u1.positives.max} per run), so operating points vary
between seeds (Tables~\ref{tab:s-usage} and~\ref{tab:s-usage-phys}).

With the U1 label raised to a re-run margin of at least 1~cm, the margin for which $A^\ast$ is defined, from
\res{mia.u1eps.positives.min} to \res{mia.u1eps.positives.max} cells per run are positive. The $A^\ast$ rule still
recovers more of them than the linear rain rule in \res{mia.u1eps.nruns.recall.astar-gt-linear} of 12 runs, and its
score keeps a higher PR-AUC than the linear-rule score in \res{mia.u1eps.nruns.dprauc.astar-vs-linear.positive} of 12
runs and than the logistic model in \res{mia.u1eps.nruns.dprauc.astar-vs-logistic.positive} of 12 runs.

\subsection{Operating points, threshold-free comparisons and incremental information}\label{app:K6}

\par\textbf{U2 operating points.}
Recalls in this paragraph are fractions of the observed exceedances. Against observed exceedances, the binary $A^\ast$
rule has recall from \res{mia.u2.astar.recall.min} to \res{mia.u2.astar.recall.max} (from
\res{mia.u2.astar.recall-pct.min}\% to \res{mia.u2.astar.recall-pct.max}\% in whole percent, shown in Fig.~\ref{fig:state-event}), with precision from
\res{mia.u2.astar.precision.min} to \res{mia.u2.astar.precision.max}. The rate-matched current margin reaches recall
from \res{mia.u2.curmargin.recall.min} to \res{mia.u2.curmargin.recall.max} (precision from
\res{mia.u2.curmargin.precision.min} to \res{mia.u2.curmargin.precision.max}) and the rate-matched predicted-peak
margin from \res{mia.u2.predmargin.recall.min} to \res{mia.u2.predmargin.recall.max} (precision from
\res{mia.u2.predmargin.precision.min} to \res{mia.u2.predmargin.precision.max}), with the larger alarm budgets stated
above. The linear rain rule reaches from \res{mia.u2.linear.recall.min} to \res{mia.u2.linear.recall.max}, below the
$A^\ast$ rule, and the fixed rule ``at or above the threshold'' from \res{mia.u2.curmargin_zero.recall.min} to
\res{mia.u2.curmargin_zero.recall.max} with precision from \res{mia.u2.curmargin_zero.precision.min} to
\res{mia.u2.curmargin_zero.precision.max}. The same model re-run with the realized rain field, a hindsight reference
rather than a usable rule, reaches recall from \res{mia.u2.rerun.recall.min} to \res{mia.u2.rerun.recall.max} with
precision from \res{mia.u2.rerun.precision.min} to \res{mia.u2.rerun.precision.max}. Its misses can come from model
error, filled inputs or unrecorded operations as well as from other physical drivers.

\par\textbf{U2 without thresholds.}
Ranking quality removes the dependence on alarm budgets (Table~\ref{tab:s-usage-ranking}, Fig.~\ref{fig:s-usage}b). Read
with their 7-day block intervals, the PR-AUC differences between the $A^\ast$ score and the current margin place the
$A^\ast$ score about level with the current margin for the Scenario GNN, slightly ahead of it for the Hybrid forecaster
(\res{mia.u2.dprauc_astar_vs_cur.hybrid} [\res{mia.u2.dprauc_astar_vs_cur.hybrid.lo},
\res{mia.u2.dprauc_astar_vs_cur.hybrid.hi}]), behind it for the Physics + state residual model and lower with an
interval including zero for the Physics-only model. The ROC-AUC of the $A^\ast$ score, from
\res{mia.u2.rocauc.astar.physonly} to \res{mia.u2.rocauc.astar.hybrid}, stays below that of the predicted-peak margin,
from \res{mia.u2.rocauc.predmargin.physonly} to \res{mia.u2.rocauc.predmargin.gnn}, for every model, although the
$A^\ast$ score uses the realized rain total and the margin only issue-time information.

\par\textbf{U3 and incremental information.}
For the same model's warning 12~h later (Fig.~\ref{fig:s-usage}c), the ROC-AUC of the $A^\ast$ score ranges from
\res{mia.u3.auc.astar.min} to \res{mia.u3.auc.astar.max}, but the predicted-peak margin alone reaches from
\res{mia.u3.auc.predmargin.min} to \res{mia.u3.auc.predmargin.max} and is higher in all twelve runs. Adding $A^\ast$ to
a validation-fitted logistic model of the margins changes the test ROC-AUC by \res{mia.u3.d.add_astar.min} to
\res{mia.u3.d.add_astar.max}, with every interval including zero, so in Miami-Dade $A^\ast$ adds nothing to U3 beyond
the model's own predicted margin. Persistence of the current warning has a ROC-AUC of 0.5 by construction, because
every cell in this denominator is unwarned at the issue time.

\subsection{Timing scans}\label{app:K7}

The main protocol starts the fixed storm shape at the issue time. A timing scan
also starts it 12~h and 24~h later and forms two summaries, the
smallest $A^\ast$ over the three starts, available at the issue time, and the start nearest the realized rain centroid
minus 12~h (oracle timing), which uses hindsight. Taking the smallest $A^\ast$ changes U1 recall by \res{mia.timing.u1.gainmin.min} to
\res{mia.timing.u1.gainmin.max} percentage points per run and the oracle timing by \res{mia.timing.u1.gainoracle.min}
to \res{mia.timing.u1.gainoracle.max}, and U2 recall changes by \res{mia.timing.u2.gainmin.min} to
\res{mia.timing.u2.gainmin.max} points (Fig.~\ref{fig:s-usage}a,d). Each variant removes its own non-monotone cells,
so the cell sets differ slightly. The U1 recall level therefore does not come from the timing of the fixed shape
alone.

\section{Observed Regions and Second Simulator}\label{app:L}

This appendix describes the six observed flood regions and nine test periods of Table~\ref{tab:xr} and the second
simulator\smbadd{, and App.~\ref{app:Lsmb} adds a synthetic tidal-barrage testbed with rule-controlled gates and pumps}. Every model was trained on its own region with the main-study recipe and
three seeds (42, 43 and 44), without transfer or fine-tuning, so the same structure name denotes different weights in
each region. Observed regions have no paired truth, so they test the cost of the scenario interface, the observational
check and threshold summaries, while the size, propagation and interaction of responses are graded on the simulators.

\subsection{Regions, data and graphs}\label{app:L1}

\par\textbf{Choice of regions and test periods.}
SF$^2$Bench~\cite{zheng2026uncovering} cuts South Florida into three spatial parts and eight multi-year splits. Part~1
is the managed coastal canal system of Miami-Dade, part~0 the upper Kissimmee chain of lakes and part~2 the inland
canals of Hendry and Collier counties. The protocol needs disjoint training, selection and test years, observed stages,
rain and operations at the issue time, astronomical tide and surge series, and rain depths in millimeters. Split S\_6
(training 2015 to 2017, selection 2018, test 2019) meets these needs in all three parts, and split S\_7 meets them with a
2022 hold-out. Split S\_4 was excluded because the benchmark's validation range overlaps its training years, S\_5
because it has no tide or surge series, and S\_0 to S\_3 because they also lack observed rain and wells in their test
years. The hold-out trains on 2020-01 to 2021-07, selects epochs on the rest of 2021, tests on 2022, which contains
hurricanes Ian and Nicole, and keeps 2023 as a second test year. Statistics, rain-unit factors and the rain forecast use
2020 and 2021 only. In the Kissimmee lakes the largest 24-h rain of 2022, during Ian, lies far above the
training maximum. Among the public datasets examined for new regions, coastal Louisiana is the only one in which rain
and surge can both be tested over 2015 to 2019, and England has rain gauges but no tide gauge for these years.

\par\textbf{The South Florida parts.}
The Kissimmee lakes graph holds \res{sf.s6p0.groups.n-canal} canal gauges, \res{sf.s6p0.groups.n-wells} wells,
\res{sf.s6p0.groups.n-rain-gauges} rain gauges and \res{sf.s6p0.groups.n-gates} gate series, and the Hendry-Collier
graph \res{sf.s6p2.groups.n-canal} canal gauges, \res{sf.s6p2.groups.n-wells} wells, \res{sf.s6p2.groups.n-rain-gauges}
rain gauges, \res{sf.s6p2.groups.n-gates} gate series and \res{sf.s6p2.groups.n-pumps} pump series. Both read the tide
gauges as inputs but have \res{sf.s6p0.groups.n-tide-edges} tide edges, because a canal gauge is coupled to the ocean
only within 30~km of a tide gauge. Their surge groups are empty, so surge is not testable there. In 2022 the graphs hold
\res{sf.s7p0d1.groups.n-canal} and \res{sf.s7p2d1.groups.n-canal} canal gauges, again without tide edges. The 2022
Miami-Dade graph has \res{s7.stations.water} canal stations, among which \res{s7.stations.shared} of the
\res{s7.stations.s6p1_water} stations of 2019 reappear with \res{s7.wells.shared} wells, so it is a
temporal check of the same system. Its 2022 surge maxima occur during Nicole and stay below the training maxima at two
tide gauges (for example \res{s7.surge.max.test.g2} against \res{s7.surge.max.train.g2}~cm), while at the third the 2022
maximum of \res{s7.surge.max.test.g1}~cm slightly exceeds the training \res{s7.surge.max.train.g1}~cm. The year has
\res{s7.surge.windows_gt20.test} windows whose 48-h mean surge change exceeds 20~cm, against
\res{s7.surge.windows_gt20.selection} on the selection segment, so it tests more surge-active windows, not surge far
beyond the training range.

\par\textbf{Houston, Louisiana and England.}
Houston uses the Harris County flood warning system~\cite{hcfcd2024fws}, with \res{hou.gauges.stage.selected} of
\res{hou.gauges.stage.candidates} stage gauges and \res{hou.gauges.rain.selected} of \res{hou.gauges.rain.candidates}
rain gauges that have at least 80\% measured hours in every year, and \res{hou.tide.gauges} NOAA tide
gauges~\cite{noaa2024coops} as ocean boundaries. The largest surge, \res{hou.surge.max_cm}~cm during Hurricane Harvey,
falls in the training period. The bayous have no gates, pumps or wells. Coastal Louisiana keeps
\res{crms.data.n_stations} of \res{crms.data.n_candidates} stations of the Coastwide Reference Monitoring
System~\cite{crms2024} with the same coverage rule, in \res{crms.data.n_basins} basins. These stations record no rain, so
hourly precipitation of the Analysis of Record for Calibration~\cite{noaa2024aorc,fall2023aorc}, a gridded analysis
rather than a gauge record, is averaged over the
\res{crms.data.n_rain_inputs} boxes of 0.1 degree that contain a station. Its hour-ending time stamp was checked against
radar rain. Tide and surge come from \res{crms.data.n_tide_gauges} NOAA gauges. The Aire and Calder catchments use
\res{uk.data.n_stations} of \res{uk.data.n_candidates} level gauges and \res{uk.data.n_rain_inputs} rain gauges of the
Environment Agency~\cite{ea2024hydrology}. No tide gauge covers these years, and stage is measured above each gauge's
local zero because no English level series has a published datum. The high-water threshold is the training p95
everywhere. The Houston storm-event comparison also uses the published Flooding Likely level at \res{hou.thr.fws} gauges.
All three regions use 168~h of history, a 48-h horizon, hourly issue times, training on 2015 to 2017, selection on 2018
and testing on 2019 in calendar years of universal time. Their ridge rain forecast has a median per-gauge $R^2$ in 2019 of
\res{hou.rainfc.median_r2.test} in Houston, \res{crms.rainfc.median_test_r2} in Louisiana and
\res{uk.rainfc.median_test_r2} in England.

\par\textbf{Physical river networks.}
The South Florida parts use the benchmark's structure graph. The other regions use physical river networks built from
hydrography, in which stations may lack edges. Houston and Louisiana use the National Hydrography Dataset Plus High
Resolution~\cite{usgs2024nhdplushr}, which carries flow direction and connectivity. A station within 500~m of a network flowline, or in
an open-water polygon on the network, is snapped, a station at the coastline becomes a shore boundary station, and any
other station keeps only its rain mix. A walk downstream from each snapped station gives a directed channel edge at the
first gauged station within 40~km, and a walk that reaches the coast gives one tide edge to a NOAA gauge.

In Houston \res{hou.phys.graph.n_snapped} of \res{hou.phys.graph.n_stations} gauges lie on a channel, giving
\res{hou.phys.graph.n_channel_edges} channel edges (median path \res{hou.phys.graph.channel_km_median}~km) and
\res{hou.phys.graph.n_tide_edges} tide edges, and \res{hou.phys.graph.n_no_edge} gauges have no edge. Smoothed elevation
drops along \res{hou.phys.graph.dir_nhd_agree} of the \res{hou.phys.graph.dir_stage_n} flow-path edges, and hourly level
changes correlate at a median of \res{hou.phys.graph.corr_edges_median} along edges against
\res{hou.phys.graph.corr_nonedge_median} for unjoined pairs within 15~km. Against the inferred Houston graph, built from
gauge names, mean levels and tide correlations, the physical graph keeps \res{hou.phys.graph.kept} of its
\res{hou.phys.graph.n_channel_edges_inferred} channel edges, adds \res{hou.phys.graph.added} and removes
\res{hou.phys.graph.removed}, including hand-listed confluences that joined a tributary to a main-channel gauge
upstream of the confluence. In Louisiana \res{crms.phys.graph.n_snapped} of \res{crms.phys.graph.n_stations} stations lie on a channel,
the graph has \res{crms.phys.graph.n_channel_edges} channel edges and \res{crms.phys.graph.n_tide_edges} tide edges, and
\res{crms.phys.graph.n_no_edge} stations have no edge. Joined stations co-move less than nearby unjoined pairs (median
correlation \res{crms.phys.graph.corr_edges_median} against \res{crms.phys.graph.corr_nonedge_median}), while tide-edge
stations follow their gauge's tide closely (median \res{crms.phys.graph.tidal_r_tide_edges_median}). The graph's content
therefore lies in its connection to the coast, since the mapped channels do not describe how marsh stations exchange
water. The inferred graph, a minimum spanning tree inside each basin, carries little beyond proximity
(\res{crms.graph.check_corr_hourly_change.edges_median} along edges against
\res{crms.graph.check_corr_hourly_change.same_basin_nonedge_pairs_le_maxkm_median} for unjoined pairs).

England uses OS Open Rivers~\cite{os2024openrivers}. Gauges snap to a link of the same river name within 500~m, or to
the nearest link within 150~m, and walking downstream joins each gauge to the next gauge on its branch. Canals are
excluded because they cross watersheds, and the \res{uk.phys.graph.off_channel_gauges} gauges that measure storage
beside the river get one lateral edge each. The graph has \res{uk.phys.graph.edges} edges in \res{uk.phys.graph.components} components, and
\res{uk.phys.graph.isolated} gauges have no edge. Hourly level changes correlate at a median of
\res{uk.phys.graph.corr.physical} along its edges against \res{uk.phys.graph.corr.inferred} along the inferred
edges and \res{uk.phys.graph.corr.random} for random unjoined pairs. On \res{uk.phys.graph.dir.downstream_later} of the
\res{uk.phys.graph.dir.n} correlated in-channel edges the downstream gauge responds later, and on
\res{uk.phys.graph.dir.downstream_earlier} earlier. The
inferred graph joins river pieces across watersheds.

\par\textbf{Effect of the graph and missing structures.}
The Scenario GNN and its reference exist on both graphs, the physics models on the physical graphs. With seeds paired,
the physical graph changes the Scenario GNN's median Nash-Sutcliffe efficiency (NSE) under $q_0$ by
\res{hou.phys.vsinf.test.gnn.sc.q0.dnse} in Houston, \res{crms.phys.vsinf.test.gnn.sc.q0.dnse} in Louisiana and
\res{uk.phys.cmp.test.gconv.dnse_phys_minus_inferred} in England, and with realized drivers the
inferred graph is better in Houston and Louisiana. Without gates, pumps or wells, the physics branch in these regions
moves water only along channel and tide edges and adds rain and storage, and a station without an edge keeps only rain
and storage. In England the surge scenarios are inert, so the issue-time reference differs from the scenario-input model
only in never seeing future rain. Without a common datum, each edge offset starts at minus the training-mean head
difference and absorbs datum offsets, so the learned offsets are not physical gradients.

\subsection{Station group formation}\label{app:L2}

\par\textbf{Station group formation across regions.}
In every region and period, per-driver station groups are formed from selection-period observations (2018 for the 2019
test periods, and the 2021 selection segment for the 2022 hold-out) together with the network graph, without any model
output. On the physical Houston graph, adding tidal-edge stations expands the surge group
from \res{hou.groups.surge} to \res{hou.phys.groups.n_surge} gauges, \res{hou.phys.groups.n_surge_tide_only} of which
enter through a tide edge although their 2018 surge correlation is below 0.15.

\subsection{Forecast skill and interface cost}\label{app:L3}

\begin{table*}[tp]
\caption{Forecast skill across the nine observed test periods: median station NSE and paired $\Delta$NSE vs.\ reference (mean\,$\pm$\,SD over three seeds).}
\label{tab:s-xr-forecast}
\centering
\setlength{\tabcolsep}{3pt}%
\resizebox{\textwidth}{!}{%
\begin{tabular}{@{}cllcccccc@{}}
\toprule
 & Test period & Persist. & Structure & NSE under $q_0$ & Reference & $\Delta$NSE & No rain forecast & Perfect-driver hindcast \\
\midrule
\multirow{24}{*}{\rotatebox[origin=c]{90}{South Florida}} & \multirow{4}{*}{Miami-Dade 2019} & \multirow{4}{*}{0.755} & Scenario GNN & 0.811$\pm$0.003 & 0.814$\pm$0.001 & $-$0.002$\pm$0.003 & 0.804$\pm$0.005 & 0.846$\pm$0.002 \\
 &  &  & Hybrid & 0.804$\pm$0.001 & 0.809$\pm$0.006 & $-$0.005$\pm$0.007 & 0.811$\pm$0.005 & 0.844$\pm$0.001 \\
 &  &  & Physics + residual & 0.774$\pm$0.002 & 0.773$\pm$0.002 & $+$0.001$\pm$0.003 & 0.772$\pm$0.002 & 0.813$\pm$0.004 \\
 &  &  & Physics-only & 0.752$\pm$0.002 & 0.766$\pm$$<$0.001 & $-$0.014$\pm$0.002 & 0.746$\pm$$<$0.001 & 0.776$\pm$0.018 \\
\cmidrule(l){2-9}
 & \multirow{4}{*}{Miami-Dade 2022} & \multirow{4}{*}{0.704} & Scenario GNN & 0.787$\pm$0.002 & 0.787$\pm$0.001 & 0.000$\pm$0.003 & 0.782 & 0.817 \\
 &  &  & Hybrid & 0.718$\pm$0.006 & 0.739$\pm$0.003 & $-$0.021$\pm$0.007 & 0.705 & 0.786 \\
 &  &  & Physics + residual & 0.612$\pm$0.095 & 0.702$\pm$0.054 & $-$0.090$\pm$0.058 & 0.607 & 0.678 \\
 &  &  & Physics-only & 0.653$\pm$0.025 & 0.703$\pm$0.033 & $-$0.050$\pm$0.021 & 0.649 & 0.755 \\
\cmidrule(l){2-9}
 & \multirow{4}{*}{Kissimmee lakes 2019} & \multirow{4}{*}{0.993} & Scenario GNN & 0.992$\pm$$<$0.001 & 0.993$\pm$$<$0.001 & 0.000$\pm$$<$0.001 & 0.992 & 0.993 \\
 &  &  & Hybrid & 0.989$\pm$0.001 & 0.989$\pm$0.001 & 0.000$\pm$0.001 & 0.989 & 0.992 \\
 &  &  & Physics + residual & 0.984$\pm$0.014 & 0.992$\pm$0.001 & $-$0.008$\pm$0.013 & 0.985 & 0.985 \\
 &  &  & Physics-only & 0.991$\pm$$<$0.001 & 0.993$\pm$$<$0.001 & $-$0.002$\pm$$<$0.001 & 0.992 & 0.992 \\
\cmidrule(l){2-9}
 & \multirow{4}{*}{Kissimmee lakes 2022} & \multirow{4}{*}{0.980} & Scenario GNN & 0.976$\pm$$<$0.001 & 0.977$\pm$0.001 & $-$0.001$\pm$0.001 & 0.975 & 0.977 \\
 &  &  & Hybrid & 0.973$\pm$0.002 & 0.973$\pm$0.002 & 0.000$\pm$0.001 & 0.973 & 0.977 \\
 &  &  & Physics + residual & 0.971$\pm$0.001 & 0.977$\pm$0.001 & $-$0.006$\pm$0.002 & 0.970 & 0.976 \\
 &  &  & Physics-only & 0.977$\pm$$<$0.001 & 0.978$\pm$$<$0.001 & $-$0.001$\pm$$<$0.001 & 0.977 & 0.979 \\
\cmidrule(l){2-9}
 & \multirow{4}{*}{Hendry-Collier 2019} & \multirow{4}{*}{0.941} & Scenario GNN & 0.938$\pm$0.003 & 0.941$\pm$$<$0.001 & $-$0.003$\pm$0.003 & 0.935 & 0.931 \\
 &  &  & Hybrid & 0.922$\pm$0.001 & 0.935$\pm$0.003 & $-$0.013$\pm$0.003 & 0.922 & 0.940 \\
 &  &  & Physics + residual & 0.933$\pm$0.001 & 0.937$\pm$0.001 & $-$0.004$\pm$0.001 & 0.931 & 0.938 \\
 &  &  & Physics-only & 0.934$\pm$$<$0.001 & 0.935$\pm$0.001 & $-$0.001$\pm$$<$0.001 & 0.934 & 0.940 \\
\cmidrule(l){2-9}
 & \multirow{4}{*}{Hendry-Collier 2022} & \multirow{4}{*}{0.903} & Scenario GNN & 0.892$\pm$0.001 & 0.897$\pm$0.002 & $-$0.005$\pm$0.001 & 0.892 & 0.906 \\
 &  &  & Hybrid & 0.870$\pm$0.003 & 0.902$\pm$0.001 & $-$0.032$\pm$0.003 & 0.868 & 0.900 \\
 &  &  & Physics + residual & 0.876$\pm$0.003 & 0.896$\pm$0.001 & $-$0.020$\pm$0.003 & 0.872 & 0.903 \\
 &  &  & Physics-only & 0.856$\pm$0.001 & 0.895$\pm$$<$0.001 & $-$0.039$\pm$0.001 & 0.858 & 0.893 \\
\midrule
\multirow{12}{*}{\rotatebox[origin=c]{90}{River networks}} & \multirow{4}{*}{Houston 2019} & \multirow{4}{*}{$-$0.021} & Scenario GNN & 0.326$\pm$0.005 & 0.412$\pm$0.002 & $-$0.086$\pm$0.005 & 0.364$\pm$0.004 & 0.771$\pm$0.002 \\
 &  &  & Hybrid & 0.316$\pm$0.005 & 0.408$\pm$0.004 & $-$0.092$\pm$0.007 & 0.374$\pm$0.007 & 0.769$\pm$0.007 \\
 &  &  & Physics + residual & 0.250$\pm$0.006 & 0.355$\pm$0.004 & $-$0.105$\pm$0.006 & 0.333$\pm$0.003 & 0.723$\pm$0.006 \\
 &  &  & Physics-only & 0.111$\pm$0.025 & 0.293$\pm$0.003 & $-$0.182$\pm$0.023 & 0.287$\pm$0.004 & 0.612$\pm$0.006 \\
\cmidrule(l){2-9}
 & \multirow{4}{*}{Louisiana 2019} & \multirow{4}{*}{0.654} & Scenario GNN & 0.768$\pm$$<$0.001 & 0.764$\pm$0.003 & $+$0.004$\pm$0.003 & 0.763$\pm$0.003 & 0.872$\pm$$<$0.001 \\
 &  &  & Hybrid & 0.756$\pm$0.006 & 0.766$\pm$0.003 & $-$0.011$\pm$0.009 & 0.752$\pm$0.003 & 0.877$\pm$0.002 \\
 &  &  & Physics + residual & 0.714$\pm$0.001 & 0.719$\pm$0.001 & $-$0.005$\pm$0.001 & 0.709$\pm$0.002 & 0.827$\pm$0.002 \\
 &  &  & Physics-only & 0.688$\pm$$<$0.001 & 0.690$\pm$$<$0.001 & $-$0.002$\pm$$<$0.001 & 0.688$\pm$$<$0.001 & 0.804$\pm$0.001 \\
\cmidrule(l){2-9}
 & \multirow{4}{*}{Aire and Calder 2019} & \multirow{4}{*}{0.523} & Scenario GNN & 0.623$\pm$0.002 & 0.609$\pm$0.004 & $+$0.014$\pm$0.002 & 0.542$\pm$0.007 & 0.872$\pm$0.001 \\
 &  &  & Hybrid & 0.613$\pm$0.004 & 0.605$\pm$0.005 & $+$0.007$\pm$0.001 & 0.539$\pm$0.010 & 0.865$\pm$0.005 \\
 &  &  & Physics + residual & 0.626$\pm$0.002 & 0.614$\pm$$<$0.001 & $+$0.012$\pm$0.002 & 0.543$\pm$0.003 & 0.838$\pm$0.003 \\
 &  &  & Physics-only & 0.546$\pm$0.009 & 0.567$\pm$0.003 & $-$0.022$\pm$0.006 & 0.539$\pm$0.001 & 0.641$\pm$0.027 \\
\bottomrule
\end{tabular}}
\end{table*}

Table~\ref{tab:s-xr-forecast} extends Table~\ref{tab:xr} with the issue-time reference, the forecast without the rain
forecast and the perfect-driver hindcast. The issue-time scenario $q_0$ uses the statistical rain forecast, and the
reference is the same architecture trained without future rain and with surge and operations held. The forecast
without rain forecast sets future rain to zero, the perfect-driver hindcast receives the realized drivers, and
persistence repeats the last observation. A negative $\Delta$NSE is a cost. The hindcast is a diagnostic reference and
no bound on skill. In Hendry-Collier 2019, for example, the Scenario GNN's hindcast (\res{sf.s6p2.fc.test.gnn.real.nse.mean})
lies below its forecast under $q_0$ (\res{sf.s6p2.fc.test.gnn.q0.nse.mean}).

\par\textbf{South Florida.}
Inland water levels are so persistent that no structure clearly beats persistence (\res{sf.s6p0.fc.test.persistence.nse}
in the Kissimmee lakes and \res{sf.s6p2.fc.test.persistence.nse} in Hendry-Collier in 2019), so their small costs say
little. The largest inland costs fall on the Hybrid forecaster and the Physics-only model in Hendry-Collier 2022
(\res{sf.s7p2d1.cost.test.hybrid.nse.mean} and \res{sf.s7p2d1.cost.test.physonly.nse.mean}), and one Physics + state
residual seed in the Kissimmee lakes 2019 is unstable. In Miami-Dade 2022 the Scenario GNN has no cost, while the physics
models lose \res{sf.s7p1d1.cost.test.physres.nse.mean} and \res{sf.s7p1d1.cost.test.physonly.nse.mean} and fall below
persistence, with references near persistence too (\res{sf.s7p1d1.fc.test.physres.ref.nse.mean} and
\res{sf.s7p1d1.fc.test.physonly.ref.nse.mean}). Two Physics + state residual seeds stopped early, with best epochs as
early as \res{s7.best_epoch.physres.min}.

\par\textbf{Houston.}
Every structure pays an NSE cost, from \res{hou.phys.acc.test.gnn.dnse} for the Scenario GNN to
\res{hou.phys.acc.test.physonly.dnse} for the Physics-only model, and every structure forecasts better without its rain
forecast. That forecast has no skill (median per-gauge $R^2$ \res{hou.rainfc.median_r2.test} in 2019 and
\res{hou.rainfc.median_r2.val} in 2018), so in $q_0$ it adds error that the reference, which never sees future rain, does
not carry. Persistence has a negative NSE (\res{hou.phys.acc.test.persistence.nse}) because bayou levels change fast,
and the perfect-driver hindcast of the Scenario GNN reaches \res{hou.phys.acc.test.gnn.real.nse}, far above its NSE
under $q_0$. The inferred-graph build shows a similar cost (\res{hou.test.gnn.dnse}), and there the Scenario GNN's
critical success index at p95 exceeds the reference's by \res{hou.test.cost.csi95}, so the cost lies in continuous error.

\par\textbf{Louisiana and England.}
In Louisiana every change lies between \res{crms.phys.acc.test.hybrid.dnse} and \res{crms.phys.acc.test.gnn.dnse}, and
the rain forecast has almost no skill. For the inferred-graph Scenario GNN, runs with a 400-epoch cap in place
of the 30-epoch recipe converge in \res{crms.conv3.gnn.converged-seeds} seeds and shrink its cost from
\res{crms.acc.test.gconv.dnse} to \res{crms.conv3.test.dnse} with verdicts unchanged, so a Louisiana cost depends on the
training budget. In England three structures match or exceed references that never see future rain. The Scenario
GNN's reference early-stopped in all \res{uk.phys.train.gconv.rt.early_stopped} seeds, and a run on the
inferred graph with early stopping switched off also shows no cost (\res{uk.ref2.test.dnse}). The rain forecast carries most of the
issue-time skill over persistence (Scenario GNN \res{uk.phys.acc.test.gconv.q0.nse} with it and
\res{uk.phys.acc.test.gconv.q0_rain0.nse} without). The Physics-only model loses \res{uk.phys.cost.test.d9phys.dnse}.
It has no neural parameters and its seeds differ only in minibatch order, so its standard deviation (SD) is not a sampling uncertainty.

\begin{table*}[tp]
\caption{Median station NSE (mean\,$\pm$\,SD over three seeds) and strict-rule verdicts (rain/surge; codes in Table~\ref{tab:s-xr-verdicts}) for Scenario GNN and three extra baselines.}
\label{tab:s-xr-baselines}
\centering
\setlength{\tabcolsep}{3pt}%
\resizebox{\textwidth}{!}{%
\begin{tabular}{@{}clccccccccc@{}}
\toprule
 & & & \multicolumn{4}{c}{NSE under $q_0$} & \multicolumn{4}{c}{Verdict, rain/surge} \\
\cmidrule(lr){4-7}\cmidrule(l){8-11}
 & Test period & Persist. & GNN & Transformer & TCN & MLP & GNN & Transformer & TCN & MLP \\
\midrule
\multirow{6}{*}{\rotatebox[origin=c]{90}{Structure graph}} & Miami-Dade 2019 & 0.755 & 0.811$\pm$0.003 & 0.804$\pm$0.003 & 0.795$\pm$0.001 & 0.769$\pm$0.011 & F/F & P/N & N/N & F/F \\
 & Miami-Dade 2022 & 0.704 & 0.787$\pm$0.002 & 0.783$\pm$0.007 & 0.736$\pm$0.002 & 0.731$\pm$0.003 & P/N & F/N & F/N & F/F \\
 & Kissimmee lakes 2019 & 0.993 & 0.992$\pm$$<$0.001 & 0.993$\pm$$<$0.001 & 0.768$\pm$0.119 & 0.991$\pm$$<$0.001 & F/-- & F/-- & F/-- & F/-- \\
 & Kissimmee lakes 2022 & 0.980 & 0.976$\pm$$<$0.001 & 0.981$\pm$0.001 & 0.926$\pm$0.053 & 0.980$\pm$$<$0.001 & F/-- & F/-- & F/-- & N/-- \\
 & Hendry-Collier 2019 & 0.941 & 0.938$\pm$0.003 & 0.942$\pm$0.001 & 0.942$\pm$0.002 & 0.940$\pm$0.001 & F/-- & F/-- & N/-- & N/-- \\
 & Hendry-Collier 2022 & 0.903 & 0.892$\pm$0.001 & 0.905$\pm$0.003 & 0.710$\pm$0.157 & 0.277$\pm$0.131 & F/-- & N/-- & N/-- & N/-- \\
\midrule
\multirow{3}{*}{\rotatebox[origin=c]{90}{Inferred}} & Houston 2019 & $-$0.021 & 0.331$\pm$0.002 & 0.393 & 0.332 & 0.338 & F/F & F/P & F/N & P/P \\
 & Louisiana 2019 & 0.654 & 0.757 & 0.780 & 0.747 & 0.733 & F/F & P/P & F/N & F/P \\
 & Aire and Calder 2019 & 0.523 & 0.621 & 0.601 & 0.624 & 0.575 & P/-- & F/-- & F/-- & P/-- \\
\bottomrule
\end{tabular}}
\end{table*}

\par\textbf{Data-driven baselines.}
Three scenario-input baselines see the same information as the Scenario GNN (Table~\ref{tab:s-xr-baselines}). A
Transformer encodes 6-h patches of each series' past and known future with shared weights, a temporal convolutional
network (TCN) applies dilated convolutions to each series, and a multilayer perceptron (MLP) reads all series flattened
together. None of them uses the graph. In the inland parts the TCN and the MLP can collapse, for example the TCN in the
Kissimmee lakes in 2019 (\res{sf.s6p0.fc.test.tcn.q0.nse.mean}) and the MLP in Hendry-Collier 2022
(\res{sf.s7p2d1.fc.test.mlp.q0.nse.mean}). The South Florida rows of Table~\ref{tab:s-xr-baselines} use the same
Scenario GNN runs as Table~\ref{tab:s-xr-forecast}. The Houston, Louisiana and England rows come from the
inferred-graph build, which holds the Scenario GNN, these baselines and an issue-time reference GNN and no physics
model, so their Scenario GNN differs from the one on the physical river network.

\subsection{Observational check per region}\label{app:L4}

\begin{table}[t]
\caption{Strict-rule observational verdicts (rain/surge) across nine observed test periods (P: pass, F: fail, N: no information, C: cannot separate; $^\ddagger$physical network, App.~\ref{app:L1}; P+R: Physics + residual).}
\label{tab:s-xr-verdicts}
\centering
\setlength{\tabcolsep}{3pt}%
\resizebox{\columnwidth}{!}{%
\begin{tabular}{@{}cllcccc@{}}
\toprule
 & & & \multicolumn{4}{c}{Verdict, rain/surge} \\
\cmidrule(l){4-7}
 & Region & Year & GNN & Hybrid & P+R & P-only \\
\midrule
\multirow{6}{*}{\rotatebox[origin=c]{90}{South Florida}} & Miami-Dade canals & 2019 & F/F & P/F & F/P & F/P \\
 & Miami-Dade canals & 2022 & P/N & F/P & N/C & F/C \\
 & Kissimmee lakes & 2019 & F/-- & P/-- & F/-- & F/-- \\
 & Kissimmee lakes & 2022 & F/-- & F/-- & F/-- & F/-- \\
 & Hendry-Collier canals & 2019 & F/-- & F/-- & F/-- & F/-- \\
 & Hendry-Collier canals & 2022 & F/-- & F/-- & F/-- & F/-- \\
\midrule
\multirow{3}{*}{\rotatebox[origin=c]{90}{\shortstack{River\\networks}}} & Houston bayous & 2019$^{\ddagger}$ & F/P & F/P & F/P & F/P \\
 & Louisiana marshes & 2019$^{\ddagger}$ & F/F & P/P & P/C & P/C \\
 & Aire and Calder rivers & 2019$^{\ddagger}$ & P/-- & F/-- & P/-- & F/-- \\
\bottomrule
\end{tabular}}
\end{table}

\begin{table*}[tp]
\caption{Strict-rule observational check across test and selection (Sel.) periods: three-seed slope ranges (pass window $[0.75,1.33]$) and verdicts (codes in Table~\ref{tab:s-xr-verdicts}; $^1$slope window, $^{\mathrm{2b}}$untrained-copy threshold).}
\label{tab:s-xr-check}
\centering
\setlength{\tabcolsep}{3pt}%
\resizebox{\textwidth}{!}{%
\begin{tabular}{@{}cll*{4}{ccc}@{}}
\toprule
 & & & \multicolumn{3}{c}{GNN} & \multicolumn{3}{c}{Hybrid} & \multicolumn{3}{c}{P+R} & \multicolumn{3}{c}{P-only} \\
\cmidrule(lr){4-6}\cmidrule(lr){7-9}\cmidrule(lr){10-12}\cmidrule(l){13-15}
 & Test period & Driver & Slope & Test & Sel. & Slope & Test & Sel. & Slope & Test & Sel. & Slope & Test & Sel. \\
\midrule
\multirow{8}{*}{\rotatebox[origin=c]{90}{South Florida}} & Miami-Dade 2019 & Rain & 0.99--1.09 & F & P & 1.01--1.11 & P & P & 0.87--0.92 & F & F & 1.06--1.14 & F & F \\
 &  & Surge & 3.65--4.70 & F & F & 1.42--1.51 & F & F & 1.10--1.12 & P & P & 1.09 & P & P \\
 & Miami-Dade 2022 & Rain & 0.76--0.91 & P & P & 0.99--1.00 & F$^{\mathrm{2b}}$ & F$^{\mathrm{1}}$ & 0.78--8.69 & N & N & 0.93--0.99 & F$^{\mathrm{2b}}$ & F$^{\mathrm{2b}}$ \\
 &  & Surge & 3.00--4.00 & N & N & 1.09--1.16 & P & C & 1.01--1.05 & C & P & 1.00--1.01 & C & C \\
 & Kissimmee lakes 2019 & Rain & 0.76--0.80 & F$^{\mathrm{2b}}$ & F$^{\mathrm{2b}}$ & 0.86--0.93 & P & P & 0.76--0.89 & F$^{\mathrm{2b}}$ & P & 0.79--0.90 & F$^{\mathrm{2b}}$ & P \\
 & Kissimmee lakes 2022 & Rain & 1.19--1.33 & F$^{\mathrm{2b}}$ & F$^{\mathrm{2b}}$ & 2.54--2.60 & F$^{\mathrm{1,2b}}$ & F$^{\mathrm{1}}$ & 1.10--1.27 & F$^{\mathrm{2b}}$ & F$^{\mathrm{1}}$ & 1.39--1.48 & F$^{\mathrm{1,2b}}$ & F$^{\mathrm{1,2b}}$ \\
 & Hendry-Collier 2019 & Rain & 0.68--0.71 & F$^{\mathrm{1}}$ & P & 0.68--0.73 & F$^{\mathrm{1}}$ & P & 0.50--0.52 & F$^{\mathrm{1}}$ & F$^{\mathrm{1}}$ & 0.60--0.64 & F$^{\mathrm{1}}$ & P \\
 & Hendry-Collier 2022 & Rain & 0.86--1.25 & F$^{\mathrm{2b}}$ & F$^{\mathrm{2b}}$ & 1.32--1.52 & F$^{\mathrm{1}}$ & F$^{\mathrm{1}}$ & 1.29--1.32 & F$^{\mathrm{2b}}$ & F$^{\mathrm{2b}}$ & 1.40--1.43 & F$^{\mathrm{1,2b}}$ & F$^{\mathrm{1,2b}}$ \\
\midrule
\multirow{5}{*}{\rotatebox[origin=c]{90}{River networks}} & Houston 2019 & Rain & 1.02--1.05 & F & F & 1.40--1.41 & F & F & 1.26--1.36 & F & F & 1.24--1.35 & F & F \\
 &  & Surge & 0.87--0.88 & P & F & 1.03--1.05 & P & P & 0.99 & P & P & 0.98--0.99 & P & P \\
 & Louisiana 2019 & Rain & 0.96--1.03 & F & F & 1.03 & P & P & 1.00--1.03 & P & P & 1.12--1.28 & P & P \\
 &  & Surge & 1.16--1.18 & F & P & 1.20--1.21 & P & P & 1.22--1.23 & C & P & 1.18--1.19 & C & P \\
 & Aire and Calder 2019 & Rain & 0.85--0.86 & P & F & 1.44--1.45 & F & P & 0.79--0.80 & P & P & 0.59--0.77 & F & P \\
\bottomrule
\end{tabular}}
\end{table*}

Table~\ref{tab:s-xr-verdicts} collects the strict-rule verdicts of the nine test periods of Table~\ref{tab:xr},
and Table~\ref{tab:s-xr-check} adds the slope ranges and the selection period behind them.
A verdict is a diagnostic of the observational check, not a grade of the explanation, which is not checkable in
observed regions (Section~\ref{sec:results-observed}). The Miami-Dade 2019 verdicts use surge group~A.
The slopes are station fixed-effect slopes, and a superscript names the failing clause where it was recorded
(App.~\ref{app:H}). The selection period is the year used for epoch selection and for forming the station groups,
which is 2018, or the end of 2021 for the 2022 hold-out. In both tables GNN, P+R and P-only denote the Scenario GNN,
the Physics + state residual model and the Physics-only model.

\par\textbf{Slopes outside the window.}
By Eq.~\eqref{eq:substitution} a slope below the window $[0.75,1.33]$ is consistent with a predicted change larger than the
observed one unless the model's error covaries negatively with its response, and a slope above it with a smaller one
unless the error covaries positively. Neither reading proves a response size. Slopes fall below the window for all four
structures in Hendry-Collier 2019 (from \res{sf.s6p2.obs.test.rain.physres.fe.slope.min} to
\res{sf.s6p2.obs.test.rain.hybrid.fe.slope.max}). They rise above it for the Hybrid forecaster in Houston and in
England, for one seed each of both physics models in Houston and for the Hybrid forecaster and the Physics-only model in
both inland parts in 2022, the year of Ian (Table~\ref{tab:s-xr-check}). The English Physics-only model fails through its one seed
that stopped early.

\par\textbf{The clause on the untrained copy.}
Several failures arise from clause~(2b) alone, which requires the trained model's sign agreement to exceed its untrained
copy's by 0.10. When an untrained copy has very few active cells above 1~cm, its sign agreement on those cells can approach unity. In the Kissimmee lakes in 2019 the untrained copies agree in sign on \res{sf.s6p0.obs.test.rain.gnn.untrained.sign}
(Scenario GNN), \res{sf.s6p0.obs.test.rain.physres.untrained.sign} and \res{sf.s6p0.obs.test.rain.physonly.untrained.sign}
of their active cells, so three structures fail clause~(2b) with slopes inside the acceptance window. In Miami-Dade 2022 the Hybrid forecaster fails
on a threshold set by \res{s7.v2.rain.hybrid.untrained.nbig} untrained cells, and the Scenario GNN in Houston and Louisiana also
fails rain through clause~(2b) with slopes inside the window.

\par\textbf{Changes between years.}
The groups were chosen on the selection period, so the comparison with it is partly in-sample. In the Kissimmee lakes
three of the four structures pass rain in 2018 but only the Hybrid forecaster in 2019, and in Hendry-Collier three pass
in 2018 and none in 2019. In England the Scenario GNN fails in 2018 and passes in 2019, while the Hybrid forecaster and
the Physics-only model do the opposite. For surge, the Houston Scenario GNN fails in 2018 and passes in 2019, and in
Louisiana three structures pass in 2018 but not in 2019. In Miami-Dade the Scenario GNN's rain pass of 2022 becomes a
fail in 2023 (slopes from \res{s7.v2.y2023.rain.gnn.slope.min} to \res{s7.v2.y2023.rain.gnn.slope.max}), and both
physics models move from \emph{cannot separate} in 2022 to a surge pass in 2023, while in Hendry-Collier the Hybrid
forecaster fails in 2022 and passes in 2023. Verdicts are therefore specific to their test period.

\par\textbf{Surge and built-in responses.}
Surge can be checked only in Miami-Dade, Houston and Louisiana. In the Kissimmee lakes in 2022 the surge group holds
\res{sf.s7p0d1.groups.surge.n} stations, below the minimum of three. In Louisiana the tide edges impose the boundary by
construction, so the untrained physics copies already meet the slope and sign clauses and the check cannot separate
learned from built-in surge responses, as for both physics models in Miami-Dade 2022. Under the pooled rule all four
structures pass rain in Miami-Dade 2019 (App.~\ref{app:H}). Among the baselines, rain passes for the Transformer in
Miami-Dade 2019 and Louisiana, for the MLP in Houston and England and for the Scenario GNN in England, and
surge passes for the Transformer and the MLP in Houston and Louisiana.

\subsection{Threshold queries and attribution}\label{app:L5}

\begin{table*}[tp]
\caption{Threshold queries and usage-task scores on unwarned rain-group cells in 2019 (three-seed ranges or seed means).}
\label{tab:s-xr-event}
\centering
\begin{tabular*}{\textwidth}{@{\extracolsep{\fill}}@{}llccccc@{}}
\toprule
Region & Structure & Not crossed (\%) & Median $A^\ast$ (mm) & U1 recall & U2 recall & U3 $\Delta$AUC \\
\midrule
\multirow{4}{*}{Miami-Dade} & Scenario GNN & 47 & 65 & 0.25 & 0.031 & $-$0.029 \\
 & Hybrid & 69 & 63 & 0.16 & 0.041 & $-$0.012 \\
 & Physics + residual & 68 & 58 & 0.21 & 0.045 & $-$0.103 \\
 & Physics-only & 81 & 56 & 0.25 & 0.042 & $-$0.135 \\
\midrule
\multirow{4}{*}{Kissimmee lakes} & Scenario GNN & 92.7--93.9 & 45--50 & 0.47--0.52 & 0.14--0.18 & $-$0.01 to $-$0.00 \\
 & Hybrid & 95.3--95.7 & 44--52 & 0.33--0.38 & 0.07--0.13 & $-$0.03 to 0.01 \\
 & Physics + residual & 92.8--98.9 & 48--64 & 0.50--0.60 & 0.00--0.15 & $-$0.41 to 0.00 \\
 & Physics-only & 93.4--94.6 & 46--49 & 0.55--0.67 & 0.10--0.11 & $-$0.00 to 0.00 \\
\midrule
\multirow{4}{*}{Hendry-Collier} & Scenario GNN & 44.6--63.0 & 77--87 & 0.46--0.64 & 0.08--0.13 & $-$0.09 to $-$0.02 \\
 & Hybrid & 57.3--65.3 & 63--65 & 0.32--0.33 & 0.04--0.08 & $-$0.08 to $-$0.01 \\
 & Physics + residual & 55.9--57.1 & 50--53 & 0.58--0.61 & 0.13--0.16 & $-$0.02 to $-$0.01 \\
 & Physics-only & 60.9--63.0 & 51--52 & 0.64--0.66 & 0.18--0.20 & $-$0.01 \\
\midrule
\multirow{4}{*}{Houston} & Scenario GNN & 0.4 & 39.5 & 0.51 & 0.24 & 0.001 \\
 & Hybrid & 4.5 & 34.5 & 0.43 & 0.28 & 0.013 \\
 & Physics + residual & 1.0 & 45.8 & 0.43 & 0.20 & $-$0.010 \\
 & Physics-only & 3.0 & 50.6 & 0.49 & 0.18 & $-$0.001 \\
\midrule
\multirow{4}{*}{Louisiana} & Scenario GNN & 81.6 & 57.1 & 0.24 & 0.09 & $-$0.032 \\
 & Hybrid & 85.1 & 57.6 & 0.21 & 0.08 & $-$0.038 \\
 & Physics + residual & 84.9 & 64.7 & 0.20 & 0.05 & $-$0.046 \\
 & Physics-only & 98.2 & 61.8 & 0.19 & 0.01 & $-$0.355 \\
\midrule
\multirow{4}{*}{Aire and Calder} & Scenario GNN & 21--28 & 16--17 & 0.65--0.70 & 0.47--0.50 & 0.07 \\
 & Hybrid & 12 & 20--21 & 0.51--0.53 & 0.41--0.42 & 0.06 to 0.07 \\
 & Physics + residual & 7--9 & 22 & 0.58 & 0.38--0.39 & 0.05 to 0.06 \\
 & Physics-only & 47--52 & 26 & 0.64--0.66 & 0.20--0.21 & $-$0.12 to $-$0.10 \\
\bottomrule
\end{tabular*}
\end{table*}

\par\textbf{Protocol and results.}
Every region follows the Miami-Dade threshold-query protocol (App.~\ref{app:K}), with added rain following a 24-h
template of the region's 20 largest training storms up to $A_{\max}$, the 99.9th percentile of training 24-h rain. This
is \res{mia.thr.amax_mm}~mm per 24~h in Miami-Dade, \res{sf.u1chk.s6p0.a-max-mm} in the Kissimmee lakes,
\res{sf.u1chk.s6p2.a-max-mm} in Hendry-Collier, \res{hou.phys.cf.a_max_mm} in Houston and \res{crms.phys.cf.a_max_mm} in
Louisiana. Table~\ref{tab:s-xr-event} reports the measures on the rain-group cells that are not warned under $q_0$ and
have a monotone rain grid, every 6~h. Not crossed is the share of these cells whose warning does not switch on within
the training rain range, and the median $A^\ast$ is taken over the crossing cells. U1 is the recall of the $A^\ast$
rule, which flags a cell when realized rain beyond the forecast reaches $A^\ast$, against the same model re-run with
realized rain. U2 is its recall of observed exceedances, and U3 is the area under the ROC curve of $-A^\ast$ minus that
of the predicted margin for the model's own warning 12~h later. Almost no
unwarned Kissimmee cell crosses within range, almost every Houston cell crosses within its Harvey-sized range, and the
Louisiana Physics-only model almost never crosses, since at stations without edges it responds through the rain
reservoir alone. In Houston the Scenario GNN's rain grid is non-monotone in \res{hou.phys.cf.test.gnn.share_nonmono}\% of
unwarned cells on average, with large differences between seeds. The $A^\ast$ rule recovers a minority of the model's
own re-run crossings in Miami-Dade and Louisiana and from about a third to about two thirds elsewhere, on small inland
denominators (from
\res{sf.u1chk.s6p0.all.positives.min} to \res{sf.u1chk.s6p0.all.positives.max} positives per run in the Kissimmee lakes).
For observed exceedances (U2) the rule uses the realized rain in hindsight. For the model's own warning 12~h later (U3),
$-A^\ast$ ranks cells about as well as the predicted margin or worse in the 2019 South Florida parts, Houston and
Louisiana. In England it ranks them better than the predicted margin for three structures, by
\res{uk.phys.u3.test.physres.d-astar-pred.auc.min} to \res{uk.phys.u3.test.gnn.d-astar-pred.auc.max} in ROC-AUC with
intervals that exclude zero in all \res{uk.phys.u3.test.gnn.d-astar-pred.n-ci-excl0} seeds, and worse for
the Physics-only model, a head-to-head difference in ranking, not an increment of information. The unstable
Kissimmee seed gives the outlying \res{sf.s6p0.u3.test.physres.d-astar-pred.auc.min}.

\par\textbf{Attribution.}
The inland parts have no tide edge, so their surge and tide Shapley values are zero to printed precision and rain is
the largest player in \res{sf.s6p0.attr.test.gnn.share-r-largest.min}\% of cells for every structure and seed. This
follows from the graph and is no evidence about surge. Mean $|\phi_R|$ ranges from
\res{sf.s6p0.attr.test.hybrid.trained.phir.meanabs.min} to \res{sf.s6p0.attr.test.physonly.trained.phir.meanabs.max}~cm
in the Kissimmee lakes, against at most \res{sf.s6p0.attr.test.physres.untrained.phir.meanabs}~cm for untrained copies.
In Houston the untrained physics copies attribute more rain (\res{hou.phys.attrib.test.rain.physres.r.untrained}~cm)
than the trained models (at most \res{hou.phys.attrib.test.rain.physonly.r}~cm), and in England they attribute
\res{uk.phys.attr.test.d9phys.phi_rain.untrained}~cm against \res{uk.phys.attr.test.d9phys.phi_rain.mean}~cm trained, so
attribution size alone does not show learning. In Louisiana the held surge outweighs rain on rain-group cells for three
structures.

\subsection{Second simulator}\label{app:L6}

\begin{figure*}[tp]
\centering
\includegraphics[width=\textwidth]{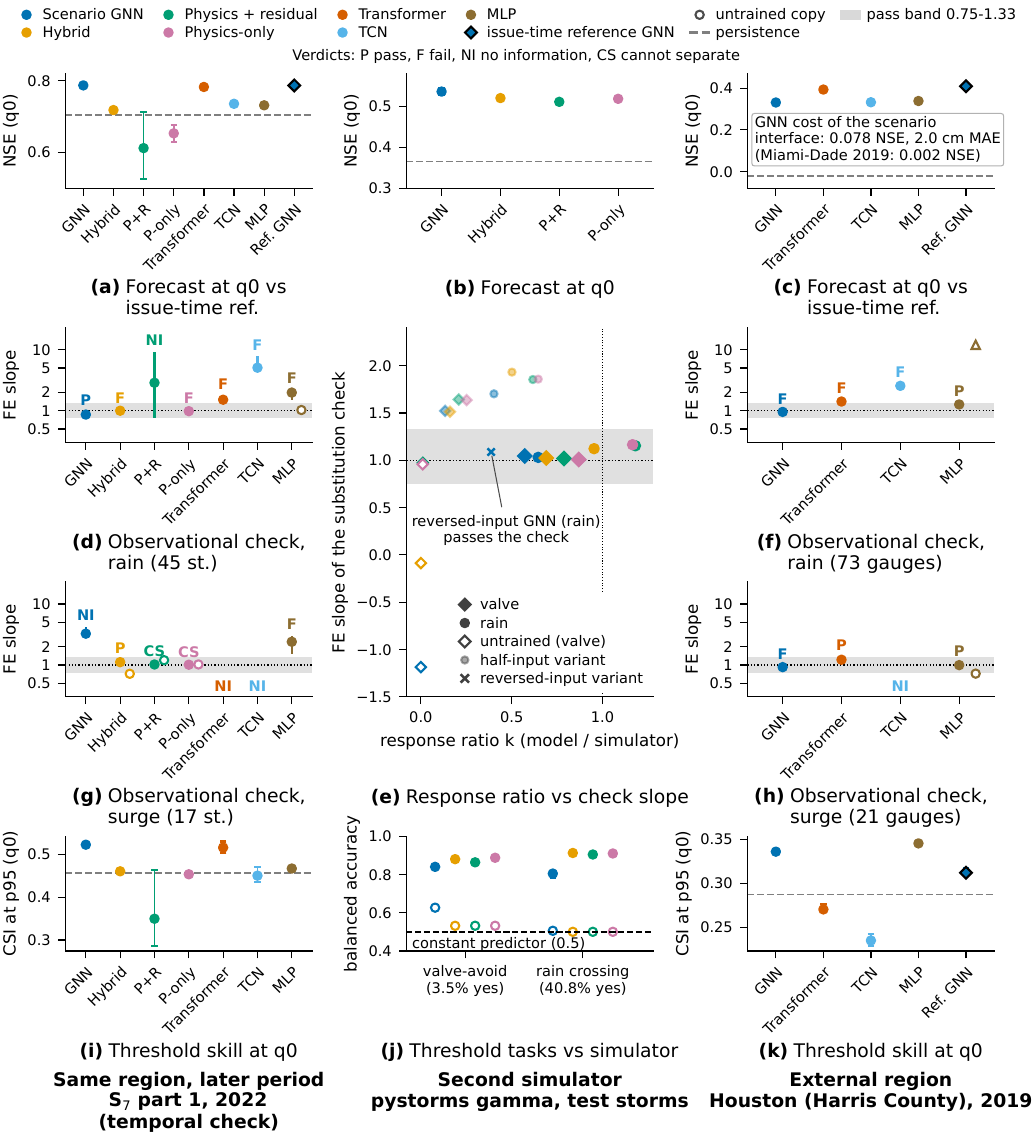}
\caption{Cross-region forecast NSE and strict-rule station fixed-effect (FE) slopes across the nine observed test periods and the simulated river-valve network (seed mean and range; P+R: Physics + residual).}
\label{fig:s-transfer}
\end{figure*}

\begin{table*}[t]
\caption{Driver responses in the extension systems: strict-rule slope ranges, verdicts (NI: no information, CS: cannot separate, F$_u$: untrained-clause fail), and simulated $k$ (mean\,$\pm$\,SD; untrained values in parentheses; second driver: surge in L1/L3, valves in L2).}
\label{tab:s-transfer-response}
\begin{tabular*}{\textwidth}{@{\extracolsep{\fill}}clcccccc@{}}
\toprule
 & & \multicolumn{3}{c}{Rain} & \multicolumn{3}{c}{Second driver} \\
\cmidrule(lr){3-5}\cmidrule(lr){6-8}
 & Model & Slope & Verdict & $k$ & Slope & Verdict & $k$ \\
\midrule
\multirow{7}{*}{\rotatebox[origin=c]{90}{\begin{tabular}{@{}c@{}}L1\\S\_7 2022\end{tabular}}} & Scenario GNN & 0.76--0.91 & P & \multirow{7}{*}{--} & 3.00--4.00 & NI & \multirow{7}{*}{--} \\
 & Hybrid & 0.99--1.00 & F$_u$ (1.00) &  & 1.09--1.16 & P &  \\
 & Physics + residual & -- & NI &  & 1.01--1.05 & CS (1.19) &  \\
 & Physics-only & 0.93--0.99 & F$_u$ (0.79) &  & 1.00--1.01 & CS (1.02) &  \\
 & Transformer & 1.50--1.69 & F &  & -- & NI &  \\
 & TCN & 4.55--7.45 & F &  & -- & NI &  \\
 & MLP & 1.53--2.08 & F &  & 1.55--2.84 & F &  \\
\midrule
\multirow{5}{*}{\rotatebox[origin=c]{90}{\begin{tabular}{@{}c@{}}L2\\\texttt{gamma}\end{tabular}}} & Scenario GNN & 1.03--1.04 & P & 0.65\,$\pm$\,0.10 (0.00) & 1.02--1.07 & P & 0.57\,$\pm$\,0.03 (0.00) \\
 & Hybrid & 1.11--1.14 & P & 0.95\,$\pm$\,0.03 (0.00) & 1.01--1.04 & P & 0.69\,$\pm$\,0.05 (0.00) \\
 & Physics + residual & 1.15--1.16 & F$_u$ (1.00) & 1.18\,$\pm$\,0.02 (0.01) & 1.01--1.02 & CS (0.97) & 0.79\,$\pm$\,0.01 (0.01) \\
 & Physics-only & 1.16--1.17 & F$_u$ (1.00) & 1.17\,$\pm$\,$<$0.01 (0.01) & 1.00--1.01 & CS (0.96) & 0.87\,$\pm$\,$<$0.01 (0.01) \\
\cmidrule(l){2-8}
 & Reversed-input GNN & 0.92--1.21 & P & 0.39\,$\pm$\,0.18 & $-$0.17 to $-$0.14 & F & $-$0.11\,$\pm$\,0.04 \\
\midrule
\multirow{4}{*}{\rotatebox[origin=c]{90}{\begin{tabular}{@{}c@{}}L3\\Houston\end{tabular}}} & Scenario GNN & 0.95--0.96 & F$_u$ (0.86) & \multirow{4}{*}{--} & 0.91--0.94 & F$_u$ (0.80) & \multirow{4}{*}{--} \\
 & Transformer & 1.34--1.42 & F &  & 1.16--1.32 & P &  \\
 & TCN & 2.42--2.60 & F &  & -- & NI &  \\
 & MLP & 1.26--1.28 & P &  & 0.97--1.02 & P &  \\
\bottomrule
\end{tabular*}
\end{table*}

\begin{table}[t]
\caption{Valve-avoid decisions on the second simulator (L2). Mean\,$\pm$\,SD over three seeds. UD, undefined (no predicted positives).}
\label{tab:s-transfer-valve}
\setlength{\tabcolsep}{3pt}%
\resizebox{\columnwidth}{!}{%
\begin{tabular}{@{}lcccc@{}}
\toprule
Predictor & \begin{tabular}[b]{@{}c@{}}Balanced\\acc.\end{tabular} & \begin{tabular}[b]{@{}c@{}}Recall\\(`avoidable')\end{tabular} & \begin{tabular}[b]{@{}c@{}}Precision\\(`avoidable')\end{tabular} & \begin{tabular}[b]{@{}c@{}}Raw\\agreement\end{tabular} \\
\midrule
Always `avoidable' & 0.50 & 1.00 & 0.04 & 0.035 \\
Never `avoidable' & 0.50 & 0.00 & UD & 0.965 \\
\midrule
Scenario GNN & 0.84\,$\pm$\,0.02 & 0.79\,$\pm$\,0.04 & 0.21\,$\pm$\,0.03 & 0.889\,$\pm$\,0.019 \\
Hybrid & 0.88\,$\pm$\,0.02 & 0.90\,$\pm$\,0.03 & 0.19\,$\pm$\,0.02 & 0.859\,$\pm$\,0.020 \\
Physics + residual & 0.86\,$\pm$\,$<$0.01 & 0.75\,$\pm$\,$<$0.01 & 0.54\,$\pm$\,0.01 & 0.969\,$\pm$\,$<$0.001 \\
Physics-only & 0.89\,$\pm$\,$<$0.01 & 0.80\,$\pm$\,$<$0.01 & 0.52\,$\pm$\,0.01 & 0.967\,$\pm$\,0.001 \\
\midrule
Untrained copies (range over 4) & 0.53--0.63 & 0.98--0.99 & 0.04--0.05 & 0.109--0.297 \\
\bottomrule
\end{tabular}}
\end{table}

\par\textbf{Extension checks.}
Figure~\ref{fig:s-transfer} and Tables~\ref{tab:s-transfer-response} and~\ref{tab:s-transfer-valve} summarize three extension checks, labeled L1 to L3: L1 is the 2022 Miami-Dade hold-out on SF$^2$Bench data, L2 is the second simulator (\emph{gamma}) described below, and L3 is the Houston build on
its inferred graph with the gauges of the Harris County flood warning system and NOAA.
In every check each model was retrained on its own system with the main-study recipe and three
seeds, without frozen transfer or fine-tuning. In the figure and in Table~\ref{tab:s-transfer-response}, a slope is the station fixed-effect slope of the
observational check, taken over basins on the second simulator, measuring residual covariance rather than response magnitude. Its verdicts follow the strict rule, with P pass and F fail as in
Table~\ref{tab:s-xr-verdicts}. The printed codes NI and CS correspond to N (no information) and C (cannot separate) in
that table, where C indicates that the untrained copy also meets the slope and sign criteria, and F$_u$ denotes a failure on the
untrained-copy clause alone. In Table~\ref{tab:s-transfer-response}, $k$ is the response ratio of Eq.~\eqref{eq:paired},
parentheses give the untrained copy's $k$, its slope for CS or its sign share for F$_u$, and -- marks $k$ where no
simulator exists and a slope where the check has no information.

\par\textbf{System.}
The pystorms \emph{gamma}
network~\cite{rimer2023pystorms} drains storage basins (Table~\ref{tab:s-inventory}), each through its own orifice
valve, to one free outfall, without an ocean boundary. Storms rescale one design-storm shape to totals between
\res{gamma.rain.min_mm} and \res{gamma.rain.max_mm}~mm, and \res{gamma.flood_share}\% of the \res{gamma.storms.total}
storms flood. They are split into \res{gamma.storms.train} training, \res{gamma.storms.val} validation and
\res{gamma.storms.test} test storms, and models read and forecast 6~h at 15-min steps. Every valve starts at a random
setting and is redrawn with probability 1/12 at every 15-min boundary, independent of the state, which avoids
closed-loop confounding but contains no control logic. The noise floor is far lower than on the main simulator (median
\res{gamma.cf.noise.median}~cm). For the \res{gamma.cf.n_issue} test issue times, 2~h before the storm peak,
\res{gamma.cf.n_runs} runs in all hold each valve at five settings or add rain, each continuous from the storm's start.
With realized drivers the four structures reach test NSE from \res{gamma.nse.real.physonly.mean} to
\res{gamma.nse.real.hybrid.mean}, and under the issue-time scenario, which holds the valves and adds no rain, from
\res{gamma.nse.q0.physres.mean} to \res{gamma.nse.q0.gnn.mean}, against \res{gamma.nse.persistence} for persistence.

\par\textbf{Operation responses.}
On this network the evaluated operation responses are those to valve settings. On the simulated coastal network the
first set of runs holds one fixed control path, and a second set of runs with random schedules evaluates gate, weir and
pump operations (App.~\ref{app:P}). The median valve response ratio is
\res{gamma.k_valve.gnn.trained} for the Scenario GNN, \res{gamma.k_valve.hybrid.trained} for the Hybrid forecaster,
\res{gamma.k_valve.physres.trained} for the Physics + state residual model and \res{gamma.k_valve.physonly.trained}
for the Physics-only model, over \res{gamma.k_valve.ncells} cells. Half-input variants reach from
\res{gamma.k_valve.gnn.half} to \res{gamma.k_valve.physonly.half}, and untrained copies at most
\res{gamma.k_valve.physres.untrained}. For an added 25.4~mm of rain the ratios are \res{gamma.k_rain.gnn.trained},
\res{gamma.k_rain.hybrid.trained}, \res{gamma.k_rain.physres.trained} and \res{gamma.k_rain.physonly.trained} in the
same order, and the reversed-input Scenario GNN has a positive rain ratio of \res{gamma.k_rain.gnn.reversed}. The
observational check gives valve slopes near one for every trained model while these ratios stay below one, and the
untrained physics copies already reach valve slopes near one, so their valve verdict is \emph{cannot separate}. Under
the strict rule the Scenario GNN and the Hybrid forecaster pass for valves and rain, the physics models fail
rain through the clause on the untrained copy, every half-input variant fails, and the reversed-input Scenario GNN passes the
rain check.

\par\textbf{Threshold decisions.}
Table~\ref{tab:s-transfer-valve} lays out the threshold decision metrics on the second simulator in the format of
Table~\ref{tab:cf-threshold}. In the valve-avoid task a cell is a basin warned at the simulated base at one of the test
issue times, 2~h before the storm peak, together with one valve. It is avoidable when some single setting of that valve,
from 0 to 1 in steps of 0.25 and held after the issue time, keeps the basin's 6-h peak below its training p95. The
constant ``never avoidable'' makes no positive prediction, so its precision is undefined (UD in the table). Only
\res{gamma.va.pos_rate}\% of the \res{gamma.va.n} cells are avoidable, so a constant ``never avoidable''
agrees with the simulator on \res{gamma.va.const_acc} of the cells, more than the Scenario GNN (at most
\res{gamma.va.gnn.trained.acc.max}) and the Hybrid forecaster (at most \res{gamma.va.hybrid.trained.acc.max}). In
balanced accuracy the trained models reach from \res{gamma.va.all.trained.ba.min} to \res{gamma.va.all.trained.ba.max},
against \res{gamma.va.all.untrained.ba.min} to \res{gamma.va.all.untrained.ba.max} for untrained copies and 0.5 for a
constant classifier. The precision of the avoidable class is \res{gamma.va.gnn.trained.precision.mean} for the Scenario
GNN and \res{gamma.va.hybrid.trained.precision.mean} for the Hybrid forecaster, against
\res{gamma.va.physres.trained.precision.mean} and \res{gamma.va.physonly.trained.precision.mean} for the physics models,
so the two data-driven models call many cells avoidable that are not. In the rain-crossing task (\res{gamma.rf.n}
unwarned basin cells) the trained models reach a balanced accuracy from \res{gamma.rf.all.trained.ba.min} to
\res{gamma.rf.all.trained.ba.max}, against at most \res{gamma.rf.all.untrained.ba.max} for untrained copies. These
results concern a random, state-independent schedule on a network without an ocean boundary, not real operating rules.

{\smbcolor\subsection{Synthetic tidal-barrage testbed}\label{app:Lsmb}

\begin{table}[t]\smbcolor
\caption{\smbcolor Physical, operational, and forcing settings of the synthetic coastal-barrage testbed (levels in meters above 2008--2016 Tanjong Pagar mean sea level; $^{\dagger}$effective Manning roughness, 0.025 in sensitivity test).}
\label{tab:s-smb-setup}
\centering
\setlength{\tabcolsep}{4pt}%
\begin{adjustbox}{max width=\columnwidth}
\begin{tabular}{@{}lll@{}}
\toprule
Setting & Value & Basis \\
\midrule
\multicolumn{3}{@{}l}{\emph{Network and structures}} \\
Subcatchments & 100 of 1~km$^2$ & documented \\
Imperviousness & 85, 70 or 30\% & derived \\
Bank level & at least +4.0~m & assumed \\
Trunk Manning $n$ & 0.030 & assumed$^{\dagger}$ \\
Reservoir area & 2.07~km$^2$ & derived \\
Gates & 3 groups of 3 & documented \\
Gate width, height & 29.75~m, 5~m & documented \\
Gate opening time & 0.42~h & documented \\
Gate sill, crest & $-$2.5~m, +2.5~m & assumed \\
Discharge coefficient & 0.65 & assumed \\
Flap gates on gates & no backflow & assumed \\
Pumps & 7 $\times$ 40~m$^3$/s at 1~m lift & documented \\
Pump shut-off lift & 3.5~m & assumed \\
Spill weir crest, width & +2.5~m, 267.75~m & derived \\
Spill weir & uncontrolled, flap gate & derived \\
\midrule
\multicolumn{3}{@{}l}{\emph{Operating rule}} \\
Operating level & 0.00~m & assumed \\
Gate opening levels & +0.10, +0.20, +0.30~m & assumed \\
Gravity margin & 0.15~m & assumed \\
Gate closing level & 0.00~m & assumed \\
Start level of pump $j$ & +0.10 + 0.05$(j-1)$~m & assumed \\
All pumps on & +0.60~m & assumed \\
Pump stop level & $-$0.05~m & assumed \\
Pump stop margin & 0.25~m & assumed \\
Minimum run time & 1~h & assumed \\
\midrule
\multicolumn{3}{@{}l}{\emph{Drivers and runs}} \\
Rain & 10 NEA gauges, 5~min & recorded \\
Rain years & 2017--2023 & recorded \\
Sea level, hourly & UHSLC Tanjong Pagar & recorded \\
Tide fit & 37 constituents, 2008--2016 & recorded \\
Runs per issue time & 50 & testbed \\
Engine & SWMM 5.2.4, dynamic wave & -- \\
\bottomrule
\end{tabular}
\end{adjustbox}
\end{table}

\begin{table*}[t]\smbcolor
\caption{\smbcolor Forecast skill (7{,}300 test windows) and rain responses (200 test issue times, 25.4~mm added rain) on the synthetic barrage testbed (mean\,$\pm$\,SD over three seeds; $A^{*}$ coverage in \% in brackets; $^{p}$persistence; $^{u}$max $|k|$ of untrained copies).}
\label{tab:s-smb}
\setlength{\tabcolsep}{4pt}%
\begin{tabular*}{\textwidth}{@{\extracolsep{\fill}}llccccc@{}}
\toprule
 & Training & Scenario GNN & Hybrid & Physics + residual & Physics-only & Reference \\
\midrule
\multirow{2}{*}{NSE} & Rule & 0.02\,$\pm$\,0.003 & 0.02\,$\pm$\,0.003 & $-$0.12\,$\pm$\,0.007 & $-$0.21\,$\pm$\,0.028 & \multirow{2}{*}{$-$0.55$^{p}$} \\
 & Random gates & 0.00\,$\pm$\,0.003 & 0.01\,$\pm$\,0.003 & $-$0.35\,$\pm$\,0.015 & $-$0.52\,$\pm$\,0.010 & \\
\multirow{2}{*}{MAE (cm)} & Rule & 5.9 & 5.8 & 7.2 & 8.0 & \multirow{2}{*}{8.9$^{p}$} \\
 & Random gates & 9.3 & 8.9 & 11.1 & 11.7 & \\
CSI p95 & Rule & 0.088 & 0.089 & 0.114 & 0.111 & 0.101$^{p}$ \\
NSE against the held base & Rule & $-$0.23 & $-$0.23 & $-$0.30 & $-$0.37 & $-$0.50$^{p}$ \\
NSE, realized drivers & Rule & 0.93 & 0.83 & 0.31 & 0.03 & -- \\
\midrule
\multirow{2}{*}{$k_{\mathrm{rain}}$} & Rule & 1.06\,$\pm$\,0.01 & 0.98\,$\pm$\,0.07 & 0.43\,$\pm$\,0.05 & 0.26\,$\pm$\,0.02 & \multirow{2}{*}{$\le$0.01$^{u}$} \\
 & Random gates & 1.05\,$\pm$\,0.02 & 0.94\,$\pm$\,0.01 & 0.34\,$\pm$\,0.04 & 0.27\,$\pm$\,0.02 & \\
$k_{\mathrm{rain}}$, half / reversed input & Rule & 0.59 / 0.16 & 0.53 / $-$0.16 & 0.22 / $-$0.11 & 0.15 / 0.00 & -- \\
$A^{*}$ error $|e|$ / signed (mm) & Rule & 4.6 / $-$2.9 [95] & 3.5 / $-$1.6 [97] & 17.1 / 17.0 [81] & 14.3 / 13.6 [56] & -- \\
\bottomrule
\end{tabular*}
\end{table*}

\begin{table*}[t]\smbcolor
\caption{\smbcolor Reservoir held-operation response ratios $k$ (gravity / mixed / pump) and threshold balanced accuracy (gravity / pump; positive share in parentheses) on the synthetic barrage testbed (200 issue times, seed means; $^{\dagger}$reservoir or simulated base; $^{u}$untrained range; $^{c}$controller baseline).}
\label{tab:s-smb-ops}
\centering
\setlength{\tabcolsep}{4pt}%
\resizebox{\textwidth}{!}{%
\begin{tabular}{@{}llccccc@{}}
\toprule
 & Training & Scenario GNN & Hybrid & Physics + residual & Physics-only & Reference \\
\midrule
\multirow{2}{*}{$k$, held gates$^{\dagger}$} & Rule & $-$0.21 / $-$0.15 / $-$0.46 & 0.21 / $-$0.46 / $-$3.08 & 0.64 / $-$0.45 / $-$1.79 & 0.47 / $-$0.23 / $-$1.15 & -- \\
 & Random gates & 0.23 / 0.00 / $-$0.16 & 0.08 / 0.04 / $-$0.13 & 0.36 / $-$0.10 / $-$0.89 & 0.35 / 0.02 / $-$0.27 & -- \\
\multirow{2}{*}{$k$, held pumps$^{\dagger}$} & Rule & $-$0.44 / $-$0.18 / $-$0.25 & $-$0.77 / $-$0.32 / $-$0.29 & 0.60 / 0.49 / 0.45 & 0.41 / 0.33 / 0.29 & -- \\
 & Random gates & $-$1.32 / $-$0.56 / $-$0.41 & $-$0.89 / $-$0.42 / $-$0.31 & 1.32 / 0.57 / 0.48 & 1.93 / 0.70 / 0.69 & -- \\
\midrule
\multirow{2}{*}{Operation-avoid (14.3 / 11.5\%)} & Rule & 0.76 / 0.67 & 0.82 / 0.75 & 0.74 / 0.75 & 0.65 / 0.64 & 0.49--0.54 / 0.50--0.52$^{u}$ \\
 & Random gates & 0.81 / 0.73 & 0.86 / 0.75 & 0.74 / 0.75 & 0.65 / 0.64 & 0.35--0.56 / 0.39--0.54$^{u}$ \\
Pumps-off harm (0.8 / 1.8\%) & Rule & 0.59 / 0.65 & 0.56 / 0.63 & 0.61 / 0.60 & 0.80 / 0.68 & 0.66--0.88 / 0.51--0.67$^{u}$ \\
Set-points $-$0.5~m, avoid (9.5\%) & Rule & 0.65 & 0.67 & 0.71 & 0.63 & 0.83 / 0.74 / 0.52$^{c}$ \\
\quad simulated base + own change$^{\dagger}$ & Rule & 0.52 & 0.52 & 0.68 & 0.64 & -- \\
\quad change with the simulator's sign (\%)$^{\dagger}$ & Rule & 14 & 26 & 70 & 72 & 94$^{c}$ \\
Gates stuck closed, harm (2.28\%) & -- & -- & -- & -- & -- & 0.78 / 0.68 / 0.50$^{c}$ \\
\bottomrule
\end{tabular}}
\end{table*}

\begin{table}[t]\smbcolor
\caption{\smbcolor Observational check on the synthetic barrage testbed across 2{,}555 storm windows (verdicts: pooled / station FE / strict; P: pass, F: fail, NI: no information; $^{\dagger}$1{,}232 responsive cells, $<$1~mm counted as no response).}
\label{tab:s-smb-check}
\centering
\setlength{\tabcolsep}{3pt}%
\begin{adjustbox}{max width=\columnwidth}
\begin{tabular}{@{}lcccc@{}}
\toprule
 & \begin{tabular}[b]{@{}c@{}}Scenario\\GNN\end{tabular} & Hybrid & \begin{tabular}[b]{@{}c@{}}Physics +\\residual\end{tabular} & \begin{tabular}[b]{@{}c@{}}Physics-\\only\end{tabular} \\
\midrule
\multicolumn{5}{@{}l}{\emph{Operations on the operation-affected group}} \\
Verdict, rule & P/P/P & P/P/F & F/F/F & F/F/F \\
Verdict, random gates & F/F/F & P/P/F & F/F/F & F/F/F \\
FE slope, rule & 1.24 & 1.22 & 0.29 & $-$1.09 \\
Sign agreement, rule & 0.82 & 0.87 & 0.38 & 0.27 \\
\midrule
\multicolumn{5}{@{}l}{\emph{Same substitution, rule}} \\
No response, \%$^{\dagger}$ & 61 & 42 & 59 & 37 \\
Same sign, \%$^{\dagger}$ & 7 & 10 & 41 & 61 \\
Opposite sign, \%$^{\dagger}$ & 32 & 47 & 1 & 2 \\
Slope, all cells & $-$1.30 & $-$0.76 & 2.41 & 3.67 \\
\midrule
\multicolumn{5}{@{}l}{\emph{Rain on the rain group, surge near the barrage, rule}} \\
Verdict, rain & P/P/P & P/P/F & P/F/F & P/F/F \\
Verdict, surge & F/F/NI & F/F/NI & F/F/NI & F/F/NI \\
\bottomrule
\end{tabular}
\end{adjustbox}
\end{table}

\par\textbf{Role and design.}
The synthetic tidal-barrage testbed provides a simulated drainage system combining an ocean boundary with state-dependent, rule-controlled gate and pump operations, allowing counterfactual responses to closed-loop hydraulic controls to be evaluated against ground truth. Modeled on the published layout of Singapore's Marina Barrage and forced by recorded Singapore rainfall and sea-level series, its conduit geometry, operating levels, and rule thresholds are synthetic design choices (Table~\ref{tab:s-smb-setup}); results therefore characterize how learned forecasters respond to a tidal boundary and rule-controlled gates and pumps rather than representing actual Singapore drainage performance.

\par\textbf{System and simulations.}
A central reservoir collects runoff from \res{smb.net.subcatchments} subcatchments through \res{smb.net.conduits} conduits and discharges to the sea through three gate groups, seven pumps, and an uncontrolled spill weir (omitted from the models' graph). A level controller drains the reservoir by gravity when reservoir stage exceeds sea level by more than 0.15~m and otherwise activates staged pumps. Specifically, the three gate groups open at reservoir levels of $+0.10$, $+0.20$, and $+0.30$~m when the reservoir-to-sea head exceeds $0.15$~m and close below $0.00$~m or under reversed head; pump $j\in\{1,\dots,7\}$ starts above $+0.10+0.05(j-1)$~m when the sea is within $0.15$~m of the reservoir or higher (all pumps run above $+0.60$~m) and stops below $-0.05$~m or when reservoir head exceeds sea level by more than $0.25$~m, with a 1-h minimum run/open duration. The \res{smb.storms.total} storms pair 2017--2023 rainfall events with co-occurring or same-season sea-level series and are partitioned by episode into \res{smb.storms.train} training, \res{smb.storms.val} validation, and \res{smb.storms.test} test storms. Each test storm is evaluated at two issue times, roughly 2~h before the rainfall peak and 3~h before low water (\res{smb.it.test} test issue times total). Of these, \res{smb.it.gravity} fall in the gravity regime (reservoir exceeding sea level by $>0.15$~m for at least two thirds of the 6-h window), \res{smb.it.pump} in the pump regime (at most one third), and \res{smb.it.mixed} in the mixed regime. Each issue time comprises 50 continuous simulations from the storm start, including a rule-controlled base run and a held base run in which all gates and pumps remain fixed at their issue-time settings. The \res{smb.nodes.scored} scored nodes are grouped into \res{smb.groups.rain} rain nodes, \res{smb.groups.near} near-barrage nodes (within 10 links of the reservoir), and \res{smb.groups.op} operation-affected nodes, each using its 95th training percentile as the high-water threshold. Metric definitions follow App.~\ref{app:E}, sea-level response components are scored where they yield at least 100 responsive cells per 100 issue times, and confidence intervals resample the \res{smb.boot.clusters} test episodes.

\par\textbf{Model training.}
The four architectures are trained using the protocol of App.~\ref{app:C4} across three seeds with a 30-epoch cap (reached by \res{smb.train.capped} of 24 runs), receiving realized gate and pump settings and sea level as future scenario inputs. The \emph{rule} training set uses rule-controlled trajectories, whereas the \emph{random gates} training set randomizes gate schedules independently of water state while keeping pumps under the level controller. The issue-time scenario $q_0$ persists gates, pumps, and surge at their issue-time values and assumes zero future rainfall.

\par\textbf{Forecast skill and rainfall response.}
Under $q_0$, all four architectures outperform persistence in NSE, and all rule-trained models also improve in MAE (Table~\ref{tab:s-smb}). Under $+25.4$~mm of added rainfall, rule-trained response ratios on the rain group reach \res{smb.t2.r0.k.gnn} [\res{smb.t2.r0.k.gnn.lo}, \res{smb.t2.r0.k.gnn.hi}] for the Scenario GNN, \res{smb.t2.r0.k.hybrid} for the Hybrid forecaster, and \res{smb.t2.r0.k.physres} and \res{smb.t2.r0.k.physonly} for the two water-balance models. Because \res{smb.t6.posrate}\% of cells cross their threshold within the rainfall grid, minimum crossing rainfall is evaluated via its conditional error $|A^\ast_{\mathrm{model}}-A^\ast_{\mathrm{sim}}|$, yielding median absolute errors of \res{smb.t6.r0.err.gnn} and \res{smb.t6.r0.err.hybrid}~mm for the data-driven models versus \res{smb.t6.r0.err.physres} and \res{smb.t6.r0.err.physonly}~mm for the water-balance models, which require more added rainfall to cross.

\par\textbf{Sea-level response under held and rule-controlled operations.}
When operations are held fixed at $t_0$, closed gates and one-way flap gates block coastal inflow at most issue times, leaving fewer than \res{smb.t3.hold.max_per100} responsive cells per 100 issue times across near-barrage nodes for sea-level rises of $7.62$--$60.96$~cm. Under closed-loop rule control, replaying the base operation schedule into the perturbed sea-level run decomposes the total stage response into the direct coastal push and the controller's compensatory pumping/gating reaction. Summed across all \res{smb.t3.cells} near-barrage cells (\res{smb.t3.n_responsive} responsive), the net stage rise of \res{smb.t3.total}~m decomposes into a direct coastal push of \res{smb.t3.direct}~m offset by \res{smb.t3.comp}~m of controller compensation (\res{smb.t3.compshare}\% of the direct push).

\par\textbf{Held operations across the network and at the reservoir.}
A held operation setting lies within training support when its combination of open gate groups, active pumps, and tide regime occurs in at least 1\% of rule-controlled training steps (the mixed regime accounts for \res{smb.k11.mixed_share_train_steps}\% of training steps, so none of its held configurations reach the 1\% support threshold). Across all scored nodes, median response ratios to held gate or pump settings remain within $\pm$\res{smb.t4.maxabs.typical} across all regimes and support strata (except for held gates in the off-support pump regime, where one physics model reaches \res{smb.t4.big.min}), even though \res{smb.t4x.far.simshare}\% of SWMM's responsive cells lie three or more links upstream of the reservoir (where model ratios stay within $\pm$\res{smb.t4x.far.maxabs.pump}).

At the reservoir node itself (Table~\ref{tab:s-smb-ops}), the rule-trained Scenario GNN and Hybrid forecaster respond to held pumps with the \emph{opposite} sign to SWMM across every tide regime, reflecting the closed-loop training correlation in which more pumps operate when reservoir stages are higher. Untrained copies of the three models containing a physical branch exhibit the correct physical pump sign (ratios \res{smb.t4x.res.pump.r0.untrained.phys.min} to \res{smb.t4x.res.pump.r0.untrained.phys.max}), which the water-balance models preserve after training but training reverses for the Hybrid model. For held gates, all models except the Scenario GNN match the simulator's sign in the gravity regime, whereas all four exhibit the opposite sign in the mixed and pump regimes (already present in the untrained physics-branch copies at \res{smb.t4x.res.gate.r0.pump.untrained.phys.min} to \res{smb.t4x.res.gate.r0.pump.untrained.phys.max}, as model gate edges omit one-way flap valves). Training on randomized gate schedules restores the correct gate sign in the gravity regime for all four architectures while leaving pump sign reversals in the data-driven models unchanged.

\par\textbf{Operation threshold decisions.}
In the operation-avoid task (identifying whether some held setting keeps a warned node below its threshold), the rule-trained water-balance models fail to warn at the held base in \res{smb.t5b.r0.initmiss.phys.min}--\res{smb.t5b.r0.initmiss.phys.max}\% of warned nodes (precision \res{smb.t5b.r0.prec.phys.min}--\res{smb.t5b.r0.prec.phys.max}; balanced accuracies in Table~\ref{tab:s-smb-ops}). In the gates-stuck-closed harm task (positive rate \res{smb.t5a.o3.rate}\% [\res{smb.t5a.o3.rate.lo}, \res{smb.t5a.o3.rate.hi}], below the 5\% evaluation threshold, as is the pumps-off harm task), a validation-fitted linear controller read-out baseline achieves a balanced accuracy of \res{smb.crb.o3.ba} [\res{smb.crb.o3.ba.lo}, \res{smb.crb.o3.ba.hi}]. In the set-point task (lowering all control levels by $0.5$~m), the linear read-out baseline reaches \res{smb.crb.o4.ba} [\res{smb.crb.o4.ba.lo}, \res{smb.crb.o4.ba.hi}]; when the Scenario GNN and Hybrid are evaluated by adding their predicted stage change to the simulated base, they achieve at most \res{smb.o4.same.r0.dd.ba.max} balanced accuracy and match the simulated direction in only \res{smb.o4.delta.r0.dd.same.min}--\res{smb.o4.delta.r0.dd.same.max}\% of responsive cells.

\par\textbf{Observational check under closed-loop operations.}
Evaluating the observational check with operations as the substituted driver (comparing predictions under realized rule-controlled operations against held issue-time operations on storms with co-occurring sea level; Table~\ref{tab:s-smb-check}), the rule-trained Scenario GNN and Hybrid pass the pooled rule (station FE slopes \res{smb.t7.r0.ctrl.slope.gnn} and \res{smb.t7.r0.ctrl.slope.hybrid}), and the Scenario GNN also passes the strict rule, whereas both water-balance models fail. Yet on the \res{smb.t7.truth.n} responsive operation-affected cells of the corresponding paired simulator substitution (median simulated change \res{smb.t7.truth.dsim_med}~cm), the Scenario GNN predicts near-zero change ($<1$~mm) in \res{smb.t7.truth.r0.none.gnn}\% of cells and the wrong sign in \res{smb.t7.truth.r0.opp.gnn}\% (Hybrid: \res{smb.t7.truth.r0.none.hybrid}\% and \res{smb.t7.truth.r0.opp.hybrid}\%), whereas the water-balance models disagree in sign in at most \res{smb.t7.truth.r0.opp.phys.max}\% of cells. Because closed-loop operations respond to water levels, a forecaster can pass the observational check even when its causal response to control actions is absent or inverted, a failure that only paired simulator interventions expose.

\par\textbf{Parameterization and roughness sensitivity.}\label{app:M6}
To prevent artificial free seiches and overbank losses from obscuring held-setting responses in SWMM, the baseline barrage network uses an effective Manning roughness of $0.030$ on the \res{smb.dev.trunk_conduits} trunk and reservoir-inlet conduits and $+4.0$~m bank elevations on \res{smb.dev.bank_conduits} conduits, excluding \res{smb.nodes.excluded} ringing nodes via a validation-split oscillation screen. All \res{smb.runs.core} core simulations and \res{smb.runs.primary} primary test runs completed without solver errors. Re-simulating all test issue times with trunk Manning roughness reduced to $0.025$ preserves every qualitative finding and decomposition sign: the summed closed-loop sea-level rise shifts modestly from \res{smb.t3.total} to \res{smb.tier1.t3.total}~m (controller compensation share from \res{smb.t3.compshare}\% to \res{smb.tier1.t3.compshare}\%), and the controller read-out baseline balanced accuracy on the gates-stuck-closed task remains stable (\res{smb.crb.o3.ba} vs.\ \res{smb.tier1.crb.o3.ba}). Extending the evaluation window from 6~h to 24~h across \res{smb.k2b.var.runs} test runs confirms that 6-h held-pump reservoir responses capture \res{smb.k2b.capture.min}--\res{smb.k2b.capture.max}\% of the 24-h ramp change without altering the model comparison.
}

\section{Hardware and Software Environment}\label{app:M}

\par\textbf{Hardware.}
Model training, scenario evaluation, and explanation queries were executed on NVIDIA A100 GPUs (40~GB or 80~GB VRAM) on shared compute nodes. Models for the second simulated network (\emph{gamma}) were trained on CPUs, and SWMM simulations for both \emph{beta} and \emph{gamma} were executed on CPUs using eight parallel worker processes. Models for the synthetic tidal-barrage testbed (App.~\ref{app:Lsmb}) were trained on NVIDIA RTX 5090 GPUs, with SWMM simulations executed across 24 parallel CPU processes.

\par\textbf{Software.}
Forecasters and evaluation pipelines were implemented in Python~3.12 with PyTorch~2.6.0 (CUDA~12.4; PyTorch~2.7.1 with CUDA~12.8 for the synthetic barrage testbed), NumPy~1.26.4, pandas~2.2.3, and SciPy~1.15.2. Hydraulic simulations used SWMM~5.2.4 via \texttt{pyswmm}~2.1.0 and \texttt{swmm-toolkit}~0.17.0~\cite{rossman2022swmm,mcdonnell2020pyswmm} with benchmark network definitions from \texttt{pystorms}~1.0.0~\cite{rimer2023pystorms}. Figures and statistical summaries were generated with \texttt{matplotlib}~3.9.2, and the manuscript was compiled with TeX~Live~2026.

\section{Gate, Weir and Pump Operations on the Coastal Network}\label{app:P}

This appendix documents the operation experiment, the second set of paired simulator runs on the simulated coastal
network (pystorms \emph{beta}; App.~\ref{app:D}). The first set of
runs on this network holds one fixed control path (App.~\ref{app:B4}). The second set tests whether a forecaster knows
what a changed operation would do. It trains the four architectures with random operation schedules as scenario inputs,
asks what holding a gate, a weir or the pump closed or open would change, and grades that kind of explanation against
paired runs (Section~\ref{sec:results-ops}). This appendix describes the operated assets, storms and
training, the grading criteria, the paired runs and measures, the responses per
operated asset, architecture and variant, the interactions and threshold decisions, the observational check, a spatial
breakdown of where the responses sit and the forecast skill with realized and held schedules.

\subsection{Operated assets, storms and training}\label{app:P1}

\par\textbf{Operated assets.}
The network, its numerical settings and its scored nodes are those of App.~\ref{app:D1}, and three operated assets now
follow an operation schedule. The orifice R2 at the outlet of the inland storage unit acts as a gate whose opening is
set between 0 (closed) and 1 (fully open). The weir W0 between J205 and J33, two links from the outfall, is set in the
same range. The pump P0, from its wet well J56 to J55, is commanded on (1) or off (0), and its start and stop depths are
set to zero, so that only the command switches it. In the first set of runs R2 and W0 stay open and P0 is held on. The
outfall keeps its prescribed stage of astronomical tide and surge behind the flap gate, and J56 and J57 stay outside
every scored set, which leaves \res{ctrl.n-scored-nodes} scored nodes.

\par\textbf{Storms and schedules.}
The \res{data.beta.n-storms} storms of 48~h keep the rain, astronomical tide and surge of the first set, drawn by the
same generator with the same seeds, and its split by storm into \res{data.beta.n-storms-train} training,
\res{data.beta.n-storms-val} validation and \res{data.beta.n-storms-test} test storms. Only the operations differ. R2
and W0 start at a setting drawn uniformly between 0 and 1 and draw a new setting with probability 1/12 at every 15-min
boundary, so a setting is held for 3~h on average. The pump starts on or off with equal probability and switches with
probability 1/12 at every boundary. Each storm's schedule comes from its own seeded generator, which leaves the rain,
tide and surge draws untouched. No schedule depends on the water state, so a known future schedule tells a model
nothing about the future state except through the operations' own effect, as for the valves of the second simulated
network (App.~\ref{app:L6}). Settings change at the 15-min boundaries and are read back from the engine every minute.
The generated storms equal those of the first set in rain, tide, surge and outlet stage and differ from them in their
depths, because the operations move the water.

\par\textbf{Training.}
Each architecture is trained as on the first set (App.~\ref{app:C4}), with seeds 42, 43 and 44, 24-step windows of
history and horizon (6~h each) and future rain and surge, except that the realized future schedule of the three assets
enters as the operation part $W$ of the scenario instead of being held. One untrained copy per architecture uses the
seed-42 configuration with initialization seed 0 (App.~\ref{app:G5}). The issue-time scenario $q_0$ holds the
operations at their last setting, adds no future rain and holds the surge at its last value.

\subsection{Grading and decision criteria}\label{app:P2}

\par\textbf{Classes.}
For each operated asset $a$ and architecture, on seed means, the operation explanation is \emph{shown wrong} if
$|k_a|<0.2$ or its sign agreement is below 0.5, \emph{supported} if $0.75\le k_a\le1.33$, its sign agreement is at
least 0.8 and the untrained copy has $|k_a|<0.2$, and \emph{supported in part} otherwise. A note \emph{built in} is
added when the untrained copy reaches $|k_a|\ge0.2$ with a sign agreement of at least 0.6, because such a response is not
credited to training. The rule is also applied to the half and reversed variants, as a test of the rule itself.

\par\textbf{Interactions, decisions and cell counts.}
An interaction counts as reproduced if its signed recall on detectable cells is at least 0.5 and its sign agreement at
least 0.7. The operation decisions count as informative if their balanced accuracy exceeds both constant classifiers
(0.5) by at least 0.1 and the level-only answer by at least 0.05, and a balanced accuracy of at least 0.6 without the
second margin reads \emph{only through the forecast level}. A class, interaction or decision resting on fewer than 20
responsive cells, 20 detectable cells or 20 cells in either simulated class would have been left without a verdict,
and none fell below these counts. The observational check \emph{misses a wrong response} if it passes a half or
reversed variant.

\subsection{Paired runs and measures}\label{app:P3}

\par\textbf{Runs.}
Each of the \res{ctrl.n-issue} test storms has one issue time, 2~h before the applied rain peak and rounded down to
0.5~h as in the first set, and a window of the next 24 steps (6~h). Every run is continuous from $t=0$ and never
restored from a saved state. Each issue time has \res{ctrl.n-runs-per-issue} runs, \res{ctrl.n-runs} in all. They are
the base run with the realized drivers and schedule, the two noise runs of App.~\ref{app:D3}, six operation alternatives
that hold one asset at 0 or at 1 from the issue time to the window end while the other two follow the schedule, the
added outlet stage $\bar s$ of 30.48~cm alone and with each alternative, and the added rain $\bar A$ of 25.4~mm alone and with
each alternative, both added as in the first set (App.~\ref{app:B3} and~\ref{app:B4}). The base runs reproduce the
dataset, every run except the whole-storm noise run equals its base run before the issue time (largest differences
\res{ctrl.check.base-vs-dataset.cm} and \res{ctrl.check.pre-issue.cm}~cm), and every setting read back from the engine
equals its command (largest error \res{ctrl.check.readback.max}).

\par\textbf{Model side.}
The models receive the same changes in their scenario input, step for step and in the units of the dataset. This is
done for the twelve trained runs, their half and reversed variants and the four untrained copies. The variants replace
the operation input of every scenario by $W_0+a(W-W_0)$ with $a=1/2$ or $a=-1$, where $W$ is the scenario's operation
input and $W_0$ holds each asset at its setting in the last step before the issue time. A model's whole operation
response is therefore halved or reversed, while rain and surge stay as in the scenario. Unlike the variants of the first
set, which scale every driver change from $q_0$ (App.~\ref{app:H5}), these alter only the operations. A scaling from the
realized scenario would leave the forecast under the realized schedule unchanged and give the observational check a
slope near 1 for every variant. The observational check uses the same definition.

\par\textbf{Measures.}
All measures use the 6-h window peak at each scored node. The noise floor $\nu$ of a cell (issue time, node) is the
largest absolute difference within the window between a noise run and the base run. Its median is
\res{ctrl.noise.median.cm}~cm and its 90th percentile \res{ctrl.noise.p90.cm}~cm, and \res{ctrl.noise.share-le-1cm}\%
of cells lie at or below 1~cm. A cell responds to an alternative when $|\Delta_{\mathrm{sim}}|>\max(c_1\nu,\delta)$ with
$c_1=2$ and $\delta=1$~cm. The response ratio $k_a$ of asset $a$ is the median of
$\Delta_{\mathrm{model}}/\Delta_{\mathrm{sim}}$ over the responsive cells of its two alternatives (Eq.~\eqref{eq:paired}).
Its sign agreement is the share of these cells in which the model change has the simulator's sign, and a model change
of exactly zero counts as a disagreement. Both are computed per seed and then averaged over seeds. The interaction of
an alternative with $\bar s$ or $\bar A$ is the second difference of Eq.~\eqref{eq:interaction} with the base run as the
reference. It is detectable where $|I_{\mathrm{sim}}|>\max(c_2\nu,\delta)$ with $c_2=4$, and its signed recall is the
share of detectable cells in which the model's interaction exceeds the same bound with the simulator's sign. The
decisions use the \res{ctrl.n-threshold-nodes} nodes whose training p95 lies more than 5~cm below the rim. For a cell
warned at the simulated base, the question is whether some setting of the asset keeps the peak below the threshold
(avoid), and for an unwarned cell whether some setting lifts it to the threshold (lift). The level-only answer reads the
same question from the model's own base forecast without any operation, calling a warned cell avoidable when that
forecast already lies below the threshold and an unwarned cell liftable when it already lies above, so that a model that
only gets the forecast level right earns no credit for the operation. As context without a rule, $k$ is also computed
for $\bar s$ alone at the near-outlet nodes and for $\bar A$ alone at the runoff-receiving nodes. The reach of an asset counts
the nodes at which the larger absolute change of its two settings, averaged over issue times, exceeds 1~cm. Forecast
skill is the median node NSE on the test storms with realized drivers, with the schedule held at its last setting and
the other drivers realized, and under $q_0$.

\subsection{Responses per operated asset}\label{app:P4}

\begin{table*}[tp]
\caption{Operation responses on the simulated coastal network across \res{ctrl.n-issue} test issue times: response ratio $k$ (mean\,$\pm$\,SD over three seeds), sign agreement (\%), and untrained $k$.}
\label{tab:s-ops}
\centering
\setlength{\tabcolsep}{4pt}%
\resizebox{0.86\textwidth}{!}{%
\begin{tabular}{@{}lccccccccc@{}}
\toprule
 & \multicolumn{3}{c}{Inland orifice R2} & \multicolumn{3}{c}{Weir W0} & \multicolumn{3}{c}{Pump P0} \\
\cmidrule(lr){2-4}\cmidrule(lr){5-7}\cmidrule(l){8-10}
 & $k$ & Sign & Untrained & $k$ & Sign & Untrained & $k$ & Sign & Untrained \\
\midrule
Scenario GNN & \res{ctrl.k.r2.gnn.trained.mean}\,$\pm$\,\res{ctrl.k.r2.gnn.trained.sd} & \res{ctrl.sign.r2.gnn.trained.mean} & \res{ctrl.k.r2.gnn.untrained} & \res{ctrl.k.w0.gnn.trained.mean}\,$\pm$\,\res{ctrl.k.w0.gnn.trained.sd} & \res{ctrl.sign.w0.gnn.trained.mean} & \res{ctrl.k.w0.gnn.untrained} & \res{ctrl.k.p0.gnn.trained.mean}\,$\pm$\,\res{ctrl.k.p0.gnn.trained.sd} & \res{ctrl.sign.p0.gnn.trained.mean} & \res{ctrl.k.p0.gnn.untrained} \\
Hybrid & \res{ctrl.k.r2.hybrid.trained.mean}\,$\pm$\,\res{ctrl.k.r2.hybrid.trained.sd} & \res{ctrl.sign.r2.hybrid.trained.mean} & \res{ctrl.k.r2.hybrid.untrained} & \res{ctrl.k.w0.hybrid.trained.mean}\,$\pm$\,\res{ctrl.k.w0.hybrid.trained.sd} & \res{ctrl.sign.w0.hybrid.trained.mean} & \res{ctrl.k.w0.hybrid.untrained} & \res{ctrl.k.p0.hybrid.trained.mean}\,$\pm$\,\res{ctrl.k.p0.hybrid.trained.sd} & \res{ctrl.sign.p0.hybrid.trained.mean} & \res{ctrl.k.p0.hybrid.untrained} \\
Physics + state residual & \res{ctrl.k.r2.physres.trained.mean}\,$\pm$\,\res{ctrl.k.r2.physres.trained.sd} & \res{ctrl.sign.r2.physres.trained.mean} & \res{ctrl.k.r2.physres.untrained} & \res{ctrl.k.w0.physres.trained.mean}\,$\pm$\,\res{ctrl.k.w0.physres.trained.sd} & \res{ctrl.sign.w0.physres.trained.mean} & \res{ctrl.k.w0.physres.untrained} & \res{ctrl.k.p0.physres.trained.mean}\,$\pm$\,\res{ctrl.k.p0.physres.trained.sd} & \res{ctrl.sign.p0.physres.trained.mean} & \res{ctrl.k.p0.physres.untrained} \\
Physics-only & \res{ctrl.k.r2.physonly.trained.mean}\,$\pm$\,\res{ctrl.k.r2.physonly.trained.sd} & \res{ctrl.sign.r2.physonly.trained.mean} & \res{ctrl.k.r2.physonly.untrained} & \res{ctrl.k.w0.physonly.trained.mean}\,$\pm$\,\res{ctrl.k.w0.physonly.trained.sd} & \res{ctrl.sign.w0.physonly.trained.mean} & \res{ctrl.k.w0.physonly.untrained} & \res{ctrl.k.p0.physonly.trained.mean}\,$\pm$\,\res{ctrl.k.p0.physonly.trained.sd} & \res{ctrl.sign.p0.physonly.trained.mean} & \res{ctrl.k.p0.physonly.untrained} \\
\midrule
Simulator, responsive cells & \multicolumn{3}{c}{\res{ctrl.sim.r2.n-responsive}} & \multicolumn{3}{c}{\res{ctrl.sim.w0.n-responsive}} & \multicolumn{3}{c}{\res{ctrl.sim.p0.n-responsive}} \\
Simulator, median change (cm) & \multicolumn{3}{c}{\res{ctrl.sim.r2.change.median.cm}} & \multicolumn{3}{c}{\res{ctrl.sim.w0.change.median.cm}} & \multicolumn{3}{c}{\res{ctrl.sim.p0.change.median.cm}} \\
\bottomrule
\end{tabular}}
\end{table*}

\par\textbf{Simulator.}
Every operated asset moves simulated window peaks (Table~\ref{tab:s-ops}). Of the (issue time, setting, node) cells,
\res{ctrl.sim.r2.share-responsive}\% respond to the inland orifice, \res{ctrl.sim.w0.share-responsive}\% to the weir and
\res{ctrl.sim.p0.share-responsive}\% to the pump, with median absolute changes of \res{ctrl.sim.r2.change.median.cm},
\res{ctrl.sim.w0.change.median.cm} and \res{ctrl.sim.p0.change.median.cm}~cm and 90th percentiles of
\res{ctrl.sim.r2.change.p90.cm}, \res{ctrl.sim.w0.change.p90.cm} and \res{ctrl.sim.p0.change.p90.cm}~cm. The simulated
peak falls in \res{ctrl.sim.r2.share-negative}\%, \res{ctrl.sim.w0.share-negative}\% and
\res{ctrl.sim.p0.share-negative}\% of the responsive cells, so every asset raises peaks in some cells and lowers them in
others. The reach is \res{ctrl.sim.r2.reach.n}, \res{ctrl.sim.w0.reach.n} and
\res{ctrl.sim.p0.reach.n} of the \res{ctrl.n-scored-nodes} scored nodes, up to \res{ctrl.sim.r2.reach.maxhops},
\res{ctrl.sim.w0.reach.maxhops} and \res{ctrl.sim.p0.reach.maxhops} links from the asset.

\par\textbf{Models.}
Every architecture and seed answers with a median response ratio of \res{ctrl.k.r2.gnn.trained.mean} to each asset
(Table~\ref{tab:s-ops}), with sign agreement from \res{ctrl.sign.w0.physonly.trained.mean}\% to
\res{ctrl.sign.p0.hybrid.trained.mean}\%, so the operation explanation is \res{ctrl.class.r2.gnn.trained} for all
twelve architecture and asset pairs. The untrained copies also have $k$ of \res{ctrl.k.r2.gnn.untrained}, and no pair
carries the note built in. The operation input is active, since holding the schedule at its last setting moves
the test forecasts by a mean of \res{ctrl.skill.physres.hold-move.cm} to \res{ctrl.skill.gnn.hold-move.cm}~cm, and the
spatial breakdown of App.~\ref{app:P7} finds responses next to each asset. Trained with the schedule as an input, the
models still respond to added rain, with $k$ from \res{ctrl.k.rain.gnn.trained.mean} (Scenario GNN) to
\res{ctrl.k.rain.physres.trained.mean} (Physics + state residual), and not to the raised outlet stage
(\res{ctrl.k.stage.gnn.trained.mean} for every architecture), as on the first set. The class rule gives the half and
reversed variants the same class, as it must when the trained response is already zero, and reversing the input lowers
the physics models' sign agreement to values from \res{ctrl.sign.w0.physres.reversed.mean}\% to
\res{ctrl.sign.r2.physonly.reversed.mean}\%.

\subsection{Interactions and threshold decisions}\label{app:P5}

\par\textbf{Interactions.}
In the simulator the effect of the raised outlet stage depends detectably on the setting of an asset in
\res{ctrl.sim.r2.int-stage.n-detectable} cells for the inland orifice, \res{ctrl.sim.w0.int-stage.n-detectable} for the
weir and \res{ctrl.sim.p0.int-stage.n-detectable} for the pump, and the effect of the added rain in
\res{ctrl.sim.r2.int-rain.n-detectable}, \res{ctrl.sim.w0.int-rain.n-detectable} and
\res{ctrl.sim.p0.int-rain.n-detectable} cells. No interaction is reproduced. The largest signed recall of a trained
architecture is \res{ctrl.int-stage.w0.gnn.recall}\% for an operation with the raised stage (Scenario GNN, weir) and
\res{ctrl.int-rain.r2.physonly.recall}\% for an operation with added rain (Physics-only model, inland orifice), against
the 50\% that the rule requires, and every untrained copy has a signed recall of
\res{ctrl.int-rain.r2.gnn.untrained-recall}\%.

\par\textbf{Threshold decisions.}
Few cells can change sides of the threshold. Some setting keeps a warned node below its threshold in
\res{ctrl.sim.r2.dec.avoidable}\% (inland orifice), \res{ctrl.sim.w0.dec.avoidable}\% (weir) and
\res{ctrl.sim.p0.dec.avoidable}\% (pump) of the \res{ctrl.sim.r2.dec.n-warned} warned cells, and lifts an unwarned node
to it in \res{ctrl.sim.r2.dec.liftable}\%, \res{ctrl.sim.w0.dec.liftable}\% and \res{ctrl.sim.p0.dec.liftable}\% of the
unwarned cells. For the avoid question the trained models reach balanced accuracy from
\res{ctrl.dec.w0.physonly.avoid.ba} to \res{ctrl.dec.p0.physres.avoid.ba}, and their level-only answers reach from
\res{ctrl.dec.w0.physonly.avoid.ba-level} to \res{ctrl.dec.p0.physres.avoid.ba-level}. No model exceeds its level-only
answer by the margin of 0.05 that the rule requires, the largest gap being \res{ctrl.dec.w0.gnn.avoid.ba} against
\res{ctrl.dec.w0.gnn.avoid.ba-level} (Scenario GNN, weir), so all twelve pairs answer only through the forecast level.
For the lift question the balanced accuracy runs from \res{ctrl.dec.r2.physres.lift.ba} to
\res{ctrl.dec.p0.gnn.lift.ba}, which is not informative.

\subsection{Observational check for operations}\label{app:P6}

\begin{figure*}[tp]
\centering
\includegraphics[width=\textwidth]{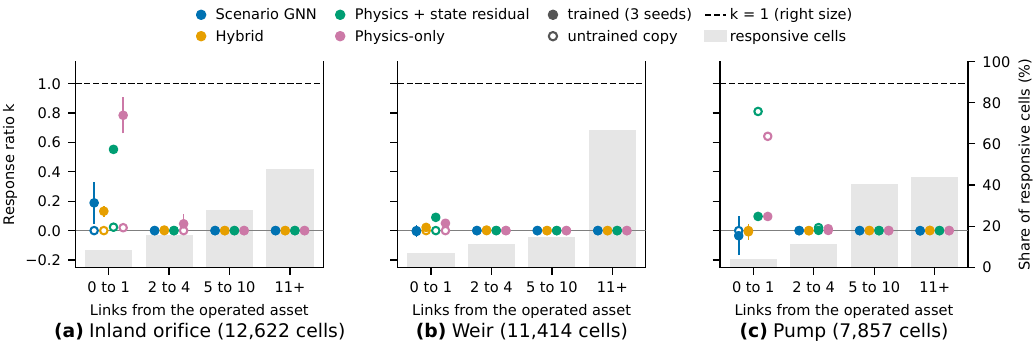}
\caption{Operation response ratio $k$ by link distance from the operated asset on the simulated coastal network (median $k$, mean\,$\pm$\,SD over three seeds).}
\label{fig:s-ops-reach}
\end{figure*}

\par\textbf{Setting.}
The observational check of Eq.~\eqref{eq:substitution} runs on the test storms with the schedule as the substituted
driver. The baseline $B$ holds the schedule at its last setting, so $\Delta_p$ is the forecast change that the realized
schedule causes and $\Delta_o$ the simulated base peak minus $B$. Its main group holds all three assets together at all
scored nodes, and three secondary groups hold one asset each while the other two follow the schedule. Both rules of App.~\ref{app:H3} apply, with $\delta_o=1$~cm, storm-block intervals from 500
resamples and, as the shuffled driver, the schedule changes of another test storm at a random offset, clipped to
settings between 0 and 1. The regression weighs only the cells in which a model predicts a change, so the cells of a
missing response carry no weight. The predicted changes it sees in the main group average from
\res{ctrl.obs.ctrl.mean-abs-dpred.min} to \res{ctrl.obs.ctrl.mean-abs-dpred.max}~cm per trained run and lie
mainly near each asset (App.~\ref{app:P7}).

\par\textbf{Main group.}
Under the pooled rule the two physics models are not separable from their untrained copies, with pooled slopes of
\res{ctrl.obs.ctrl.physres.trained.orig.slope} (Physics + state residual) and \res{ctrl.obs.ctrl.physonly.trained.orig.slope}
(Physics-only model) against \res{ctrl.obs.ctrl.physres.untrained.orig.slope} and
\res{ctrl.obs.ctrl.physonly.untrained.orig.slope} for the copies, which also meet the slope and sign criteria. The
Scenario GNN and the Hybrid forecaster fail only through the shuffle criterion, in
\res{ctrl.obs.ctrl.gnn.trained.orig.n-seeds-shuffle-fail} and \res{ctrl.obs.ctrl.hybrid.trained.orig.n-seeds-shuffle-fail}
of three seeds, with pooled slopes of \res{ctrl.obs.ctrl.gnn.trained.orig.slope} and
\res{ctrl.obs.ctrl.hybrid.trained.orig.slope} inside the window. Under the strict rule the Scenario GNN fails,
with a fixed-effect slope of \res{ctrl.obs.ctrl.gnn.trained.v2.slope} below the window, and the other three
architectures are uninformative, because their mean predicted change stays below 0.1~cm.

\par\textbf{One asset held.}
In the secondary groups the pooled rule passes \res{ctrl.obs.perasset.trained.orig.n-pass} of the
\res{ctrl.obs.perasset.n-pairs} architecture and asset pairs, all of which the paired runs show wrong. They are the
inland orifice for the Hybrid forecaster and both physics models (pooled slopes from
\res{ctrl.obs.r2.physres.trained.orig.slope} to \res{ctrl.obs.r2.hybrid.trained.orig.slope}) and the weir for all four
architectures (from \res{ctrl.obs.w0.gnn.trained.orig.slope} to \res{ctrl.obs.w0.hybrid.trained.orig.slope}). Every pump
group fails through the shuffle criterion in all three seeds, and the strict rule calls every secondary group
uninformative.

\par\textbf{Altered-input variants.}
Neither rule passes a half or reversed variant, in the main group or in any secondary group. Under the criterion of App.~\ref{app:P2}, which counts only a pass of a half or reversed variant as a miss, the observational
check therefore has no miss here, although its pooled rule passes \res{ctrl.obs.perasset.trained.orig.n-pass} of the
\res{ctrl.obs.perasset.n-pairs} trained architecture and asset pairs of the secondary groups, which the paired runs
show wrong. In the main group the pooled rule rejects the half variants of the Scenario GNN and the two physics
models only through the shuffle criterion (in \res{ctrl.obs.ctrl.gnn.half.orig.n-seeds-shuffle-fail},
\res{ctrl.obs.ctrl.physres.half.orig.n-seeds-shuffle-fail} and
\res{ctrl.obs.ctrl.physonly.half.orig.n-seeds-shuffle-fail} of three seeds), with pooled slopes from
\res{ctrl.obs.ctrl.gnn.half.orig.slope} to \res{ctrl.obs.ctrl.physonly.half.orig.slope} inside the window. In the main
group it also rejects the half variant of the Hybrid forecaster through its slope of
\res{ctrl.obs.ctrl.hybrid.half.orig.slope} above the window and every reversed variant through a negative slope (from
\res{ctrl.obs.ctrl.hybrid.reversed.orig.slope} to \res{ctrl.obs.ctrl.physonly.reversed.orig.slope}). In the same group
the strict rule calls every half variant and the reversed Hybrid forecaster uninformative and fails the other
reversed variants. The observational check thus rejects inputs altered to be wrong, while only the paired runs settle
the operation responses of the trained architectures, which they show wrong.

\subsection{Spatial breakdown of operation responses}\label{app:P7}

\par\textbf{Breakdown by link distance.}
While the all-node median response ratio $k$ is \res{ctrl.k.r2.gnn.trained.mean} for every asset and architecture, splitting the
responsive cells by the link distance of their node from the nearer of the asset's two end nodes, in bands
of 0 to 1, 2 to 4, 5 to 10 and 11 or more links (Fig.~\ref{fig:s-ops-reach}), locates where the models respond. The band of 0 to 1
holds the end nodes and their direct neighbors, with \res{ctrl.ph.r2.n-le1link} responsive cells for the inland orifice,
\res{ctrl.ph.w0.n-le1link} for the weir and \res{ctrl.ph.p0.n-le1link} for the pump. Within it the response
ratio to the inland orifice is \res{ctrl.ph.k.r2.physonly.trained.le1link} for the Physics-only model,
\res{ctrl.ph.k.r2.physres.trained.le1link} for the Physics + state residual model,
\res{ctrl.ph.k.r2.gnn.trained.le1link} for the Scenario GNN and \res{ctrl.ph.k.r2.hybrid.trained.le1link} for the
Hybrid forecaster, against at most \res{ctrl.ph.k.r2.physres.untrained.le1link} for the untrained copies. In the same
band the weir reaches at most \res{ctrl.ph.k.w0.physres.trained.le1link} and the pump
\res{ctrl.ph.k.p0.physres.trained.le1link} for each physics model, whose untrained copies already respond to
the pump with \res{ctrl.ph.k.p0.physres.untrained.le1link} and \res{ctrl.ph.k.p0.physonly.untrained.le1link}. Two or
more links away, where \res{ctrl.ph.r2.share-ge2links}\%, \res{ctrl.ph.w0.share-ge2links}\% and
\res{ctrl.ph.p0.share-ge2links}\% of the responsive cells lie, the median response ratio of every architecture and
asset is at most \res{ctrl.ph.k.r2.physonly.trained.ge2links.maxabs} in magnitude.

\par\textbf{Changes above 1~cm.}
The responses concentrate near the asset but do not stop at its end nodes. Two to four links from the inland
orifice the architectures move from \res{ctrl.ph.r2.share-gt1cm.2to4links.min}\% to
\res{ctrl.ph.r2.share-gt1cm.2to4links.max}\% of the responsive peaks by more than 1~cm (seed means), from the
weir from \res{ctrl.ph.w0.share-gt1cm.2to4links.min}\% to \res{ctrl.ph.w0.share-gt1cm.2to4links.max}\% and from the
pump from \res{ctrl.ph.p0.share-gt1cm.2to4links.min}\% to \res{ctrl.ph.p0.share-gt1cm.2to4links.max}\%, and five or
more links away they move none (\res{ctrl.ph.r2.gnn.share-gt1cm.ge5links.max}\% for every architecture and asset).
Counted per node, the models move the mean peak by more than 1~cm at no more than \res{ctrl.ph.reach.r2.model.n-max},
\res{ctrl.ph.reach.w0.model.n-max} and \res{ctrl.ph.reach.p0.model.n-max} nodes within
\res{ctrl.ph.reach.r2.model.maxhops}, \res{ctrl.ph.reach.w0.model.maxhops} and \res{ctrl.ph.reach.p0.model.maxhops} links
of the inland orifice, the weir and the pump, all of them nodes that the simulator also reaches, against the
simulator's \res{ctrl.sim.r2.reach.n}, \res{ctrl.sim.w0.reach.n} and \res{ctrl.sim.p0.reach.n} nodes up to
\res{ctrl.sim.r2.reach.maxhops}, \res{ctrl.sim.w0.reach.maxhops} and \res{ctrl.sim.p0.reach.maxhops} links away. This
is the pattern of the raised outlet stage in the first set, which the models reproduce only at the junctions next to
the outlet (App.~\ref{app:I2}).

\par\textbf{Reading.}
The breakdown locates where the models respond. On the second simulated network, trained with random valve schedules,
every architecture learns the response to valve settings partly in size (response ratio from
\res{gamma.k_valve.gnn.trained} to \res{gamma.k_valve.physonly.trained}, App.~\ref{app:L6}). The two networks differ in
size, operated assets, storms and ocean boundary.

\subsection{Forecast skill with realized and held schedules}\label{app:P8}

\par\textbf{Skill.}
With realized drivers and schedule, the median node NSE on the test storms is, for the Scenario GNN,
\res{ctrl.skill.gnn.real.nse.mean}\,$\pm$\,\res{ctrl.skill.gnn.real.nse.sd}, for the Hybrid forecaster
\res{ctrl.skill.hybrid.real.nse.mean}\,$\pm$\,\res{ctrl.skill.hybrid.real.nse.sd}, for the Physics + state residual
model \res{ctrl.skill.physres.real.nse.mean}\,$\pm$\,\res{ctrl.skill.physres.real.nse.sd} and for the Physics-only
model \res{ctrl.skill.physonly.real.nse.mean}\,$\pm$\,\res{ctrl.skill.physonly.real.nse.sd}, against
\res{ctrl.skill.persistence.nse} for persistence. With the schedule held at its last setting and the
other drivers realized, the seed means print the same values (\res{ctrl.skill.gnn.ctrl-hold.nse.mean},
\res{ctrl.skill.hybrid.ctrl-hold.nse.mean}, \res{ctrl.skill.physres.ctrl-hold.nse.mean} and
\res{ctrl.skill.physonly.ctrl-hold.nse.mean}), and no seed of any architecture differs by more than
\res{ctrl.skill.physres.real-minus-hold.maxabs}, reached by the Physics + state residual model. Under the issue-time
scenario $q_0$, which also removes the future rain and holds the surge, the NSE falls to \res{ctrl.skill.gnn.q0.nse.mean},
\res{ctrl.skill.hybrid.q0.nse.mean}, \res{ctrl.skill.physres.q0.nse.mean} and \res{ctrl.skill.physonly.q0.nse.mean}.
Forecast accuracy therefore leaves the missing operation response on this network undetected, and the paired runs
expose it.

\end{document}